\documentclass[final,5p,times,twocolumn,authoryear]{elsarticle}

\usepackage{amsmath}
\usepackage{amssymb}
\usepackage{lipsum}
\usepackage[hyphens]{url}
\usepackage{breakurl}
\usepackage{ulem}
\usepackage{xcolor}

\usepackage[utf8]{inputenc}
\usepackage[T1]{fontenc}
\usepackage[english]{babel}

\usepackage{supertabular}
\usepackage{booktabs}
\usepackage{placeins}
\usepackage{graphicx}
\usepackage{caption}
\usepackage{float}
\usepackage{color, colortbl}
\usepackage{geometry}

\usepackage{makecell}
\usepackage[figuresright]{rotating}
\usepackage{pdflscape}

\usepackage[colorlinks,
            linkcolor=red,
            anchorcolor=blue,
            citecolor=green
            ]{hyperref}

\usepackage{threeparttable}
\usepackage{afterpage}
\usepackage{placeins}

\begin{document}

\begin{frontmatter}

%% Title, authors and addresses

%% use the tnoteref command within \title for footnotes;
%% use the tnotetext command for theassociated footnote;
%% use the fnref command within \author or \affiliation for footnotes;
%% use the fntext command for theassociated footnote;
%% use the corref command within \author for corresponding author footnotes;
%% use the cortext command for theassociated footnote;
%% use the ead command for the email address,
%% and the form \ead[url] for the home page:
%% \title{Title\tnoteref{label1}}
%% \tnotetext[label1]{}
%% \author{Name\corref{cor1}\fnref{label2}}
%% \ead{email address}
%% \ead[url]{home page}
%% \fntext[label2]{}
%% \cortext[cor1]{}
%% \affiliation{organization={},
%%            addressline={}, 
%%            city={},
%%            postcode={}, 
%%            state={},
%%            country={}}
%% \fntext[label3]{}

\title{Deep Learning for Wildfire Risk Prediction: Integrating Remote Sensing and Environmental Data}

%% use optional labels to link authors explicitly to addresses:
%% \author[label1,label2]{}
%% \affiliation[label1]{organization={},
%%             addressline={},
%%             city={},
%%             postcode={},
%%             state={},
%%             country={}}
%%
%% \affiliation[label2]{organization={},
%%             addressline={},
%%             city={},
%%             postcode={},
%%             state={},
%%             country={}}

\author[inst1,inst2]{Zhengsen Xu}
\author[inst2,inst3]{Jonathan Li *}
\author[inst4]{Sibo Cheng}
\author[inst5]{Xue Rui}
\author[inst6]{Yu Zhao}
\author[inst2]{Hongjie He}
\author[inst7]{Haiyan Guan}
\author[inst8]{Aryan Sharma}
\author[inst9]{Matthew Erxleben}
\author[inst3]{Ryan Chang}
\author[inst1]{Linlin Xu*}

\affiliation[inst1]{organization={Schulich School of Engineering, Department of Geomatics Engineering, University of Calgary},%Department and Organization
            addressline={2500 University Dr NW}, 
            city={Calgary},
            postcode={T2N1N4}, 
            state={Alberta},
            country={Canada}}  

\affiliation[inst2]{organization={Department of Geography and Environmental Management, University of Waterloo},%Department and Organization
            addressline={100 University Avenue}, 
            city={Waterloo},
            postcode={N2L3G1}, 
            state={Ontario},
            country={Canada}}
            
\affiliation[inst3]{organization={Department of Systems Design Engineering, University of Waterloo},%Department and Organization
            addressline={100 University Avenue}, 
            city={Waterloo},
            postcode={N2L3G1}, 
            state={Ontario},
            country={Canada}}
            
\affiliation[inst4]{
            organization={CEREA, ENPC, EDF R\&D, Institut Polytechnique de Paris},
            state={Ile-de-France},
            country={France}}

\affiliation[inst5]{organization={School of Emergency Management, Nanjing University of Information Science and Technology},
            addressline={219 Ningliu Road},
            city={Nanjing},
            postcode={210044},
            country={China}}
            
\affiliation[inst6]{organization={Division of Geoinformatics, KTH Royal Institute of Technology},
            addressline={Teknikringen 10a},
            cite={Stockholm},
            postcode={100 44},
            country={Sweden}}
            
\affiliation[inst7]{organization={School of Remote Sensing and Geomatics Engineering, Nanjing University of Information Science and Technology},
            addressline={219 Ningliu Road},
            city={Nanjing},
            postcode={210044},
            country={China}}
            
\affiliation[inst8]{organization={Department of Software Engineering, University of Waterloo},%Department and Organization
            addressline={100 University Avenue}, 
            city={Waterloo},
            postcode={N2L3G1}, 
            state={Ontario},
            country={Canada}}
            
\affiliation[inst9]{organization={Department of Management Science and Engineering, University of Waterloo},%Department and Organization
            addressline={100 University Avenue}, 
            city={Waterloo},
            postcode={N2L3G1}, 
            state={Ontario},
            country={Canada}}

\begin{abstract}
Wildfires pose a significant threat to ecosystems, wildlife, and human communities, leading to habitat destruction, pollutant emissions, and biodiversity loss. Accurate wildfire risk prediction is crucial for mitigating these impacts and safeguarding both environmental and human health. This paper provides a comprehensive review of wildfire risk prediction methodologies, with a particular focus on deep learning approaches combined with remote sensing. We begin by defining wildfire risk and summarizing the geographical distribution of related studies. In terms of data, we analyze key predictive features, including fuel characteristics, meteorological and climatic conditions, socioeconomic factors, topography, and hydrology, while also reviewing publicly available wildfire prediction datasets derived from remote sensing. Additionally, we emphasize the importance of feature collinearity assessment and model interpretability to improve the understanding of prediction outcomes. Regarding methodology, we classify deep learning models into three primary categories: time-series forecasting, image segmentation, and spatiotemporal prediction, and further discuss methods for converting model outputs into risk classifications or probability-adjusted predictions. Finally, we identify the key challenges and limitations of current wildfire-risk prediction models and outline several research opportunities. These include integrating diverse remote sensing data, developing multimodal models, designing more computationally efficient architectures, and incorporating cross-disciplinary methods—such as coupling with numerical weather-prediction models—to enhance the accuracy and robustness of wildfire-risk assessments.
\end{abstract}

%%Graphical abstract
%\begin{graphicalabstract}
%\includegraphics{grabs}
%\end{graphicalabstract}

%%Research highlights
%\begin{highlights}
%\item Research highlight 1
%\item Research highlight 2
%\end{highlights}

\begin{keyword}
%% keywords here, in the form: keyword \sep keyword, up to a maximum of 6 keywords
Wildfire \sep Risk Prediction \sep Remote Sensing \sep Deep Learning \sep Review

%% PACS codes here, in the form: \PACS code \sep code

%% MSC codes here, in the form: \MSC code \sep code
%% or \MSC[2008] code \sep code (2000 is the default)

\end{keyword}

\end{frontmatter}

%\tableofcontents

%% \linenumbers

%% main text

\section{Introduction}
\label{introduction}

Wildfires are uncontrolled fires that occur in vegetated landscapes \citep{huidobro2024and, NWCG_2006}. They can ignite naturally or be caused by human activities and are generally classified into planned fires, which are deliberately set for ecological or experimental purposes, and accidental fires, which result from natural causes or human negligence \citep{fillmore2024factors, huidobro2024and}. Wildfires occur across diverse environments, including forests, grasslands, farmlands, and peatlands, where they play a significant role in shaping global ecosystems. They influence biodiversity, air quality, human health, and carbon balance, with both beneficial and detrimental effects \citep{moritz2014learning, stojanovic2016loss, bowd2021prior, holder2023hazardous, lopez2024molecular}. On the one hand, wildfires contribute to ecosystem dynamics by promoting nutrient cycling, facilitating species turnover, and enhancing vegetation regeneration \citep{bond2005global, pausas2019wildfires}. On the other hand, their uncontrolled occurrence poses substantial risks, including biodiversity loss, air pollution, threats to human settlements, and economic disruptions \citep{moritz2014learning, bowd2021prior, yang_projections_2022, lopez2024molecular}.

The magnitude of wildfire-related disasters has escalated in recent decades. For instance, the 2019–2020 "Black Summer" wildfires in Australia burned approximately 18.6 million hectares, caused 34 fatalities, destroyed over 5,900 structures, and resulted in an estimated economic loss exceeding \$110 billion \citep{naderpour_wildfire_2020, haque2021wildfire}. In North America, the 2018 Camp Fire in California led to 85 fatalities and the destruction of nearly 19,000 structures, while the 2023 Maui wildfires resulted in nearly 100 deaths and \$550 million in damages \citep{marris2023hawaii, hertelendy2024seasons}. In Canada, wildfire frequency, size, and intensity have been increasing since the mid-20th century due to climate warming \citep{cunningham2024increasing, mulverhill2024wildfires}. From 1959 to 2015, the total burned area and the number of large wildfires (over 200 hectares) have significantly risen, and the wildfire season has extended by approximately two weeks \citep{hanes2019fire, jain2024drivers}. The 2023 wildfire season set a record, with 18.5 million hectares burned nationwide, displacing 232,000 people, destroying hundreds of homes, and releasing 647 Tg C into the atmosphere \citep{erni2024mapping, byrne2024carbon, jain2024drivers}. Investigations suggest that climate change will further shorten fire return intervals and expand burned areas in vulnerable regions \citep{parks_warmer_2020, holdrege_wildfire_2024}.

In response to these challenges, wildfire risk prediction has garnered increasing attention from governments, academia, and industry. Risk prediction involves delineating research scope, selecting and processing data, establishing methodologies, and understanding predictions. Despite the growing reliance on machine learning for wildfire risk prediction, a systematic review of these components remains lacking. This paper addresses this gap by providing a comprehensive review of the multi-source data and deep learning methods used in wildfire risk prediction, with a particular emphasis on feature attribution techniques and emerging deep learning models.

First, wildfire ignition and spread result from complex interactions among multiple factors, including fuel conditions, climate, weather, socio-economic variables, terrain, hydrology, and historical wildfire occurrences \citep{fuller1991forest}. This study systematically examines the definitions, proxies, frequency of use, and calculations (see Appendix) of these features, alongside an overview of publicly available datasets relevant to wildfire prediction.

Second, an essential consideration in wildfire risk modeling is the evaluation of feature collinearity and model interpretability. Highly correlated features can obscure individual contributions and degrade model performance, especially for linear models \citep{li_predictive_2022}, while model interpretability enhances the understanding of ignition and spread dynamics \citep{abdollahi_explainable_2023}. This paper introduces key techniques for assessing collinearity and improving interpretability in machine learning-based wildfire models.

Third, the development of predictive algorithms varies widely, ranging from traditional wildfire danger rating systems and statistical models to modern machine learning and deep learning approaches. While previous studies have reviewed conventional machine learning techniques \citep{jain_review_2020}, this paper focuses on deep learning, which has seen rapid adoption and is emerging as a crucial tool for wildfire prediction. We categorize deep learning-based models into three primary task types: time-series forecasting, image segmentation and classification, and spatiotemporal prediction. A structured review of these models is presented, highlighting their applications and limitations.

Finally, integrating multi-source geospatial data with deep learning models for wildfire risk prediction presents both opportunities and challenges. This paper discusses the limitations of existing algorithms and explores potential advancements in multimodal and architecturally efficient models, as well as the possible potential of new data sources (such as microwave, LiDAR, and nightlight data) and cross-domain weather forecasting models. The fusion of these diverse datasets with deep learning techniques has the potential to enhance predictive accuracy and improve early warning systems. Given the increasing wildfire risk under climate change, developing robust, interpretable, and computationally efficient models is crucial for informing mitigation strategies, resource allocation, and adaptive management policies \citep{erni2024mapping, mulverhill2024wildfires}.

\section{Wildfire Risk Definition}

\begin{figure}
    \centering
    \includegraphics[width=1\linewidth]{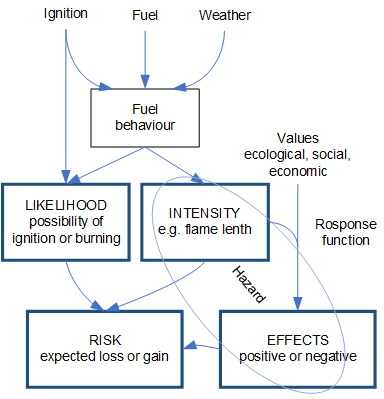}
    \caption{The generalized wildfire risk framework proposed by \cite{miller2012review}}
    \label{fig:risk definition}
\end{figure}

In the majority of published wildfire risk methodology studies, the concept of wildfire risk is not clearly defined, leading to instances of confusion regarding the interchangeable use of terms such as risk, hazard, and danger. Consequently, this paper initiates by presenting a definition of wildfire risk. \cite{miller2012review}, based on the Society for Risk Analysis's definition of risk, proposed a comprehensive fire risk framework (Fig. \ref{fig:risk definition}), which divides wildfire risk into three components: likelihood, intensity, and effects. This framework emphasizes the impact of wildfires on human activities and the environment rather than solely considering the probability of occurrence. For instance, low-intensity and low-impact wildfires might not significantly affect areas of concern, meaning that a high likelihood of wildfire occurrence does not necessarily equate to high wildfire risk. Notably, this framework does not explicitly incorporate the stochastic ignition caused by topographic and socioeconomic factors, which slightly diverges from the trends observed in the latest research.

Additionally, their study distinguishes between commonly used terms in wildfire risk research, such as risk, hazard, exposure, threats, vulnerability, and fire danger. When summarizing related work, we focus only on studies concerning wildfire risk and danger, excluding those on hazard, exposure, and threats. The former primarily describes the outlook for fire occurrence over a certain period, while the latter tends to evaluate the hazard of wildfires (Fig~\ref{fig:risk definition}). 

Consequently, if wildfire risk is defined solely based on the likelihood of fire occurrence without accounting for its potential harm, it can be simplified to the probability of wildfire ignition or spread at a given location and time \citep{deeming1972national}. This interpretation aligns with the approach taken by many past studies, where risk has often been described as the probability of wildfire occurrence. Moreover, this definition is consistent with the wildfire risk framework established by the European Commission \citep{depicker2020wildfire}.

More specifically, these studies often treat wildfire risk assessment as a conditional probability, where wildfire risk is defined as the probability of a wildfire igniting or spreading under the influence of factors such as fuel, terrain, and weather conditions. In this context, wildfire risk equates to wildfire probability \citep{yang2004hierarchical}. In addition, some definitions emphasize the role of causal agents and their frequency in influencing wildfire risk \citep{hardy2005wildland}. The importance of "causal agents" is highlighted—without these agents, the existing hazard would have zero potential to cause harm. Causal probability refers to the conditions that trigger wildfires, such as lightning or human activities \citep{deeming1972national}.

Based on these definitions, the specific form of wildfire risk varies across studies. This paper categorizes wildfire risk definitions into four types based on the algorithms used and the research objectives: explicit definitions, implicit definitions, wildfire ignition risk, and wildfire burning risk. On the one hand, definitions are categorized according to whether they explicitly represent various wildfire influencing factors and the contributions of historical wildfire records. These categories include explicit definitions based on historical observed wildfire occurrence data, influencing factors, and simulation data, as well as data-driven machine learning implicit definitions. On the other hand, wildfire prediction research is generally divided into studies on wildfire ignition risk and wildfire burning risk. Wildfire ignition risk is typically estimated using ignition data, while wildfire burning risk refers to the risk of a fire encountering a particular place. Although this risk partly depends on ignition, it is also influenced by subsequent wildfire spread and fire suppression efforts and is generally derived from historical burn data or simulation methods \citep{miller2012review}. 

In this survey, it is important to note that the aforementioned definitions of wildfire risk differ from the concept of wildfire detection. Wildfire risk refers to the probability of a wildfire occurring, whereas wildfire detection focuses on identifying already ignited wildfires. Nevertheless, accurate wildfire detection can provide valuable insights for wildfire risk prediction, particularly by serving as historical fire occurrence data for constructing risk assessment or prediction models. Additionally, near-real-time wildfire detection \citep{10553270, zhang2024assessing} can offer crucial guidance for predicting fire spread and ignition risk.

\subsection{Explicit and Implicit Definition Methods}

This subsection focuses on explicit definition methods, which include directly using historical observation data to define wildfire risk, defining wildfire risk based on the weights of influencing factors, and combining these factors to simulate wildfire distribution and define risk through simulation results. It is worth noting that this subsection does not cover the details of the simulation algorithms, but rather how to convert simulation results into wildfire risk.

Firstly, the simplest and most direct method in wildfire risk estimation is using historical fire occurrences to predict future risk. For example, \cite{rubi_performance_2023} classified regions where one or more fires occurred within an hour as "high risk," while other areas were classified as "low risk."  Similarly, \cite{umunnakwe2022data} estimated the spatiotemporal ignition probability using a similar function.

\begin{equation}
    m_{g,j} \approx  \frac{n_{g,j}}{N}
\end{equation}
where $m_{g,j}$ represents the fire ignition probability in grid cell $g$ during period $j$; $n_{g,j}$ and $N$ are the total number of wildfires and the total number of considered periods, respectively. \cite{lall2016application} assessed wildfire risk in Cape Town, South Africa, by defining daily risk ratings based on the average number of wildfires per day and their corresponding standard deviation over seven years. Specifically, fire occurrences below the average were rated as 1 or lower, those between the average and one standard deviation above were rated as 2 or moderate, those between one and two standard deviations above were rated as 3 or higher, and occurrences more than two standard deviations above the average were rated as 4 or extreme. An artificial neural network (ANN) model was then established using temperature, humidity, wind speed, and precipitation as inputs to predict wildfire risk.

% \cite{xu2023spatio} modeled wildfire risk using a marked spatiotemporal Hawkes process, with the conditional intensity function before location $k$ and time $t$.

% \begin{equation}\label{hawkes conditional intensity function}
%     \begin{aligned}
%         & \lambda(t,k,m|\mathcal{H}_{t})= \\ 
%         & \lim_{\Delta t,\Delta u\to 0} \frac{\mathbb{E}[\mathbb{N}(|t, t+\Delta t) \times B(k,\Delta k) \times B(m, \Delta m))|\mathcal{H_{t}}]}{\Delta \times B(k,\Delta k) \times B(m, \Delta m) }
%     \end{aligned}
% \end{equation}
% where $B(a,r)$ is a sphere with center $a$ and radius $r$, and $\mathcal{N}$ is the counting measure.

The above models based on historical data do not consider the conditions for wildfire ignition or spread, thus having poor robustness and generalization ability. Consequently, some studies have begun to incorporate or solely use factors such as fuel, terrain, weather conditions, and wildfire triggers as proxies to estimate ignition risk. Essentially, these methods can be defined as the contributions of various drivers to wildfire occurrence at specific times and locations. For example, \cite{castro_modeling_2003} used shrub moisture content, specifically live fuel moisture, as a proxy variable for wildfire risk in Spain. More complexly, the probability of wildfire occurrence can be defined as a conditional probability, representing the likelihood of a wildfire occurring given specific temporal, spatial, and environmental influencing factors. A typical example is the logistic model proposed by \cite{preisler2004probability}:

\begin{equation}\label{eq1}
          p_1 = \text{Prob}(N_{x,y,t} = 1 | U_{x,y,t}) = \frac{\exp(\theta_{x,y,t})}{1 + \exp(\theta_{x,y,t})}
\end{equation}
where \(N_{x, y, t}\) and \(U_{x,y, t}\) represent, respectively, whether a wildfire occurs and the collection of explanatory variables at location \((x, y)\) and time \(t\). \(\theta_{x, y, t}\) is a parameter estimated based on the explanatory variables, incorporating spatial and temporal dependencies. Considering the spatial, temporal, and nonlinear correlations of the explanatory variables, \(\theta_{x, y, t}\) can be expressed as a combination of linear and nonlinear terms:

\begin{equation}
\begin{aligned}
\theta_{x,y,t} = \text{logit}(p_{1k}) = & \ \text{fuel}_k + g_1(x_k, y_k) + g_2(\text{day}_k) + g_3(\text{elev}_k) \\
& + g_4(\text{BI}_k) + g_5(\text{FP}_k) + g_6(\text{KB}_k) + g_7(\text{TH}_k) \\
& + g_8(\text{Wspeed}_k) + g_9(\text{RH}_k) + g_{10}(\text{DBT}_k) \\
& + \text{Weather}
\end{aligned}
\end{equation}
here, \(g_i(\cdot)\) are nonparametric smoothing functions, typically estimated through generalized additive models, to capture nonlinear relationships. Variables such as \(\text{fuel}_k\) and \(\text{elev}_k\) represent explanatory factors, including fuel models, weather conditions (e.g., relative humidity, temperature), and fire danger indices (e.g., Burning Index (BI), Fire Potential Index (FP)), while \((x_k, y_k)\) are spatial coordinates, with \(g_1(x_k, y_k)\) as a bivariate smoothing function for spatial nonlinear effects. In this context, \(k\) denotes a unique spatiotemporal unit, corresponding to a specific location \((x, y)\) and time \(t\), representing an individual voxel (e.g., a 1 km\(^2\) grid cell) and a specific day where wildfire occurrence probability is modeled.

Several other models have been employed to quantify the relationships between wildfire drivers and wildfire risk as well. For example, \cite{sadasivuni_wildfire_2013} defined wildfire risk as the interaction between human activity and fuel distribution, integrating population and fuel layers through additive operations as inputs for a gravity model within a spatial interaction framework, with predictions classified into five risk levels. Similarly, \cite{gentilucci_analysis_2024} used a weight-of-evidence model to categorize wildfire drivers and classify wildfire risk into low and high levels based on regression of factor weights. Additionally, \cite{oliveira2021reassessing} and \cite{bergonse_predicting_2021} classified land use, land cover, slope, elevation, and aspect, calculated likelihood ratios for each category, and assessed fire susceptibility by summing weighted values across categories. This process can be mathematically expressed as follows:

\begin{equation}\label{eq2}
    \begin{cases}
    & Lr_{i}=\frac{S_{i}/S}{N_{i}/N}, \\
    & Lr_{j}=\sum_{i=0}^{n}X_{ij}Lr_{i}
    \end{cases}
\end{equation}
where $Lr_{i}$ represents the likelihood ratio of the variable corresponding to category $i$; $S_{i}$ and $S$ represent the number of burning pixels associated with category $i$ and the total number of burning pixels; $N_{i}$ and $N$ represent the number of pixels belonging to category $i$ and the total number of pixels in the study area; the total likelihood ratio $Lr_{j}$ of each grid cell $j$ is obtained by summing the $Lr_{i}$ values of all $n$ variables. Similarly, South Africa's official wildfire danger rating system, the Lowveld model \citep{meikle1987fire, lall2016application}, defines wildfire risk using the burning index, wind correction factor (WCF), and rain correction factor (RCF), where BI is a function of dry-bulb temperature (T) and relative humidity (RH):

\begin{equation}
    \label{Lowveld model south africa}
    \begin{cases} 
        & BI=((T-35)-(\frac{35-T}{30}))+((100-RH)\times 0.37)+30 \\ 
        & FDI=(BI+WCF)\times RCF 
    \end{cases}
\end{equation}
here, FDI refers to the fire danger index, while WCF and RCF are specified in Notice 1099 of 2013, issued in compliance with South Africa's National Veld and Forest Fire Act \citep{southafrica2013fire}. \cite{srivastava_geo-information_2014} differed from the aforementioned studies by simultaneously considering both causal and anti-causal factors, including the distribution of check dams, firelines, and fire water towers, using an expert knowledge system to quantify the likelihood of forest fires.

Simulation methods based on the above influencing factors are also commonly used in wildfire risk modeling. For instance, \cite{ager_comparison_2010} utilized the minimum travel time (MTT) fire spread algorithm, calculating the burn probability of each pixel through the burn frequency in 10,000 simulations. The simulation results defined wildfire risk as burn probability. Similarly, \cite{salis_application_2021} used MTT but calculated burn probability through a combination of different simulation results rather than frequency. \cite{ujjwal2022probability} used the Spark model and parameters such as temperature, relative humidity, wind speed, and wind direction to model wildfire risk in Tasmania, Australia, defining risk categories as low or high based on the spread area of wildfires at different locations. Another typical example of a simulation method is the Monte Carlo algorithm. For example, \cite{adhikari_developing_2021} defined wildfire risk through the overlap of multiple Monte Carlo algorithm predictions. Specifically, they predicted wildfire boundaries within 30 hours of ignition under various random weather and fuel moisture conditions using the Monte Carlo algorithm. The spatial overlap of all Monte Carlo predictions then generated risk levels.

On the other hand, wildfire ignition and spread are highly
random and exhibit a strong nonlinear relationship with various factors \citep{leuenberger2018wildfire}. Data-driven machine learning methods are frequently employed in wildfire risk prediction research due to their capability to effectively model nonlinear functions using large-scale datasets \citep{jain_review_2020}. While these methods are powerful for prediction, they do not explicitly quantify the contribution of individual factors to overall wildfire risk, unlike the explicit definition methods discussed earlier. As a result, they are classified as implicit definition methods. Further details on these methods will be provided in Section \ref{method}.

It is important to note that the wildfire risk defined by these machine learning models represents a pseudo-probability rather than a true probability density \citep{pelletier_wildfire_2023}. This pseudo-probability arises from the models' limitations in explicitly defining risk. For example, in XGBoost, the predicted pseudo-probability is derived from the inverse log transformation of the sum of terminal leaf weights. In SVM, it is estimated by calculating the distance to the classification decision boundary and using these distances for probability estimation. In Random Forest, it is the average of the voting outcomes from multiple decision trees. For models such as Multilayer Perceptron (MLP), Long Short-Term Memory (LSTM), and Convolutional Neural Network (CNN), it is typically the probability estimate from the activation function of the final layer.

In addition to the influence of method selection on predicted probabilities, these models are highly sensitive to the construction of the label sample set. Various preprocessing strategies—such as historical resampling of wildfire data—can substantially affect the resulting probability estimates. Therefore, it is essential to first establish a clear definition of wildfire risk rather than relying solely on binary labels. Furthermore, the application of post-processing techniques, such as linking explicit and implicit definitions of risk, can improve both the interpretability and accuracy of model outputs. This aspect will be further discussed in Section~\ref{calibration}.

\subsection{Wildfire Ignition and Burning Risk}
\label{Wildfire Ignition and Burning Risk}

Wildfire ignition refers to the first occurrence of a fire event during a wildfire. According to \cite{badia2006spatial}, to combat forest fires most effectively, the importance of assessing wildfire ignition risk should be emphasized. Similarly, \cite{catry2009modeling} highlighted that it is not sufficient to consider only factors influencing wildfire spread and suppression difficulty, such as fuel, weather, and terrain. They cited the definitions provided by the Food and Agriculture Organization and the Glossary of Wildland Fire Terminology, which define wildfire risk as fire ignition risk, meaning "the chance of a fire starting as determined by the presence and activity of any causative agent." Furthermore, estimating wildfire ignition risk is crucial for wildfire management, as it guides decisions related to fuel treatment and the allocation of suppression resources. Generally, wildfire ignition risk can be modeled using historical ignition data \citep{miller2012review} and used to generate broader or spatially continuous ignition risk maps. Therefore, the following section will focus on how various studies have identified or defined ignition points and converted these records into probability values or modeled dependent variables.

In the study of wildfire ignition risk in Central Spain, \cite{romero2008gis} utilized wildfire records provided by the Sección de Defensa Contra Incendios Forestales of the Madrid regional government. These records included the coordinates of ignition points, land cover of the burned area, and the time of the fire. The researchers then defined the probability of wildfire ignition during the fire season as a prior probability and calculated the posterior probability based on this prior probability. Finally, they generated predicted wildfire risk levels by comparing the ratio of prior to posterior probabilities. 

In the study by \cite{catry2009modeling}, they first cleaned and corrected a dataset of 137,204 ignition points selected from the official wildfire dataset provided by Portuguese Forest Services, ultimately retaining 127,409 fire ignition locations. To construct a balanced dataset, they randomly selected a number of non-ignition points across the country, 1.5 times the number of fire ignition locations. The ignition and non-ignition points were then encoded as binary variables, forming the final dependent variable dataset for fire ignition. The researchers used frequency analysis and the natural breaks method to convert the historical fire locations dataset into observed and expected frequencies, which were used to train a logistic regression model. In this model, the expected frequency of fire ignition was calculated proportionally to the land area of each observed frequency interval. The output of the logistic regression model was classified into six risk levels. Similarly, \cite{ager2014wildfire}, in their study of wildfire ignition and burning risk in Sardinia, Italy, and Corsica, France, defined the dependent variable of their regression model as the frequency of wildfire ignition on a daily basis, using wildfire ignition area as a condition for burning risk. It is noteworthy that ignition points in Corsica were assumed to be the centroids of burned areas. \cite{braun2010forest} used forest inventory data from Ontario's Ministry of Natural Resources to identify wildfire ignition areas and employed generalized additive models, coordinate offsets, and cubic spline interpolation to generate spatially continuous wildfire ignition probability maps, i.e., risk maps.

In the study of wildfires in Belgium, \cite{depicker2020wildfire} defined the probability of wildfire occurrence in a given area as the average probability of ignition within a calendar year:

\begin{equation}
    P(I|C_{i})=\frac{P(I)P(C_{i}|I)}{P(C_{i})}
\end{equation}
where $P(C_{i})$ and $P(C_{i}|I)$ represent the ratio of the area of ignition unit $C_{i}$ to the total study area and the ratio of the number of ignitions to the total study area, respectively, while $P(I|C_i)$ denotes the ratio of the number of ignitions to the number of units over a one-year span. To mitigate the issue of class imbalance, smaller areas were merged into larger ones:

\begin{equation}
    P_{A}=1- {\textstyle \prod_{i=1}^{n}} (1-P(I|C_{i}))^{N_{i}}
\end{equation}
where $n$ and $N_{i}$ represent the number of environments within a larger area $A$ and the number of grid cells within environment $i$, respectively. In the study of wildfire ignition risk in Galicia, Spain, \cite{calvino2017interacting} analyzed the differences between historical ignition points obtained from the Spanish Forest Fire Statistics and an equal number of randomly selected locations, examining the relationships between wildfire occurrences and factors such as the wildland-urban interface (WUI), land use/land cover (LULC), and elevation. Similarly, \cite{ying2021relative}, in their analysis of wildfire ignition causes in Yunnan Province, China, used historical records from the Yunnan Forestry Bureau for the period 2003-2015. These records, like those in previous studies, included information on the date, location, size, and cause of each ignition. In total, 5,145 ignition events were recorded, and approximately 1.5 times as many non-ignition points were randomly selected. The relationship between wildfire risk and influencing factors was modeled based on the presence or absence of ignition, rather than frequency.

\begin{figure*}[!htbp]
    \centering
    \includegraphics[width=1\linewidth]{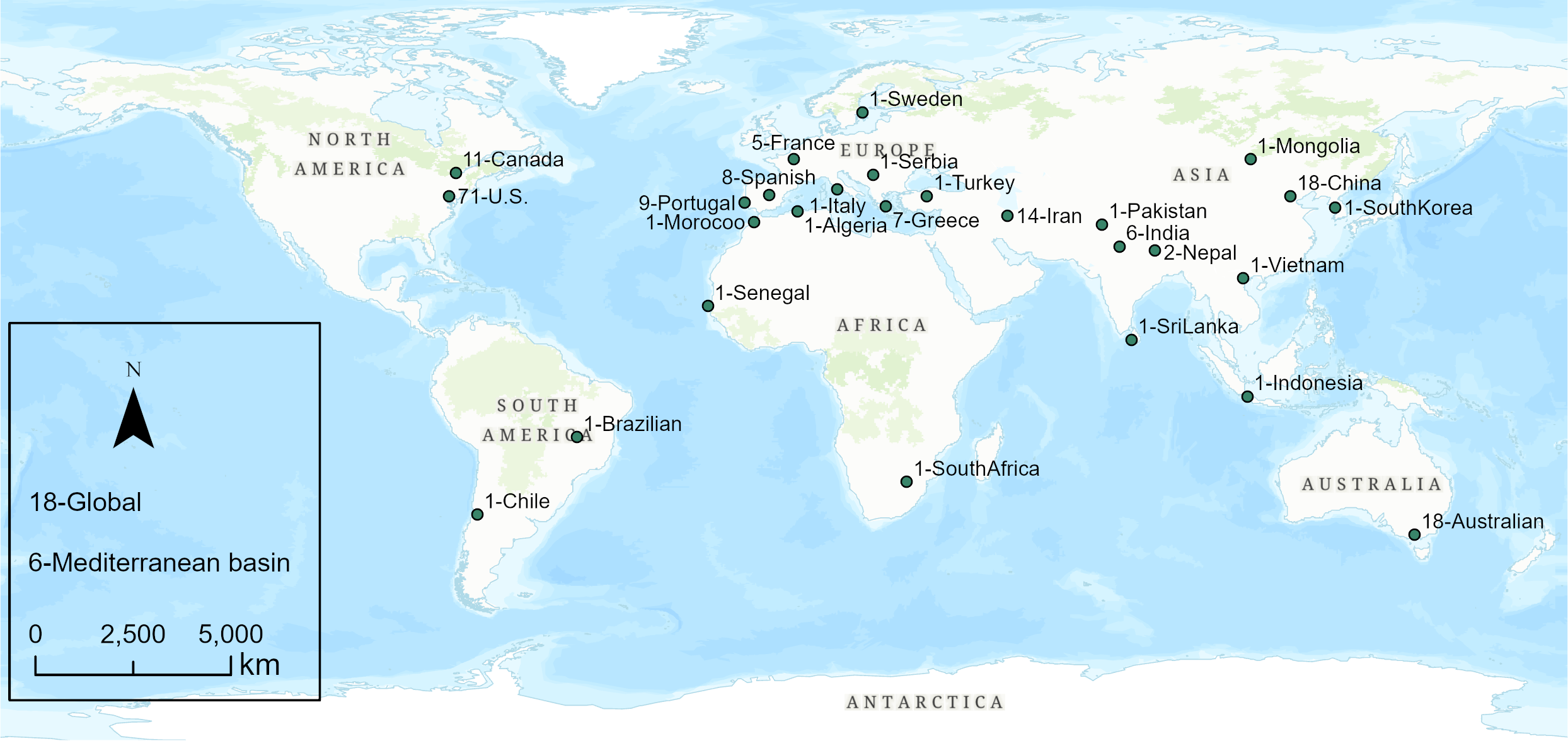}
    \caption{The spatial distribution of included research.}
    \label{fig:reseach_area_distribution}
\end{figure*}

It is important to note that the studies mentioned above utilized ground-based wildfire databases rather than the more commonly used MODIS or VIIRS fire/hotspot data in wildfire risk prediction research. This may be due to the relatively low spatiotemporal resolution of these data, making it difficult to detect ignition points with very low intensity and duration. \cite{benali2016determining, barber2024canadian} compared the consistency of wildfire field records in Portugal, Greece, Alaska, California, and southeastern Australia with MODIS active fire products (MCD14ML). They proposed a clustering algorithm to calculate the ignition and end dates of a wildfire in the MCD14ML product and used the Nash–Sutcliffe Model Efficiency Index to estimate the consistency between satellite-estimated fire dates and ground records. The study found that satellite data could not estimate the start and end dates of most fires smaller than 500 ha. Similarly, \cite{coskuner2022assessing} compared the performance of MODIS and S-NPP VIIRS active fire or hotspot detection products—specifically, MCD14ML V6 and VNP14IMG—in detecting wildfires across five different land cover types in Turkey. These detections were compared with records from Turkey’s ground-based wildfire database. The authors found that both MODIS and VIIRS had very low detection capabilities for small fires (<1 ha) (<3.5\%). They also cited studies by \cite{fusco2019detection} and \cite{ying2019wildfire} showing that MODIS fire products had high omission rates in the United States and China. Specifically, \cite{fusco2019detection} reported detection rates for MODIS products in the United States ranging from 3.5\% to 23.4\%, with detection rates increasing with fire size. Similarly, \cite{ying2019wildfire} found that in Yunnan Province, China, only 11.10\% of 5,145 ground-recorded wildfire events were detected by MODIS fire products, with most omissions attributed to the small size of the fires. These analyses suggest that the commonly used MODIS and VIIRS fire and hotspot products are not suitable for estimating ignition locations and events, nor for corresponding ignition risk estimation.

In contrast to wildfire ignition risk, wildfire burning risk is more broadly defined as the risk of a fire encountering a particular location, influenced by both ignition and spread. Therefore, most studies do not distinguish between ignition and spread when defining wildfire burning risk but consider both simultaneously. In general, there are two methods for defining wildfire burning risk: using historical burn data through explicit or implicit methods, or defining it based on simulation or spread models \citep{finney1998farsite, finney2002fire, scheller2007design, miller2008evaluating, tolhurst2008phoenix, tymstra2010development, chen2024incorporating}. These methods have been discussed in the previous sections of this chapter, so they will not be repeated here. It is important to note that in choosing between these two methods, \cite{miller2012review} suggest that for smaller fire extents (e.g., 1-10 hectares), ignition risk assessment can accurately reflect the likelihood of fire occurrence. However, for larger fire extents (e.g., 20-50 $km^2$), it is necessary to consider using burning risk to represent wildfire risk.

\section{Research Area distribution}

This paper reviews 138 recent wildfire risk prediction studies conducted globally using data-driven methods. Figure \ref{fig:reseach_area_distribution} illustrates the geographic distribution and quantity of these studies. Among these, 18 studies address global-scale wildfire risks, while 6 focus specifically on the Mediterranean basin.

At the continental scale, North America and Asia exhibit the highest levels of research activity. In North America, the United States accounts for 71 studies, followed by Canada with 11. In Asia, China leads with 18 studies, followed by Iran and India with 14 and 6 studies, respectively. In Europe, wildfire research is concentrated in the western and southern regions, with notable contributions from Spain, Portugal, Greece, and France, while smaller-scale studies have been conducted in Italy, Sweden, Serbia, and Turkey. In contrast, Africa has seen limited wildfire research, with only one study each from South Africa, Algeria, Morocco, and Senegal. Similarly, South America contributes minimally, with one study each from Brazil and Chile. In Oceania, research is dominated by Australia, which accounts for 18 studies, while no other countries in the region are represented.

The geographic concentration of wildfire research raises critical questions about whether these efforts align with global wildfire trends, particularly in terms of frequency, burned area, and carbon emissions. Africa, often referred to as the "continent of fire," has accounted for 70\% of the global burned area over the past two decades, contributing approximately 29\% of global carbon emissions \citep{giglio2013analysis, giglio2018collection, de2024impact}. Despite its prominence, Africa remains underrepresented in wildfire risk prediction studies, as shown in Figure ~\ref{fig: features frequency}. Similarly, South America, which contains approximately 864 million hectares of forest—accounting for 21\% of the global forest area and 57\% of the world's primary forests \citep{fao2011state, keenan2015dynamics}—has experienced a rising trend in wildfire frequency and burned area. This increase is primarily driven by deforestation and climate change. During the 2023–2024 wildfire season, the Amazon region, encompassing Brazil, Bolivia, Peru, and Colombia, experienced historically high burned areas, particularly in Amazonas, Loreto, La Paz, and Beni. Fire activity in June and October 2023 was the highest recorded since 1998. Southern South America, including Chile, also recorded significant wildfire activity during the same season. The burned area exceeded 4,000 square kilometers, marking the second largest in 20 years \citep{jones202data}. Despite these critical developments, data-driven wildfire research in South America remains sparse, as shown in Figure ~\ref{fig: features frequency}.

In northern boreal regions, Canada has been a focal point of wildfire research. However, studies on wildfire risk in Russia remain scarce, despite estimates suggesting that Siberian wildfire emissions surpass those of Canada \citep{de2013comparison}. Between 1998 and 2010, Russia's average annual burned area and carbon emissions were 8.23 million hectares and 121 teragrams of carbon per year, respectively \citep{shvidenko2011impact}. Recent studies indicate a growing prevalence of high-intensity, late-season fires in Siberian boreal forests, including the Arctic, with projections suggesting annual fire emissions may reach 325 teragrams of carbon by 2027 \citep{ponomarev2023wildfire}. These findings underscore the urgent need to advance wildfire risk prediction in Russia’s northern regions.

In conclusion, the geographic distribution of wildfire research highlights significant disparities. While North America and Asia dominate wildfire studies, critical regions such as Africa, South America, and parts of Russia remain underrepresented. Addressing these gaps is vital to reducing geographic inequalities and advancing climate justice \citep{jones2024state}. Expanding research to underrepresented regions, fostering cross-regional studies, and strengthening international collaboration are crucial to understanding wildfire behavior across diverse ecosystems and climates. Additionally, the reliance on localized datasets limits the generalizability of existing predictive models. Future efforts should prioritize the development of unified data platforms, facilitate open data sharing, and rigorously validate models across diverse regions to enhance their reliability and applicability.

\section{Data}

Essentially, wildfires are uncontrolled processes involving the continuous combustion of organic matter in landscapes \citep{duff2017revisiting}. The ignition and burning processes of wildfires are influenced by factors such as fuel conditions, meteorological and climatic conditions, human interventions, topographical and hydrological characteristics, and historical wildfire occurrences \citep{fuller1991forest}. Therefore, in this section, the data used in wildfire risk prediction are categorized into these types, and their components, definitions, and proxy methods are introduced. 

It is important to note that most existing wildfire risk prediction studies rely on proxies, which can only indirectly reflect the state of driving factors. Therefore, in Section ~\ref{appendix}, we provide an overview of recent advancements in the calculation methods for these driving factors, with a particular focus on fuel conditions.

\subsection{Fuel Conditions}
\label{fuel}

\cite{duff2017revisiting} suggests that wildfire behavior is largely dependent on the characteristics of the fuel that sustains it. Unlike other factors influencing wildfire ignition and spread, such as meteorological conditions, climate, topography, and hydrology, fuel can be more easily controlled \citep{chuvieco2012integrating}. Some studies even propose that fuel is the only landscape element that can be managed to directly affect fire behavior \citep{thompson2013quantifying, fernandes2003review}, emphasizing the importance of fuel conditions in wildfire risk evaluations. Fuel refers to any combustible material, including live and dead vegetation \citep{mclauchlan2020fire, leite2024leveraging}, with four main variables affecting wildfire: Fuel Moisture Content (FMC), fuel load, fuel type, and fuel continuity. These parameters directly influence fire ignition, intensity, and spread rate.

\subsubsection{Fuel Moisture Content}

FMC refers to the ratio of the weight of water to the dry weight of a sample, as shown in Eq. \ref{fuel moisture content} \citep{mbow_spectral_2004, chuvieco_short-term_1999}. FMC can be divided into Live Fuel Moisture Content (LFMC) and Dead Fuel Moisture Content (DFMC), both critical for determining fuel combustibility \citep{yebra_global_2013}.

\begin{equation} 
  \label{fuel moisture content} 
  FMC=\frac{\text{Weight of water in sample}}{\text{Dry weight of the sample}}\times 100\% 
\end{equation}

LFMC is influenced by plant ecophysiological factors and environmental conditions such as soil moisture and humidity, which regulate water uptake and transpiration \citep{castro_modeling_2003}. DFMC, on the other hand, is determined by fuel size, weather, and microclimate \citep{nolan2016predicting}, as well as the fuel's biochemical composition \citep{viney1991review}. Dead fuel, unlike live fuel, cannot regulate moisture actively and reaches an equilibrium with the surrounding atmosphere known as Equilibrium Moisture Content (EMC). The time needed to reach EMC depends on the fuel size, which is classified into fine fuels like twigs and leaves (1-hour fuels), coarser branches (10-hour and 100-hour fuels), large trunks (1000-hour fuels), and surface litter \citep{catchpole2001estimating}. Soil moisture is also considered in some cases \citep{rothermel1986modeling}. DFMC plays a crucial role in fire ignition and rapid spread early in the fire season, while LFMC significantly impacts the persistence and spread of fires.

It is important to note that estimating FMC remains a significant challenge due to the complexity of real-world conditions. For example, accurately isolating the effects of vegetation dry matter, soil moisture, and atmospheric and canopy structural interference on backscatter, optical reflectance, and emissivity remains difficult. Additionally, the availability of in situ FMC measurements is still limited, with sparse spatial distribution, which constrains the development of data-driven algorithms. 

To address these challenges, practical wildfire risk prediction systems often employ proxy parameters to represent DFMC and LFMC. For DFMC, these proxies include drought indices such as the Keetch-Byram Drought Index, meteorological variables like temperature and relative humidity, and composite indices such as the Fire Weather Index (FWI). These alternatives are easier to obtain and integrate into predictive models, while still capturing the indirect effects of DFMC on fire behavior \citep{vitolo_era5-based_2020, rothermel1986modeling}.

For LFMC, proxy parameters such as vegetation indices (e.g., NDVI and its variants), meteorological variables (e.g., fuel moisture code, meteorological drought indices, and temperature), and composite indices (e.g., the Fire Weather Index) are commonly used, as shown in Figure \ref{fig: features frequency}. These proxies provide easier-to-obtain and more consistent inputs while indirectly capturing LFMC's influence on wildfire behavior, thus offering a practical alternative for large-scale, long-term modeling efforts \citep{chuvieco2002estimation, dasgupta2007evaluating, ruffault2018well, VINODKUMAR2021108503, krueger_using_2022, balaguer2022semi, 10638113}.

\subsubsection{Fuel Load}

Fuel load refers to the amount of combustible biomass available per unit area which influences wildfire intensity and spread \citep{mcnorton_global_2024}. It can be classified as live or dead, or by size into fine fuels (e.g., grass, shrubs, twigs) and coarse fuels (e.g., trunks, logs) \citep{gould2011quantifying}. Dead fine fuel plays a critical role in fire ignition and spread, while canopy fuel load fuels crown fires, increasing both fire intensity and spread rate \citep{cruz2003assessing, arellano2018potential, walker2020fuel}. Fuel load measurements are typically obtained through direct sampling, physical models, or remote sensing techniques. The integration of SAR, LiDAR, and multispectral data has significantly enhanced the accuracy of fuel load estimation.

In practical wildfire risk prediction, directly estimating fuel load is often challenging due to its inherent complexity and data limitations, which may lead to inaccurate estimations. Consequently, the use of proxies rather than direct fuel load estimation is more common in wildfire risk prediction, as it simplifies the process and avoids the accumulation of errors.

First, due to its definitional similarity to fuel load, particularly its emphasis on above-ground biomass, some studies have employed above-ground biomass as a proxy for fuel load \citep{eames2021instantaneous, vega2022modelling, mcnorton_global_2024}. Additionally, other studies have utilized alternative proxies for fuel load. For instance, \cite{su_modeling_2021}, in their study on wildfire driving factors in certain tropical rainforest regions of southern China, used fractional vegetation cover (FVC) as a proxy for fuel load. FVC represents the percentage of an area covered by green vegetation relative to the total statistical area. In practical applications, MODIS NDVI products are often used as proxies for FVC. Furthermore, \cite{helman_phenology-based_2015} applied time-series analysis to MODIS NDVI data to distinguish between woody and herbaceous vegetation in Mediterranean forest regions. Their findings suggested that the long-term mean woody NDVI and its trend could serve as alternative indicators for wildfire fuel assessment.

\subsubsection{Fuel Types}

Fuel types categorize fuels based on shared characteristics, such as vegetation type, canopy height, fuel load, and biomass, which influence wildfire behavior under specific conditions \citep{arroyo2008fire}. Classification systems for fuel types vary globally. For instance, the U.S. National Fire Danger Rating System (NFDRS), Northern Forest Fire Laboratory (NFFL), and Canada's Fire Behaviour Prediction (FBP) system are widely used for wildfire risk prediction and management \citep{cohen1985national, albini1976estimating, canada1992development}.

Table~\ref{fuel_classification_system} presents a comparison of major fuel classification systems. The NFDRS includes 20 fuel types, NFFL defines 13 types, and the U.S. Fuel Characteristics Classification System (FCCS) has 216 fuel types. In Australia, McArthur’s system categorizes fuel into three types, while the FBP system in Canada uses 16 fuel types. The Prometheus system, applied in Mediterranean Europe, defines seven types \citep{arroyo2008fire}.

\begin{table}[h!]
\centering
\caption{Comparison of Fuel Classification Systems \citep{arroyo2008fire}}
\label{fuel_classification_system}
\begin{threeparttable}
\resizebox{\linewidth}{!}{
\begin{tabular}{l c c}
\toprule

\textbf{Fuel classification system} & 
\textbf{Number of fuel types} & 
\textbf{Country of application} \\ 

\midrule

NFDRS fuel types & 
20 &
USA \\ 

NFFL fuel types & 
13\tnote{a} & 
USA \\ 

FCCS & 
216\tnote{b} & 
USA \\ 

McArthur fuel types &
3 &
Australia \\ 

FBP fuel types & 
16 & 
Canada \\ 

Prometheus fuel types &
7 &
Europe\tnote{c} \\

\bottomrule
\end{tabular}
}
\begin{tablenotes}
\footnotesize
\item[a] Option of developing custom fuel types.
\item[b] New fuelbeds added periodically.
\item[c] Mediterranean countries.
\end{tablenotes}
\end{threeparttable}
\end{table}

In addition to the detailed classification of fuel types, most wildfire risk assessment studies based on multi-source remote sensing data utilize simplified LULC maps as proxies for fuel type to analyze the influence of different land cover types and human activities on wildfire risk. For instance, \cite{jaafari_wildfire_2019} employed logistic regression to assess the influence of various explanatory variables—including topography, elevation, Topographic Wetness Index (TWI), meteorological factors, and land use—on wildfire risk in the Zagros ecological zone of western Iran. The study identified land use as the most significant factor, with dry farmland and forests being the areas most susceptible to wildfires. Moreover, the type and continuity of vegetation were found to be critical determinants of large fire occurrences.

Similarly, \cite{ager2014wildfire} analyzed wildfire occurrence and scale in Sardinia (Italy) and Corsica (France). Their research revealed that, at the beginning of the the fire season, the probability of large fires occurring in agricultural lands in southern Sardinia was ten times higher than in forested and shrubland areas near roads. However, as the fire season progressed, the latter exhibited a higher wildfire probability. Conversely, \cite{calvino2016wildfire} investigated the relationship between WUI and LULC in northern Spain and found that primary forests and agricultural areas exhibited the lowest wildfire risk.

Likewise, \cite{donovan2020land} considered five land use types—grasslands, woody vegetation, croplands, rangelands and hayfields, and developed areas—in an assessment of wildfire occurrence risk in the Great Plains ecoregion of the United States. Their findings indicated that woody vegetation and grasslands were most prone to both wildfire ignition and large fires, whereas other land use types exhibited lower susceptibility. Among them, croplands had the lowest probability of large fires, and as crop coverage increased, the likelihood of large fire occurrences decreased.

\subsubsection{Fuel Continuity}

Since fire lines spread both horizontally and vertically, fuel continuity is generally considered in terms of both vertical and horizontal directions. Vertical canopy continuity primarily refers to the vertical structure of fuels, also known as ladder fuels. These are composed of low-lying shrubs, seedlings, upper canopy trees, and intermediate understory vegetation. In areas where vertical fuels are present, this structure can cause low-intensity surface fires to escalate into severe crown fires \citep{agee2005basic, menning2007fire}. There has been significant research on ladder fuels, including methods for precise field measurements of fuel structure \citep{kilgore1975crown, pye2003prescribed} and approaches combining field measurements with expert knowledge \citep{menning2007fire}. However, due to the complexity of field measurement, the more common proxy for vertical fuel continuity is the estimation of canopy base height (CBH) \citep{scott2001assessing}.

CBH is defined as the height above which there is sufficient canopy fuel per unit volume to support vertical fire spread into the canopy \citep{scott2001assessing}. It has widespread application in various wildfire behavior models \citep{hall2006considerations}. CBH can be estimated using allometric equations based on species, diameter at breast height (DBH), and tree height, or it can be retrieved from remote sensing data. However, due to the limitations of passive optical remote sensing signals and synthetic aperture radar (SAR) in penetrating the canopy, studies using multi-angle observations or bidirectional reflectance distribution function (BRDF) retrievals to obtain three-dimensional vegetation structure information are rare in wildfire prediction \citep{pisek_estimation_2015, canisius_retrieving_2007}. More commonly, LiDAR is used to estimate CBH at the individual tree or local scale \citep{andersen2005estimating, luo2018simple, 9765478}. Some studies suggest that while the definition of CBH is clear, its practical application is complicated by the lack of a clear fuel threshold per unit volume and the potential neglect of fine fuels, which can lead to errors in modeling canopy fire risk \citep{cruz2004modeling, kramer2014quantifying}. As a result, a quick and simple method to estimate ladder fuels is by estimating the coverage area or coverage rate of low-lying fuels as a representation of ladder fuel \citep{clark2009decision, wing2012prediction}.

In addition to vertical continuity, horizontal fuel continuity is also crucial. There are various definitions of horizontal continuity. For example, \cite{bhlitem136922} defines horizontal continuity as the degree of variation in the physical characteristics of fuels within a given area, with an emphasis on the impact of fuel type. Conversely, a more commonly used definition in describing horizontal continuity focuses on the spacing between fuels \citep{ritter2023vertical}. Tree spacing significantly affects the intensity and scale of wildfires \citep{kim2016exploring}. For instance, \cite{qadir_predicting_2021} showed in a study on wildfire risk prediction in Nepal that areas with high forest cover density have a higher risk of fires. In contrast, when there is canopy discontinuity, fire spread may slow down or even be extinguished \citep{taneja_effect_2021}. For example, \cite{kim2016exploring} used the Fire Dynamics Simulator (WFDS) to model fire behavior in Korean pine (Pinus densiflora) and found that when the spacing between trees is 6 meters or more, most crown fires stop spreading within 100 meters. Therefore, fuel management techniques that fragment fuels can be used to suppress wildfires \citep{srivastava_geo-information_2014, harrison2021understanding}. 

Despite the demonstrated importance of horizontal fuel continuity, our review indicates that it has received comparatively less attention than vertical continuity. This discrepancy is likely due to the frequent use of proxies rather than direct measurements. For example, many crown fire spread models rely on canopy bulk density to approximate horizontal fuel continuity \citep{wagner1993prediction, cruz2005development}. Additional proxies, such as fractional vegetation cover, canopy cover, vegetation cover density, and various vegetation indices, are also commonly employed.

In summary, fuel conditions play a crucial role in wildfire risk modeling. However, directly using fuel condition descriptors derived from remote sensing or multi-source geospatial data in large-scale and long-term wildfire risk prediction models is uncommon. Instead, as illustrated in Figure~\ref{fig: features frequency}, most models rely on raw multispectral or microwave data and meteorological parameters, or indices derived from these sources, which have been shown to correlate with fuel conditions. This practice is driven by several considerations:

First, accurately assessing fuel conditions is inherently complex and often requires integrating data from multiple sensors. Models designed for this purpose frequently exhibit poor generalization, making them applicable only to specific regions or temporal scales. Additionally, the availability of certain sensors poses a limitation. For instance, estimating vertical canopy structure relies heavily on LiDAR data. However, satellite-based LiDAR canopy measurements are constrained by spatial and temporal resolution, while airborne LiDAR is limited by coverage and high costs.

Second, most wildfire risk assessment studies necessitate long-term time-series data, which are typically sourced from low-cost, widely available datasets such as multispectral or microwave satellite data. Due to constraints in spatial resolution and canopy penetration, these data often fail to accurately retrieve detailed fuel condition descriptors. Consequently, inaccuracies in inversion processes may lead to cumulative errors, further impacting prediction accuracy.

Finally, the growing popularity of data-driven wildfire risk prediction algorithms reduces the reliance on explicitly defined fuel condition descriptors. These algorithms, capable of bypassing manual feature engineering, directly learn the necessary abstractions from raw data or features. This adaptability minimizes the need for explicit fuel condition parameters while maintaining high predictive performance.

\subsection{Weather and Climate Conditions}

Weather and climate factors are major drivers of wildfire activity, influencing the occurrence, burned area, and fire behavior \citep{swetnam1990fire, bessie1995relative, abatzoglou2013relationships}. Weather conditions are central to almost all global fire danger risk systems. High temperatures, low relative humidity, insufficient precipitation, and strong winds have been proven to be key factors in the ignition of fires in the short term \citep{keane_mapping_2001, zacharakis_integrated_2023}, and they also affect fire size and spread \citep{abatzoglou2011relative}. On the other hand, climate conditions generally refer to the long-term weather conditions of a region and their variations. Some studies specifically define these as the collective atmospheric conditions months or even years before the start of the fire season, emphasizing the use of prior climate conditions to convert fuel conditions into wildfire potential \citep{abatzoglou2013relationships}. This approach enables predictions of large-scale fire risks over seasonal, annual, or even multi-decadal time scales.

\subsubsection{Weather Conditions}

Short-term weather conditions focus primarily on predicting the impact of weather conditions over quarters, months, weeks, or days on fire risk. These include short-term temperature, wind, relative humidity, soil moisture, precipitation, and lightning, as well as corresponding indices like the Palmer Drought Severity Index (PDSI) and the FWI. Temperature and various humidity conditions affect fuel conditions, such as evapotranspiration and moisture content, which in turn influence the wildfire risk. Wind direction, wind speed, and other related metrics like Mean Wind Power Density (MWPD) and Mean Wind Speed (MWS) affect fire spread and intensity, making them commonly used parameters in wildfire prediction. Unlike other parameters, lightning can act as an ignition factor for wildfires.

Short-term weather parameters are widely used in wildfire risk prediction. For example, \cite{tavakkoli_piralilou_google_2022} used temperature, precipitation, MWPD, and MWS as parameters in their wildfire prediction model for the forested areas of Gilan Province, Iran. In another study conducted in Pakistan, \cite{kanwal_data-driven_2023} utilized 16 meteorological parameters, including monthly average precipitation, evapotranspiration, wind speed, soil temperature, humidity, heat flux, albedo, average land surface temperature, soil moisture, actual evapotranspiration, moisture deficit, downward surface shortwave radiation, precipitation accumulation, minimum/maximum temperature, and vapor pressure. An assessment of their importance indicated that heat flux, evapotranspiration, and vapor pressure were relatively more important compared to other meteorological parameters, although the differences in importance were not significant. \cite{bergado_predicting_2021} used seven meteorological factors in their wildfire prediction study for Victoria, Australia: daily maximum temperature, daily minimum relative humidity, daily total solar radiation, daily total precipitation, daily average wind speed, daily average wind direction, and lightning frequency. The lightning data was derived from an annual lightning climatology dataset that provided estimates of lightning density for each day of the year. Their study found that total precipitation, lightning density, and surface temperature consistently held high weight in all models. In contrast, \cite{quan_improving_2023} used various meteorological variables from the ERA5 dataset, including precipitation, relative humidity, temperature, wind speed, and the Keetch-Byram Drought Index, to predict wildfire risk in parts of the western Tibetan Plateau in China. The study revealed that relative humidity was the most critical factor in the occurrence of forest and grassland fires in the region. Higher relative humidity leads to higher DFMC, thereby reducing the likelihood of ignition.

\subsubsection{Climate Conditions}

Climate conditions primarily refer to annual or multi-year meteorological parameters such as temperature, humidity, precipitation, and evaporation. Unlike short-term meteorological conditions that affect fuel status temporarily, climate conditions play a crucial role in determining fuel types, distribution, life cycles, and other factors that influence fuel characteristics, such as fuel load, type, and continuity \citep{aldersley_global_2011, jaafari_bayesian_2017, tavakkoli_piralilou_google_2022}. These fuel characteristics respond to climatic patterns accumulated before the start of the fire season \citep{chen_how_2016}. Consequently, climate variables are often the primary drivers of long-term wildfire activity \citep{pollina_climatology_2013, jain_review_2020}. Furthermore, climate conditions explain strong seasonal variations in wildfire risk in certain regions \citep{martell_modelling_1989}. Intensified climate change in recent years has raised concerns about its long-term impacts on large-scale wildfires \citep{prior_wildfire_2013, jolly2015climate, jain2024drivers}.

Climate conditions are widely used in wildfire risk prediction. Specifically, \cite{chen_how_2016} used large-scale sea surface temperature anomalies to predict the severity of the fire season in South America on a seasonal time scale. Their study found that using a single ocean climate index could predict about 48\% of the burned area globally for up to three months or longer before the peak burning month. In a study by \cite{jaafari_wildfire_2019} in the Zagros ecoregion of western Iran, they used three climate conditions: annual temperature, rainfall, and wind effects. However, the results showed that only the annual temperature parameter was relatively important. Similarly, \cite{naderpour_wildfire_2020} studied the northern beaches of New South Wales, Australia, using three climate factors: annual rainfall, annual humidity, and annual wind speed. \cite{holdrege_wildfire_2024} studied the Great Basin sagebrush regions across 13 western U.S. states and used three climate factors: annual mean temperature, annual precipitation, and the proportion of summer (June to August) precipitation to total annual precipitation. The study found that areas with low summer precipitation proportion, medium to high annual precipitation, and high temperatures had the highest wildfire probability. Similarly, \cite{aldersley_global_2011} used annual cumulative precipitation, annual lightning frequency, monthly precipitation frequency, monthly average temperature, and climate data from the British Atmospheric Data Centre (BADC) to study the drivers of monthly fire area on a global and regional scale. They found that climate factors such as high temperatures, moderate precipitation, and drought were the most important drivers globally. \cite{mansuy_contrasting_2019} conducted a study across the U.S. and Canada. They found that climate variables were the primary control factors controlling burned area in most ecoregions, both inside and outside protected areas, surpassing landscape and anthropogenic factors. The climate data used in their study included annual average precipitation, annual average relative humidity, the Hargreaves climate moisture deficit (an annual measure of energy and moisture), and degree days below 18°C. \cite{abdollahi_explainable_2023} conducted a wildfire prediction study in the Gippsland region of Victoria, Australia, and found that humidity, wind speed, and rainfall were the most influential factors in wildfire prediction. \cite{zhao_important_2021} used data from over 600 meteorological stations in forest, grassland, and shrubland areas to calculate two drought indices, the Keetch-Byram Drought Index and the Standardized Precipitation Index. They then used global-scale PDSI data to mask these indices and assess long-term wildfire risk in China. The results showed that drought indices had a stronger correlation with seasonal and annual fire occurrences compared to climate indicators such as temperature and precipitation.

From the analysis above, it is evident that various forms of temperature, wind, humidity, and evaporation data are commonly used as meteorological or climate variables in wildfire prediction models. A few studies have combined both meteorological and climate factors to analyze their roles in the occurrence, spread, and scale of wildfires. For example, \cite{abatzoglou2011relative} found in their study on the influence of weather and climate factors on the occurrence and scale of individual wildfires in Alaska that prior climate conditions had no significant effect on the final fire size, whereas post-ignition weather conditions, such as precipitation, had a greater impact over days to weeks. Thus, the authors suggest using short-term weather factors in wildfire risk assessment models and spread models. From another perspective, weather variables are more important than climate variables for small-scale and short-term wildfire risk and spread assessments.

\subsection{Socio-Economic Factors}

In the early development stages of wildfire risk assessment, considerable attention was given to factors influencing wildfire spread and suppression difficulty \citep{mcarthur1958the, mcarthur1966weather, hollis2024introduction}. However, numerous studies have shown that the proportion of fires caused directly or indirectly by human activities surpasses those triggered by natural factors, such as lightning \citep{FAO_human_2006}. Indirect human activities typically refer to the impact of human actions on climate change, which in turn influences wildfire risk. This effect can be modeled by analyzing the influence of climate on wildfire risk. Furthermore, the impact of indirect human activities on wildfire risk also includes changes in vegetation types, land use patterns, and landscape fragmentation, as well as alterations in fuel load and fuel continuity due to activities like grazing, planting, and harvesting \citep{harris2023humans}. These changes can significantly affect local wildfire risk. Direct human activities, on the other hand, primarily involve ignition and suppression. 

The risk of wildfires caused by direct human ignition is particularly significant. For instance, from 1992 to 2012, 84\% of wildfires in the United States were human-induced \citep{balch2017human}. Similarly, \cite{vilar2016modeling} found that 95\% of wildfires in Mediterranean Europe between the 1980s and 2010s were caused by human activities. Notably, human factors not only increase the frequency of fires but also extend the fire season and lead to more random distributions of fire locations \citep{balch2017human}. This randomness introduces greater uncertainty into wildfire risk prediction, as wildfires may occur in areas with infrequent lightning or where lightning does not coincide with high temperatures, dryness, and strong winds \citep{abatzoglou2016controls}.

\cite{chuvieco1989application} first introduced human activity variables into wildfire risk research by considering proximity to road networks and areas with high human activity, such as recreational zones. Since then, an increasing number of studies have incorporated human activity variables into wildfire risk prediction models. These variables include the distance of various locations from nearby roads, power lines, villages, and cities, as well as road density, population density, and land use. The assumption underlying the inclusion of these variables is that wildfire frequency is negatively correlated with distance from human infrastructure and positively correlated with road density and population size \citep{gilreath_validation_2006, martinez_human-caused_2009}. In addition to these easily describable features, there are more complex factors influencing wildfire risk, such as the potential for urban development to increase fuel loads in rural areas, or the possibility that controlled burns for agricultural purposes in remote areas may increase the likelihood of uncontrolled wildfires \citep{martinez_human-caused_2009}. Several studies have confirmed these assumptions. For example, \cite{jaafari_wildfire_2019} found a strong correlation between high wildfire probability and road network density in the Zagros ecoregion of western Iran. \cite{ghorbanzadeh_forest_2019}, in their study of the forest areas in Amol County, Mazandaran Province, northern Iran, similarly found that human activities near major roads significantly influenced wildfire risk. The interaction between human settlements and vulnerable infrastructure in areas with high or moderate forest fire susceptibility leads to the emergence of high-risk zones on wildfire risk maps.

Therefore, most wildfire risk prediction studies consider the socio-economic factors mentioned above. For instance, \cite{sadasivuni_wildfire_2013} generated a population interaction map based on forest resources and human settlement patterns and studied wildfire risk in Mississippi, southeastern United States. They found that areas with dense fuel and sparse populations were most likely to experience wildfires. \cite{malik_data-driven_2021} mapped wildfire risk in a densely vegetated area of approximately 63 $km^2$ between Monticello and Winters, California, using the location of power lines as a proxy for socio-economic factors, given that power lines can serve as ignition sources during high wind events. \cite{nami_spatial_2018} considered LULC and proximity to roads and settlements when mapping wildfire risk in the Hyrcanian ecoregion of northern Iran. They found that the probability of fire occurrence was highly dependent on human infrastructure and related activities. Additionally, the study revealed a positive correlation between fire occurrence and landscapes within 4,000 meters of human settlements, and a negative correlation for areas beyond 4 kilometers. \cite{rubi_performance_2023} conducted a study in Brazil's Federal District, dividing the region into zones based on the coverage of different weather stations and calculating the distance from each zone's center to the nearest road and building as proxies for socio-economic factors. \cite{bergado_predicting_2021} carried out a wildfire prediction study in Victoria, Australia, and found that land cover categories (such as farmland, forests, mines, and quarries) and distance from power lines were relatively low in importance. In their wildfire risk assessment in Guangdong Province, China, Jiang et al. considered socio-economic factors such as gross domestic product (GDP), distance from roads, and population density. The results showed that, aside from meteorological factors, GDP was the most significant factor in wildfire risk. Similarly, \cite{jiang_wildfire_2024} in their wildfire risk assessment in Guangdong Province, China, also utilized GDP, distance to highways, and population density as socio-economic factors. However, their results indicated that the contribution of socio-economic factors to wildfire risk was not significant. Furthermore, in a study conducted in the western part of the Tibetan Plateau in China, \cite{quan_improving_2023} found that although government reports indicated that the majority (about 95\%) of wildfires in the region were caused by human activities, the importance of distance from roads and residential areas was relatively low compared to meteorological factors.

Current research on the impact of human activities on wildfire risk primarily focuses on using the distribution density of population, roads, and other infrastructure. It also utilizes the straight-line distance of different locations within the study area from infrastructure as input features for models. However, the consideration of these distances is typically limited to simple linear measurements, without accounting for human accessibility. This limitation may reduce model performance in regions with rugged terrain. Additionally, current research on human factors mainly focuses on human-induced wildfires, with little consideration given to the mitigating effects of firefighting facilities, such as fire stations, firebreaks, and lookout towers.

\subsection{Terrain and Hydrological Features}

Terrain factors used in wildfire prediction studies include elevation, slope, aspect, hill shade, and planar curvature. Terrain indirectly determines mesoscale and microscale climate and airflow \citep{coen_wrf-fire_2013}, thereby impacting fuel conditions and wildfire propagation. For example, to some extent, an increase in slope leads to faster fire spread \citep{ghorbanzadeh_wildfire_2018}. Steeper slopes result in faster fuel preheating and ignition rates, and compared to flat terrain, the rate of fire spread on a 20° slope increases by about four times \citep{lecina-diaz_extreme_2014}. The slope aspect influences prevailing winds, humidity, solar radiation, and plant species distribution. In many regions of the Northern Hemisphere, north-facing slopes are typically cooler and more moist than south-facing slopes, making south-facing slopes at higher risk of forest fires \citep{sayad_predictive_2019}. Curvature and valley depth are geomorphic indicators of soil moisture and vegetation distribution \citep{kalantar_forest_2020}. In larger valleys, downslope-driven cold air pooling has a significant impact on vegetation density and composition \citep{kiefer_role_2015}. Additionally, elevation can also influence human activities, as lower elevations tend to have higher human population density and are more prone to wildfires compared to higher elevation areas \citep{guo_what_2016}. In addition, the main hydrological feature is the distance between fuel and water bodies, soil moisture, and the TWI. TWI represents soil saturation and surface runoff rates. Land with higher TWI is more saturated, and high moisture content can prevent fire ignition \citep{achu_machine-learning_2021}.

In addition, \cite{ghorbanzadeh_forest_2019} studied the forestry area of Amol County in the Mazandaran province of northern Iran and used slope, slope aspect, and altitude as terrain factors. They found that steep areas were generally more prone to forest fires. \cite{nami_spatial_2018} considered terrain factors such as slope, slope aspect, altitude, planar curvature, TWI, and topographic roughness index (TRI) when mapping wildfire risk in the Hyrcanian ecological region in northern Iran. Their further collinearity analysis revealed significant collinearity between TWI and slope with other parameters, leading to their exclusion from subsequent modeling tasks. A study conducted by \cite{tran_enhancing_2023} on the island of Oahu in the Hawaiian Islands also utilized terrain and hydrological features such as elevation, slope, aspect, plan curvature, profile curvature, valley depth, TWI, and proximity to rivers. The assessment of their significance indicates that slope, elevation, and TWI are of secondary importance compared to proximity to roads and temperature in relation to fire risk. \cite{gentilucci_analysis_2024} also used TWI in a study in the Marche region of central Italy. Their research shows that areas of the terrain with less accumulation of moisture are most susceptible to fires. 

Soil moisture information can serve as a complement or substitute for drought indices and is commonly used in wildfire risk prediction models. It can be utilized in modeling fuel moisture and other related factors \citep{krueger_using_2022}. For example, \cite{bakke_data-driven_2023} conducted a study in Fennoscandia, Finland, to explore the primary hydro-meteorological factors influencing wildfire occurrence. They found that shallow soil moisture anomalies were the main influencing factor. Similarly, in the study by \cite{vissio_testing_2023} on the prediction of summer burning area in Italy, the ERA5 soil moisture reanalysis data product was used. The research found that using only a single variable, namely surface soil moisture, for predicting wildfire burn area yielded good results. This is because surface soil moisture can serve as a proxy for the combined effects of temperature and precipitation. In the study by \cite{nur_spatial_2023} focusing on Sydney, Australia, the MODIS Terra Climate dataset was used, which provided drought indices and soil moisture data. However, due to the low resolution of the dataset (4 km), the importance of soil moisture as a factor was not fully reflected. In research conducted by \cite{yahia_prediction_2023}, to improve the resolution of soil moisture, temperature vegetation dryness index (TVDI), perpendicular drought index (PDI), optical trapezoid model (OPTRM), and modified normalized difference water index (MNDWI) were used as proxies for soil moisture. These variables, along with a Gaussian Naive Bayes (GNB) classifier, were employed to generate wildfire risk maps. Additionally, the NDVI can be used to estimate soil moisture anomalies \citep{tucker_red_1979, liu_wildfire_2010}. However, when modeling moisture anomalies using NDVI, it can only directly detect surface soil moisture. To obtain deep soil moisture, a physical or empirical relationship must be employed. In addition, it is worth noting that the estimation of soil moisture based on optical or SAR remote sensing imagery still faces significant challenges. This is due to the influence of factors such as tree canopy obstruction, absorption, reflection, and scattering, which result in a mixed signal received by the sensor containing information on both soil and fuel moisture \citep{pelletier_wildfire_2023}.

\subsection{Wildfire Historical Records}

Wildfire historical records are indispensable in all studies and form the data foundation for wildfire risk prediction models. The sources and scales of wildfire historical records vary widely. For example, in Canada, many provinces, cities, or forestry management agencies maintain their own wildfire historical records. Additionally, the Canadian Natural Resources Department maintains its wildfire information data system, such as the Canadian Wildfire Information System. Commonly used public global or regional scale wildfire historical records include the National Burned Area Composite dataset (NBAC), the Canadian National Fire Database (CNFDB), the National Institute for Space Research database (INPE), the Fire Planning and Analysis Fire Occurrence Database (FPA-FOD), the Global Fire Emissions Database (GFED), and FireCCI. In regions without systematic wildfire records, most studies use satellite-derived thermal anomalies or wildfire products from MODIS, Suomi-National Polar-orbiting Partnership (S-NPP), National Oceanic and Atmospheric Administration (NOAA) VIIRS, GOES, AVHRR, Sentinel, Landsat, Himawari, FY-4, among others.

Specifically, \cite{pelletier_wildfire_2023} used the NBAC dataset from the Canadian Forest Service for their study on the northern Canadian forest peatlands. \cite{liang_neural_2019} utilized the CNFDB dataset for wildfire research in Alberta, Canada, which includes fire location (latitude and longitude points), ignition date, extinguishment date, burned area, and cause of fire provided by provincial, regional, and the Canadian Parks Service firefighting agencies. \cite{rubi_performance_2023} employed the INPE database for their study in the Federal District of Brazil. The INPE fire database is primarily based on thermal anomaly data collected from satellites such as AQUA, TERRA, NOAAs-15, 16, 17, 18, and 19, METEOSAT-02, and GOES-12. It covers the geographic location and timing of fires (latitude, longitude, and date/time) since the year 2000. \cite{wang_identifying_2021} used the FPA-FOD dataset to assess the drivers of wildfires in the continental United States, which includes discovery date, burned area, and geographical location (longitude and latitude) of wildfires from 1992 to 2020. The dataset intentionally excludes prescribed fires that escaped and required suppression response. Despite the possibility of missing data on smaller fires, the omission is tolerable for analytical purposes since the largest 5\% of all fires account for the majority of the burned area. \cite{song_global_2020} utilized the Global Fire Emissions Database version 4.1 (GFEDv4) for predicting global monthly wildfire risks throughout the year. This dataset covers monthly wildfire areas globally from 1997 to 2016, with a spatial resolution of 0.25° x 0.25°. It also includes daily burned area data globally from August 2000 to 2015. \cite{bakke_data-driven_2023} employed the FireCCI dataset in their study in Fenno-Scandinavia, Finland, which provides global-scale pixel products at spatial resolutions of 250m or 300m, or gridded products at a resolution of 0.25 x 0.25 degrees, with a temporal resolution equal to or less than one month.

Satellite thermal anomaly products, such as MODIS MCD14 and MCD64A1 products \citep{GIGLIO2003273, GIGLIO201631} are commonly used in wildfire risk prediction studies. The MCD14 series dataset has a spatial and temporal resolution of 1km and one day, respectively, offering a long time series (from 2006 to present) and providing near-real-time wildfire hotspot maps. The MCD64A1 dataset has a higher spatial resolution of 500m, although it does not offer near-real-time products. In specific studies, \cite{abdollahi_explainable_2023} used MODIS fire and thermal anomaly data in their wildfire prediction research in the Gippsland region of Victoria, Australia, along with the NPWS Fire History - Wildfire and Prescribed Burning dataset provided by the New South Wales Department of Climate Change, Energy, the Environment, and Water. \cite{ghorbanzadeh_forest_2019} acquired polygon data for 34 wildfire areas through field surveys and assessed data for the forested regions of Amol County, Mazandaran Province, northern Iran using MODIS data. \cite{jaafari_wildfire_2019} and \cite{nami_spatial_2018} also employed historical data, field surveys, and validated using MODIS hotspot products in their studies in the western Zagros ecoregion and the northern Hyrcanian ecoregion of Iran, respectively, to enhance the reliability of historical data. \cite{kanwal_data-driven_2023} in their study in Pakistan used MODIS data from the Fire Information for Resource Management System (FIRMS) as wildfire historical data, setting a 60\% confidence threshold to reduce false alarm errors.

\begin{figure*}[h]
    \centering
    \includegraphics[width=1\linewidth]{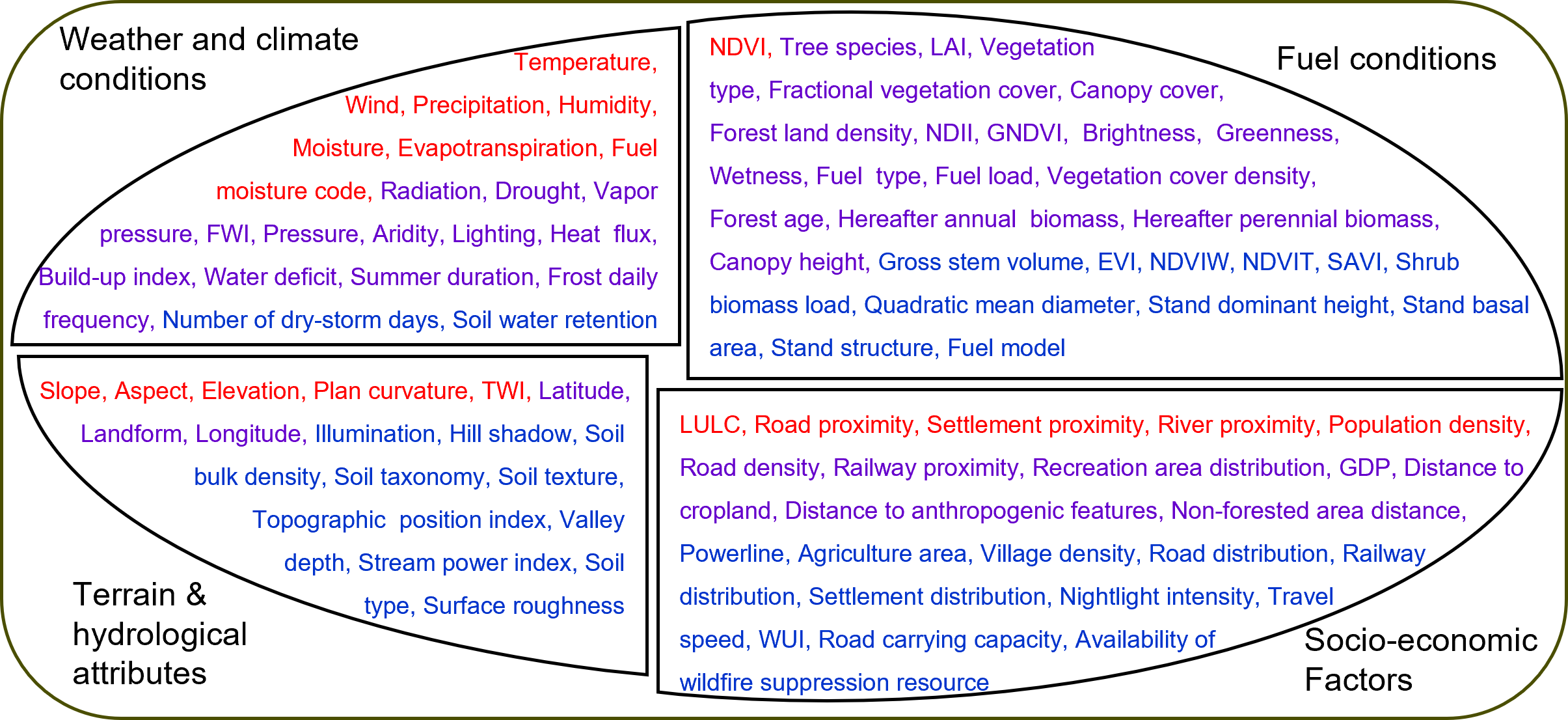}
    \caption{Feature utilization frequencies is classified into four categories:  weather and climate conditions, fuel conditions, terrain and hydrological attributes, and socio-economic variables. The color scheme in this visual representation signifies the prevalence of these features, with red, purple, and blue denoting utilization frequencies exceeding 20, 10, and 0 occurrences, respectively.}
    \label{fig: features frequency}
\end{figure*}

Additionally, some studies argue that MODIS products may overlook fires that are smaller in scale or shorter in duration \citep{hantson_strengths_2013, benali2016determining, fusco2019detection, ying2019wildfire}. Therefore, other satellite data are also used either in conjunction with MODIS data or independently for fire hotspot detection or burned area mapping, and the generation of historical datasets. For instance, the study by \cite{zhang_deep_2021} used the GFED dataset, which utilizes the MODIS MCD64A1 dataset and VNPIMG14ML dataset \citep{schroeder2014new, hall2024glocab}, to mitigate errors of commission and omission in agricultural fire detection in the MODIS dataset. Moreover, some studies have begun to explore using other high temporal resolution data to build global or regional wildfire detection datasets, such as the Sea and Land Surface Temperature Radiometer (SLSTR) sensor on Sentinel-3A/B \citep{xu2023sentinel}.

\subsection{Frequency of Different Data Utilization}

In the "Fuel Conditions" category, the NDVI is the most frequently utilized variable, with a frequency of 27, underscoring NDVI's critical role in assessing vegetation health and fuel availability. Other significant variables include Tree Species, which appears 8 times, and LAI and Vegetation Type, each with a frequency of 6. These variables are essential for understanding the characteristics of biomass that contribute to wildfire risk. The presence of numerous variables with lower frequencies, such as Enhanced Vegetation Index (EVI) and NDWI, each used only once, highlights the complexity of this category. This diversity of variables reflects the multifaceted nature of fuel dynamics, requiring a broad range of factors to be considered in fire risk modeling, as depicted in Fig. \ref{fig: features frequency}.

In the "Climate and Weather Conditions" category, Temperature emerges as the dominant variable, with a frequency of 65. Precipitation follows closely with 53 occurrences, and Wind is also significantly utilized, with a frequency of 42. These variables are fundamental in understanding fire behavior, particularly in their influence on fire ignition, spread, and intensity. Humidity, used 27 times, and Moisture, used 16 times, further contribute to this category, reflecting the central role of atmospheric conditions in fire risk assessments. Other variables, such as Fuel Moisture Code, utilized 11 times, and Vapor Pressure, with 9 occurrences, are less frequently employed but remain crucial in specific contexts. The data presented in Fig. \ref{fig: features frequency} emphasizes that climate variables, particularly Temperature and Precipitation, are among the most critical factors in fire dynamics, with a wide array of related variables also contributing to detailed climatic modeling.

Land Use and Land Cover stands out as the most frequently utilized socio-economic variable, appearing 32 times, underscoring its importance in assessing how human activities and changes in land use influence fire risk. Road Proximity, with 23 occurrences, and Settlement Proximity, with 18, follow, indicating the significance of infrastructure and human settlement patterns in fire studies. Population Density, appearing 16 times, and Road Density, with a frequency of 4, also play roles, although to a lesser extent. The data in Fig. \ref{fig: features frequency} suggest that while variables like Gross Domestic Product (GDP) and Distance to Cropland—each used 5 times—are relevant, they are more context-specific. This analysis indicates that socio-economic factors are essential for understanding the human impact on fire dynamics, particularly in areas where land use and human infrastructure intersect with natural landscapes.

\afterpage{
\newpage

\begin{landscape}
\newgeometry{left = 0.15 cm, top=18cm, right=0.15cm, bottom=0.12cm}
\begin{table}[!htbp]
% \centering
\caption{Open-access Wildfire Spread Prediction Datasets}
\label{oa_datasets_description}
\small 
\renewcommand{\arraystretch}{1.8}
% \begin{supertabular}{\textwidth}{p{2cm} p{1.25cm} p{2cm} p{3cm} p{5.8cm} p{3cm} p{2.5cm} p{2cm}}
\begin{supertabular}{p{2.2cm} p{2.3cm} p{4cm} p{6.5cm} p{3cm} p{2.3cm} p{2.5cm}}
 \hline
  \textbf{Dataset} & 
  \textbf{Spatial-temporal Coverage and Resolution} & 
  \textbf{Fuel Conditions} &
  \textbf{Climate and Weather Conditions} &
  \textbf{Socio-economic Factors} &
  \textbf{Terrain and Hydrological Features} &
  \textbf{Historical Records} \\
\hline

  \textcolor{blue}{\href{https://zenodo.org/records/6475592}{FireCube}} &
  Greece and Eastern Mediterranean (1 km), 2009-2021 (Daily) &
  \textcolor{red}{MODIS}: EVI (MOD13A2, 16d, 1 km), LAI (MOD12A2H, 500 m, 8d), NDVI (16d, 1 km) &
  \textcolor{red}{ERA5-Land}: Avg, max, min of dew point temperature, relative humidity, surface pressure, air temperature, total precipitation, U and V wind components (1d, 9 km); \textcolor{red}{MODIS}: Total evapotranspiration (MOD16A2, 8d, 500 m), day and night LST (MOD11A1, 1d, 1 km), FPAR (8d, 500 m); \textcolor{red}{Europe-EDO}: Soil moisture index anomaly, Soil moisture (10d, 5 km) &
  \textcolor{red}{Corine Land Cover}: LULC (2006, 2012, 2018); \textcolor{red}{WorldPop}: Population density (2009-2021), distance from roads, distance from waterways &
  \textcolor{red}{EU-DEM}: Elevation, aspect, roughness, slope (2016, 30m) &
  \textcolor{red}{EFFIS \& MODIS}: Ignition points, burn area, daily fire count \\  
\hline
  \textcolor{blue}{\href{https://www.kaggle.com/datasets/fantineh/next-day-wildfire-spread}{Next Day Wildfire Spread}} &
  USA (1 km), 2012-2020 (Daily) &
  \textcolor{red}{VIIRS}: NDVI (VNP13A1) (8d, 500 m) &
  \textcolor{red}{GRIDMET}: Wind direction and speed, minimum and maximum temperatures, humidity, precipitation (1d, 4 km), drought index, energy release component (ERC) (5d, 4 km) &
  \textcolor{red}{GPWv4}: Population density (1 km) &
  \textcolor{red}{SRTM}: Elevation (30 m) &
  \textcolor{red}{MODIS}: MOD14A1 V6 (1d, 1 km) \\  
\hline

  \textcolor{blue}{\href{https://wildfire-modeling.github.io/}{WildfireDB}} &
  California, USA (375 m), 2012-2018 (Daily) &
  \textcolor{red}{LANDFIRE}: Canopy base density, base height, cover, canopy height, and existing vegetation cover, height, and type (2012, 2014, 2016, 30 m) &
  \textcolor{red}{NOAA Ground Station Data}: Temperature (avg, min, max), total precipitation, atmospheric pressure, and wind speed (1d, 5787 stations) & 
  $ $ &
  Elevation and slope (2016, 30 m) &
  \textcolor{red}{VIIRS}: Active Fire (1d, 375 m) \\
\hline

   \textcolor{blue}{\href{https://github.com/SebastianGer/WildfireSpreadTS}{WildfireSpreadTS}} &
  USA (375m), 2018-2021 (Daily) &
  \textcolor{red}{VIIRS}: EVI (VNP13A1) and NDVI (8d, 500 m) &
  \textcolor{red}{GRIDMET}: Surface temperature (min, max), total precipitation, wind speed and direction, specific humidity, PDSI (1d, 4.6 km); \textcolor{red}{Global Forecast System}: Temperature and wind (avg, direction, speed) (1d, 27.83 km) &
  \textcolor{red}{MODIS}: Land Cover Type Yearly Global product (MCD12Q1, 1y, 500 m) &
  \textcolor{red}{SRTM}: Elevation, slope, and aspect (2013, 30m) &
  \textcolor{red}{VIIRS}: Active Fire (1d, 375 m) \\
\hline

  \textcolor{blue}{\href{https://osf.io/f48ry/}{CFSDS}} &
  Canada (180m), 2002-2021 (Daily) &
  \textcolor{red}{SCNFI}: Percentage deciduous component, percentage coniferous component, biomass in tonnes/ha, and crown closure percentage (30m); Peatland presence and class (250m); \textcolor{red}{ERA5-Land}:  FFMC, DMC, DC, ISI, BUI (1d, 0.1deg) &
  \textcolor{red}{ERA5-Land}:  The maximum daily temperature, 24 hr precipitation, noon wind speed, noon relative humidity, and 24hr maximum vapour pressure deficit (1d, 0.1deg) &
  \textcolor{red}{CanVec Transport Features}: Road density, distance to road &
  \textcolor{red}{ASTER}: Elevation, slope, aspect, and TWI; \textcolor{red}{National Ecological Framework for Canada}: Ecozones &  
  \textcolor{red}{National Burned Area Composite}; \textcolor{red}{VIIRS}: VNP14IMGT and VJ114IMGTDL\_NRT (375m); \textcolor{red}{MODIS}: MCD14ML (1km) \\
\hline

  \textcolor{blue}{\href{https://zenodo.org/records/8055879}{SeasFire Cube}} &
  Global (0.25deg), 2001-2021 (8d) &
  \textcolor{red}{MODIS}: NDVI (MCD15A2, 8d, 500m), LAI (MOD13C1, 16d, 5600m) &
  \textcolor{red}{ERA5}: Mean sea level pressure, total precipitation, relative humidity, vapor pressure deficit, sea surface temperature, skin temperature, wind speed, 2m temperature (mean, min, max), surface net solar radation, surface net solar radation, surface solar radation downward, volumetric soil water (level1-4), land-sea mask (0.25deg); \textcolor{red}{Copernicus CEMS}: Drought code (max, avg), fire weather index (max, avg) (1d, 0.25deg); \textcolor{red}{CAMS}: Carbon dioxide emmision from wildfire, fire radative power (1d, 0.1deg); \textcolor{red}{MODIS}: LST (MOD11C1, 1d, 0.05deg); \textcolor{red}{NOAA climate indices} &
  \textcolor{red}{GPWv4}: Population density (1km); \textcolor{red}{ESA CCI}: LULC (7d, 300m/1km); \textcolor{red}{Biomes} &
   &
  \textcolor{red}{FireCCI}: Burned areas (area, mask, fraction) (250m/300m); \textcolor{red}{GFED}: Burned areas (area, mask) (1month, 0.25deg); \textcolor{red}{GWIS}: Burned areas (area, mask) \\

\hline

\end{supertabular}
\end{table}
\end{landscape}
% \clearpage
\restoregeometry
}

In the "Terrain and Hydrological Features" category, Slope is the most frequently used variable, appearing 46 times. This reflects the critical role of slope in influencing fire spread and intensity. Aspect, with a frequency of 34, and Elevation, with 32, are also key factors in understanding how topography affects fire behavior. These variables are particularly important in regions with complex terrain, where physical landscape features significantly impact fire dynamics. Plan Curvature, used 11 times, and TWI, used 10 times, are less frequently applied but remain relevant in specific contexts, often related to hydrological modeling. Other variables like Latitude, used 5 times, and Landform, with 3 occurrences, indicate their more specialized application. The analysis of these features highlights that, while terrain and hydrological factors are crucial for understanding the physical environment's impact on fires, certain variables like Slope, Aspect, and Elevation consistently prove to be more influential.

Overall, when comparing the categories, climate and weather conditions emerge as the most frequently utilized, with Temperature and Precipitation leading the way. Fuel conditions follow closely, with NDVI being the most prominent indicator of vegetation health and fuel availability. Socio-economic factors, while important, are used more selectively, with LULC, Road Proximity, and Settlement Proximity being the primary variables. Finally, terrain and hydrological features, particularly Slope, Aspect, and Elevation, are critical for understanding how the physical landscape influences fire behavior. However, it is worth noting that the frequency of use of the aforementioned features is not solely dependent on their importance but also on their ease of availability. For example, in the "Fuel Conditions" category, NDVI is more readily accessible compared to other features; similarly, variables like Temperature, LULC, Slope, Aspect, and Elevation are also more easily obtained.

\subsection{Open-Source Wildfire Spread Prediction Datasets}

With the growing reliance on machine learning methods for wildfire risk prediction, the need for standardized datasets to establish, evaluate, and compare models has become evident. To address this, several open-source wildfire spread datasets have been released, providing varying spatial, temporal, and thematic coverage for research. These datasets support the development of machine learning-based wildfire risk prediction models and offer standardized benchmarks for evaluating model generalization. 

The datasets presented in Table ~\ref{oa_datasets_description}, including FireCube \citep{prapas_2022_6475592, kondylatos2023mesogeos}, Next Day Wildfire Spread \citep{huot2022next}, WildfireDB \citep{singla2020wildfiredb}, WildfireSpreadTS \citep{gerard2023wildfirespreadts}, CFSDS \citep{barber2024canadian}, and SeasFire Cube \citep{alonso_2023_8055879}, encompass various factors critical to wildfire spread prediction. These factors include spatial-temporal coverage of fuel conditions, climate and weather conditions, socio-economic factors, terrain, and historical records. Each dataset differs in the spatial-temporal coverage, and resolution of these factors, providing diverse perspectives and data for wildfire modeling.

First, the datasets vary significantly in their spatial and temporal resolutions. For instance, FireCube and Next Day Wildfire Spread focus on coarse-resolution data with a spatial resolution of 1 km for daily observations. WildfireDB offers more localized coverage at 375 m but over a broader temporal range (2012-2018). The CFSDS dataset provides highly detailed data at 180m resolution, though its temporal span is longer (2002-2021). SeasFire Cube offers the broadest coverage globally, albeit at a coarser resolution of 0.25 degrees and 8-day intervals.

Regarding fuel conditions, the datasets primarily utilize vegetation indices such as MODIS EVI and NDVI. FireCube combines MODIS EVI, LAI, and NDVI to provide a detailed view of vegetation. Similarly, Next Day Wildfire Spread and WildfireSpreadTS rely on VIIRS NDVI, offering 8-day data at 500m resolution. WildfireDB stands out by incorporating LANDSFIRE data, which includes specific details like canopy base density and height, providing a more comprehensive fuel description, although at a lower temporal resolution. CFSDS uses specialized Canadian forest inventory data, adding complexity by considering forest composition and peatland presence.

For climate and weather conditions, most datasets rely on reanalysis products. FireCube integrates ERA5-Land data, covering a broad range of climatic variables, including temperature, precipitation, and soil moisture. Next Day Wildfire Spread and WildfireSpreadTS use GRIDMET data for similar parameters, with WildfireSpreadTS further incorporating Global Forecast System data for temperature and wind predictions. WildfireDB uses NOAA Ground Station data, focusing on ground-based observations, which may offer more accuracy but less spatial coverage.

The inclusion of socio-economic factors varies widely among the datasets. FireCube incorporates multiple socio-economic variables, including LULC, population distribution, and proximity to roads and waterways, offering a comprehensive view of human impact on wildfire spread. Next Day Wildfire Spread and SeasFire Cube focus solely on population density, while WildfireSpreadTS considers only LULC. WildfireDB does not include socio-economic factors, limiting its ability to account for human influences on wildfire ignition.

Terrain and hydrological features are primarily derived from DEM products, providing data on elevation, slope, and aspect. FireCube uses the EU-DEM, while Next Day Wildfire Spread and WildfireSpreadTS rely on SRTM products. CFSDS employs ASTER data for terrain variables, while WildfireDB and SeasFire Cube use various unspecified sources for elevation data. Despite different products, all datasets achieve similar resolutions of around 30m.

Historical wildfire records are crucial for validating wildfire spread and risk prediction models. FireCube incorporates both MODIS and EFFIS records, offering detailed fire occurrence accounts. Next Day Wildfire Spread relies on the MOD14A1 product, while WildfireDB and WildfireSpreadTS use VIIRS data at 375m resolution. SeasFire Cube integrates multiple sources, including FireCCI, GWIS, and MODIS, to provide a comprehensive historical dataset, though at a coarser resolution.

It is important to note that we only consider time series wildfire observation datasets stored in a spatially explicit format and encompassing multiple wildfire impact factors. Datasets that do not have a clearly defined spatial distribution \citep{sayad_predictive_2019}, datasets containing only historical wildfire extent \citep{short2014spatial, andela2019global, artes2019global, lizundia2020spatio, william_w_hargrove_2022_6522071, gincheva2024monthly}, fire detection datasets \citep{chino2015bowfire, toulouse2017computer, qad6-r683-20,  mou2020era, el2023wildfire}, and simulation datasets \citep{wang2024firebench}, are not included. 

It is worth noting that, as shown in Table~\ref{oa_datasets_description}, the number of available open-source wildfire spread datasets is limited, with most coverage concentrated in the United States and Mediterranean, while other regions, such as the wildfire-prone Canada and Australia, are underrepresented. Additionally, some studies using these datasets have overly simplistic considerations of fuel conditions, such as relying solely on vegetation indices as proxies. This approach fails to accurately and comprehensively describe the state of fuel load, moisture content, and continuity. Similarly, using MODIS or VIIRS to record historical burned areas lacks precision. These datasets may may overlook small fire spots, offer low spatial and temporal resolution, and be prone to false alarms \citep{zubkova2024assessment}. Furthermore, there is a lack of quantitative evaluation of how the accuracy of different products impacts the accuracy of wildfire spread predictions.

\section{Features Collinearity and Attribution}

The relationship between wildfire risk and its driving factors has long been a focal point in wildfire prediction research. Understanding the interactions between wildfire occurrence and factors such as fuel conditions, climate, meteorological variables, terrain characteristics, and human activities is critical for model design, data selection, and supporting wildfire management decisions \citep{aldersley_global_2011}. However, the inherent "black-box" nature of deep learning models poses significant challenges in interpreting the contribution of individual factors to wildfire risk predictions \citep{castelvecchi2016can}. This section systematically reviews commonly used feature attribution methods in deep learning-based wildfire risk prediction models, highlighting their potential and limitations.

Furthermore, wildfire risk prediction models often incorporate multiple driving factors, some of which may exhibit a high degree of multicollinearity. Highly correlated independent variables can introduce redundancy, complicating the isolation and interpretation of their individual contributions to the dependent variable \citep{chen_spatial_2019}. Systematically assessing multicollinearity and quantifying the individual contributions of these factors are essential for improving model reliability and providing actionable insights for effective wildfire management. Additionally, commonly used features in wildfire risk prediction models—such as precipitation, temperature, wind speed, terrain attributes, and vegetation indices—often overlap across studies. Investigating whether these features share inherent collinearity, independent of spatial and temporal distributions, is particularly valuable. This analysis could guide feature selection in future research, especially in data-constrained scenarios, ultimately enhancing the efficiency and accuracy of model development.

\subsection{Assessment of Feature Collinearity}

Multicollinearity refers to the presence of approximate linear dependencies among two or more predictor variables, meaning that one predictor can be linearly represented by a combination of others \citep{senawi2017new}. This phenomenon leads to several issues, including inflated standard errors of coefficients, coefficients with signs inconsistent with theoretical expectations, statistically insignificant parameters despite strong correlations between predictor variables and the outcome, and excessively large correlation coefficients relative to the explanatory power of the model \citep{math10081283}.

Under multicollinearity, regression methods often produce redundant, unstable models with poor generalization performance, making it difficult to isolate the marginal effects of individual predictors \citep{katrutsa2017comprehensive, math10081283}. Although deep learning models—which do not rely on linear parameter estimation—are less affected in terms of predictive performance, multicollinearity can still adversely impact the accuracy and stability of interpretability in predictive analysis. This, in turn, reduces the overall efficiency of model interpretation and analysis \citep{salih2024perspective}.

To evaluate collinearity in wildfire risk prediction, common metrics such as the Variance Inflation Factor (VIF) and tolerance are widely employed \citep{hair1986multivariate, liao_variance_2012}. VIF measures the extent to which the variance of a regression coefficient is inflated due to collinearity, while tolerance quantifies the proportion of variance in a predictor that is not explained by other predictors in the model. These metrics are critical for identifying and mitigating collinearity-related issues, ensuring robust model development and interpretation. The formulas for these metrics are as follows:

\begin{equation}
    \label{VIF}
    \begin{cases} 
        & \text{VIF} = \frac{1}{\text{Tolerance}} \\
        & \text{Tolerance} = 1 - R^2
    \end{cases}
\end{equation}
where \( R^2 \) is the coefficient of determination when the predictor is regressed against all other predictors. A high VIF value (typically above 10) or a low tolerance value (below 0.1) indicates significant collinearity, suggesting that one predictor is largely explained by others.

VIF had been widely used in wildfire risk predictions. For example, \cite{nami_spatial_2018} conducted a collinearity analysis using VIF and tolerance in their study on wildfire risk mapping in the northern Hyrcanian ecoregion of Iran. They found significant collinearity between the TWI and slope with other parameters, leading to their exclusion from subsequent modeling efforts. In contrast, \cite{jaafari_wildfire_2019} assessed the collinearity of various predictors using VIF and tolerance in their study in the western Zagros ecoregion of Iran. They discovered that the VIF values for TWI and slope were below 5, and tolerance values were above 0.2, indicating insignificant collinearity with other predictors \citep{liao_variance_2012}. \cite{hong_predicting_2019}, in their modeling of wildfire susceptibility in Huichang County, China, used VIF and tolerance to assess the collinearity of multiple predictors and found no significant collinearity between slope and other features.

In another study, \cite{li_predictive_2022} examined the drivers of forest fires in Yunnan Province, China, and found that six meteorological factors (daily average atmospheric pressure, daily minimum atmospheric pressure, daily average temperature, daily minimum temperature, daily average surface temperature, and daily minimum surface temperature) had VIF values exceeding 10 and tolerance values below 0.1, indicating significant correlations. The study also used the Pearson correlation algorithm to analyze the correlations of the remaining variables, finding a correlation coefficient of 0.77 between surface temperature and air temperature.

Similarly, in their wildfire prediction model for Maui Island in Hawaii, \cite{rezaie_development_2023} analyzed the correlation and collinearity among independent variables (altitude, slope, aspect, valley depth, TWI, slope length, plan curvature, distance from rivers, distance from roads, NDVI, average monthly rainfall, average annual wind speed, and average annual temperature) using Pearson correlation coefficients, VIF, and tolerance. They found the strongest correlations between average annual temperature and altitude, and the most significant collinearity between average annual temperature and distance from roads. However, the collinearity and correlations among all variables were not statistically significant.

In summary, various studies have revealed differing levels of multicollinearity among features, highlighting the challenge of drawing consistent conclusions about the relationships between specific features in wildfire risk prediction models. Nevertheless, feature multicollinearity analysis remains critical as it facilitates dimensionality reduction, thereby accelerating model training and inference. More importantly, addressing multicollinearity can significantly improve the accuracy, reliability, and stability of feature attribution, particularly when employing attribution methods that assume feature independence \citep{salih2024perspective}.

\subsection{Feature Attribution}

In wildfire risk prediction algorithms, explicit methods can predict wildfire risk by defining the contributions of different features to the ignition and spread of wildfires. However, these methods often have poor fitting capabilities and struggle to handle the nonlinearity and randomness characteristic of wildfire risk prediction tasks. On the other hand, machine learning models, such as deep learning and ensemble learning models, can extract patterns from highly complex datasets and demonstrate superior predictive performance. However, their decision-making processes are opaque, earning them the label of "black box models."

In wildfire risk prediction, this lack of transparency undermines the credibility of the predictions and makes it difficult for researchers and fire management agencies to understand the factors inducing wildfire ignition and spread. This, in turn, hinders the understanding of wildfire mechanisms and the prevention and management of wildfires. The interpretation of the decision-making process in machine learning models is generally referred to as explainable AI (XAI). In this paper, we define XAI as the attribution of causal responsibility or feature attribution \citep{josephson1996abductive}, as XAI in wildfire risk prediction algorithms mainly focuses on explaining the influence of different wildfire risk factors on model predictions. Currently, common XAI methods in wildfire risk prediction include Permutation Feature Importance (PFI), Explainable Feature Engineering \citep{ali2023explainable}, and SHapley Additive exPlanations (SHAP).

Firstly, PFI quantitatively analyzes feature importance by randomly permuting each feature and evaluating the change in model performance. For example, \cite{rubi_performance_2023} used PFI to assess the contributions of meteorological, fuel, terrain, and socioeconomic factors in models such as ANN, SVM, NB, KNN, LR, LogLR, and AdaBoost in their study in the Federal District of Brazil. They found that analyzing individual variables might lead to contradictory results. For instance, NDVI was more important for models like AdaBoost and ANN, whereas its influence was relatively low or nonexistent in other models such as LR and SVM. This finding highlights the necessity of considering multiple variables in predictive modeling and the varying importance of different models. Additionally, \cite{shadrin_wildfire_2024} evaluated the impact of specific feature values on the final test results by setting them to zero during model testing. Similarly, \cite{chen_estimation_2024} calculated the importance of each feature by measuring the variation in prediction error when feature values were perturbed.

Secondly, Explainable Feature Engineering \citep{ali2023explainable} methods have also been applied to wildfire risk prediction. For example, \cite{rezaie_development_2023} used the Information Gain Ratio (IGR) to select the optimal subset of variables in their wildfire prediction model for Maui, Hawaii. They found that the distance to roads had the highest IGR value of 0.903, indicating its significant impact on mapping fire-prone areas. The next important variables were annual average temperature (0.724), elevation (0.715), and slope (0.665). Similarly, \cite{zhang_forest_2019} used IGR to evaluate the importance of different factors in assessing forest fire risk in Yunnan Province, China. They found that temperature and wind speed contributed the most to the risk, while the importance of distance to roads and rivers was the lowest. In another example, \cite{hong_predicting_2019} modeled the spatial distribution of wildfire susceptibility in Huichang County, China, using the Analytic Hierarchy Process to assign weights to a logistic regression model and found that elevation was the most influential indicator of fire occurrence, followed by land use, NDVI, and distance to human settlements.

Finally, SHAP analysis \citep{shapley1953stochastic}, which is based on game theory, is more commonly used, especially for explaining complex models, because it considers both feature importance and feature interactions while offering higher computational efficiency. SHAP explains the impact of each feature on model predictions by assigning SHAP values to them. Common SHAP methods include Kernel SHAP, Deep SHAP, and Tree SHAP \citep{lundberg_unified_2017}. For instance, to address the challenge of explaining deep learning models, \cite{abdollahi_explainable_2023} introduced SHAP to study the contribution of different features to the model. The authors found that relatively high NDVI, temperature, and elevation, as well as relatively low NDMI and precipitation, increased wildfire risk, while relatively high humidity reduced this risk. They also identified elevation, NDMI, and precipitation as the three most critical factors influencing wildfire occurrence.

Similarly, \cite{wang_identifying_2021} used SHAP to evaluate the drivers of large-scale wildfires in the contiguous United States and found that coordinate variables (including longitude and latitude) and local meteorological variables (such as ERC, RH, temperature, and VPD) were important predictors of burned area across the domain. This is because coordinate variables carry crucial geographic information that broadly reflects climate, land use, human activity, and other key geographic factors, helping to distinguish temporal characteristics of different fire regimes and thus aiding in predicting burned areas. Using SHAP's local interpretability, they also explained the driving factors of large burned areas in different regions and months within the contiguous United States. For example, ERC was identified as the most important indicator of large burned areas in the western US.

\cite{iban2024shap} conducted a local SHAP importance analysis in the İzmir region of Turkey and found that wind speed, LULC, slope, temperature, and NDVI were the five most important influencing factors. Wind speeds above 3.5 m/s, forests in LULC, slopes greater than 8 degrees, annual average temperatures above 14 degrees, and NDVI values below 0.2 all promoted wildfires, while higher NDVI reduced the likelihood of wildfire occurrence. In the wildfire risk assessment for Hawai'i Island, \cite{tran2024improving} considered 14 factors, including meteorological, topographic, anthropogenic, and vegetation-related variables: elevation, slope, aspect, plan curvature, profile curvature, TWI, valley depth, annual average wind speed, NDVI, annual average precipitation, annual average temperature, distance to roads, distance to rivers, and land use. SHAP analysis revealed that the distance from a road, annual temperature, and elevation were the most influential factors, with wildfire risk increasing as distance from roads decreased and temperature increased. Additionally, low- to mid-elevation areas were generally associated with a higher likelihood of fire occurrence due to their relatively high temperatures and accessibility.

\cite{kondylatos_wildfire_2022} conducted a SHAP analysis for the Eastern Mediterranean region and found that soil moisture index, humidity indicators, temperature variables, NDVI, and wind speed were the most important factors influencing wildfire risk. Notably, the study found that the relationship between NDVI and fire risk was U-shaped, meaning that both very low and very high NDVI levels were associated with lower wildfire risk.

Currently, explainability analysis in wildfire prediction faces several challenges, such as insufficient exploration of alternative explainability methods, dependency on models and explanation techniques, unclear attribution analysis, and the uncertainty of the explanations provided.

Firstly, techniques like activation maximization analysis \citep{erhan2009visualizing}, Partial Dependence Plot \citep{greenwell2017pdp}, Integrated Gradients, and other additive feature attribution methods beyond SHAP—such as Local Interpretable Model-agnostic Explanations (LIME), DeepLIFT, and Layer-Wise Relevance Propagation—have not been adequately explored in the context of wildfire risk prediction. The limited application of these XAI methods leaves gaps in our understanding of how different approaches might enhance model transparency in this domain.

Secondly, many studies select the best-performing model through comparative experiments and then conduct explainability analysis based on that model. The explanations provided by these methods often estimate how a specific ML model derives predictions from input data \citep{good2012common}. However, treating these analyses as direct insights into real-world phenomena can lead to misleading or erroneous conclusions, particularly if the model’s learned decision rules do not align with the actual underlying data relationships \citep{jiang2024interpretable}. For instance, using a single explainability method to interpret different ML models, or using different explainability methods to interpret the same ML model, might result in varying or even contradictory observations. Therefore, any inferences drawn from these post-hoc explanations should be approached with caution \citep{ploton2020spatial}.

Furthermore, wildfire prediction studies often employ multiple input variables and model the relationships between these variables and wildfire risk. While such models may perform well, they do not necessarily explain whether causal relationships exist between independent and dependent variables, as the modeled relationships often result from correlations among various features \citep{rogger2017land, molnar2020general}. An example of this is multicollinearity among features, which is very common in Earth sciences due to the highly complex interactions and interdependencies between geological, meteorological, ecological, and additional factors in the Earth system \citep{jiang2024interpretable}. For instance, features commonly used in wildfire risk prediction, such as NDVI, soil moisture, precipitation, temperature, and evapotranspiration, often exhibit complex interrelationships. A crucial condition for ML models to produce valid causal effect estimates is that their input variables must be independent of unobserved confounders. Therefore, in most cases, ML should not be considered a definitive source of causal knowledge \citep{jiang2024interpretable}. This highlights the need for further research into the application of causal ML \citep{tesch2023causal} in wildfire risk prediction.

Lastly, while explainability analysis enhances model transparency, these explanations are still subject to uncertainty, unreliability, and low robustness due to the influence of model selection, choice of explainability method, and the correlations within input data \citep{jiang2024interpretable}. Thus, it is important to explore how insights from related fields can be applied to quantitatively assess the validity of explainability analysis results in wildfire risk assessment models \citep{bommer2024finding}.

\section{Deep Learning Methods for Wildfire Risk Prediction}

\label{method}

Currently, methods for wildfire risk prediction include fire rating systems, statistical models, traditional machine learning, and deep learning approaches. Fire rating systems, extensively applied in regions like Canada, the USA, and Australia, combine empirical and mathematical modeling based on fluid dynamics, combustion, energy transfer, and factors like fuel, weather, and terrain conditions \citep{rezaie_development_2023, mell_physics-based_2007}. These models are interpretable and generalizable, especially when training data is limited. However, their reliance on assumptions regarding soil moisture, fuel load, landscape structure, and other factors can reduce prediction accuracy \citep{bar_massada_effects_2011}.

For more specific analyses, statistical models use regression to establish relationships between wildfire risks and influencing factors such as fuel conditions, meteorology, and terrain. While simple and interpretable, they have limited capacity for handling the nonlinearity and randomness typical in wildfire prediction tasks \citep{li_predictive_2022}. Traditional machine learning methods (e.g., Random Forests, SVM, ensemble models, and Bayesian methods) improve upon statistical models by using data-driven techniques and manual feature engineering to identify patterns in historical fire data. These methods operate non-parametrically and are better suited for non-linearities \citep{oliveira_wildfire_2021}, yet they present challenges in variable interpretation compared to statistical approaches. Over the past decade, the focus in wildfire modeling has shifted from statistical algorithms to machine learning \citep{jain_review_2020, oliveira_wildfire_2021}. Decision tree methods, such as Random Forests (RF), are particularly popular due to their accuracy and interpretability, as highlighted in studies like \cite{malik_data-driven_2021}. Nevertheless, some researchers argue that no single model is universally superior, suggesting a need for model combinations to enhance accuracy \citep{jaafari_wildfire_2019, razavi_termeh_flood_2018}.

Traditional machine learning methods often use shallow architectures that fail to exploit spatial patterns effectively, limiting feature extraction and classification accuracy \citep{zhang_hybrid_2018}. In contrast, deep learning models, despite their higher data requirements, can automatically extract complex features through hierarchical structures, reducing the need for manual feature engineering and enhancing generalizability. Their larger parameter capacity also allows them to manage large datasets and minimize overfitting risks. Although deep learning has become increasingly popular for wildfire risk prediction, reviews such as \cite{jain_review_2020} and \cite{zacharakis_integrated_2023} have only partially covered models like Recurrent Neural Networks (RNNs) and CNNs without a comprehensive systematization.

This section reviews the latest advancements in deep learning-based wildfire risk prediction models, which are typically framed as segmentation or classification tasks. These models use historical wildfire data and influencing factors as inputs, with future wildfire occurrences or risk levels as ground truth. Models are categorized into time series prediction, image semantic segmentation and classification, and spatio-temporal prediction. Time series models (RNN, LSTM, Gated recurrent unit (GRU), Transformers) predict wildfire risks using 1D feature vectors, while semantic segmentation and classification models (CNNs, Transformers) employ 2D feature maps to produce 2D risk predictions. Spatio-temporal models integrate these approaches using 2D time series data.

Deep learning models for spatio-temporal prediction often combine CNNs, Transformers, GCNs, and graph neural networks (GNNs) with RNNs, LSTMs, or GRUs, enhancing resolution with higher-resolution data over extended sequences. Despite advances in GPU technology supporting larger data scales, these models still struggle with efficiency and large spatiotemporal scales compared to newer models like Mamba \citep{gu_mamba_2023}. Additionally, many models use multi-branch networks that rely solely on images or nodes, missing the benefits of multimodal approaches like incorporating textual information.

While previous reviews (e.g., \cite{jain_review_2020, zacharakis_integrated_2023}) have examined machine learning in wildfire science, they often omit post-processing techniques essential for converting model outputs into actionable risk information. Directly using these outputs can be misleading due to class imbalances and variability between models. This section also introduces methods for classifying and calibrating deep learning wildfire risk predictions to address these challenges.

\subsection{Wildfire Risk Rating and Probability Calibration}
\label{calibration}

Although \cite{jain_review_2020, zacharakis_integrated_2023}  systematically reviewed the application of various machine learning methods in wildfire risk research, they did not provide a description of the post-processing techniques required to convert the predictions of machine learning models into actual wildfire risk. Directly using the output of these models as the probability of wildfire risk may lack theoretical foundation and stability. This is because wildfire historical data sets typically suffer from severe class imbalance, where the spatial and temporal scope of fire occurrences is considerably smaller compared to non-fire events. To expedite model convergence, many machine learning datasets are often constructed with a balanced class ratio, typically 1:1, 1:1.5, or 1:2 \citep{huot_deep_2021, kondylatos_wildfire_2022} between fire and non-fire instances. This may lead to an overestimation of false positives during model evaluation \citep{phelps2021guidelines, phelps_comparing_2021, bakke_data-driven_2023}. However, actual wildfire risk probabilities are usually much lower. For example, in wildfire risk assessments for the Mediterranean region and the United States, \cite{ager_wildfire_2014} and \cite{preisler2004probability} used probabilities in the thousandths range, respectively.

Using the probability output of machine learning models as an indicator of wildfire risk could be misleading to non-experts. Therefore, many studies convert continuous wildfire risk probabilities into risk levels to support wildfire management, such as conveying public warnings and guiding fire suppression measures. Among the methods for risk level classification, the Jenks natural breaks classification \citep{jenks1971error} is most commonly used. For instance, \cite{xie2022wildfire} trained an ensemble machine learning model using binary wildfire ignition records and then classified the predicted probabilities into five groups using the Jenks method. Similarly, \cite{iban2022machine} and \cite{moayedi2023wildfire} also classified the hazard maps generated by machine learning models into five categories using the Jenks method. Along the same lines, \cite{chen_estimation_2024} used the Jenks method to divide the estimated potential rate of spread and fire radiative power into five levels for wildfire risk assessment. 

Other methods have also been employed for risk level classification. For example, \cite{liang_neural_2019} used the Kennard-Stone method combined with standardized duration and fire size to categorize the wildfire risk predicted by an RNN into five levels. \cite{jalilian2023forest} applied natural discontinuity classification methods to group the probabilities predicted by machine learning models into three classes. \cite{bjaanes2021deep} and \cite{chicas2022modelling} divided the wildfire ignition probability maps produced by the models into five levels using an equal-interval method, with intervals of 0.2. Similarly, \cite{trucchia_defining_2022} manually defined five risk levels.

In comparison to methods that predict risk classifications or directly utilize confidence levels from deterministic models as risk indicators, there remains a paucity of research focused on recalibrating probability outputs to obtain actual wildfire risk probabilities through post-processing and pre-processing of model predictions and inputs. For instance, \cite{pelletier_wildfire_2023} constructed a semi-balanced dataset for model training, in which the numbers of burned and never-burned sites were 3,268 and 60,146, respectively. They then used the method proposed by \cite{elkan2001foundations} (Eq. (\ref{rectification})) to convert the wildfire probabilities predicted by a time series XGBoost model into actual probability values:

\begin{equation}
    \label{rectification}
    p'=\frac{BR2\times (p-p\times BR1)}{BR1-p\times BR1+p\times BR2-BR1\times BR2}
\end{equation}
where $p'$ represents the revised probability for a prediction, $BR1$ is the base rate of the predicted probability estimates (the average of predicted probabilities for that class), and $BR2$ denotes the base rate of the actual probability estimates derived from the target population. Here, the target population refers to the average probability of fires across all observations within the training dataset, including the unburned observations that were not utilized in model training. After comparing the fire probabilities before and after calibration, they found that the calibrated probabilities more accurately reflected the true likelihood of wildfire occurrence.

Similarly, \cite{phelps_comparing_2021} used the function proposed by \cite{dal2015calibrating} to correct the neural network's probability output:
\begin{equation}
    \label{rectification_2}
    p_{k}=\frac{\pi \gamma_{k}}{\pi \gamma_{k}-\gamma_{k}+1} 
\end{equation}
where $p_{k}$ and $\gamma_{k}$ represent the wildfire probabilities modeled from the original and sampled distributions, respectively, and $\pi$ denotes the proportion of non-fire observations sampled. Furthermore, this study also employed Platt Scaling and Offset methods to calibrate tree-based methods (bagged classification trees/random forest) and statistical models (logistic regression and logistic generalised additive models), respectively. The research ultimately found that uncalibrated machine learning models severely overestimate fire probabilities (e.g., predicted values several orders of magnitude higher than actual probabilities).

\cite{10.1145/3651671.3651708} propose a preprocessing method for calibrating wildfire ignition prediction probabilities. The core concept involves transforming binary fire events (occurrence or non-occurrence) into continuous fire risk intensity curves during the preprocessing phase, then mapping these through logistic regression models to produce calibrated probability predictions. This approach addresses class imbalance issues in wildfire risk prediction while avoiding the overestimation problems caused by sampling and uncalibrated models. Specifically, the research first applies a Laplacian convolution kernel to diffuse daily fire event sequences, thereby reducing the significance of isolated fire events while enhancing risk values for dates adjacent to continuous fire events. Subsequently, the authors calculate fire influence values for a given day using multi-scale sliding windows (Eq. (\ref{risk calibration 1})), incorporating both short-term and long-term trends.

\begin{equation}
\label{risk calibration 1}
\text{Influence}(J) = \sum_{t=1}^{T} \left( \frac{1}{2t+1} \sum_{i=-t}^{t} \text{Fires}_{J+i} \right)
\end{equation}
where, $T$ and $J$ represent the maximum time window length and the current day, respectively, while $\text{Fires}_{J+i}$ indicates the number of fire events on day $J+i$. Subsequently, $\text{Influence}(J)$ serves as input to a logistic regression model, which learns the relationship between this value and actual fire occurrences (0/1), yielding calibrated probability outputs. Evaluation of this method demonstrates excellent calibration performance. Assessment metrics include Log Loss and Expected Calibration Error (ECE). Compared to uncalibrated traditional classification models, this approach significantly reduces ECE (from approximately 10\% to 2\%--4\%) while maintaining low Log Loss, proving that its predictions achieve a favorable balance between accuracy and reliability \citep{10.1145/3651671.3651708}.

In addition to the aforementioned methods, \cite{ZHANG2025104563} introduced a two-stage parametric calibration framework specifically designed to address distributional shifts in deep learning-based drought detection, which is conceptually transferable to wildfire probability calibration. This two-step process corrects overconfident predictions by explicitly addressing spatial heterogeneity and sampling-induced class imbalance. In the first stage, they divided the study area into $N$ climate zones and fitted zone-specific logistic calibration functions to mitigate label prior shift. For each zone $K_i$, the calibrated probability $\hat{p}$ is computed as:

\begin{equation}
\label{zhang_stage1}
\hat{p} = \frac{1}{1 + \exp(-a_i \cdot \log\frac{p}{1-p} - b_i)},
\end{equation}
where $p$ is the original model output (logit), and $(a_i, b_i)$ are the learned parameters optimized by minimizing negative log-likelihood on zone-specific training samples. In the second stage, to correct likelihood shift across unseen validation data, they introduced a bias-corrected transformation:

\begin{equation}
\label{zhang_stage2}
\tilde{p} = \frac{\beta \hat{p}}{\beta \hat{p} - \hat{p} + 1},
\end{equation}
where $\beta$ is tuned by minimizing the ECE over binned confidence intervals. Experimental results demonstrated that this method achieved the lowest calibration error (ECE reduced from 4.06\% to 0.31\%), outperforming common techniques such as temperature scaling and Platt scaling.

Notably, research on calibrating confidence scores from deterministic models into well-calibrated probability predictions remains limited in deep learning-based wildfire risk studies. This may be attributed to the growing preference for probabilistic deep learning methods—such as Bayesian neural networks—that inherently estimate predictive distributions. Compared to deterministic models, probabilistic approaches typically exhibit superior calibration performance. They yield outputs that more accurately approximate true event probabilities while simultaneously quantifying both data-related (aleatoric) and model-related (epistemic) uncertainties. Related research on uncertainty estimation will be discussed in Section~\ref{uncertainty estimation}.

\color{black}

\subsection{Time Series Prediction}
\subsubsection{Recurrent Neural Networks}
As discussed in previous sections, the occurrence and spread of wildfires are influenced by various factors, including meteorological conditions, fuel status, and historical wildfire data. These factors exhibit cumulative effects \citep{flannigan2013global}. Therefore, utilizing time series data enables models to capture the gradual changes in influencing factors, thus facilitating the prediction of future wildfire risks. The most classical methods in time series prediction tasks are RNNs, which include RNN, LSTM, and GRU.

RNN is a basic neural network structure with recurrent connections, designed to process sequential data. RNN integrates input from the current time step with output from the previous time step to model and predict sequence data. A simple RNN structure consists of an input layer, hidden layer, and output layer. Input data are time series data, i.e., feature vectors with dimensions $f \times d$. The hidden layer includes several hidden units, where each hidden unit outputs $\bold{h}_{t-m}$, the 'memory' at any intermediate time $t-m (m<n)$, as a weighted sum of its previous 'memory' and current time step's input features:

% \begin{figure}
%     \centering
%     \includegraphics[width=1\linewidth]{images/rnn_architecture.png}
%     \caption{The architecture of RNN where (a) and (b) represent the overall architecture and detailed hiddenlayer calculation.}
%     \label{fig:classic_RNN_network}
% \end{figure}

\begin{equation}
    \begin{cases}
    & \bold{h}_{t-m}=\sigma_{h}(\bold{W}_{IN}\bold{x}_{t-m}+\bold{W}_{HH}\bold{h}_{t-(m+1)}+\bold{b}_{h}) \\
    & \bold{y}_{t-m}=\sigma_{o}(\bold{W}_{HO}\bold{h}_{t-m}+\bold{b}_{o})
    \end{cases}
\end{equation}
where $\bold{W}_{IN}$ and $\bold{W}_{HH}$ represent weight matrices for processing input to the hidden layer and from hidden layer to hidden layer, respectively; $\bold{x}_{t-m}$ and $\bold{h}_{t-(m+1)}$ represent the input feature vector at time $t-m$ and 'memory' from the previous time step, respectively; $\bold{b}_{h}$ and $\bold{b}_{o}$ represent bias vectors for the hidden units and output layer, respectively. Ultimately, the output $\bold{y}_{t}$ at time $(t-m (m<n))$ is a weighted sum of all hidden units. By iteratively repeating the above process, each timestep of the entire time series data ($t-n, t-(n-1), ..., t-1$) is predicted. Finally, a loss function (Eq. \ref{loss_function}) estimates the discrepancy between the prediction and labels, and model weight parameters are iteratively updated through backpropagation.

\begin{equation}
\label{loss_function}
 \iota(\bold{y}, \bold{GT})= {\textstyle \sum_{t-n}^{t-1}} \iota_{t}((\bold{y}_{t}, \bold{GT}_{t})
\end{equation}
where $\bold{GT}_{t}$ represents the label value at time $t$.

Classic RNN networks and their variants have been employed in wildfire risk prediction. For instance, \cite{cheng2008integrated} designed a spatiotemporal Elman RNN-based prediction framework integrating historical wildfire and weather data to predict the annual average burned area in Canadian forests. Experiments demonstrated a prediction error of less than 0.5ha on areas of 500ha, validating the feasibility of using RNN for fire prediction. Similarly, a comparative study of ten ANNs based on meteorological data for predicting wildfire incidence in Australia found that Elman RNN exhibited the best performance, with average accuracy, sensitivity, and specificity exceeding 93\% \citep{dutta2013deep}. Furthermore, a network composed of a dynamic autoencoder and RNN was used to predict the next month's burned area in five U.S. regions, where the dynamic autoencoder was employed to transform multidimensional time series data into RNN input features, and dual decoders were used for predicting fire spot likelihood over different time spans, i.e., one week and one month. Experiments showed that their method was slightly superior to comparative methods, generative network, and GRU \citep{chavalithumrong_learning_2021}.

While \cite{jain_review_2020} and others have highlighted the limited use of classical RNNs, our findings are consistent with these observations. The underutilization of RNNs may be attributed to their difficulties in handling long sequence data, where they are prone to gradient vanishing or exploding, making it challenging to capture long-term dependencies \citep{hochreiter1997long, gers2000learning}. To address these issues, LSTM networks were developed. LSTMs incorporate three gating units: input gate, forget gate, and output gate, which control the flow and retention of information. These gates have trainable weights and decide whether to pass, forget, or output information based on the current input and the output from the previous time steps. This gating mechanism effectively resolves the issues of vanishing and exploding gradients, thereby enabling better handling of long sequences and capturing of long-term dependencies. Consequently, LSTMs are more prevalently applied in wildfire risk prediction than classical RNNs.

Studies utilizing LSTM for wildfire risk prediction often model various risk factors, including weather and climate conditions, fuel conditions, and socio-economic factors. For instance, \cite{liang_neural_2019} found that LSTM outperformed RNN and Back Propagation Neural Network (BPNN) in predicting the class of wildfires in Alberta, Canada, based on meteorological conditions, achieving an overall accuracy of 90.9\%. Similar findings were reported by \cite{9456113}, who assessed the wildfire risk in Indian forests using meteorological parameters and LSTM methods. The accuracy and RMSE were 94.77\% and 37.5\%, respectively, surpassing other benchmark methods like CNN and SVM. Additionally, \cite{kondylatos_wildfire_2022} compared RF, XGBoost, and deep learning methods (LSTM and ConvLSTM) and found that deep learning significantly outperformed traditional machine learning methods in predicting next day wildfires in the Mediterranean using ERA5-Land meteorological data, MODIS NDVI, and diurnal LST data along with soil moisture, terrain data, land cover types, and distances from infrastructure. \cite{li2023attentionfire_v1} further integrated attention mechanisms and deep learning interpretability into an LSTM network, using monthly-scale climate, socio-economic, fuel variables, and oceanic indices to predict short-term (1–4 months) and long-term (5–8 months) wildfire burning areas in tropical regions. Experimental results demonstrated that the enhanced LSTM model outperformed traditional machine learning methods and the baseline LSTM model. Furthermore, \cite{cheng2022data,cheng2024deep} have combined LSTM with reduced order modeling techniques to improve the prediction accuracy for both regional and global scale wildfire systems. Additionally, interpretability analysis revealed that precipitation and vapor pressure deficit were critical driving factors, while oceanic indices provided more significant contributions to long-term wildfire burning area predictions.

Moreover, some studies use only historical wildfire occurrence data combined with LSTM to predict future regional wildfire risks. For example, \cite{kadir2023wildfire} used historical MODIS wildfire data combined with LSTM to predict the spatiotemporal distribution of wildfires in Indonesia for the following year, with a success rate exceeding 90\% and an error rate of only 6.94\%. Notably, to reduce the amount of data, this study grouped several days' worth of data into single days for training and testing. Furthermore, \cite{hu_developing_2023} combined LSTM with autoencoders, using historical wildfire data from 1992-2018 to predict the level of wildfire occurrence in high-risk areas such as California. The autoencoder was employed to generalize wildfire events from the overall dataset, enhancing the accuracy of anomaly event predictions through reconstruction errors.

Similar to the LSTM, the GRU addresses the challenges of gradient vanishing and explosion inherent in traditional RNNs through its gating mechanisms. GRU reduces the number of gating units found in LSTM, containing only a reset gate and an update gate which simplifies the model architecture and enhances computational efficiency \citep{chung2014empirical}. Despite its simpler configuration, GRU retains the capability to model long-term dependencies with performance comparable to that of LSTM \citep{jin2020urban}.

% \begin{align*}
%     \label{gru}
%     r_{t} = \sigma(W_{r}[h_{t-1}, x_{t}] + b_{r}) \tag{18} \\
%     z_{t} = \sigma(W_{z}[h_{t-1}, x_{t}] + b_{z}) \tag{19} \\
%     \tilde{h}_{t} = \tanh(W_{h}[r_{t} h_{t-1}, x_{t}] + b_{h}) \tag{20} \\
%     h_{t} = (1 - z_{t}) h_{t-1} + z_{t} \tilde{h}_{t} \tag{21} \\
% \end{align*}

% \begin{figure}
%     \centering
%     \includegraphics[width=1\linewidth]{images/GRU_architecture.png}
%     \caption{Schematic representation of the GRU architecture highlighting its reset and update gates.}
%     \label{fig:gru_architecture}
% \end{figure}

Studies such as \cite{10455879} have compared the performance of LSTM, BiLSTM, CNN-LSTM, Stacked-LSTM, and GRU in predicting wildfire risk in the Indonesian region of Kalimantan, finding that GRU slightly outperformed the other models, though the advantage is not significant. In contrast, \cite{gopu2023comparative} observed better results with LSTM in a comparative study of wildfire prediction in the Montesinho Natural Park, Portugal. Additionally, \cite{chavalithumrong_learning_2021} found that models integrating a dynamic autoencoder with RNN outperformed standard GRU models, suggesting that enhanced RNN architectures can offer superior performance.

\subsubsection{Transformers for Wildfire Risk Prediction}

Transformers utilize multi-head self-attention mechanisms to calculate attention scores, enabling them to capture dependencies at any position within the sequence \citep{vaswani2017attention}. The attention mechanism is defined as:
\begin{equation}
\text{Attention}(Q, K, V) = \text{softmax}\left(\frac{QK^T}{\sqrt{d_k}}\right)V 
\end{equation}
where \(d_k\) represents the dimensionality of the key. The architecture of a Transformer comprises encoders, decoders, positional encoding, and multiple attention mechanisms.

Transformers operate on sequences transformed into tokens, which are then embedded with positional encoding to maintain the sequence order absent in traditional RNNs. The sequence of embedded tokens is processed by the encoder, which transforms it into a context vector using layers of multi-head self-attention and feed-forward networks. The decoder similarly processes the context vector to predict the output sequence. The ability of Transformers to process sequence data without relying on sequential processing allows for parallel computation during training, significantly speeding up the learning process compared to RNNs \citep{vaswani2017attention}. This makes them particularly effective for long-sequence predictions or complex scenarios requiring modelling of long-range dependencies \citep{zerveas_transformer-based_2021, prapas_televit_2023}.

Transformers been adapted for several predictive tasks in wildfire risk prediction. For example, \cite{miao_time_2023} developed a Transformer model with a window-based attention mechanism to predict forest fire risks in Chongli District, Beijing, China, using time-series data of meteorological, topographical, vegetation, and anthropogenic factors. The window mechanism confines attention to a fixed-size window centered on each element, reducing computational complexity and memory usage. Similarly, \cite{cao2024forest} employed a Transformer to predict next-day wildfire risks in Quanzhou County, Guangxi Province, China, based on three days of meteorological, topographical, and human activity data. The model's performance, analyzed using IGR, outperformed LSTM, RNN, and SVM in terms of generalization, noise resistance, and overfitting mitigation.

% \begin{figure}[!h]
%     \centering
%     \includegraphics[width=1\linewidth]{images/transformer.png}
%     \caption{Schematic diagram of the Transformer architecture, highlighting its main components.}
%     \label{fig:Transformer_architecture}
% \end{figure}

From the analysis above, it is evident that both RNNs and GRUs are less frequently used in wildfire risk prediction, which includes forecasting occurrences, susceptibility, and spread. In contrast, LSTM networks have a relatively higher usage rate. RNNs, compared to LSTMs, are particularly challenged by issues such as vanishing and exploding gradients and exhibit weaker capabilities in modeling long-distance dependencies. Furthermore, although some studies argue that GRUs perform comparably to LSTMs \citep{viswanathan2019sequence, gao2020short, liu2021short, gao2021stock, zarzycki2022advanced}, the simplified structure of GRUs may result in inferior performance when processing highly complex sequences or capturing intricate relationships \citep{cahuantzi2023comparison}, especially in wildfire risk prediction tasks that do not require high real-time processing capabilities or are intended for edge computing devices. Recently, however, with the advent of Transformers, these models are gaining popularity in wildfire risk prediction due to their complex architecture, larger parameter capacity, and the advantages of parallel computation.

\subsection{Image Semantic Segmentation and Classification Methods}

Although RNNs and Transformers excel at capturing temporal dependencies for sequence-to-sequence learning tasks, these methods primarily classify individual pixels' temporal sequence data. However, wildfire occurrences are not only influenced by local conditions but also by neighboring areas (e.g., wind propagation). When RNNs use 1D data inputs, the spatial structure of the image is often lost \citep{8960629}. Consequently, several studies have explored the use of image semantic segmentation and classification methods, such as CNNs, GNNs, and Transformers, to predict wildfire risk at time $t$ using the influencing factors (e.g., fuel, weather) from time $t'<t$. These methods rely on 2D image inputs, which explicitly model the spatial dependencies of input features, while also implicitly modeling the influence of current driving factors on future wildfire risk by having different input features and predicted time steps. In wildfire risk prediction, CNNs are commonly used, with GNNs and Transformers increasingly adopted in recent studies.

\subsubsection{Convolutional Neural Networks}

\begin{figure}[!b]
    \centering
    \includegraphics[width=1\linewidth]{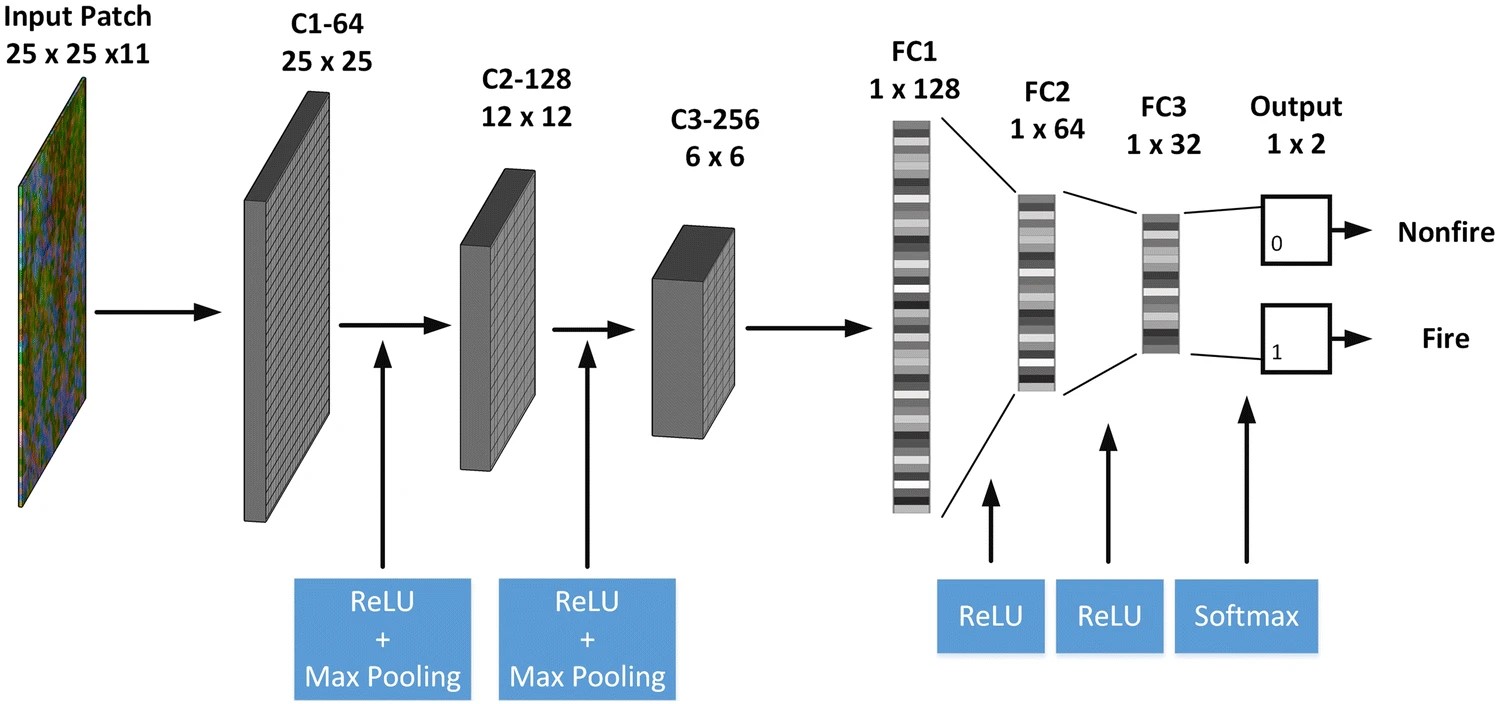}
    \caption{The architecture of CNN classification network where the center pixel is classified using surrounding information \citep{zhang_forest_2019}.}
    \label{fig:CNN_architecture}
\end{figure}

CNNs are foundational network structures in deep learning, designed to extract local spatial features through convolution operations, making them well-suited for capturing the spatial information necessary for wildfire risk prediction, as shown in Fig~\ref{fig:CNN_architecture}. For example, \cite{zhang_forest_2019} employed a modified AlexNet \citep{krizhevsky2012imagenet} using annual fire point maps and springtime mean influencing factors for Yunnan Province, China, to classify image patches as wildfire-prone areas. The model outputs a probability map of wildfire risk through softmax mapping in the final CNN layer. When comparing the CNN model's performance with RF, SVM, MLP, and kernel logistic regression models, the authors found that CNN achieved the highest overall accuracy. They also analyzed the contributions of various wildfire drivers using the information gain ratio, identifying temperature, wind speed, surface roughness, precipitation, and elevation as the most influential factors.

Building upon this, the authors extended the modified AlexNet to generate a global quarterly wildfire susceptibility map \citep{zhang_deep_2021}. They compared its performance with MLP-2D, CNN-1D, and MLP-1D models, finding that CNN-2D and MLP-1D models outperformed CNN-1D and MLP-2D. Additionally, they applied an improved permutation importance (PI) method and partial dependence plots (PDPs) to investigate the interpretability of the CNN-2D model. The PI analysis highlighted that maximum monthly temperature, soil temperature, NDVI, and soil moisture were significant factors affecting the model. Furthermore, PDP analysis revealed a negative relationship between cumulative precipitation and wildfire probability, while maximum monthly temperature and soil temperature positively influenced wildfire occurrence.

Moreover, using a similar CNN approach, they predicted the spatiotemporal variation of wildfire risk under different climate scenarios, discovering that global warming would lead to an increase in burned areas and a northward shift of wildfire-prone regions \citep{zhang2024current}. In another study, \cite{kanwal_data-driven_2023} explored CNN-1D and CNN-2D classification models along with other machine learning algorithms for evaluating seasonal wildfire risks in Pakistan, finding that CNN-2D outperformed CNN-1D and other machine learning algorithms.

In addition, some studies have employed deeper or structurally optimized CNN networks. For instance, \cite{oak2024novel} utilized VGG16 with ReLU, BatchNorm, Dropout, and a sigmoid layer to predict wildfire susceptibility in Quebec, Canada. Comparative experiments demonstrated that this model outperformed other architectures like Xception, VGG16, ResNet50, and Inception V3. Similarly, \cite{nur2022creation} applied a meta-heuristic optimization algorithm to fine-tune the hyperparameters of a CNN model for wildfire susceptibility mapping in the Plumas National Forest, USA. Their IGR analysis revealed that land use, drought index, and maximum temperature had the greatest impact on wildfire susceptibility, and the model’s performance improved significantly after optimization. 

Beyond classification-based methods, some studies have implemented end-to-end models for higher-resolution predictions. For example, \cite{hodges2019wildland} proposed a deep convolutional inverse graphic network comprising convolution and deconvolution layers to simulate wildfire spread over homogeneous and heterogeneous landscapes for 6-hour and 24-hour intervals, demonstrating the high efficiency of CNNs in wildfire spread prediction. Similarly, \cite{santopaolo2021forest} employed an end-to-end CNN-based image segmentation model to predict wildfire risk in Sicily, Italy, and California, USA, using the average values of driving factors over the preceding eight days.

Furthermore, \cite{allaire2021emulation} developed a hybrid architecture combining convolutional and fully connected layers to emulate the output of a numerical fire spread simulator based on the Rothermel model of fire propagation. Similarly, \cite{bergado_predicting_2021} designed an AllConvNet, a fully convolutional network, to predict wildfire burn probability for the next 7 days in Victoria, Australia. This model achieved higher prediction accuracy compared to logistic regression, SegNet, and MLP models. Feature extraction and normalization of logistic regression coefficients revealed that total rainfall, lightning density, and surface temperature were among the most influential variables.

Additionally, \cite{li2023predicting} introduced spatial and channel attention mechanisms to enhance an end-to-end CNN model’s focus on critical features while suppressing irrelevant ones for predicting the next moment's wildfire spread. This model also incorporated the burning map from the previous time step, outperforming other CNN models and ConvLSTM. Similarly, \cite{shadrin_wildfire_2024} used the MA-Net segmentation model, which includes position attention and multi-scale attention blocks, to predict wildfire extent for the next 1-5 days. Unlike previous studies, they included short-term weather forecasts along with various wildfire drivers. Experimental results showed that their model accurately predicted wildfire spread within 3 days, with the wind components, land cover, NDVI, EVI, and LAI being the most influential factors.

Lastly, \cite{marjani2024application} combined CNN with atrous spatial pyramid pooling to model wildfires at different scales and shapes, adopting an end-to-end approach for next-day wildfire spread prediction. They used linear regression and Grad-CAM heatmap correlations to demonstrate how different dilation rates help extract both large-scale and small-scale wildfires. Similarly, \cite{jiang2023wfnet} employed a three-branch network to process static DEM and fuel data, dynamic wind speed, temperature, humidity data, and the previous time step's wildfire status. Their experiments showed that this model’s inference efficiency significantly surpassed that of cellular automata and Farsite models, accurately predicting burned areas in real wildfire scenarios.

\subsubsection{Other Methods}

In addition to CNNs, other image segmentation methods, such as GNNs and Transformers, have also been applied to wildfire risk prediction. GNNs, compared to CNNs, are better suited for handling non-Euclidean structured data \citep{9395439}, while Transformers' multi-head self-attention mechanisms are more effective in modeling long-range semantic dependencies within images \citep{vaswani2017attention}. For example, \cite{jiang2022modeling} proposed using an irregular graph network to simulate wildfire spread in variable-scale landscapes, addressing the issue of fixed-grid data being unable to adaptively differentiate landscape heterogeneity. The model adjusts graph node density based on terrain complexity to achieve uniformity in nodes and edges at varying scales. Comparative experiments with cellular automata and FARSITE demonstrated that the proposed model exhibits competitive accuracy and efficiency.

Additionally, \cite{chen2022knowledge} incorporated a GNN into the knowledge graph framework proposed by \cite{ge2022spatio} to enable automated learning and prediction of burn areas. Specifically, the model first constructs a knowledge graph in triples and uses RotaE to compute representation vectors for each entity. These entity vectors are associated with nodes in the graph, and the vectors are subsequently fed into the GNN to facilitate automatic updates. Finally, link prediction algorithms are employed to predict the burn areas associated with the recorded nodes. Experiments conducted in Portugal's Montesinho Natural Park demonstrated that this algorithm significantly outperformed traditional machine learning algorithms such as RF, SVM, and MLP.

Moreover, \cite{li2024wildfire} introduced spatial attention mechanisms and focal modulation into a U-shaped encoder-decoder network based on Transformers, known as Swin Unet. This end-to-end image segmentation approach utilizes wildfire risk driving factors from the previous day to predict wildfire risk for the following day across the continental United States. Comparative experiments using the extended Next Day Wildfire Spread dataset \citep{huot2022next} showed that this model outperforms baseline models, including the standard UNet and Swin Unet. In addition, \cite{prapas_televit_2023} introduced a TeleViT model that incorporates teleconnections to model long-range spatial interactions globally, aiming to predict subseasonal to seasonal wildfire patterns. This model combines fine-grained local-scale inputs with coarse-grained global-scale data, improving prediction accuracy for global wildfire patterns up to four months in advance.

While deep learning-based wildfire risk prediction methods often overlap in their reliance on similar wildfire driving data, they do not typically distinguish between wildfire spread simulation and broader wildfire risk prediction, as defined in Section \ref{Wildfire Ignition and Burning Risk} of this manuscript. Wildfire spread simulation models are generally tailored to specific scenarios, such as predicting the dynamics of fire spread from known or predefined ignition points in localized regions like parks or mountainous areas \citep{kondylatos2023mesogeos, jiang2023wfnet, li2023predicting, marjani2024application}. In contrast, wildfire risk prediction models aim to identify areas with potential ignition points or high susceptibility to fire spread, explicitly incorporating the stochastic nature of ignition events. This approach allows for a comprehensive assessment of fire risk by considering both ignition probability and subsequent spread potential. Moreover, wildfire risk prediction accommodates a wider range of spatial scales, from local parks to provincial, national, continental, or even global levels \citep{santopaolo2021forest, bergado_predicting_2021, chen2022knowledge, prapas_televit_2023}. This adaptability, coupled with the ability to model stochastic ignition events, highlights the broader applicability and versatility of risk prediction models for large-scale wildfire management.

Additionally, the training process for deep learning-based spread simulation models often involves datasets of discrete historical wildfire events combined with associated driving data, which can limit generalization across diverse wildfire scenarios \citep{huot2022next, prapas_2022_6475592, kondylatos2023mesogeos}. Consequently, while spread simulation methods offer precise fire spread modeling for individual events, their predictive performance tends to be less robust when applied to large-scale wildfire risk assessments \citep{gerard2023wildfirespreadts}. This distinction underlines the need for improved methodologies explicitly designed for large-scale and long-term wildfire risk prediction, ensuring scalability and reliability across varying geographies and timeframes.

\subsection{Spatiotemporal Prediction Methods}

The previously discussed time series prediction methods and image segmentation/classification methods each have their strengths in capturing temporal and spatial dependencies, respectively. However, they are unable to simultaneously handle both spatial and temporal dependencies and capture the complex cross-spatiotemporal interaction patterns. For instance, \cite{prapas2021deep} pointed out that focusing only on spatial scale modeling could overestimate a model's real-world performance. To integrate spatial structure and time series data, many studies have attempted to combine CNN, GNN, and Transformer models with RNN, LSTM, and GRU methods. This transforms the task of time series prediction or spatially localized optimization between consecutive wildfire states \citep{jiang2023wfnet} into spatiotemporal sequence prediction tasks. There are three primary approaches to achieving this combination: separated spatial and temporal modeling, coupled spatiotemporal modeling, and spatially enhanced coupled spatiotemporal modeling.

Specifically, separated spatial and temporal modeling involves abstracting spatial features into 1D feature vectors using spatial modeling methods, which are then input into temporal prediction models. Coupled spatiotemporal modeling converts the 1D tensors in RNNs into 3D tensors, incorporating spatial dimensions (rows and columns) and using convolutions or graphs (instead of weight matrix multiplication) to determine a grid cell's future state based on its neighbors' current and past states. Lastly, spatially enhanced coupled spatiotemporal modeling introduces CNNs or graph-based methods to extract spatial features before the spatiotemporal modeling stage.

\subsubsection{Separated Spatial and Temporal Modeling Methods}

\begin{figure}[h]
    \centering
    \includegraphics[width=1\linewidth]{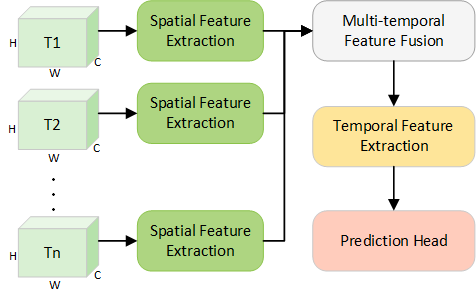}
    \caption{The diagram of the separated spatial and temporal modeling method.}
    \label{fig:separated_st_architecture}
\end{figure}

The approach of separately extracting spatial and temporal features is widely adopted in wildfire risk prediction. A typical architecture of such a spatiotemporal model is illustrated in Fig.~\ref{fig:separated_st_architecture}. The model takes as input spatial data composed of multiple driving factors across \( n \) time steps (\( T_1 \) through \( T_n \)), represented as tensors with dimensions height (\( H \)), width (\( W \)), and number of drivers (\( C \)). Each temporal slice is processed by parallel spatial feature extraction modules, either with shared or independent weights. The resulting spatial features are then aggregated through a multi-temporal feature fusion stage, during which the original spatial resolution \( H \times W \) is typically reduced to a global representation (e.g., \( 1 \times 1 \)). This fused representation is then passed through a temporal feature extraction module to capture dynamic patterns across the time sequence. Finally, the integrated spatiotemporal features are input into a prediction head to produce the final wildfire risk assessment.

Typically, multiple CNNs are employed to process multidimensional spatial images at each time step, with dimensionality reduction applied before feeding the data into the LSTM. Finally, the output from the LSTM is mapped to output probabilities using a CNN, or directly without it, along with the application of a sigmoid function. This architecture effectively leverages both the spatial pattern recognition capabilities of CNNs and the sequential modeling strengths of LSTMs to capture the complex spatiotemporal dynamics inherent in wildfire systems.

This method is widely used in wildfire risk prediction. For example, \cite{jin2020ufsp} proposed enhancing spatial feature extraction by employing CNNs and GCNs to extract spatial information, which is then passed through a feed-forward network to unify the data into a 1D vector before being input into a GRU. The output 1D vector is then restored to its original size using deconvolution layers, performing urban fire situation prediction through an end-to-end semantic segmentation approach. Compared to traditional machine learning methods like RF and XGBoost, as well as GRUN and Conv-GRUN, this model showed significant performance improvements.

Similarly, \cite{huot_deep_2021} used the Next Day Wildfire Spread dataset and categorized wildfire risk prediction into two tasks: next-day wildfire probability prediction and final burning area prediction. They experimented with four models: convolutional autoencoder and residual UNet for segmentation, and convolutional autoencoder and residual UNet combined with LSTM for temporal prediction. The results showed that, among the segmentation models, the convolutional autoencoder outperformed UNet. When combining the two models with LSTM for final burning area prediction, there was no significant performance difference between the models.

\cite{zhang2022dynamic} designed and compared four deep neural networks with identical structures but different depths, combining 2D CNN and LSTM models to predict global monthly burned area maps using seven meteorological and two biophysical predictors from the previous 12 months. Each model reduced spatial dimensions through convolution and max-pooling operations and used fully connected layers to flatten the input patches into 1D features before feeding them into LSTM layers. The primary difference among the models was the number of CNN and LSTM layers. Ultimately, the CNN-LSTM network with two convolutional layers and two LSTM layers was selected for its balance of performance and efficiency. The results showed that the combination of convolution and LSTM methods outperformed standalone LSTM and CNN models.

Furthermore, \cite{marjani2024cnn} combined CNN with bidirectional LSTM (BiLSTM) for a study in Australia, using wildfire driving factors from the previous four days and wildfire masks to predict next-day wildfire spread. The method first employed four CNN layers to extract spatial features, which were then flattened and input into a three-layer BiLSTM. The model produced binary outputs through dilated convolution, dense layers, and sigmoid activation functions. Comparative experiments demonstrated that this approach outperformed LSTM and Conv-LSTM methods, and the study also confirmed that using longer time steps improved prediction capabilities. In another study, \cite{li_wildland_2021} not only introduced a CNN before the LSTM but also incorporated an attention mechanism within the LSTM to assign different weights to the hidden states, reducing information loss and improving prediction accuracy. Applied in Portugal's Montesinho Natural Park, the model used various meteorological factors from the FWI, historical burned areas, and time and location data to predict future wildfire risk. Comparisons with SVM, XGBoost, neural networks, RNN, and basic LSTM showed that deep learning models significantly outperformed machine learning models, with LSTM surpassing RNN and the CNN-attention LSTM performing best overall.

\cite{rosch2024data} applied graph convolution combined with GRU to develop Spatiotemporal Graph Neural Networks (ST-GNNs) for predicting wildfire spread in Portugal and the Mediterranean region. These models used an H3 grid system built from weather data, land cover, fuel types, DEM, and fire weather indices. However, the models exhibited high false positive rates.

In another study, \cite{bhowmik_multi-modal_2023} proposed a U-shaped LSTM network combining UNet and LSTM, using seven days of meteorological, environmental, vegetation, and geological data to predict wildfire risk. The U-shaped LSTM network first used a three-layer UNet encoder to extract semantic information, which was then flattened and input into an LSTM. The LSTM output was passed through a UNet decoder to produce end-to-end pixel-level predictions of wildfire heatmaps for the next 24 hours to two weeks. This model achieved over 97\% accuracy in predicting large-scale wildfires in California, outperforming CNN models.

Similarly, \cite{10029946_2022} conducted research on wildfire risk prediction in the western United States, using a convolutional encoder to reduce the dimensionality of input time series data into vectors, which were then fed into a GRU model. A dual deconvolution decoder was designed to restore image size, predicting wildfire risk at different time steps (one, two, three, and four weeks). Compared to logistic regression, generative networks, and GRU, the combination of CNN and GRU showed significant performance improvement.

Lastly, \cite{marjani2023firepred} developed a spatiotemporal prediction model combining CNNs and RNNs to predict wildfire spread at time step $t+1$, using environmental variables and burned areas from time steps $t$ and $t-1$. Data were divided into three blocks: hourly block (6-hour averaged wind speed and direction), daily block (temperature, precipitation, and wildfire burning map), and constant variables (land cover type, population, terrain, etc.). Each block was processed by CNNs and flattened into 1D features, then fed into RNNs with 64 and 31 neurons, respectively. Constant variables were similarly processed and flattened. The RNN outputs were concatenated with the constant variables and passed through convolution, reshape, dropout, and sigmoid layers to produce a burning probability map. Comparative results showed the superiority of this spatiotemporal modeling approach over CNN and the deep convolutional inverse graphics network proposed by \cite{hodges2019wildland}. Additionally, experiments assessing data availability, noise addition, and Monte Carlo dropout were conducted to evaluate model uncertainty.

\subsubsection{Coupled Spatiotemporal Modeling Methods}

Compared to the separated spatial and temporal modeling methods, time-series prediction algorithms like ConvLSTM are also widely used. For instance, \cite{burge2021convolutionallstmneuralnetworks} evaluated the feasibility of using ConvLSTM to simulate wildfire propagation dynamics on three simulated datasets with varying terrain, wind, humidity, and vegetation density distributions. The experiments demonstrated that ConvLSTM's predictive capability surpassed that of CNN. In the study by \cite{prapas2021deep}, which compared traditional machine learning models (RF), time series prediction algorithms (LSTM), image classification methods (CNN), and spatiotemporal modeling methods (ConvLSTM) for next-day wildfire danger prediction across Greece, several key findings emerged. For the RF model, training data was constructed by averaging each driving factor for the 10 days preceding the prediction date at each spatial location, resulting in a dataset with dimensions \(f \times 1\), where \(f\) is the number of driving factors. The LSTM model used time series input, with data dimensions of \(f \times t\), where \(t\) represents the time series length (10 days). For the image classification task, the spatial data was constructed as spatial patches centered on the target point, with dimensions \(f \times h \times w\), where \(h\) and \(w\) are the height and width of the image. Lastly, for the spatiotemporal prediction task using ConvLSTM, the dataset dimensions were \(f \times h \times w \times t\). The results showed that LSTM and ConvLSTM performed better than RF and CNN. Evaluating the binary classification results, LSTM outperformed ConvLSTM overall, but when using the area under the ROC curve as the evaluation metric, ConvLSTM slightly surpassed LSTM, with AUC scores of 0.926 and 0.920, respectively.

Similarly, \cite{khalaf2024performance} compared ConvLSTM with cellular automata and FlamMap in simulating wildfire spread in Iran's Golestan National Park. The results indicated that ConvLSTM had the highest accuracy in predicting burned areas, but for spread rate evaluation, the CA algorithm performed better than the other models. In another study, \cite{michail_seasonal_2024} compared temporally-enabled GNN, GRU, and ConvLSTM using the SeasFire Datacube \citep{alonso_2023_8055879}, finding that GNN was better at capturing global information, thus improving the understanding of complex patterns. The study also emphasized the importance of long time series information and a large receptive field in global wildfire risk prediction, especially for seasonal forecasts. It showed that longer input sequences provided more reliable predictions, and integrating spatial information to capture the spatiotemporal dynamics of wildfires improved model performance. Therefore, to improve predictions over longer forecast horizons, it is essential to consider a larger spatial receptive field.

\cite{MASRUR2023119, masrur2024capturing} further proposed two ConvLSTM models based on spatiotemporal attention to predict wildfire spread over the next 10 days using the previous 10 days' NDVI, wind components, and soil moisture. These models were designed to capture local-to-global and short-to-long-term spatiotemporal dependencies. The models were trained and tested on both simulated and real-world datasets. Pairwise self-attention was used to calculate attention scores for input variables, while patchwise self-attention replaced convolution operations in ConvLSTM. Comparisons on simulated datasets showed that the pairwise self-attention model performed better, while the standard ConvLSTM outperformed the patchwise self-attention model. Moreover, they employed integrated gradients \citep{sundararajan2017axiomatic} to enhance the interpretability of the contributions of time steps and driving factors to the model's predictions. The spatial patterns from integrated gradients confirmed the importance of capturing both local and global spatial dependencies. However, in real-world scenarios, replacing ConvLSTM's convolution operations with patchwise self-attention significantly improved the standard ConvLSTM's prediction accuracy and model transferability, surpassing even the pairwise self-attention ConvLSTM in both aspects.

Finally, \cite{eddin2023location} designed two separate convolutional branches to process dynamic variables (those changing over time) and static variables (those remaining constant throughout the study period). To address the causal effects of static variables on dynamic ones, they incorporated a location-aware adaptive normalization layer that adjusts the activation maps in the dynamic branch based on static features. They also employed sinusoidal encoding of Julian days to provide the model with explicit temporal information. Their experiments on the FireCube dataset \citep{prapas_2022_6475592, kondylatos2023mesogeos}, which covers parts of the Mediterranean, demonstrated that the proposed model's predictions of wildfire spread at time \( t+1 \) outperformed baseline models, including LSTM, ConvLSTM, RF, and XGBoost. In a similar vein, \cite{dong2024deep} proposed an architecture that models dynamic and static variables separately during the encoding phase, leading to improved accuracy in predicting fire radiative power.

\subsubsection{Spatially Enhanced Coupled Spatiotemporal Modeling Methods}

\begin{figure}[!h]
    \centering
    \includegraphics[width=1\linewidth]{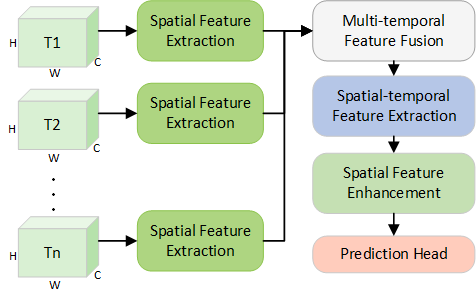}
    \caption{A typical spatial feature-enhanced coupled spatiotemporal wildfire risk prediction model diagram.}
    \label{fig:spatially_enhaced}
\end{figure}

In addition, there are studies that combine convolutional layers with ConvLSTM to further emphasize spatial feature extraction. Similar to the separated spatial and temporal sequence wildfire risk prediction models, the spatially enhanced coupled spatiotemporal wildfire risk prediction method, as shown in Fig.~\ref{fig:spatially_enhaced}, first extracts spatial features using methods such as convolution operations. Dimensionality reduction is then applied before feeding the data into a temporal model, such as ConvLSTM. The output of the temporal model can then be processed by a convolutional block followed by a sigmoid mapping, or directly mapped using a sigmoid function, depending on the specific design. Compared to the separated spatial and temporal modeling approach, this method tends to preserve a certain level of spatial information throughout the modeling process, which may help retain finer-grained spatial patterns in the final prediction.

For example, \cite{deng2023wildfire} used terrain, vegetation, climate, and human activity factors in Yunnan Province, China, to build a 3D CNN and ConvLSTM hybrid model for next-day wildfire prediction. During data preprocessing, Pearson correlation analysis was used to exclude the minimum and average temperatures from the dataset. Variance inflation factors and tolerance were then employed to assess multicollinearity among wildfire driving factors. Information gain ratio analysis revealed that precipitation and slope had the highest impact. In the proposed model, 3D CNN was primarily responsible for extracting spatial features and reducing the input patch's spatial dimensions, which were then fed into a many-to-many ConvLSTM for prediction. The results showed that the combined 3D CNN and ConvLSTM model outperformed both standalone ConvLSTM and 3D CNN models.

Similarly, \cite{burge2023recurrent} built upon the model proposed by \cite{hodges2019wildland} by introducing fuel embedding layers and inserting ConvLSTM between the encoder and decoder for end-to-end wildfire spread simulation. The experimental results demonstrated that incorporating LSTM improved prediction accuracy.

Furthermore, \cite{ji2024global}, following a similar branch structure to \cite{eddin2023location} and \cite{jiang2023wfnet}, proposed a static location-aware model based on ConvLSTM. This model aimed to integrate global static features (sine and cosine values of longitude and latitude, along with climate variables) with dynamic variables (e.g., meteorological, human, and vegetation cover factors) to predict global wildfire probability on sub-seasonal to seasonal scales. In this model, global static features were first convolved and max-pooled, and the resulting feature maps were multiplied with climate variables for data fusion. SKNet was then employed to extract higher-order features, which were upsampled and merged with the static features generated by the encoder. Finally, upsampling and skip connections were used to restore the image size and supplement the spatial details of global static location variables. The dynamic and static variables were concatenated and input into ConvLSTM. The hidden state output from the last time step of ConvLSTM was combined with the static and global features, and a \(1 \times 1\) convolution was applied to produce the final wildfire probability prediction. Comparing accuracy, recall, F1 score, precision, and Kappa coefficient with different variants of Vision Transformer and ConvLSTM demonstrated the effectiveness of incorporating global location information and climate variables for enhancing predictive capabilities.

Similarly, \cite{he2024deep} developed a next-day wildfire risk prediction model for eastern China by combining CNN, ConvLSTM, and Vision Transformer models, also employing a branch structure. In this model, eight CNN modules were used to model eight meteorological variables, while another CNN module modeled terrain, vegetation, and anthropogenic factors from the last day of each time step. The sequence feature maps generated by the temporal meteorological variables were input into LSTM, and its output was merged with the other variables' CNN outputs. Channel attention, spatial attention, and CNN were used to extract higher-order features. To emphasize the high-resolution structural features of vegetation and scattered small settlements, Landsat data were used to generate 30-meter monthly high-resolution NDVI products, and a Vision Transformer was applied to extract these structural features. The flattened outputs from the Vision Transformer and CNN were concatenated and classified using an ANN binary classifier. Ablation studies demonstrated the effectiveness of incorporating attention mechanisms, high-resolution spatial structure features, and Vision Transformer. Sensitivity analysis also confirmed that meteorological factors were the most important among wildfire drivers.

Finally, \cite{chen2024explainable} designed an end-to-end encoder-decoder network structure similar to that of \cite{bhowmik_multi-modal_2023} and \cite{10029946_2022}, combining GCN and ConvLSTM. The GCN component was introduced to model the long-distance spatial dependencies required for global wildfire risk prediction, while mitigating the impact of missing data from ocean regions. Explicitly modeling information flow through the graph edges improved the interpretability of the prediction results. Specifically, the Earth’s land surface pixels, as simulated by the JULES-INFERNO model, were treated as graph nodes, and the correlation coefficients between nodes were used to establish the graph’s edges. Spatial dependencies were modeled using a spatial graph convolution model, followed by LSTM for long-term temporal dependencies. The model was trained and validated using 30 years of simulated data from 1961 to 1990 in the JULES-INFERNO dataset, and its performance was compared against LSTM, ConvLSTM, and convolutional autoencoder with LSTM. The proposed model demonstrated superior performance, with ConvLSTM being the best among the comparative models. Additionally, the study used the Louvain method to identify potential wildfire communities, and integrated gradient analysis revealed that short-term (one-month) wildfire risk is primarily driven by lightning, while long-term (11-month) risk is more influenced by temperature, followed by lightning. Node importance analysis also showed that even geographically distant locations could have strong interactions regarding wildfire risk. Similarly, \cite{zhao2024causalgraphneuralnetworks} aimed to improve model interpretability by combining causal GNN with LSTM, comparing its performance with LSTM and GRU in predicting future wildfire risk.

This section reviews recent advancements in deep learning methods applied to wildfire risk prediction, emphasizing their evolution from temporal and spatial modeling to spatiotemporal sequence modeling and the integration of emerging technologies. Due to their hierarchical structures and flexible architectures \citep{lecun2015deep}, deep learning models excel at capturing nonlinear relationships, enabling them to model the complex interactions and latent patterns within high-dimensional spatiotemporal data sourced from meteorological variables, vegetation indices, topographical features, and socioeconomic factors. Unlike traditional statistical models that rely on explicit feature engineering and linear assumptions, deep learning models uncover intricate dependencies that are difficult to detect using linear or shallow frameworks. This capability is particularly crucial in wildfire risk prediction, where the interdependencies among driving factors often exhibit nonlinear behaviors that vary dynamically across both temporal and spatial scales \citep{jain_review_2020}.

The evolution of deep learning models in this domain demonstrates significant strides in both predictive accuracy and computational efficiency \citep{cheng2024deep}. Early applications relied on RNNs and LSTM networks to process wildfire time series data, while CNNs and GCNs were used to extract spatial features. Recent innovations have merged these approaches into hybrid spatiotemporal models, enabling simultaneous analysis of spatial and temporal dependencies. These advancements have been accompanied by the introduction of unique design elements tailored to address specific challenges in wildfire risk prediction. For example, dynamic-static feature separation enables models to distinguish between temporally varying inputs, such as meteorological variables, and static attributes, such as topography and land cover, improving prediction efficiency \citep{eddin2023location, shadrin_wildfire_2024, jiang2023wfnet, dong2024deep}. Explicit modeling of temporal and spatial dependencies, often through attention mechanisms and spatiotemporal encoding, allows these models to capture complex interactions across time steps or spatially distant regions, thereby enhancing their ability to predict wildfire propagation over large areas \citep{ji2024global}. Additionally, graph-based representations of geographical regions, where nodes correspond to areas and edges represent interactions such as wind patterns or vegetation continuity, provide a new framework for modeling long-range dependencies and spatial interactions inherent in wildfire dynamics \citep{jiang2022modeling, chen2022knowledge, chen2024explainable}.

\section{Promising Deep Learning Methods and Data}

In recent years, the model architectures and the types and quantity of driving features employed in wildfire risk prediction have largely stabilized. From a modeling perspective, given the complex nonlinear and spatiotemporal relationships between various driving factors and wildfire occurrence and spread, a key trend in deep learning-based wildfire prediction is enhancing the capability to model long-range spatiotemporal dependencies. This aims to capture subtle and weak correlations effectively. More specifically, wildfire risk prediction has evolved from purely temporal sequence forecasting to spatiotemporal sequence prediction and spatially enhanced spatiotemporal modeling. These advancements focus on increasing both the temporal and spatial scales of the model, enabling the incorporation of higher-resolution data over longer time sequences while expanding the dataset size. Current research has explored wildfire risk prediction models based on RNNs, CNNs, transformers, and GNNs.

However, compared to more recent models such as Mamba \citep{gu_mamba_2023}, these approaches still suffer from inefficiencies or limited capability in modeling large-scale spatiotemporal relationships. Additionally, many of these models rely on multi-branch networks that primarily use image-based or node-based representations to capture long-range semantic dependencies, without exploring the potential efficiency gains offered by multimodal approaches, such as incorporating textual descriptions. Furthermore, the applicability of cross-disciplinary solutions in addressing the inherent challenges of existing wildfire risk prediction models—such as data dependency, poor interpretability, difficulty in uncertainty quantification, and challenges in handling stochasticity, multi-source heterogeneous data, and complex factor interactions—remains an open question.

From a data perspective, large-scale wildfire risk prediction studies primarily prioritize data availability, commonality, and ease of processing. This has resulted in an over-reliance on meteorological data from various sources, particularly ground station measurements, ERA-5 reanalysis data, and multiple MODIS-derived products. While MODIS provides high temporal resolution and long-term large-scale observations, its passive optical remote sensing capabilities are inherently limited in penetrating vegetation canopies. Consequently, MODIS-based assessments of fuel conditions are often imprecise, and the sensor’s inability to capture data under all-weather, all-day conditions significantly constrains the applicability of wildfire risk prediction models trained on such data. A key question that remains is whether alternative data sources can improve the accuracy of wildfire risk prediction, particularly given the fundamental reality that data acquisition capabilities in Earth sciences far exceed current data processing capacities \citep{karniadakis2021physics}.

In response to these challenges, this section systematically reviews recent advancements in deep learning methods—particularly those employing multimodal approaches—for segmentation, classification, and prediction tasks. Additionally, it provides a brief overview of the Mamba model, a parameter-efficient deep learning framework, to facilitate the development of computationally efficient wildfire risk prediction models in future research. Given the similarities between meteorological forecasting and wildfire risk prediction—both of which involve multiple driving factors, nonlinear coupling, and dynamic spatiotemporal modeling—this section also summarizes recent developments in meteorological forecasting models. The goal is to explore whether cross-disciplinary methodologies can enhance deep learning-based wildfire risk prediction by introducing physical constraints, improving uncertainty quantification, and incorporating lightweight model designs to enhance reliability and accuracy. Furthermore, this section discusses underutilized remote sensing data types in wildfire risk prediction, particularly microwave and LiDAR data, which offer penetrative capabilities, as well as high-spatial-resolution nighttime light data, which can enable dynamic nighttime wildfire monitoring. These discussions aim to provide a broader data reference for future wildfire risk research.

\subsection{Promising Deep Learning Models}

\subsubsection{Multimodal Models}

Multimodal learning is a universal approach to building AI models, enabling the extraction and association of information from various sources, such as different sensors \citep{baltruvsaitis2018multimodal}. Unlike unimodal image-based learning, multimodal learning can acquire, interpret, and reason across multiple information sources, resembling human perception \citep{10123038}. A key advantage of multimodal learning over unimodal approaches is its ability to more easily embed domain knowledge or abstract, knowledge-driven perspectives \citep{baltruvsaitis2018multimodal}. Currently, multimodal learning has been extensively studied in image segmentation and classification \citep{10153685, 10147258, nasir2023multi, Chng_2024_CVPR, Wang_2024_CVPR, Xu_2023_ICCV}, data fusion \citep{khan2023multimodal, kalamkar2023multimodal}, generation, and visual-language pretraining models \citep{liu2023grounding, Li_2023_CVPR, oquab2024dinov2learningrobustvisual}. Given that this research focuses on exploring the potential applications of segmentation and classification methods in wildfire risk prediction, we will delve into Referring Image Segmentation (RIS) in subsequent sections.

RIS refers to models that understand natural language descriptions and correspond them to specific objects in images at the pixel level, outputting an object mask \citep{10460426}. RIS requires processing both text and visual modalities, with the latter typically being high-dimensional and exhibiting strong spatial characteristics encapsulating attributes such as color, texture, shape, and size. In contrast, natural language is generally low-dimensional and continuous, abstractly describing these visual features \citep{ji2024survey}. The key challenge of RIS lies in designing appropriate multimodal fusion methods to establish a relationship between language and visual modalities, thereby bridging the modality gap between the two \citep{Chng_2024_CVPR}. Generally, fusion methods can be categorized into early fusion and late fusion, commonly known as decoder-fusion and encoder-fusion, respectively \citep{10460426}.

Decoder fusion extracts visual and language features separately using different encoders, such as CNNs for visual data and LSTMs for language data, and designs specific fusion modules to combine these features at the decoder stage. For example, \cite{jain-gandhi-2022-comprehensive} utilized CNNs to extract multi-stage image features and LSTMs for text feature extraction, integrating these features through a hierarchical cross-modal aggregation module. In another approach, \cite{9878961} designed a knowledge transfer method to adapt image-level knowledge from Contrastive Language-Image Pretraining (CLIP) into RIS for pixel-level segmentation tasks, implementing a Neck module to fuse outputs from both the text and image encoders. Similarly, \cite{yan2023mmnetmultimasknetworkreferring} introduced a Fusion Neck module to combine image and text encoding information, alongside a multi-query generator module to aid segmentation.

Encoder fusion, in contrast, integrates multimodal features at the encoder stage using auxiliary losses or alignment modules \citep{Chng_2024_CVPR}. This method is considered more effective \citep{Feng_2021_CVPR, Yang_2022_CVPR} and has gained traction in recent studies \citep{10460426}. For example, \cite{Xu_2023_ICCV} introduced a Bridger module, based on self-attention and cross-attention mechanisms, to facilitate cross-modal information exchange between each level of the text encoder and image encoder. Furthermore, the study designed a task decoder with hierarchical and global alignment to further fuse multimodal information and generate the mask prediction. Similarly, \cite{10458079} employed BERT \citep{devlin-etal-2019-bert} for text feature extraction and Swin Transformer \citep{Liu_2021_ICCV} for image feature extraction, combining them with a self-attention-based language-guided cross-scale enhancer and a pixel-word attention module for feature fusion. Moreover, \cite{Shah_2024_CVPR} incorporated a Gaussian Enhancement Module between textual and visual encoders to extract both local and global visual-language relationships, enhancing overall representation. In a similar vein, \cite{Chng_2024_CVPR} designed a Cross-modal Alignment Module between multiple layers of text and image encoders, addressing the granularity mismatch between the two modalities through bidirectional interactions. Additionally, they introduced a mask encoder branch and a Transformer-based masked token decoder to predict the mask.

In summary, multimodal tasks offer greater flexibility and intuitive interaction by incorporating natural language to guide the model, compared to unimodal tasks. Multimodal information provides additional contextual data that unimodal models often lack, allowing the language encoder to construct a relation graph for numerical geospatial relations \citep{10458079}. Therefore, multimodal models, especially RIS, hold significant potential in wildfire risk prediction by capturing subtle spatiotemporal semantic dependencies that are difficult to model using existing methods.

\subsubsection{Efficient Deep Learning Models}

The aforementioned deep learning-based methods for spatial, temporal, and spatiotemporal modeling face challenges in capturing long-term and spatial dependencies. As a result, when these methods are applied to wildfire risk prediction on a regional scale, they often neglect the influence of climate factors outside the study area. Furthermore, global-scale wildfire risk prediction is limited by computational capacity, resulting in relatively low spatial resolution in the predictions. Additionally, whether modeling wildfire risk on a regional or global scale, the time series of driving factors considered is often short (a few days) or constructed using monthly or yearly averages for long-term sequences. This calls for more efficient models capable of handling longer temporal scales and higher spatial resolution.

One potential solution is Mamba: linear time sequence modeling with selective state spaces \citep{gu_mamba_2023}, a parameter-efficient model designed to handle long sequence modeling \citep{xu2024visualmambasurveynew} such as videos or remote sensing imagery over long temporal scales. Mamba's proposal is primarily aimed at addressing the computational complexity found in prevalent transformer-based networks. Specifically, the self-attention mechanism in transformers involves a computational complexity of $O(n^2)$, where $n$ represents the sequence length. Mamba reduces this complexity from $O(n^2)$ to $O(n)$. Compared to the Transformer network, which uses the self-attention mechanism, the RNN—a previously dominant approach to sequence modeling—suffers from short memory when handling long sequences. Moreover, while RNNs are efficient in the inference stage, they are less efficient during the training phase. Mamba is built on top of the State Space Model (SSM), which, similar to RNNs, is used for sequence modeling. The state-space model can be represented as:

\begin{align}
\text{State Equation:} \quad & x_t = A x_{t-1} + B u_t + w_t
\label{SSM_state}
\\
\text{Observation Equation:} \quad & y_t = C x_t + D u_t + v_t
\label{SSM_obser}
\end{align}
Where $x_t$ is the state vector at time $t$, $u_t$ is the control/input vector, $y_t$ is the observation/output vector, $A$ is the state transition matrix, $B$ is the input matrix, $C$ is the observation matrix, $D$ is the direct transmission matrix (optional, sometimes set to 0), and $w_t$ and $v_t$ are process and observation noise, respectively, often assumed to be Gaussian noise and were often removed from most SSM equations.

By revising the vanilla State Space Model (SSM), as shown in Eq. \ref{SSM_state} and \ref{SSM_obser}, the Structured State Space Model for Sequences (S4) was proposed by \cite{gu2021efficiently}. Specifically, since the data used in language and image processing are discrete, the SSM needed to be adapted into a discretized version. The zero-order hold technique was applied to discretize the SSM. Furthermore, the SSM equations were reformulated into a combination of recurrent and convolutional structures. This allows the S4 models to be trained in a CNN manner while being tested in a RNN manner, thus improving efficiency in both the training and inference stages.

To address the short memory issue typically found in RNNs, S4 employs the HiPPO matrix to replace matrix $A$ in Eq. \ref{SSM_state}. This modification enables the model to store long sequence information in a compact and sparse matrix format. However, one limitation of the S4 models is that the matrices $A$, $B$, and $C$ are not data-dependent, which restricts their adaptability — this property is known as Linear Time Invariance (LTI).

To enhance flexibility, a gate-like operation was introduced to matrix $A$, allowing it to become data-dependent in a parameter-efficient manner. Similarly, matrices $B$, $C$, and the time step were adapted to be data-dependent using linear projection functions. Since all parameters are now data-dependent or dynamic, \cite{gu_mamba_2023} proposed a parallel scan algorithm to efficiently operate the model on GPUs, enabling SSM models to scale for parallel computation.

While Mamba demonstrated effectiveness, especially with longer sequence lengths, it still did not fully capitalize on the power of GPUs. To address this, \cite{gu_mamba_2023} introduced Mamba2, which applies a state-space duality approach. This new model leverages GPU capabilities more effectively, further improving the computational efficiency while maintaining high performance in long sequence modeling.

\subsubsection{Numerical Weather Prediction Models}

Wildfire risk prediction is a complex forecasting system that necessitates the consideration of multiple interacting driving factors. Similarly, numerical weather prediction (NWP) is a highly nonlinear, multiscale system requiring the integration of diverse influencing variables. Early NWP models, such as the Integrated Forecast System (IFS) developed by the European Centre for Medium-Range Weather Forecasts (ECMWF) and the Weather Research and Forecasting (WRF) model from the National Center for Atmospheric Research (NCAR), were primarily based on physical equations supplemented by empirical parameterization schemes. These models exhibit strong capabilities in handling cross-scale atmospheric dynamics, including global circulation and local weather interactions, multiscale numerical simulations (ranging from hourly nowcasting to seasonal forecasting), stochastic process modeling, and multiphysical coupling analyses (e.g., the interactions between temperature, humidity, and wind fields). 

With the recent breakthroughs in deep learning, the meteorological forecasting community has increasingly explored data-driven approaches to reduce computational complexity, enhance efficiency, and improve predictive accuracy by continuously integrating newly available data \citep{sønderby2020metnetneuralweathermodel, lam2023learning, fourcast2023, waqas2024artificial}. Notably, as in other Earth science applications, deep learning circumvents the need to solve partial differential equations (PDEs) or radiative transfer equations by directly learning statistical relationships from historical weather data. This enables predictions to be generated in a single forward pass, significantly reducing computational demands \citep{lam2023learning}. Furthermore, the parallel processing capabilities of GPUs allow for substantial acceleration in model inference \citep{sønderby2020metnetneuralweathermodel}. Another notable trend in meteorological forecasting is the emergence of hybrid modeling paradigms that integrate physical mechanisms with data-driven intelligence. By replacing traditional parameterization schemes (e.g., subgrid-scale turbulence closure, cloud microphysics, and computationally intensive radiative transfer modules) with neural networks, these hybrid models achieve substantial improvements in computational efficiency and predictive performance while preserving physical constraints. 

Intuitively, both meteorological forecasting and wildfire risk prediction share fundamental similarities as complex systems driven by multiple interacting factors. Consequently, insights from physical and deep learning-based meteorological models could provide valuable guidance for developing physics-informed, data-driven wildfire risk prediction models. This is particularly relevant since existing studies integrating physics-based constraints with data-driven methodologies in wildfire risk prediction remain limited. Therefore, this section reviews the latest advancements in NWP modeling and their potential implications for wildfire risk prediction. Specifically, it highlights key developments in autoregressive modeling, lightweight architectures, uncertainty quantification, and hybrid models that integrate physics-based constraints with data-driven approaches.

\paragraph{Autoregressive Modeling in Meteorological Forecasting}

Atmospheric evolution is inherently dynamic and time-dependent, making meteorological forecasting a temporal sequence prediction task reliant on past weather conditions to estimate future states. Due to the chaotic nature of atmospheric motion, end-to-end forecasting models tend to exhibit error accumulation over time. To mitigate this issue, autoregressive (AR) strategies are commonly employed, wherein forecasts are iteratively propagated to predict future states. For instance, GraphCast generates global weather predictions at 6-hour intervals for up to 10 days \citep{lam2023learning}. Similarly, the diffusion model-based GenCast adopts an iterative denoising approach to forecast weather conditions 15 days in advance, using a 12-hour prediction interval. To alleviate error accumulation inherent in autoregressive methods, Pangu-Weather introduces a Hierarchical Temporal Aggregation strategy, training four sub-models to predict weather at varying time scales (1, 3, 6, and 12 hours) rather than employing a fixed prediction step, thereby reducing accumulated forecasting errors \citep{bi2023accurate}.

\paragraph{Lightweight Model Designs in Meteorological Forecasting}

Similar to wildfire risk prediction, meteorological forecasting requires extensive input variables to predict large-scale future states, leading to substantial computational costs. Traditional uncertainty quantification in NWP relies on ensemble forecasting, which involves generating multiple weather trajectories under different initial conditions—an approach that significantly increases computational demands. Consequently, various lightweight model designs have been proposed, including specialized Transformer architectures, graph neural networks, and Fourier neural operators.

For instance, MetNet first applies CNNs and pooling layers to downscale past 90-minute meteorological radar (MRMS) and satellite (GOES-16) data, reducing computational complexity. Subsequently, ConvLSTM processes temporal sequences, while an axial self-attention module captures large-scale spatial dependencies while mitigating the quadratic computational burden of traditional Transformer architectures. Notably, this model is primarily applicable to short-range precipitation forecasting, limiting its broader utility. To address this, MetNet-2 introduces a Context Aggregation Module and Lead Time Conditioning, expanding its spatial coverage from $1024 \times 1024$ km over an 8-hour forecast period to $2048 \times 2048$ km over 12 hours \citep{espeholt2022deep}. Despite these improvements, MetNet-2 remains constrained to short-term, localized forecasts.

To enable global medium-range forecasting of multiple meteorological variables and extreme weather events, GraphCast leverages a graph-based approach \citep{lam2023learning}. With 36.7 million parameters, GraphCast produces 10-day global forecasts within one minute using a single Google Cloud TPU V4. Algorithmically, GraphCast follows a classic encoder-processor-decoder architecture: the encoder maps past 6-hour reanalysis or forecast data onto a multi-scale graph structure (where each node represents spatially distinct scales), the processor employs a 16-layer GNN to model both local and long-range dependencies, and the decoder transforms graph features back to geospatial grids for final weather predictions. While GraphCast outperforms ECMWF’s HRES NWP model on 90.3\% of 1,380 evaluation targets, it remains computationally expensive.

To further enhance efficiency, FourCastNet incorporates Adaptive Fourier Neural Operators (AFNO) for global 7-day weather forecasting and extreme event simulation \citep{fourcast2023}. Similar to Transformers, AFNO initially partitions input data into patches and assigns positional encodings to preserve spatial information. To mitigate the quadratic complexity of standard Transformers, AFNO employs 2D fast Fourier transforms to transition patches into the frequency domain, where global spatial dependencies are modeled, followed by fully connected layers to facilitate cross-channel interactions. Additionally, AFNO introduces Spectral Mixing MLP and an adaptive gating mechanism to dynamically adjust frequency components and optimize feature mixing.

\paragraph{Uncertainty Quantification in Meteorological Forecasting}
\label{uncertainty estimation}

Given uncertainties in observations, model formulations, and the chaotic nature of weather systems, uncertainty quantification (UQ) is fundamental to meteorological forecasting, particularly for probabilistic predictions. UQ provides reliable confidence estimates and enhances early warnings for extreme weather events \citep{waqas2024artificial}. Traditional NWP models frequently employ stochastic differential equations (SDEs) to model random processes such as precipitation probabilities and temperature fluctuations. A similar approach can be adapted for wildfire risk modeling by formulating ignition probability as a stochastic process:

\begin{equation}
\label{SDE}
dP_{\text{ignition}}(x, t) = \mu(x, t)dt + \sigma(x, t)dW_t + \lambda(x, t)dN_t
\end{equation}
where $P_{\text{ignition}}(x, t)$ represents the wildfire ignition probability at location $x$ and time $t$. The deterministic term $\mu(x, t)dt$ accounts for predictable factors such as fuel conditions, wind speed, temperature, and humidity, while the stochastic fluctuation term $\sigma(x, t)dW_t$ incorporates random disturbances modeled as Wiener processes (standard Brownian motion). Additionally, the jump process term $\lambda(x, t)dN_t$ captures discrete, abrupt events such as lightning strikes or explosions, represented as a Poisson process. 

Beyond stochastic differential equations, several other methods are widely used for uncertainty quantification in meteorological forecasting. Ensemble prediction, a common approach in both deep learning-based and traditional NWP models, generates multiple weather trajectories using perturbed initial conditions (PICs). The spread among these trajectories, along with confidence intervals and probability density estimations, provides a measure of uncertainty. Ensemble forecasting only accounts for variations in initial conditions and does not capture uncertainties in the underlying dynamical system. 

To address this limitation, GenCast employs generative AI (diffusion models) for probabilistic weather prediction. Its core idea is to generate multiple plausible weather trajectories via noise injection and refine them progressively using a 3D Graph Transformer denoising network. This approach not only eliminates the high computational cost of multiple inference runs in ensemble forecasting but also simultaneously captures uncertainties in both initial conditions and weather dynamics. Additionally, the 3D Graph Transformer employs sparse computations, mitigating the substantial computational burden associated with grid-based data processing in conventional Transformers. Through different denoising stages, GenCast directly produces probabilistic weather distributions along with uncertainty assessments.

Furthermore, \cite{waqas2024artificial} highlighted several other techniques for uncertainty quantification in meteorological forecasting, including Bayesian neural networks, Gaussian processes, Monte Carlo methods, and probabilistic loss functions such as negative log-likelihood. They also recommended statistical post-processing methods, such as Quantile Regression Forests (QRF) and variational Bayesian techniques, to reduce systematic biases. Moreover, Generative Adversarial Networks (GANs) have been proposed to learn error distributions, thereby improving confidence interval estimations in weather prediction models.

\paragraph{Hybrid Physics-Guided and Data-Driven Models}

Despite significant advancements in modeling multiphysical and multiscale atmospheric systems using PDEs and Navier-Stokes equations, the computational cost remains prohibitively high. For example, traditional medium-range (7–10 days) weather forecasting using conventional NWP models requires hours of computation on supercomputers \citep{bi2023accurate, price2025probabilistic}. In contrast, deep learning-based approaches can achieve comparable accuracy within seconds \citep{bi2023accurate, lam2023learning}. Data-driven methods are particularly advantageous in addressing the challenge of abundant observational data yet limited processing capabilities in Earth science. However, purely data-driven models often suffer from poor generalization due to limited modeling capacity or observational biases, leading to reduced interpretability and potential inconsistencies with physical laws \citep{karniadakis2021physics, shi2025deeplearningfoundationmodels}. 

To address these challenges, hybrid models integrating physics-based constraints with data-driven learning have been proposed. These models leverage prior knowledge from physics and mathematics to ensure physically consistent predictions, enhance interpretability, improve robustness, and potentially reduce data dependency. Generally, three types of inductive biases can be introduced into data-driven models: observational bias, inductive bias, and learning bias. These biases incorporate physical principles either implicitly or explicitly through observational data, architectural design, or loss function constraints \citep{karniadakis2021physics}.

A common approach to incorporating physics-based constraints is the use of ERA-5 reanalysis data as input for deep learning-based forecasting models, representing an example of observational bias. Since observational data inherently encapsulate physical laws, training deep learning models on these datasets enables the learning of physically constrained functions or operators. In addition to ERA-5 data, the AI Forecasting System (AIFS) also utilizes ECMWF 4D-Var data for high-quality initialization states \citep{lang2024aifsecmwfsdatadriven}. Compared to random initialization, this approach integrates multi-temporal observations with numerical model simulations. It reduces inconsistencies and minimizes errors introduced by single-source observations, thereby ensuring that the initial state of the weather model closely represents real-world atmospheric conditions.

Beyond these specific examples, data assimilation techniques can also be considered as a form of observational or learning bias. Data assimilation algorithms refine state variable estimates by integrating prior model simulations with observational data \citep{cheng2022data}. Traditional NWP models frequently employ multi-source heterogeneous datasets, combining surface station data, radiosonde measurements, short-term NWP outputs, and satellite observations. However, these datasets are often sparse, incomplete, and contain noise. Directly using such observations without refinement can introduce errors. Data assimilation techniques enable the integration of these heterogeneous datasets with numerical models by performing spatial and temporal interpolation, optimizing initial conditions, and reducing forecast uncertainty, thereby improving prediction accuracy. Common data assimilation methods in meteorology include Kalman filtering, 4D-Var, and particle filtering, which facilitate the fusion of observational data with model predictions \citep{rabier2005overview}. Similarly, data assimilation techniques can be applied in wildfire risk prediction to supplement or correct missing data.

Notably, data assimilation is not a one-time operation but a continuous process. By periodically assimilating new observations, models can iteratively refine their state estimates at each time step, thereby mitigating the accumulation of initial errors and maintaining accuracy in dynamic environments \citep{cheng2022data}. In this context, data assimilation can be interpreted as a form of learning bias. Wildfire prediction models could adopt a similar approach by integrating heterogeneous data sources such as satellite fire detections (e.g., MODIS/VIIRS thermal anomalies), UAV monitoring, ground-based sensors, and social media reports to dynamically update wildfire conditions. For instance, SAR radar data could be used to estimate fuel moisture, while UAV-based infrared imaging could assess fuel distribution, enabling high-frequency updates of fuel conditions and improving short-term prediction accuracy. In wildfire prediction tasks, \cite{cheng2022data} demonstrated that data assimilation significantly reduced systematic biases in background forecasts and improved long-term reliability by incorporating real-time observations. Their experimental results indicated that data assimilation reduced the relative root-mean-square error (R-RMSE) by approximately 50$\%$, leading to substantial improvements in long-term prediction accuracy.

A notable example of inductive bias in deep learning-based meteorological forecasting is the NeuralGCM model, which employs differentiable physics solvers to resolve large-scale fluid dynamics and thermodynamic processes in the atmosphere, including gravitational forces and Coriolis effects \citep{kochkov2024neural}. Additionally, NeuralGCM is designed with neural network modules dedicated to single-column physical processes such as cloud formation, radiative transfer, and precipitation, while maintaining a clear separation between horizontal dynamics and local physical processes. This architectural design ensures that the neural network focuses on small-scale physical interactions, while large-scale atmospheric dynamics remain strictly governed by physical equations, preventing inconsistencies commonly observed in purely data-driven models.

In future wildfire risk prediction models, additional forms of inductive and learning biases could be explored. For example, constraints on fire spread speed based on wind velocity, temperature, humidity, and fuel load could be imposed to prevent physically unrealistic fire propagation rates. Alternatively, classical fire behavior models, such as the Rothermel model (fireline intensity equation) or PDE-based fire propagation models (e.g., level-set methods), could be incorporated as regularization terms to ensure that neural network outputs comply with energy release rates and fire spread direction constraints. Another approach involves leveraging physics-informed neural networks to directly solve fire propagation equations, such as using combustion rate equations as learning constraints. Additionally, techniques from large-eddy simulation (LES) in meteorological modeling could be adapted to develop hybrid models for wildfire turbulence diffusion, potentially integrating deep reinforcement learning to optimize turbulence parameterization.

\subsection{Promising Multi-source Remote Sensing Data}

From a data perspective, the majority of wildfire risk prediction studies predominantly rely on satellite-based multispectral data, particularly various MODIS-derived products, to obtain real-time or near-real-time information on fuel conditions, wildfire status, and human activities. The primary advantage of multispectral data lies in its ability to provide long-term observational data across extensive spatial scales. However, these products are inherently limited by their spatial and temporal resolution, as well as their inability to penetrate through vegetation canopies, clouds, fog, and smoke. Consequently, they struggle to deliver precise fuel condition assessments and wildfire state parameters crucial for risk prediction, creating significant obstacles particularly for short time-scale forecasting tasks such as next-day wildfire risk prediction. To overcome these limitations, integrating ground-based or unmanned aerial vehicle (UAV) platforms with satellite data, or fusing multispectral data with UAV, satellite and airborne LiDAR, microwave, and nighttime light data, can substantially enhance prediction accuracy. These multi-source data fusion approaches enable more comprehensive characterization of wildfire environments by combining complementary information across various sensing modalities and spatial-temporal resolutions.

Several review studies have already examined UAV-based wildfire management \citep{yuan2015survey, bouguettaya2022review, moumgiakmas2021computer, boroujeni2024comprehensive}. Notably, these surveys indicate that UAV-related wildfire research has predominantly focused on detection rather than risk prediction \citep{boroujeni2024comprehensive}. The potential of UAV-based wildfire risk prediction lies in its capacity to utilize various onboard sensors, including high-resolution multispectral cameras, LiDAR, radar, and acoustic and meteorological sensors, to provide highly accurate fuel condition assessments and environmental monitoring \citep{boroujeni2024comprehensive, 10681400}. While UAV-based data offer advantages in terms of resolution and precision, from a methodological perspective, wildfire risk prediction algorithms utilizing UAV data do not fundamentally differ from those relying on spaceborne multispectral and geospatial datasets.

Beyond UAV data, microwave, spaceborne LiDAR, and nighttime light data play crucial roles in wildfire risk assessment by offering more accurate fuel condition evaluations and records of wildfire occurrences. Each of these data sources provides unique insights: microwave data (e.g., SAR) can penetrate clouds and smoke, enabling the assessment of wildfire, fuel moisture, and soil moisture dynamics; LiDAR data offer detailed three-dimensional structural information of fuel, facilitating accurate estimations of fuel load, continuity, and classification; nighttime light data are sensitive to thermal anomalies and human activities, making them valuable for identifying potential wildfire ignition sources and estimating carbon emissions. Despite their immense potential in wildfire risk assessment models, studies leveraging these data sources remain relatively scarce, warranting further exploration in future research.

\subsubsection{Microwave Data}

SAR satellites offer a significant advantage over optical sensors in wildfire risk assessment due to their ability to penetrate clouds and smoke, which often hinder optical observations. Moreover, SAR data provide valuable information on surface roughness and dielectric properties, which vary significantly with changes in fuel characteristics and soil moisture \citep{Meyer2019CHAPTER2S}. In wildfire and vegetation-related studies, the C-band (3.75–7.5 cm) and L-band (15–30 cm) SAR are the most commonly used. The C-band exhibits limited penetration capability, making it more sensitive to the upper canopy layer and finer vegetation structures such as small branches or leaves. In contrast, the L-band can penetrate sparse to moderately dense vegetation canopies, allowing for the analysis of structural attributes such as trunk biomass and larger branches. Furthermore, L-band SAR has been shown to be more sensitive to changes in biomass, sub-canopy vegetation, and soil conditions, particularly in high-biomass environments \citep{coops2002eucalypt, millard2018soil, coops2021modelling}. 

Apart from SAR, passive microwave remote sensing also provides large-scale soil and vegetation moisture estimates, which are critical factors in wildfire ignition and spread modeling. Notably, passive microwave products are relatively scarce, and in wildfire risk prediction, soil moisture and drought indices derived from meteorological data are more commonly used as proxies.

Currently, microwave data are primarily employed in wildfire monitoring, burning map generation, and severity assessment, rather than as direct input for wildfire risk prediction models. For instance, \cite{addison2018utilizing} utilized L-band ALOS PALSAR data to assess wildfire burn severity in the western United States, leveraging SAR’s sensitivity to surface roughness. Their results demonstrated that SAR-based assessments outperformed optical indices such as the differenced normalized burn ratio (dNBR). Similarly, \cite{tariq2021quantitative} leveraged SAR’s penetration capability and sensitivity to vegetation structural changes by employing Sentinel-1 C-band SAR data to evaluate post-prescribed burning surface fuel and understory vegetation changes in Croajingolong National Park, southeastern Australia. Furthermore, \cite{hu2023gan} explored the application of GANs to translate SAR images into optical representations and generate corresponding burn severity indices. In another study, \cite{horton2024wildfire} developed a wildfire monitoring algorithm based on SAR-derived vegetation index variations, identifying wildfire occurrence through changes in vegetation structure. 

Regarding real-time wildfire spread monitoring, \cite{zhang2021learning} integrated Sentinel-1 SAR and Sentinel-2 multispectral data to develop an approach for near-real-time wildfire propagation tracking. Similarly, \cite{9468971} compared the wildfire detection capabilities of Sentinel-1 C-band SAR, Sentinel-2 multispectral data, Sentinel-3 sea surface temperature data, and MODIS products. Their findings indicated that SAR data significantly enhanced wildfire detection accuracy under cloudy conditions.

In addition to wildfire detection and severity mapping, a limited number of studies have investigated the potential of SAR’s penetration capability and all-weather data acquisition for wildfire risk prediction. For example, \cite{nur2022creation} utilized Sentinel-1 SAR data to generate damage proxy maps, which were subsequently used as target labels for wildfire risk prediction models. Notably, this study primarily focused on SAR-based wildfire burning map generation rather than direct risk assessment. In contrast, \cite{pelletier_wildfire_2023} employed Sentinel-1 C-band SAR backscatter data as input features for wildfire risk prediction in Canadian treed peatlands. Their findings demonstrated that incorporating SAR data into a Sentinel-2 multispectral model improved the prediction of peatland fire probability. Importantly, they also observed that in dense vegetation areas, C-band SAR backscatter exhibited only a weak correlation with fire probability, leading them to recommend its application primarily in sparse or open peatlands. Likewise, \cite{fernandez2022predicting} incorporated Sentinel-1 SAR data into a wildfire severity prediction model for southern Europe, using pre-fire C-band SAR cross-ratio as a complementary proxy for fuel load and moisture. Although their study found a negative correlation between SAR-derived indicators and fuel conditions, Sentinel-1-based features exhibited lower predictive performance compared to Sentinel-2 multispectral indices. As a result, these SAR-derived features were not included in their optimized model.

Beyond their direct application in wildfire mapping, monitoring, severity assessment, and risk prediction, both active and passive microwave data have been extensively validated for their effectiveness in vegetation condition assessment, including FMC and fuel load estimation, as discussed in Section~\ref{fuel}. Since these variables are crucial explanatory factors in wildfire risk assessment models, leveraging microwave data to derive indices related to fuel conditions or improving the accuracy of vegetation assessments presents a promising avenue for enhancing wildfire risk prediction. Notably, research in this area remains relatively underexplored.

\subsubsection{LiDAR Data}

LiDAR data share similarities with active microwave remote sensing in that both actively emit signals and receive backscattered or reflected signals. This capability allows them to operate in an all-weather, all-day setting while providing penetration through fuel layers to retrieve structural information. Comparisons between optical sensors and SAR sensors in fuel load estimation have highlighted the importance of canopy structural information for accurate fuel load retrieval \citep{aragoneses2024mapping}. Unlike optical or SAR-based fuel load estimation methods, airborne LiDAR provides more detailed three-dimensional structural information on fuel distribution \citep{andersen2005estimating, Hermosilla2014}. Consequently, airborne and spaceborne LiDAR sensors offer the potential to enhance wildfire risk prediction models, assess wildfire damage levels, and evaluate the effectiveness of fuel management strategies by directly and accurately estimating fuel load and fuel continuity at various scales \citep{hillman2021comparison}.

For example, \cite{skowronski_three-Dimensional_2011} utilized LiDAR systems to estimate canopy fuel load in pine-dominated forests in New Jersey, USA, predicting canopy bulk density and canopy fuel weight. Similarly, \cite{chen_development_2017} combined LiDAR data with historical wildfire occurrences, canopy density, elevation, and fuel type to estimate ground fuel load in the jarrah forests of Western Australia. \cite{bright2017prediction} compared the effectiveness of airborne LiDAR and Landsat time-series data for estimating canopy and surface fuel load in Grand County, Colorado. Their findings indicated that Landsat-based improvements in fuel load estimation were relatively limited. Expanding upon this, \cite{labenski_quantifying_2023} integrated Sentinel-2 multispectral data with airborne LiDAR-derived forest structure and composition information from two forest regions in southwestern Germany, employing a random forest model to predict and map surface fuel distribution. Moreover, \cite{d2021machine} incorporated Sentinel-1 SAR data alongside Sentinel-2 multispectral and airborne LiDAR data to estimate dead fine fuel load in southeastern Italy. Their study revealed that, compared to NDVI and C-band SAR data, LiDAR-derived variables exhibited superior performance in dead fine fuel load estimation. Additionally, their findings supported prior observations that woody components contribute more significantly to fuel load, consistent with \cite{li2020assessment}. Furthermore, \cite{comesana2024wildfire} explored the integration of multispectral point cloud data with deep learning models to refine the Prometheus fuel model, aiming to establish a more comprehensive fuel classification standard that better reflects variations in wildfire behavior, thereby enhancing fuel management practices.

NASA’s recently developed Global Ecosystem Dynamics Investigation (GEDI) full-waveform LiDAR sensor has already demonstrated broad applicability in large-scale aboveground biomass estimation and canopy structure analysis \citep{myroniuk2023combining, liang2023quantifying, aragoneses2024mapping, guerra2024impact}. This advancement holds promise for improving wildfire simulation, assessing wildfire-induced vegetation damage, and understanding burn severity. However, relatively few studies have directly employed GEDI LiDAR data for fuel load estimation. One such study by \cite{leite2022large} utilized UAV simulations and on-orbit GEDI data to assess stratified fuel load across three ecosystem layers—grassland, savanna, and forest—in the Brazilian tropical savanna. Their findings suggested that GEDI’s performance in estimating ground and low-lying vegetation fuel load was suboptimal and significantly affected by terrain conditions.

Beyond fuel parameter retrieval, a limited number of studies have explored the application of LiDAR-derived information in downstream tasks such as wildfire spread modeling and hazard assessment. For instance, \cite{botequim2019improving} integrated crown fuel characteristics derived from low-density airborne small-footprint full-waveform LiDAR with stand structural indices to construct spatial wildfire simulations, demonstrating that LiDAR data significantly improved estimates of key forest structural characteristics, including canopy height, tree density, and fuel load. In a laboratory-scale study, \cite{marcozzi2023application} employed terrestrial laser scanning point cloud data to characterize three-dimensional fuel structures and evaluate the relationship between fire behavior simulations and LiDAR-derived fuel data resolution. Meanwhile, \cite{hoe2018multitemporal} investigated wildfire severity in the Klamath Mountains ecoregion of southwestern Oregon by combining pre- and post-fire multitemporal LiDAR data with Landsat-derived relative differenced normalized burn ratio (RdNBR). Their results indicated that incorporating LiDAR data significantly improved the accuracy of basal area mortality estimation. Additionally, ground-based LiDAR, due to its high vertical resolution, can precisely capture the spatial distribution of wildfire smoke plumes. By analyzing reflectance intensity and pulse time delays, it is possible to infer aerosol particle size, concentration, and optical properties, offering valuable insights into wildfire smoke dispersion and pollutant transport pathways \citep{ye2022assessing, deng2022wildfire, huang2024assessment}.

Despite these advantages, LiDAR-based research in wildfire applications has primarily focused on improving the accuracy of fuel structure retrieval rather than directly integrating these results into wildfire risk prediction models. Several factors may contribute to this gap. First, compared to conventional 2D optical imagery, LiDAR data are inherently more complex and specialized, requiring researchers to possess specific technical expertise and training. Second, the availability of spaceborne LiDAR data is limited, and airborne LiDAR acquisitions are costly, making large-scale, long-term data collection challenging. As a result, the scarcity of LiDAR data has constrained its application in wildfire risk prediction. Future research should aim to incorporate higher-quality fuel structure parameters derived from LiDAR data to enhance prediction accuracy while reducing model uncertainty.

\subsubsection{Nighttime Light Data}

In addition to microwave and LiDAR data, nighttime light data also hold potential for improving wildfire risk prediction due to their sensitivity to thermal anomalies, human activities, and lightning events. These data can be leveraged for wildfire detection and mapping, as well as for modeling the relationships between human activities, lightning, and wildfire occurrences. Furthermore, nighttime light data can serve as indicators of combustion efficiency, aiding in the estimation of wildfire-induced carbon emissions and aerosol composition, as well as in identifying sources of air pollution \citep{10081011}.

The most commonly used nighttime remote sensing data sources include the Visible Infrared Imaging Radiometer Suite (VIIRS) onboard S-NPP, NOAA-20, and NOAA-21. VIIRS is a 22-band scanning radiometer covering wavelengths from 412 nm to 12 $\mu m$ m and is regarded as the successor to MODIS for Earth science data product generation \citep{s20226442}. VIIRS is considered an optimal balance between spatial and temporal resolution \citep{chuvieco2020satellite}, offering a spatial resolution of 375 m in its five imagery bands (I-bands) and 750 m in its 16 moderate resolution bands (M-bands) and the day/night band (DNB). Compared to MODIS, VIIRS provides improvements in nighttime detection capabilities and higher spatial resolution. Specifically, VIIRS features seven shortwave radiation bands (M07–M13 and I4) dedicated to nighttime wildfire detection, addressing MODIS's limitation of having only a single 3.96 $\mu m$ MWIR detection band, which is insufficient for characterizing biomass burning emissions \citep{7498622, 10081011}. Additionally, VIIRS's enhanced spatial resolution enables better detection of smaller wildfires \citep{schroeder2014new}.

Extensive research has been conducted on wildfire detection using VIIRS data. Since the initial implementation of VIIRS-based wildfire detection algorithms utilizing I-bands and contextual techniques, significant advancements have been made, yielding high spatial resolution (375 m) and strong consistency with MODIS products \citep{schroeder2014new}. For example, to address limitations in detecting small and low-intensity fires, the Fire Identification, Mapping, and Monitoring Algorithm (FILDA) series introduced advanced methodologies. FILDA-1 mitigated the "bundling" effect by incorporating additional spectral bands to enhance fire detection sensitivity \citep{7498622}, while FILDA-2 integrated the DNB and M-bands to improve the detection of low-intensity fires and enhance the retrieval accuracy of fire radiative power (FRP) \citep{WANG2020111466, 10081011}. These advancements have significantly improved VIIRS-based fire monitoring, providing more reliable tools for wildfire detection and emissions analysis. Additionally, \cite{rs15061541} employed VIIRS's VNP14IMG product to construct training labels for a random forest-based wildfire detection model using Himawari-8 data. Similarly, \cite{10551624} utilized VIIRS VNP46A2 data in combination with a random forest model for forest fire detection in southwestern China. Their study demonstrated that VIIRS outperformed MODIS and other products in detecting lower-temperature wildfires, making it a valuable complementary data source for wildfire detection.

Beyond wildfire detection, nighttime light data can serve as a proxy for human activity, making them suitable as input features in data-driven models for analyzing human-induced wildfire risk. Compared to conventional socioeconomic indicators such as GDP, population distribution, energy consumption, land use, and accessibility, nighttime light data offer advantages including higher spatiotemporal resolution, frequent updates, extensive coverage, and spatial continuity, making them more effective in capturing human activity patterns \citep{Chen31122023, rs15030598}. Specifically, \cite{Chen_2022} employed annual composite nighttime light images as indicators of human settlement and electricity infrastructure density. By integrating these data with fuel, topographic, and meteorological variables, they analyzed the influence of anthropogenic and lightning ignition on wildfire probability in California. Similarly, \cite{rs15030598} utilized nighttime light data as a proxy for socioeconomic factors, compensating for the limitations of statistical data in capturing the spatiotemporal heterogeneity of human activity. By combining nighttime light data with other drivers, they investigated the relationship between wildfire probability and various influencing factors in Anhui Province, China. Their study revealed that socioeconomic variables exhibited the highest correlation with wildfire occurrence, while vegetation, topographic, and meteorological factors played a secondary role. Furthermore, \cite{Chen31122023} developed a human activity factor based on nighttime light intensity within forest buffer zones. By integrating this factor with other wildfire drivers, they assessed its contribution to wildfire risk across China. Their analysis demonstrated that nighttime light data had greater importance than topographic, meteorological, and fuel-related factors, and incorporating nighttime light data improved model accuracy by 6\%.

While numerous studies have explored the application of nighttime light data in wildfire detection and risk prediction, their broader potential in wildfire-related research remains underexplored. Beyond directly or indirectly incorporating nighttime light data as a socioeconomic driver in wildfire risk models, a limited number of studies have investigated the use of nighttime light data for wildfire hazard assessment, particularly in terms of wildfire impact on human settlements. For instance, \cite{yue2023assessment} employed nighttime light indices to represent urban development levels (vulnerability) rather than as a direct socioeconomic driver in a wildfire risk prediction model. This approach enabled them to identify high-risk areas where wildfires posed potential threats to economic assets and human populations in Guilin, China. Notably, overall, research on wildfire hazard prediction based on VIIRS and other nighttime light products remains limited. Future studies could leverage nighttime light data to assess wildfire hazards, particularly in the WUI. Additionally, long-term analysis of population density, WUI expansion, and urban dynamics could provide valuable insights into their relationships with wildfire risk and wildfire-induced damages.

\color{black}

\section{Conclusion}

Wildfire risk prediction has evolved through the integration of diverse data sources and the development of various modeling approaches. However, a comprehensive and systematic review of these data and methodologies remains insufficient, particularly in the domain of deep learning, which has seen rapid advancements and increasing applications in wildfire risk assessment. This paper provides a thorough review of wildfire driving factor datasets, feature collinearity analysis, model interpretability, and deep learning-based wildfire prediction models, with a particular emphasis on the role of remote sensing in wildfire risk assessment.

The development of wildfire risk prediction models necessitates the integration of multiple input data sources, along with reliable ground truth data for model training and validation. These input data typically include fuel characteristics, meteorological and climatic conditions, topography, hydrology, and socioeconomic factors. Our review indicates that most studies prioritize meteorological and climatic variables, followed by fuel and topographical conditions, while socioeconomic and hydrological factors remain underexplored despite their potential impact. To facilitate deep learning-based wildfire risk assessment, this study also introduced publicly available wildfire burn prediction datasets, including FireCube, Next Day Wildfire Spread, Wildfire DB, WildfireSpreadTS, CFSDS, and the SeasFire Datacube. Furthermore, to enhance model reliability and interpretability, we discussed feature collinearity assessment techniques such as Variance Inflation Factor (VIF), tolerance, and Pearson correlation coefficients, alongside interpretability methods for deep learning models, including Permutation Feature Importance, Explainable Feature Engineering, and SHapley Additive exPlanations (SHAP).

In terms of modeling approaches, wildfire risk prediction has historically relied on traditional machine learning methods and wildfire danger rating systems. While these conventional approaches have been extensively reviewed, deep learning-based models, despite their increasing adoption, have lacked a structured and systematic analysis. This paper categorizes deep learning models into three primary types: time-series forecasting, image segmentation and classification, and spatiotemporal prediction. Our review highlights an emerging trend towards the utilization of longer time-series data and the development of higher spatiotemporal resolution models to improve wildfire risk assessments.

Looking forward, we identify several key directions for future research. First, the integration of novel and underutilized remote sensing data sources, such as microwave, LiDAR, and nighttime light data, holds significant potential for improving the accuracy and robustness of wildfire risk prediction models. Second, multimodal learning approaches, which leverage heterogeneous data types such as geospatial imagery, meteorological records, and textual descriptions, may offer enhanced predictive performance by capturing complex interactions among wildfire-driving factors. Additionally, the development of computationally efficient deep learning architectures, such as Mamba, presents an opportunity to model long-term spatiotemporal dependencies more effectively while reducing computational costs. Lastly, cross-disciplinary approaches, particularly the adaptation of techniques from numerical weather prediction models, may provide insights into uncertainty quantification, physical constraints, and data assimilation strategies that could improve wildfire risk assessment models. By addressing these challenges and opportunities, future research can contribute to more reliable, interpretable, and scalable deep learning models for wildfire risk prediction.

\appendix
\label{appendix}
\section{Fuel Conditions Estimation}

Through the above review, we identify that one of the key limitations in wildfire risk prediction is the inaccurate characterization of fuel conditions. Given data scarcity, most studies rely on readily available multispectral and meteorological data, along with indices derived from these sources, to describe fuel conditions. Therefore, we have examined the latest advancements in fuel condition retrieval, with a particular focus on fuel moisture content and fuel load, aiming to provide support for the development of more accurate wildfire risk prediction models.

\subsection{Fuel Moisture Content}

\subsubsection{Dead Fuel Moisture Content}

DFMC is a crucial input in nearly all wildfire risk assessment models \citep{matthews2013dead}. For example, estimates of DFMC are integral to current operational fire prediction systems, including the FWI and the National Fire Danger Rating System in the United States \citep{vitolo_era5-based_2020}. Estimating DFMC involves calculating EMC based on factors like temperature, relative humidity, and fuel response time, combined with precipitation data and drying algorithms \citep{nieto2010dead}. DFMC estimation methods can generally be divided into four categories: measurements at meteorological stations, field sampling, estimates based on meteorological data, and indirect retrieval using multi-source remote sensing data.

DFMC can be measured using standard wooden dowels installed 10–12 inches above the ground at meteorological stations \citep{rakhmatulina2021soil}. However, the sparse distribution of DFMC measurement stations and the limited accuracy of these measurements make it difficult to achieve large-area coverage, even with interpolation of atmospheric parameters. Field sampling of DFMC, on the other hand, provides highly accurate measurements and serves as a valuable data source for both physical process models and empirical modeling. Despite its accuracy, field measurements require careful consideration of various environmental factors, including terrain, vegetation type, ecosystem structure, climate, and soil characteristics. This complexity makes field sampling time-consuming, costly, and spatially constrained \citep{duff2017revisiting}. 

Given these limitations, many studies have focused on integrating field measurements with meteorological station data, numerical weather prediction models, or meteorological reanalysis data to develop empirical and mechanistic models for estimating FMC \citep{matthews2013dead}. Among these, physical process models primarily rely on energy and water balance equations. The most widely used model for DFMC estimation is the Nelson Dead Fuel Moisture Model, proposed by \cite{nelson_jr_prediction_2000}, which predicts DFMC by simulating heat conduction and moisture transfer between a wooden cylinder, the surrounding atmosphere, and external conditions. This model requires hourly input data on temperature, humidity, radiation, and precipitation to provide DFMC estimates. Additionally, \cite{mcnorton_global_2024} designed a new method for estimating DFMC by grouping fuels and simplifying the Nelson model. Other models used for DFMC estimation include bulk litter layer models, Byram’s diffusion equation-based models, and complete process-based models \citep{matthews2013dead}. 

Empirical methods for estimating DFMC rely on observed relationships between measured DFMC and factors such as air temperature, relative humidity, wind speed, soil moisture content, and vapor pressure deficit. These data-driven methods include statistical techniques \citep{ferguson2002measuring, ray2010predicting, cawson2020estimation, rakhmatulina2021soil}, but their primary limitation is that they are entirely data-driven, making it difficult to explain the relationship between driving factors like temperature, heat and water vapor fluxes, radiation, and precipitation, and DFMC. Consequently, some studies have started to combine data-driven approaches with physical process models \citep{fan2021physics, fan2024process}.

Apart from the meteorological data-based FMC estimation methods mentioned above, indices such as the Cellulose Absorption Index (CAI), the Lignocellulose Absorption Index (LCAI), the Normalized Difference Tillage Index (NDTI), the Normalized Difference Vegetation Index (NDVI), and the Shortwave Infrared Normalized Difference Residue Index (SINDRI) have been shown to correlate with DFMC \citep{zormpas2017dead, zacharakis_integrated_2023}. This forms the theoretical basis for DFMC estimation using remote sensing. However, research on DFMC estimation using multi-source remote sensing data is still limited \citep{mccandless2020enhancing}.

\subsubsection{Live Fuel Moisture Content}

In early studies, the LFMC was measured similarly to the DFMC through field sampling and weighing methods, with interpolation or inversion used to estimate the entire study area \citep{bianchi_live_2015}. While this method is highly accurate locally, it is time-consuming, labor-intensive, and not scalable to landscape, regional, or global levels \citep{yebra2013global}. As such, many studies favor multi-source geospatial data for large-scale and near real-time LFMC estimation, utilizing the growing capabilities of Earth observation technologies for high-resolution data \citep{rao2020sar, capps2021modelling}. These contemporary approaches utilize environmental variables such as precipitation and soil moisture, along with spectral, thermal infrared, and microwave data, often in combination with machine learning algorithms, to enhance prediction accuracy \citep{bowyer2004sensitivity, yebra2013global}. 

First, similar to DFMC, LFMC can be estimated on a large scale in near real-time using meteorological parameters. This is because plant water regulation is influenced by environmental factors such as precipitation, soil moisture, and evapotranspiration \citep{bowyer2004sensitivity}. For example, the NFDRS used by the US Forest Service calculates LFMC using empirical formulas based on precipitation, temperature, and relative humidity \citep{Bradshaw_1984}. Moreover, \cite{castro_modeling_2003} modeled the live fine fuel moisture content (LFFMC) of Cistus spp. in the Catalonia region of Spain using factors such as temperature, soil water availability, and atmospheric water content. Similarly, \cite{ruffault2018well} compared six drought indices' ability to estimate LFMC across six different Mediterranean shrubs in southern France, finding that drought indices were effective only at stand scale, with indices simulating long-term drought dynamics (DC, RWC, and KBDI) providing better LFMC predictions. \cite{viegas2001estimating} used DC, BUI, and DMC from the FWI to estimate LFFMC in various plant species in central Portugal and Catalonia. Furthermore, \cite{vinodkumar_continental-scale_2021} successfully predicted LFMC across Australia using the lag relationship between soil moisture changes and vegetation moisture responses combined with regression analysis and physical models.

Meteorological data can also be combined with other datasets to estimate LFMC. For instance, \cite{miller_projecting_2023} proposed a method using a multimodal Multi-tempCNN model that incorporates multitemporal meteorological data, MODIS MCD43A4 BRDF-Adjusted Reflectance data, and climate zone, topography, and location information. This method enables large-scale LFMC estimation at medium resolution (500 m) over periods longer than two weeks. Similarly, \cite{mcnorton_global_2024} integrated soil moisture data and LAI to estimate LFMC, validating the estimates using MODIS data. In the western United States, \cite{wang2023estimation} used meteorological indicators, soil moisture data, and vegetation indices to estimate LFMC, finding that inversion accuracy varied with vegetation cover type.

The advantage of meteorological parameter-based LFMC estimation methods is the ability to generate near real-time LFMC estimates or predictions using data from weather stations, meteorological satellites, or numerical weather forecasting \citep{capps2021modelling}. However, LFMC is more challenging to estimate using meteorological parameters compared to DFMC. This is because plants actively regulate their water content through absorption and transpiration, with significant differences in physiological characteristics between plant species and a greater influence from phenological conditions and medium-term meteorological conditions \citep{pellizzaro2007seasonal, ruffault2018well, jolly2018pyro}.

Therefore, compared to LFMC estimation methods based on meteorological data or drought indices, most studies prefer using multi-source remote sensing data to estimate LFMC \citep{rao2020sar, capps2021modelling, quan2021global, rodriguez2023modelling}. Remote sensing technology, as a low-cost means of large-scale, long-term dynamic Earth observation, also enables near real-time, high-resolution LFMC estimation in highly heterogeneous real-world scenarios \citep{quan_global_2021, zhu_live_2021, verbesselt_estimation_2002}. Additionally, the exponential growth of Earth observation missions \citep{ustin2021current} has provided new opportunities for LFMC estimation \citep{leite2024leveraging}. Studies on LFMC estimation using multi-source satellite remote sensing primarily utilize multispectral reflectance data, thermal infrared data, and active and passive microwave data, combined with data-driven empirical or machine learning models, radiative transfer-based physical models, or hybrid models that integrate physical constraints with data-driven approaches \citep{yebra2013global, mcnorton_global_2024}.

The theoretical basis for LFMC inversion methods using multispectral reflectance data lies in the presence of water absorption bands in the near-infrared (~750-1100 nm) and shortwave infrared (~1100-2500 nm) spectral regions. Plant stress due to drought leads to changes in chlorophyll concentration and leaf internal structure, allowing the relationship between vegetation spectral reflectance (i.e., spectral characteristics) and LFMC to be established \citep{knipling1970physical, kriedemann_photosynthetic_1983, bowyer2004sensitivity, yebra2013global, sow2013estimation, zacharakis_integrated_2023, leite2024leveraging}. Thus, LFMC can be inverted by simulating the relationship between spectral characteristics under different water content levels and observed spectral characteristics \citep{hao2007retrieval, yebra2008estimation, yebra2013global, marino2020investigating, quan2021global}. While this physical method is generally more robust, it is also more complex in terms of parameterization \citep{yebra2013global}. Alternatively, empirical methods based on spectral reflectance or spectral indices, including the NDVI, enhanced vegetation index (EVI), soil-adjusted vegetation index (SAVI), normalized difference infrared index (NDII), water index (WI), normalized difference water index (NDWI), visible atmospheric resistance index (VARI), Leaf Water Content Index (LWCI) \citep{hunt1987measurement}, Global Vegetation Moisture Index (GVMI), and Shortwave Infrared Water Stress Index (SIWSI) \citep{fensholt_derivation_2003}, are simpler and only require fitting the relationship between in situ LFMC measurements and spectral reflectance characteristics. These methods can perform as well as or better than physical models at local scales and across specific vegetation types \citep{cunill2022live}.

For example, \cite{chuvieco2002estimation} compared the ability of shortwave infrared bands, NDVI, NDII, LWCI, the wetness and greenness components of the Tasseled Cap transformation, and spectral derivatives to model LFMC in Mediterranean grasslands and shrubs. They found that shortwave infrared bands and indices using these bands performed better due to the high correlation between shortwave infrared and LFMC, attributed to water absorption in leaves. Near-infrared and red bands also showed good correlation with LFMC, particularly for herbaceous plants, where drought stress led to changes in chlorophyll content and structure. The study recommended further use of thermal bands and meteorological data to improve modeling accuracy. Additionally, \cite{myoung2018estimating} used Pearson correlation analysis to compare the correlations between 16-day NDVI and EVI products (MOD13Q1 and MYD13Q1) and NDWI, NDII, and VIARI calculated from MOD09A1 reflectance data and in situ LFMC measurements from the National Fuel Moisture Database. The study ultimately selected EVI and introduced daily minimum temperature data to establish an empirical model for estimating LFMC in southern California. This choice was because EVI effectively captured changes in chlorophyll concentration related to water content, showing better response than other vegetation indices and shortwave infrared bands under water stress, particularly for Mediterranean vegetation. Introducing temperature data improved the model's ability to capture excessive water loss during extreme heat, correcting overestimates of minimum LFMC. Similarly, \cite{garcia2020live} found that EVI performed better than other ratio vegetation indices, such as NDII, NDVI, and VARI, for LFMC modeling in Eastern California and Oregon.

\cite{marino2020investigating} used NDVI, NDII, GVMI, NDWI, EVI, SAVI, VARI, and $VI_{green}$ from MCD43A4 and MOD09GA products, as well as the same indices calculated from Sentinel-2 data, to estimate the LFMC of Mediterranean shrubs. The main difference between MOD09GA and MCD43A4 and the MOD13Q1 and MYD13Q1 products used by \cite{myoung2018estimating} is that the former are daily data, which do not smooth out short-term LFMC changes. This feature is more conducive to capturing abnormal LFMC values under extreme conditions and to estimating LFMC for shrubs and grasslands that are more sensitive to short-term environmental changes. Similar to \cite{myoung2018estimating}, \cite{marino2020investigating} also found that using vegetation indices to estimate LFMC tends to overestimate lower LFMC values. However, \cite{marino2020investigating} found that VARI had the highest accuracy in LFMC modeling compared to other vegetation indices. Combining VARI with NDVI can further improve the accuracy of the model. This is because VARI and NDVI are based on visible (blue, green, red) and near-infrared reflectance data, which are suitable for measuring changes in greenness and LAI related to leaf drying and can indirectly reflect leaf water content. Similarly, \cite{stow2006time, peterson2008mapping, caccamo2011monitoring} found that VARI performed better than other vegetation indices or reflectance bands in estimating LFMC for shrubs in Southern California, Mediterranean climate regions, and southeastern Australia. Although these studies did not show that any particular vegetation index is the best choice in all scenarios, they emphasized the importance of chlorophyll (or greenness) response to water stress.

Water availability is a key parameter for plant transpiration. Therefore, changes in LFMC affect not only reflectance spectral bands but also canopy temperature (ST). When plants dry out, transpiration and latent heat transfer decrease, which in turn increases sensible heat \citep{kozlowskiphy, yebra2013global}. As a result, introducing thermal infrared bands (2500-14000 nm) into LFMC estimation can improve model performance under certain conditions. It is worth noting that in most studies, land surface temperature (LST) is used as a proxy for ST. \cite{chuvieco2004combining} used NDVI and ST data from the AVHRR sensor to estimate LFMC for grasslands and shrubs in Cabañeros National Park, central Spain, based on a linear regression model. The study found that NDVI and ST were positively and negatively correlated with LFMC, respectively, and that the introduction of ST significantly improved the model's performance. \cite{cunill2022live} combined MODIS spectral bands, vegetation indices, land surface temperature, and day of the year as predictors, using a random forest algorithm to estimate LFMC in the Mediterranean basin wildlands. They found that the combination of LST and day of the year had the highest importance, and that when the temperature exceeded 20°C, LST was negatively correlated with LFMC. However, in LFMC estimation experiments based on MODIS LST products, LST did not show a significant advantage over other factors. For example, \cite{sow2013estimation} compared the performance of the NDVI-to-LST ratio and GVMI, WI, NDWI, NDII, and MSI in estimating LFMC and equivalent water thickness (EWT) in three study areas in Senegal, finding that the NDVI-to-LST ratio had no significant advantage in correlating with LFMC and EWT compared to other vegetation indices. It is worth noting that this study did not compare the performance of using NDVI alone to model LFMC or EWT. \citep{mccandless2020enhancing} used machine learning methods to model LFMC across the continental United States using MODIS reflectance data, meteorological data, and terrain data. The random forest method was used to evaluate the importance of various predictors, finding that the importance of LST was only 6\%, lower than the importance of altitude (22\%), soil saturation (17\%), accumulated evapotranspiration (11\%), and slope (7\%). Moreover, some other studies also indicate that canopy temperature is an indicator of plant water stress \citep{gonzalez2012almond, rahimzadeh2012comparative, yebra2013global}.

Although spectral features can model canopy LFMC, they are more sensitive to dry matter such as canopy structure, resulting in lower robustness \citep{rao_sar-enhanced_2020}. Additionally, these spectral features couple dry matter and water content information, so when both change due to water stress, there can be issues with estimating LFMC using vegetation indices and other spectral features \citep{danson2004estimating, yebra_global_2013}. Moreover, spectral features are affected by lighting conditions and clouds or rain, making it difficult to provide full-time or near-real-time large-scale, long-term series LFMC mapping \citep{rao_sar-enhanced_2020}. Compared to spectral reflectance, microwave remote sensing is more capable of penetrating clouds and is more sensitive to water content \citep{rao_sar-enhanced_2020}. For example, active microwave sensors are sensitive to vegetation structure or dielectric properties (including vegetation moisture) through backscatter signals, and their penetration capability can be used to model soil moisture, thus monitoring vegetation water stress \citep{dorigo2017esa}. Passive microwave sensors can also be used to estimate LFMC through soil moisture time lags or vegetation optical depth (VOD) \citep{lu2021evaluation, forkel_estimating_2023}. Commonly used microwave sensors primarily operate in the P-band ($\approx$ 70 cm), L-band ($\approx$ 23 cm), C-band and X-band ($\approx$ 3 cm), and Ku-band ($\approx$ 1.6-2.5 cm). Among them, the backscatter signal of P and L bands is primarily coupled with backscatter from the soil and lower canopy, while the backscatter signal of C, X, and Ku bands mainly comes from the interaction between microwave radiation and the upper canopy.

\cite{tanase2015monitoring} used vertical-vertical (VV), vertical-horizontal (VH), horizontal-vertical (HV), and horizontal-horizontal (HH) polarization L-band radar backscatter signals to model the relationship between backscatter intensity, polarization decomposition signals, radar vegetation index (RVI) \citep{6112792}, and canopy height with LFMC and EWT. Experiments in Australian forests demonstrated that the modeling capability of L-band radar data was superior to spectral reflectance data. The penetration of L-band data enables modeling of the lower canopy and surface fine fuel moisture content. However, when the forest is high and dense, and both lower canopy and soil moisture are low, there will be a large error in estimating LFMC for the lower canopy. \cite{wang2019assessment} explored the performance of C-band Sentinel-1A (5.4 GHz) VV and VH polarization data in estimating LFMC in Texas forests. The model coupled an empirical bare soil backscatter model with the Water Cloud Model (WCM) \citep{de2001semi} to establish an LFMC lookup table, directly estimating LFMC, and compared this LFMC with results obtained from Landsat 8 OLI spectral reflectance data using the empirical partial least squares (PLSR) method and measured LFMC data. The experiment showed that the accuracy of LFMC inversion based on Sentinel-1A data was higher than that based on Landsat 8 OLI data and that it could reproduce the temporal variation trend of LFMC. \cite{lu_evaluation_2021} used multi-source microwave soil moisture products ECV\_SM and SMAP L-band Level 4 surface and root zone soil moisture data, using a robust linear regression model with a time lag to assess LFMC across the continental United States. The study showed that LFMC was most correlated with soil moisture conditions 60 days before LFMC sampling in most regions.

Considering the influence of soil moisture content and canopy dry matter on radar backscatter, reflectance, or emission signal intensity, some studies have integrated SAR and multispectral reflectance data, along with vegetation cover, canopy height, and soil moisture data, to estimate LFMC. For instance, \cite{jia2019estimating} employed soil moisture data obtained from the SMAP L-band radiometer, combined with VARI calculated from MODIS MCD43A4 reflectance data, growing degree days (GDD), and precipitation, to estimate LFMC in Southern California using a regression model. This study found that using cumulative soil moisture data combined with cumulative growing degree days yielded better performance in LFMC estimation compared to using MODIS VARI combined with cumulative growing degree days. The superior performance is attributed to a more accurate representation of the water cycle processes between vegetation and soil.

\cite{rao_sar-enhanced_2020} designed a knowledge-assisted neural network to estimate LFMC in the western United States, utilizing Sentinel-1 VV and VH polarized SAR backscatter data, Landsat-8 red, green, blue, near-infrared (NIR), and shortwave infrared (SWIR) optical reflectance data, along with NDVI, NDWI, and near-infrared spectral indices derived from the reflectance data. To separate the contributions of vegetation and soil moisture to the backscatter signal, the study proposed a knowledge-based method and a data-driven approach. The former method leverages the differential sensitivity of VV and VH polarizations to soil and vegetation moisture, using the difference between VV and VH polarizations to separate the contributions of soil and vegetation to backscatter intensity. The latter approach involves training the deep learning model to learn the separation of vegetation moisture and soil moisture contributions by incorporating soil type as an input. The study also established a ratio of backscatter signals to spectral reflectance, utilizing the high water sensitivity of backscatter signals and the high spectral reflectance of dry matter, alongside the aforementioned vegetation indices to separate the contributions of dry matter and water content to LFMC estimation.

Similarly, \cite{xie2022retrieval} employed Sentinel-1 data along with MODIS MCD43A4 nadir BRDF-adjusted reflectance (NBAR) data, original band reflectance data from Landsat-8 OLI (red, green, blue, NIR, and SWIR channels), and vegetation indices such as NDVI, NIRV, and NDWI. The study used canopy height and land cover type as auxiliary data to estimate LFMC in the western United States. These data were interpolated to unify spatial and temporal resolutions before being input into an integrated model, where LFMC was predicted through end-to-end training.

Furthermore, \cite{forkel_estimating_2023} achieved global LFMC estimation by combining Ku-band, X-band, and C-band VOD data derived from passive microwave technology with MODIS LAI, AVHRR vegetation cover data, LULC, and meteorological and topographical data. This study demonstrated that the combination of Ku-VOD and LAI as predictors significantly improved LFMC estimation performance. Conversely, \cite{chaparro2024vegetation} conducted LFMC inversion experiments in the western United States, showing that when a model effectively separates vegetation structure and dry matter information, the accuracy of LFMC estimation based on VOD is independent of canopy height, land cover type, and radar backscatter signals.

While the aforementioned LFMC estimates based on multispectral and microwave remote sensing data have achieved satisfactory accuracy, it is noteworthy that some studies suggest that the inclusion of microwave data does not necessarily enhance LFMC inversion accuracy. For example, \cite{tanase2022characterizing} explored the estimation of LFMC in Mediterranean vegetation using C-band Sentinel-1 and Sentinel-2 multispectral data combined with canopy height, vegetation cover, and topographical data. The study found that incorporating radar data did not improve model performance and thus only recommended its use in areas with significant cloud cover.

Research has also been conducted on the effectiveness of different data sources for LFMC estimation. For instance, \cite{fan2018evaluation} compared the performance of LFMC estimation in southern France using ECV\_SM, root zone soil moisture from ECV\_SM, microwave polarization difference index (MPDI) calculated from C, X, Ku, K, and Ka band passive microwave data at 6.9, 10.7, 18.7, 23.8, and 36.5 GHz, VOD derived from AMSR-E C and X band data, and vegetation indices such as NDVI, SAVI, VARI, NDWI, NDII, and GVMI calculated from MODIS data. The MPDI and VOD were averaged over 15 days prior to in-situ measurements. The study found that the optical vegetation indices VARI and SAVI outperformed the microwave-derived indices and other optical indices. Among the microwave indices, X-band VOD showed the highest consistency with in-situ measured LFMC.

\subsection{Fuel Load}

The methods for estimating fuel load are similar to those for estimating FMC, including direct sampling and the use of physical models or multispectral and microwave remote sensing data. Direct measurement involves collecting field samples, weighing, drying them, and then estimating biomass in the area, which is considered the most accurate assessment method and can provide a data foundation for physical or empirical model evaluations. First, fuel load estimation methods based on ground-measured data combined with vegetation or forest growth models can predict or estimate fuel load by forecasting future vegetation growth, forest structure, and more. For example, \cite{lee2022prediction} used a Weibull function and mortality model based on forest inventory data to predict the fuel load of Korean red pine (Pinus densiflora) for the next 20 years in Korea. Meanwhile, \cite{nolan2022framework} modeled above-ground biomass in the shrub layer of eucalyptus forests in southeastern Australia using allometric equations, considering it as fuel load. 

While ground sampling offers high accuracy, it is inefficient and costly. In contrast, remote sensing technology provides an efficient and comprehensive method for assessing fuel load. For example, \cite{saatchi_estimation_2007} used airborne SAR data and a semi-empirical algorithm to estimate the distribution of forest biomass and canopy fuel load in Yellowstone National Park. Additionally, \cite{duff_predicting_2012} developed a biophysical model to explore the feasibility of using satellite-based NDVI to estimate forest fuel load in southeastern Australia. \cite{li2020assessment} estimated changes in live fuel load in shrubs and grasslands in the Owyhee Basin, southern Idaho, using MODIS NPP products combined with the Biome-BGC model, and calculated changes in dead fuel load based on the estimates. The study found that even with the same productivity, the fuel load of shrubs was higher than that of grasslands.

In addition to fuel load estimation using physical or semi-empirical models based on remote sensing data, empirical model-based methods are simpler and more commonly used. For instance, \cite{arellano2018potential} used the random forest method and Sentinel-2 multispectral and vegetation index data to estimate the canopy fuel load of Pinus radiata and Pinus pinaster stands in northwestern Spain. However, due to the low penetration capability of optical sensors, the accuracy of canopy fuel load estimation in this study was limited. Furthermore, \cite{li2021forest} compared the effectiveness of L-band SAR (ALOS-PALSAR) and Landsat 7 ETM+ in estimating fuel load in coniferous mixed forests and found that L-band SAR outperformed optical data in all three fuel load types, with estimation accuracy further improved when both were integrated. \cite{li_forest_2022} used a data-driven random forest model to correlate long-term spatiotemporal characteristics derived from Landsat 8 OLI and Sentinel-1 data with field-measured fuel load, estimating the fine fuel load (FFL) of pine forests in some areas of Sichuan Province, China. The study found that combining SAR and multispectral data improved FFL estimation accuracy, and temporal features were more important than spatial features.

The comparison between optical and SAR sensor-based fuel load estimation accuracy highlights the importance of canopy structure information in remote sensing-based fuel load estimation. Compared to the fuel load assessment methods based on optical or SAR imagery mentioned earlier, airborne LiDAR provides more detailed information about vegetation's three-dimensional structure \citep{aragoneses2024mapping}. Therefore, airborne and satellite-based LiDAR sensors can more directly and accurately estimate fuel load at different scales. For instance, \cite{skowronski_three-Dimensional_2011} used a LiDAR system to estimate canopy fuel load in pine-dominated forests in New Jersey, USA, predicting crown bulk density and canopy fuel weight. \cite{chen_development_2017} effectively estimated ground fuel load in Western Australia's Jarrah forests by combining LiDAR data with previous year fire data, canopy density, elevation, and fuel types. \cite{bright2017prediction} compared the performance of airborne LiDAR and Landsat time-series data in estimating forest canopy and surface fuel load in a study area in Grand County, Colorado, USA, finding that Landsat provided limited improvement in fuel load estimation. \cite{labenski_quantifying_2023} further combined Sentinel-2 multispectral data with airborne LiDAR data to extract forest structure and composition information for two forest areas in southwestern Germany, using a random forest model to predict and map surface fuel load. In another study, \cite{d2021machine} incorporated Sentinel-1 SAR data into a fuel load estimation model based on Sentinel-2 multispectral data and airborne LiDAR data for southeastern Italy, estimating dead fine fuel load. The study found that compared to NDVI and C-band SAR data, LiDAR variables had a stronger capacity to estimate dead fine fuel load. The study also found that woody vegetation contributed more to fuel load, consistent with the findings of \cite{li2020assessment}.

In addition, the more recent NASA GEDI full-waveform LiDAR sensor has been widely used for large-scale above-ground biomass estimation and vegetation canopy structure analysis \citep{myroniuk2023combining, liang2023quantifying, aragoneses2024mapping, guerra2024impact}, aiding wildfire simulation, assessment of wildfire impacts on vegetation, and understanding wildfire severity. However, only a few studies have directly used GEDI LiDAR data to estimate fuel load. For example, \cite{leite2022large} used unmanned aerial vehicle (UAV) simulations and GEDI data to study the ability to stratify fuel load in grasslands, savannas, and forests within tropical savannas in Brazil. The study found that GEDI performed poorly in estimating fuel load for ground and low-lying vegetation, and was significantly affected by terrain.

It is important to note that the aforementioned studies do not explicitly distinguish between live and dead fuel. Notably, dead fuel load is a critical component of forest ecosystems and a key determinant of wildfire behavior \citep{rothermel1972fire, woodall2013biomass}, warranting special attention. In particular, fine dead fuel is emphasized because it dries rapidly and is highly flammable, making it the primary carrier of fire ignition and spread \citep{cruz2003assessing, elia2020likelihood, d2021machine}. Additionally, dead and down woody debris (DWD) is another significant contributor to wildfires, particularly influencing surface fire intensity and the transition from surface to crown fire \citep{alexander1982calculating, hanes2021dead}. Once surface fires reach the canopy, wildfire intensity and suppression difficulty increase substantially under high wind speeds \citep{schroeder1977fire, hanes2021dead}. However, estimating dead fuel load remains challenging. Apart from fine fuels such as annual grasses in open grasslands, forest components like litter, duff, and DWD are often invisible in optical remote sensing data, particularly in dense forest regions.

\cite{d2021machine} noted that dead fuel load estimation is typically conducted using standard fuel parameter models \citep{rothermel1972fire, ottmar2007overview} or field sampling. However, our investigation reveals that alternative approaches, including estimation based on net primary productivity (NPP) or above-ground biomass (AGB), field sampling with extrapolation, and remote sensing-based retrieval, are also widely employed.

The theoretical basis for estimating fuel load using NPP lies in the fact that NPP represents the rate of vegetation growth and is equivalent to litter production \citep{matthews1997global}. Based on this principle, \cite{clarke2016investigation} used the BIOS2 modeling environment to simulate fuel load (fine litter) and NPP in Australia. To establish the relationship between these variables, Australia was divided into six climatic regions, and the Pearson correlation coefficient was used to determine the statistical relationship between annual NPP and annual fine litter in each region. Subsequently, the Community Atmosphere-Biosphere Land Exchange (CABLE) land surface model was used to simulate NPP under various climate scenarios. Using the previously derived NPP–fuel load relationship, fuel load was estimated from NPP across the study regions. Finally, the estimated fuel load values were expressed in g C m$^{-2}$ and converted to t ha$^{-1}$, assuming a carbon fraction of 47\%.

Furthermore, \cite{mcnorton_global_2024} developed a long-term global-scale fuel load dataset by integrating the Net Ecosystem Exchange (NEE) model with satellite observations. The study first used the ESA Biomass Climate Change Initiative (ESA-CCI) v3 dataset as the initial global above-ground biomass (AGB) estimate for 2010. Subsequently, AGB was dynamically updated using the ECLand land surface model, ERA-5 meteorological data, and ESA-CCI AGB changes from 2010 to 2018.  The model further partitions AGB into foliage and wood and classifies each component into live and dead fractions. A static live-to-dead biomass ratio is assigned to each of the 20 vegetation types in the ECLand model based on literature estimates. Finally, the dead foliage and dead wood fuel loads are estimated using the following equations:

\begin{equation}
TF_{dead} = \frac{DFP}{100} \times \left( AGB - (TW_{live} + TF_{live} + SW_{live}) \right)
\end{equation}

\begin{equation}
TW_{dead} = \left( 1 - \frac{DFP}{100} \right) \times \left( AGB - (TW_{live} + TF_{live} + SW_{live}) \right)
\end{equation}
where \( TF_{dead} \) and \( TW_{dead} \) represent the dead foliage and dead wood fuel loads (kg m$^{-2}$), respectively. \( AGB \) denotes the above-ground biomass (kg m$^{-2}$), obtained from satellite observations and land surface models. \( TW_{live} \) and \( TF_{live} \) correspond to the live wood and live foliage fuel loads, respectively, with \( TW_{live} \) representing short-lived seasonal wood mass (e.g., twigs) and \( TF_{live} \) determined using leaf mass per unit area (LMA) and LAI. \( SW_{live} \) refers to live structural wood, including long-lived biomass such as tree trunks. The parameter \( DFP \) represents the dead foliage percentage, which is vegetation-type dependent and remains constant over time. 

These approaches establish relationships between NPP or AGB and fuel load; however, they often rely on idealized assumptions. For instance, \cite{clarke2016investigation} assumed a linear relationship between NPP and fine litter while neglecting mechanistic changes in litterfall and decomposition. Similarly, \cite{mcnorton_global_2024} adopted static live-to-dead biomass ratios and simplified fuel decomposition processes. In contrast, although field sampling methods are expensive and time-consuming, they provide more accurate estimates of dead fuel load. 

To improve estimation accuracy, some studies have integrated field sampling with allometric equations to quantify dead fuel load for specific species. For example, \cite{deltell2024allometric} conducted an investigation in Ayora, eastern Valencia Province on the Iberian Peninsula, focusing on nine Mediterranean species at different developmental stages. They measured basal stem diameter, height, maximum crown diameter, the perpendicular diameter to the maximum crown diameter, and biomass for both live and dead fuels. Using plant structural variables as predictors, species- and development-stage-specific allometric equations were constructed to estimate live fuel, dead fuel, and total fuel load.

In recent years, some countries have conducted large-scale national inventories of dead fuel load. \cite{woodall2013biomass} presented results from a nationwide forest downed and dead woody materials (DWM) inventory in the United States, using FIA plots and systematic sampling. Their study summarized total carbon storage and biomass density by state and common tree species while assessing relationships between DWM and factors such as climate, aboveground live biomass, and relative density of live trees. Similarly, \cite{hanes2021dead} conducted field measurements of DWD fuel load across 766 National Forest Inventory plots and additional sites across Canada. Their study recorded key variables, including latitude, tree age, species, diameter at breast height (DBH), stand density, and soil drainage class. Using field sampling data combined with these explanatory variables and bioclimatic factors, they developed a multiple linear regression model to generate a nationwide distribution map of DWD fuel load across different forest types and ecological zones in Canada.

Large-scale field inventories of dead fuel load provide reliable data for understanding fuel characteristics but are not suitable for real-time wildfire risk prediction. Additionally, extrapolation models often rely on idealized assumptions, leading to inaccuracies in dead fuel load estimation \citep{hanes2021dead}. Consequently, recent studies have explored the use of remote sensing data to provide real-time and repeatable dead fuel load estimates for regional scale. Fundamentally, these studies aim to establish relationships between ground inventory data and one or more remote sensing datasets or derived indices.

Passive optical remote sensing images cannot retrieve fuel loads beneath the canopy due to vegetation occlusion and are unable to provide estimates for horizontal and vertical fuel layers within forest stands. Therefore, LiDAR, which captures vegetation's three-dimensional (3D) structure, is widely used for fuel load estimation. For instance, \cite{lopes2020estimating} proposed a method for estimating coarse woody debris (CWD) volume by integrating high-resolution optical imagery (5 cm GSD) with multispectral LiDAR (including SWIR, NIR, and green channels). Their approach involved first segmenting optical imagery using a random forest classifier to extract visible CWD layers (vCWD) and then estimating canopy-covered CWD distribution using LiDAR-derived active vegetation indices (aNDVI, aNBR). A two-stage zero-adjusted gamma (ZAGA) model was constructed, where a binomial submodel (input variables: vCWD, wetland probability) predicted the probability of CWD occurrence, and a continuous submodel (input variables: vCWD², canopy closure) estimated CWD volume, optimized using the AICc criterion. Experimental results demonstrated that incorporating LiDAR significantly improved model performance compared to optical imagery alone, achieving $R^2$ = 0.72 (RMSE = 0.198 m³/100m²) in independent validation. However, limitations remained, including noise in aNDVI due to the low point density (1.51 pts/m²) of the LiDAR green channel and missing ground points in areas with high canopy closure. Additionally, the model's reliance on wetland probability and maximum tree height variables requires further validation in heterogeneous forests.

\cite{lin2024use} further investigated the capability of airborne LiDAR alone for estimating four types of dead fuel loads (1-hour, 10-hour, 100-hour, and litter) in southern Italy's Apulia region. Their study developed multiple linear regression models relating airborne LiDAR-derived height and canopy cover indices to field-sampled dead fuel loads. Regression analysis results showed a predictive $R^2$ between 0.521 and 0.569, with nRMSE values ranging from 0.110 to 0.169. These findings suggest that airborne LiDAR may not accurately estimate dead fuel load, particularly given the study's LiDAR resolution of 3 pts/m² and the influence of fine fuel size, terrain complexity, environmental factors, and the simplicity of the multiple linear regression model. However, the study identified canopy structural parameters as critical predictors of dead fuel load, consistent with the findings of \cite{lopes2020estimating, hanes2021dead}, reinforcing the importance of vegetation 3D structure in fuel load estimation.

Building on airborne LiDAR data, \cite{d2021machine} incorporated Sentinel-2 multispectral and Sentinel-1 SAR data to estimate 1-hour dead fine fuel load in southeastern Italy's Apulia region. Sentinel-1 SAR's VH and VV polarization data were used for surface reflectance analysis, Sentinel-2 multispectral data for vegetation index calculations, and airborne LiDAR for generating canopy height models (CHM), canopy cover, and vegetation density. RF, support vector machine (SVM), and multiple linear regression (MLR) models were developed to estimate 1-hour dead fine fuel load using these explanatory variables. The study found that, compared to NDVI and C-band SAR data, LiDAR-derived variables (canopy height and cover) contributed more to fine fuel load estimation. Among the models, RF performed best, achieving RMSE = 0.09 t/ha, $R^2$ = 0.50, and a Pearson correlation coefficient $r$ = 0.71. 

Both \cite{lin2024use} and \cite{d2021machine} highlight that even with airborne LiDAR, model performance may not meet the high accuracy requirements for fine dead fuel load estimation. However, compared to \cite{lopes2020estimating}, which focused on larger dead fuels (CWD), LiDAR demonstrated higher accuracy in retrieving CWD volume ($R^2$ = 0.72, RMSE = 0.198 m³/100m²). These findings suggest that while LiDAR effectively estimates coarse woody debris, its performance in fine dead fuel load retrieval remains limited, necessitating further improvements in data integration and modeling approaches.

Compared to dead fuel load estimation methods that rely solely on or integrate LiDAR, \cite{9553105} highlighted that LiDAR is cost-prohibitive, making it difficult to analyze fuel dynamics across spatial and temporal scales. To address this limitation, their study in southwestern Sichuan, China, explored the use of Sentinel-1 VH and VV polarization data, Sentinel-2 reflectance data, and topographic variables (elevation, slope, and aspect) to estimate five types of forest dead fuel load (FDFL)—1-hour, 10-hour, 100-hour, litter, and the total dead fuel load—using a random forest regression approach. The results indicated that 1-hour and 10-hour FDFL exhibited the highest correlation with SAR data, while 100-hour FDFL showed similar correlations with both SAR and optical data. Litter was more strongly associated with Sentinel-2 data, whereas the total FDFL was highly negatively correlated with Sentinel-1 VH polarization and visible bands. In terms of predictive performance, 1-hour fuel exhibited the highest accuracy with $R^2$ = 0.57 and MSE = 0.18 tons/ha, while 10-hour and 100-hour fuel achieved $R^2$ values of 0.40 and 0.29, respectively. Overall, C-band SAR combined with multispectral data was still insufficient for accurate dead fuel load estimation.

From the above review, it is evident that whether using AGB-, NPP-, or allometric equation-based retrieval methods, field sampling extrapolation, or LiDAR-based inversion, all these approaches fundamentally estimate dead fuel load indirectly through vegetation biomass or three-dimensional structural information (e.g., canopy cover, canopy height, leaf area index (LAI), gap fraction, and understory density), rather than directly capturing the three-dimensional or spectral characteristics of dead fuel load for quantitative estimation. This limitation likely stems from the inability of passive optical imagery to penetrate the canopy, restricting its capacity to retrieve sub-canopy fuel distribution. Meanwhile, although LiDAR and SAR offer all-weather capabilities, they exhibit low sensitivity to fine dead fuels, with SAR further constrained by environmental factors such as soil moisture. Consequently, direct estimation of dead fuel load is expected to remain a significant challenge in the future. A more feasible solution under data-limited conditions may be to develop a comprehensive understanding of the relationships between climate, topography, vegetation type, structure, and developmental stages with dead fuel load.

%Thanks to ...

%% The Appendices part is started with the command \appendix;
%% appendix sections are then done as normal sections
%\appendix

%\section{Appendix title 1}
%% \label{}

%\section{Appendix title 2}
%% \label{}

%% If you have bibdatabase file and want bibtex to generate the
%% bibitems, please use
%%
\bibliographystyle{elsarticle-harv} 
% \bibliography{main_ref_new}
\bibliography{ref_new}

\begin{thebibliography}{503}
\expandafter\ifx\csname natexlab\endcsname\relax\def\natexlab#1{#1}\fi
\providecommand{\url}[1]{\texttt{#1}}
\providecommand{\href}[2]{#2}
\providecommand{\path}[1]{#1}
\providecommand{\DOIprefix}{doi:}
\providecommand{\ArXivprefix}{arXiv:}
\providecommand{\URLprefix}{URL: }
\providecommand{\Pubmedprefix}{pmid:}
\providecommand{\doi}[1]{\href{http://dx.doi.org/#1}{\path{#1}}}
\providecommand{\Pubmed}[1]{\href{pmid:#1}{\path{#1}}}
\providecommand{\bibinfo}[2]{#2}
\ifx\xfnm\relax \def\xfnm[#1]{\unskip,\space#1}\fi
%Type = Article
\bibitem[{Abatzoglou and Kolden(2011)}]{abatzoglou2011relative}
\bibinfo{author}{Abatzoglou, J.T.}, \bibinfo{author}{Kolden, C.A.}, \bibinfo{year}{2011}.
\newblock \bibinfo{title}{Relative importance of weather and climate on wildfire growth in interior alaska}.
\newblock \bibinfo{journal}{International Journal of Wildland Fire} \bibinfo{volume}{20}, \bibinfo{pages}{479--486}.
%Type = Article
\bibitem[{Abatzoglou and Kolden(2013)}]{abatzoglou2013relationships}
\bibinfo{author}{Abatzoglou, J.T.}, \bibinfo{author}{Kolden, C.A.}, \bibinfo{year}{2013}.
\newblock \bibinfo{title}{Relationships between climate and macroscale area burned in the western united states}.
\newblock \bibinfo{journal}{International Journal of Wildland Fire} \bibinfo{volume}{22}, \bibinfo{pages}{1003--1020}.
%Type = Article
\bibitem[{Abatzoglou et~al.(2016)Abatzoglou, Kolden, Balch and Bradley}]{abatzoglou2016controls}
\bibinfo{author}{Abatzoglou, J.T.}, \bibinfo{author}{Kolden, C.A.}, \bibinfo{author}{Balch, J.K.}, \bibinfo{author}{Bradley, B.A.}, \bibinfo{year}{2016}.
\newblock \bibinfo{title}{Controls on interannual variability in lightning-caused fire activity in the western us}.
\newblock \bibinfo{journal}{Environmental Research Letters} \bibinfo{volume}{11}, \bibinfo{pages}{045005}.
%Type = Article
\bibitem[{Abdollahi and Pradhan(2023)}]{abdollahi_explainable_2023}
\bibinfo{author}{Abdollahi, A.}, \bibinfo{author}{Pradhan, B.}, \bibinfo{year}{2023}.
\newblock \bibinfo{title}{Explainable artificial intelligence ({XAI}) for interpreting the contributing factors feed into the wildfire susceptibility prediction model}.
\newblock \bibinfo{journal}{SCIENCE OF THE TOTAL ENVIRONMENT} \bibinfo{volume}{879}, \bibinfo{pages}{163004}.
\newblock \URLprefix \url{https://www.webofscience.com/api/gateway?GWVersion=2&SrcAuth=DynamicDOIArticle&SrcApp=WOS&KeyAID=10.1016%2Fj.scitotenv.2023.163004&DestApp=DOI&SrcAppSID=USW2EC0D7EfQ4jdthTkoDVukmMmGx&SrcJTitle=SCIENCE+OF+THE+TOTAL+ENVIRONMENT&DestDOIRegistrantName=Elsevier}, \DOIprefix\doi{10.1016/j.scitotenv.2023.163004}. \bibinfo{note}{num Pages: 14 Place: Amsterdam Publisher: Elsevier Web of Science ID: WOS:000973651700001}.
%Type = Article
\bibitem[{Achu et~al.(2021)Achu, Thomas, Aju, Gopinath, Kumar and Reghunath}]{achu_machine-learning_2021}
\bibinfo{author}{Achu, A.L.}, \bibinfo{author}{Thomas, J.}, \bibinfo{author}{Aju, C.D.}, \bibinfo{author}{Gopinath, G.}, \bibinfo{author}{Kumar, S.}, \bibinfo{author}{Reghunath, R.}, \bibinfo{year}{2021}.
\newblock \bibinfo{title}{Machine-learning modelling of fire susceptibility in a forest-agriculture mosaic landscape of southern {India}}.
\newblock \bibinfo{journal}{Ecological Informatics} \bibinfo{volume}{64}, \bibinfo{pages}{101348}.
\newblock \URLprefix \url{https://www.sciencedirect.com/science/article/pii/S1574954121001394}, \DOIprefix\doi{10.1016/j.ecoinf.2021.101348}.
%Type = Article
\bibitem[{Addison and Oommen(2018)}]{addison2018utilizing}
\bibinfo{author}{Addison, P.}, \bibinfo{author}{Oommen, T.}, \bibinfo{year}{2018}.
\newblock \bibinfo{title}{Utilizing satellite radar remote sensing for burn severity estimation}.
\newblock \bibinfo{journal}{International journal of applied earth observation and geoinformation} \bibinfo{volume}{73}, \bibinfo{pages}{292--299}.
%Type = Article
\bibitem[{Adhikari et~al.(2021)Adhikari, Xu, Hodza and Minckley}]{adhikari_developing_2021}
\bibinfo{author}{Adhikari, B.}, \bibinfo{author}{Xu, C.}, \bibinfo{author}{Hodza, P.}, \bibinfo{author}{Minckley, T.}, \bibinfo{year}{2021}.
\newblock \bibinfo{title}{Developing a geospatial data-driven solution for rapid natural wildfire risk assessment}.
\newblock \bibinfo{journal}{Applied Geography} \bibinfo{volume}{126}, \bibinfo{pages}{102382}.
\newblock \URLprefix \url{https://www.sciencedirect.com/science/article/pii/S0143622820314818}, \DOIprefix\doi{10.1016/j.apgeog.2020.102382}.
%Type = Article
\bibitem[{Agee and Skinner(2005)}]{agee2005basic}
\bibinfo{author}{Agee, J.K.}, \bibinfo{author}{Skinner, C.N.}, \bibinfo{year}{2005}.
\newblock \bibinfo{title}{Basic principles of forest fuel reduction treatments}.
\newblock \bibinfo{journal}{Forest ecology and management} \bibinfo{volume}{211}, \bibinfo{pages}{83--96}.
%Type = Article
\bibitem[{Ager et~al.(2014a)Ager, Day, McHugh, Short, Gilbertson-Day, Finney and Calkin}]{ager_wildfire_2014}
\bibinfo{author}{Ager, A.A.}, \bibinfo{author}{Day, M.A.}, \bibinfo{author}{McHugh, C.W.}, \bibinfo{author}{Short, K.}, \bibinfo{author}{Gilbertson-Day, J.}, \bibinfo{author}{Finney, M.A.}, \bibinfo{author}{Calkin, D.E.}, \bibinfo{year}{2014}a.
\newblock \bibinfo{title}{Wildfire exposure and fuel management on western {US} national forests}.
\newblock \bibinfo{journal}{Journal of Environmental Management} \bibinfo{volume}{145}, \bibinfo{pages}{54--70}.
\newblock \URLprefix \url{https://linkinghub.elsevier.com/retrieve/pii/S0301479714002916}, \DOIprefix\doi{10.1016/j.jenvman.2014.05.035}.
%Type = Article
\bibitem[{Ager et~al.(2014b)Ager, Preisler, Arca, Spano and Salis}]{ager2014wildfire}
\bibinfo{author}{Ager, A.A.}, \bibinfo{author}{Preisler, H.K.}, \bibinfo{author}{Arca, B.}, \bibinfo{author}{Spano, D.}, \bibinfo{author}{Salis, M.}, \bibinfo{year}{2014}b.
\newblock \bibinfo{title}{Wildfire risk estimation in the mediterranean area}.
\newblock \bibinfo{journal}{Environmetrics} \bibinfo{volume}{25}, \bibinfo{pages}{384--396}.
%Type = Article
\bibitem[{Ager et~al.(2010)Ager, Vaillant and Finney}]{ager_comparison_2010}
\bibinfo{author}{Ager, A.A.}, \bibinfo{author}{Vaillant, N.M.}, \bibinfo{author}{Finney, M.A.}, \bibinfo{year}{2010}.
\newblock \bibinfo{title}{A comparison of landscape fuel treatment strategies to mitigate wildland fire risk in the urban interface and preserve old forest structure}.
\newblock \bibinfo{journal}{Forest Ecology and Management} \bibinfo{volume}{259}, \bibinfo{pages}{1556--1570}.
\newblock \URLprefix \url{https://www.sciencedirect.com/science/article/pii/S0378112710000514}, \DOIprefix\doi{10.1016/j.foreco.2010.01.032}.
%Type = Book
\bibitem[{Albini(1976)}]{albini1976estimating}
\bibinfo{author}{Albini, F.A.}, \bibinfo{year}{1976}.
\newblock \bibinfo{title}{Estimating wildfire behavior and effects}. volume~\bibinfo{volume}{30}.
\newblock \bibinfo{publisher}{Department of Agriculture, Forest Service, Intermountain Forest and Range~…}.
%Type = Article
\bibitem[{Aldersley et~al.(2011)Aldersley, Murray and Cornell}]{aldersley_global_2011}
\bibinfo{author}{Aldersley, A.}, \bibinfo{author}{Murray, S.J.}, \bibinfo{author}{Cornell, S.E.}, \bibinfo{year}{2011}.
\newblock \bibinfo{title}{Global and regional analysis of climate and human drivers of wildfire}.
\newblock \bibinfo{journal}{Science of The Total Environment} \bibinfo{volume}{409}, \bibinfo{pages}{3472--3481}.
\newblock \URLprefix \url{https://www.sciencedirect.com/science/article/pii/S0048969711005353}, \DOIprefix\doi{10.1016/j.scitotenv.2011.05.032}.
%Type = Article
\bibitem[{Alexander(1982)}]{alexander1982calculating}
\bibinfo{author}{Alexander, M.E.}, \bibinfo{year}{1982}.
\newblock \bibinfo{title}{Calculating and interpreting forest fire intensities}.
\newblock \bibinfo{journal}{Canadian Journal of Botany} \bibinfo{volume}{60}, \bibinfo{pages}{349--357}.
%Type = Article
\bibitem[{Ali et~al.(2023)Ali, Abuhmed, El-Sappagh, Muhammad, Alonso-Moral, Confalonieri, Guidotti, Del~Ser, D{\'\i}az-Rodr{\'\i}guez and Herrera}]{ali2023explainable}
\bibinfo{author}{Ali, S.}, \bibinfo{author}{Abuhmed, T.}, \bibinfo{author}{El-Sappagh, S.}, \bibinfo{author}{Muhammad, K.}, \bibinfo{author}{Alonso-Moral, J.M.}, \bibinfo{author}{Confalonieri, R.}, \bibinfo{author}{Guidotti, R.}, \bibinfo{author}{Del~Ser, J.}, \bibinfo{author}{D{\'\i}az-Rodr{\'\i}guez, N.}, \bibinfo{author}{Herrera, F.}, \bibinfo{year}{2023}.
\newblock \bibinfo{title}{Explainable artificial intelligence (xai): What we know and what is left to attain trustworthy artificial intelligence}.
\newblock \bibinfo{journal}{Information fusion} \bibinfo{volume}{99}, \bibinfo{pages}{101805}.
%Type = Article
\bibitem[{Allaire et~al.(2021)Allaire, Mallet and Filippi}]{allaire2021emulation}
\bibinfo{author}{Allaire, F.}, \bibinfo{author}{Mallet, V.}, \bibinfo{author}{Filippi, J.B.}, \bibinfo{year}{2021}.
\newblock \bibinfo{title}{Emulation of wildland fire spread simulation using deep learning}.
\newblock \bibinfo{journal}{Neural networks} \bibinfo{volume}{141}, \bibinfo{pages}{184--198}.
%Type = Misc
\bibitem[{Alonso et~al.(2023)Alonso, Gans, Karasante, Ahuja, Prapas, Kondylatos, Papoutsis, Panagiotou, Mihail, Cremer, Weber and Carvalhais}]{alonso_2023_8055879}
\bibinfo{author}{Alonso, L.}, \bibinfo{author}{Gans, F.}, \bibinfo{author}{Karasante, I.}, \bibinfo{author}{Ahuja, A.}, \bibinfo{author}{Prapas, I.}, \bibinfo{author}{Kondylatos, S.}, \bibinfo{author}{Papoutsis, I.}, \bibinfo{author}{Panagiotou, E.}, \bibinfo{author}{Mihail, D.}, \bibinfo{author}{Cremer, F.}, \bibinfo{author}{Weber, U.}, \bibinfo{author}{Carvalhais, N.}, \bibinfo{year}{2023}.
\newblock \bibinfo{title}{{SeasFire Cube: A Global Dataset for Seasonal Fire Modeling in the Earth System}}.
\newblock \URLprefix \url{https://doi.org/10.5281/zenodo.8055879}, \DOIprefix\doi{10.5281/zenodo.8055879}.
%Type = Article
\bibitem[{Andela et~al.(2019)Andela, Morton, Giglio, Paugam, Chen, Hantson, Van Der~Werf and Randerson}]{andela2019global}
\bibinfo{author}{Andela, N.}, \bibinfo{author}{Morton, D.C.}, \bibinfo{author}{Giglio, L.}, \bibinfo{author}{Paugam, R.}, \bibinfo{author}{Chen, Y.}, \bibinfo{author}{Hantson, S.}, \bibinfo{author}{Van Der~Werf, G.R.}, \bibinfo{author}{Randerson, J.T.}, \bibinfo{year}{2019}.
\newblock \bibinfo{title}{The global fire atlas of individual fire size, duration, speed and direction}.
\newblock \bibinfo{journal}{Earth System Science Data} \bibinfo{volume}{11}, \bibinfo{pages}{529--552}.
%Type = Article
\bibitem[{Andersen et~al.(2005)Andersen, McGaughey and Reutebuch}]{andersen2005estimating}
\bibinfo{author}{Andersen, H.E.}, \bibinfo{author}{McGaughey, R.J.}, \bibinfo{author}{Reutebuch, S.E.}, \bibinfo{year}{2005}.
\newblock \bibinfo{title}{Estimating forest canopy fuel parameters using lidar data}.
\newblock \bibinfo{journal}{Remote sensing of Environment} \bibinfo{volume}{94}, \bibinfo{pages}{441--449}.
%Type = Article
\bibitem[{Aragoneses et~al.(2024)Aragoneses, Garc{\'\i}a, Ruiz-Benito and Chuvieco}]{aragoneses2024mapping}
\bibinfo{author}{Aragoneses, E.}, \bibinfo{author}{Garc{\'\i}a, M.}, \bibinfo{author}{Ruiz-Benito, P.}, \bibinfo{author}{Chuvieco, E.}, \bibinfo{year}{2024}.
\newblock \bibinfo{title}{Mapping forest canopy fuel parameters at european scale using spaceborne lidar and satellite data}.
\newblock \bibinfo{journal}{Remote Sensing of Environment} \bibinfo{volume}{303}, \bibinfo{pages}{114005}.
%Type = Article
\bibitem[{Arellano-P{\'e}rez et~al.(2018)Arellano-P{\'e}rez, Castedo-Dorado, L{\'o}pez-S{\'a}nchez, Gonz{\'a}lez-Ferreiro, Yang, D{\'\i}az-Varela, {\'A}lvarez-Gonz{\'a}lez, Vega and Ruiz-Gonz{\'a}lez}]{arellano2018potential}
\bibinfo{author}{Arellano-P{\'e}rez, S.}, \bibinfo{author}{Castedo-Dorado, F.}, \bibinfo{author}{L{\'o}pez-S{\'a}nchez, C.A.}, \bibinfo{author}{Gonz{\'a}lez-Ferreiro, E.}, \bibinfo{author}{Yang, Z.}, \bibinfo{author}{D{\'\i}az-Varela, R.A.}, \bibinfo{author}{{\'A}lvarez-Gonz{\'a}lez, J.G.}, \bibinfo{author}{Vega, J.A.}, \bibinfo{author}{Ruiz-Gonz{\'a}lez, A.D.}, \bibinfo{year}{2018}.
\newblock \bibinfo{title}{Potential of sentinel-2a data to model surface and canopy fuel characteristics in relation to crown fire hazard}.
\newblock \bibinfo{journal}{Remote Sensing} \bibinfo{volume}{10}, \bibinfo{pages}{1645}.
%Type = Article
\bibitem[{Arroyo et~al.(2008)Arroyo, Pascual and Manzanera}]{arroyo2008fire}
\bibinfo{author}{Arroyo, L.A.}, \bibinfo{author}{Pascual, C.}, \bibinfo{author}{Manzanera, J.A.}, \bibinfo{year}{2008}.
\newblock \bibinfo{title}{Fire models and methods to map fuel types: The role of remote sensing}.
\newblock \bibinfo{journal}{Forest ecology and management} \bibinfo{volume}{256}, \bibinfo{pages}{1239--1252}.
%Type = Article
\bibitem[{Art{\'e}s et~al.(2019)Art{\'e}s, Oom, De~Rigo, Durrant, Maianti, Libert{\`a} and San-Miguel-Ayanz}]{artes2019global}
\bibinfo{author}{Art{\'e}s, T.}, \bibinfo{author}{Oom, D.}, \bibinfo{author}{De~Rigo, D.}, \bibinfo{author}{Durrant, T.H.}, \bibinfo{author}{Maianti, P.}, \bibinfo{author}{Libert{\`a}, G.}, \bibinfo{author}{San-Miguel-Ayanz, J.}, \bibinfo{year}{2019}.
\newblock \bibinfo{title}{A global wildfire dataset for the analysis of fire regimes and fire behaviour}.
\newblock \bibinfo{journal}{Scientific data} \bibinfo{volume}{6}, \bibinfo{pages}{296}.
%Type = Article
\bibitem[{Asif et~al.(2021)Asif, Sarker, Chakrabortty, Ryan, Ahamed, Saha, Badal, Das, Ali, Moyeen, Islam and Tasneem}]{9395439}
\bibinfo{author}{Asif, N.A.}, \bibinfo{author}{Sarker, Y.}, \bibinfo{author}{Chakrabortty, R.K.}, \bibinfo{author}{Ryan, M.J.}, \bibinfo{author}{Ahamed, M.H.}, \bibinfo{author}{Saha, D.K.}, \bibinfo{author}{Badal, F.R.}, \bibinfo{author}{Das, S.K.}, \bibinfo{author}{Ali, M.F.}, \bibinfo{author}{Moyeen, S.I.}, \bibinfo{author}{Islam, M.R.}, \bibinfo{author}{Tasneem, Z.}, \bibinfo{year}{2021}.
\newblock \bibinfo{title}{Graph neural network: A comprehensive review on non-euclidean space}.
\newblock \bibinfo{journal}{IEEE Access} \bibinfo{volume}{9}, \bibinfo{pages}{60588--60606}.
\newblock \DOIprefix\doi{10.1109/ACCESS.2021.3071274}.
%Type = Article
\bibitem[{Badia-Perpiny{\`a} and Pallares-Barbera(2006)}]{badia2006spatial}
\bibinfo{author}{Badia-Perpiny{\`a}, A.}, \bibinfo{author}{Pallares-Barbera, M.}, \bibinfo{year}{2006}.
\newblock \bibinfo{title}{Spatial distribution of ignitions in mediterranean periurban and rural areas: the case of catalonia}.
\newblock \bibinfo{journal}{International journal of Wildland fire} \bibinfo{volume}{15}, \bibinfo{pages}{187--196}.
%Type = Article
\bibitem[{Bakke et~al.(2023)Bakke, Wanders, Van Der~Wiel and Tallaksen}]{bakke_data-driven_2023}
\bibinfo{author}{Bakke, S.J.}, \bibinfo{author}{Wanders, N.}, \bibinfo{author}{Van Der~Wiel, K.}, \bibinfo{author}{Tallaksen, L.M.}, \bibinfo{year}{2023}.
\newblock \bibinfo{title}{A data-driven model for {Fennoscandian} wildfire danger}.
\newblock \bibinfo{journal}{Natural Hazards and Earth System Sciences} \bibinfo{volume}{23}, \bibinfo{pages}{65--89}.
\newblock \URLprefix \url{https://www.duo.uio.no/handle/10852/108719}, \DOIprefix\doi{10.5194/nhess-23-65-2023}. \bibinfo{note}{accepted: 2024-02-27T18:29:41Z}.
%Type = Article
\bibitem[{Balaguer-Romano et~al.(2022)Balaguer-Romano, D{\'\i}az-Sierra, De~C{\'a}ceres, Cunill-Camprub{\'\i}, Nolan, Boer, Voltas and de~Dios}]{balaguer2022semi}
\bibinfo{author}{Balaguer-Romano, R.}, \bibinfo{author}{D{\'\i}az-Sierra, R.}, \bibinfo{author}{De~C{\'a}ceres, M.}, \bibinfo{author}{Cunill-Camprub{\'\i}, {\`A}.}, \bibinfo{author}{Nolan, R.H.}, \bibinfo{author}{Boer, M.M.}, \bibinfo{author}{Voltas, J.}, \bibinfo{author}{de~Dios, V.R.}, \bibinfo{year}{2022}.
\newblock \bibinfo{title}{A semi-mechanistic model for predicting daily variations in species-level live fuel moisture content}.
\newblock \bibinfo{journal}{Agricultural and Forest Meteorology} \bibinfo{volume}{323}, \bibinfo{pages}{109022}.
%Type = Article
\bibitem[{Balch et~al.(2017)Balch, Bradley, Abatzoglou, Nagy, Fusco and Mahood}]{balch2017human}
\bibinfo{author}{Balch, J.K.}, \bibinfo{author}{Bradley, B.A.}, \bibinfo{author}{Abatzoglou, J.T.}, \bibinfo{author}{Nagy, R.C.}, \bibinfo{author}{Fusco, E.J.}, \bibinfo{author}{Mahood, A.L.}, \bibinfo{year}{2017}.
\newblock \bibinfo{title}{Human-started wildfires expand the fire niche across the united states}.
\newblock \bibinfo{journal}{Proceedings of the National Academy of Sciences} \bibinfo{volume}{114}, \bibinfo{pages}{2946--2951}.
%Type = Article
\bibitem[{Baltru{\v{s}}aitis et~al.(2018)Baltru{\v{s}}aitis, Ahuja and Morency}]{baltruvsaitis2018multimodal}
\bibinfo{author}{Baltru{\v{s}}aitis, T.}, \bibinfo{author}{Ahuja, C.}, \bibinfo{author}{Morency, L.P.}, \bibinfo{year}{2018}.
\newblock \bibinfo{title}{Multimodal machine learning: A survey and taxonomy}.
\newblock \bibinfo{journal}{IEEE transactions on pattern analysis and machine intelligence} \bibinfo{volume}{41}, \bibinfo{pages}{423--443}.
%Type = Article
\bibitem[{Bar~Massada et~al.(2011)Bar~Massada, Syphard, Hawbaker, Stewart and Radeloff}]{bar_massada_effects_2011}
\bibinfo{author}{Bar~Massada, A.}, \bibinfo{author}{Syphard, A.D.}, \bibinfo{author}{Hawbaker, T.J.}, \bibinfo{author}{Stewart, S.I.}, \bibinfo{author}{Radeloff, V.C.}, \bibinfo{year}{2011}.
\newblock \bibinfo{title}{Effects of ignition location models on the burn patterns of simulated wildfires}.
\newblock \bibinfo{journal}{Environmental Modelling \& Software} \bibinfo{volume}{26}, \bibinfo{pages}{583--592}.
\newblock \URLprefix \url{https://www.sciencedirect.com/science/article/pii/S1364815210003233}, \DOIprefix\doi{10.1016/j.envsoft.2010.11.016}.
%Type = Article
\bibitem[{Barber et~al.(2024)Barber, Jain, Whitman, Thompson, Guindon, Parks, Wang, Hethcoat and Parisien}]{barber2024canadian}
\bibinfo{author}{Barber, Q.E.}, \bibinfo{author}{Jain, P.}, \bibinfo{author}{Whitman, E.}, \bibinfo{author}{Thompson, D.K.}, \bibinfo{author}{Guindon, L.}, \bibinfo{author}{Parks, S.A.}, \bibinfo{author}{Wang, X.}, \bibinfo{author}{Hethcoat, M.G.}, \bibinfo{author}{Parisien, M.A.}, \bibinfo{year}{2024}.
\newblock \bibinfo{title}{The canadian fire spread dataset}.
\newblock \bibinfo{journal}{Scientific data} \bibinfo{volume}{11}, \bibinfo{pages}{764}.
%Type = Article
\bibitem[{Barmpoutis et~al.(2020)Barmpoutis, Papaioannou, Dimitropoulos and Grammalidis}]{s20226442}
\bibinfo{author}{Barmpoutis, P.}, \bibinfo{author}{Papaioannou, P.}, \bibinfo{author}{Dimitropoulos, K.}, \bibinfo{author}{Grammalidis, N.}, \bibinfo{year}{2020}.
\newblock \bibinfo{title}{A review on early forest fire detection systems using optical remote sensing}.
\newblock \bibinfo{journal}{Sensors} \bibinfo{volume}{20}.
\newblock \URLprefix \url{https://www.mdpi.com/1424-8220/20/22/6442}, \DOIprefix\doi{10.3390/s20226442}.
%Type = Article
\bibitem[{Benali et~al.(2016)Benali, Russo, S{\'a}, Pinto, Price, Koutsias and Pereira}]{benali2016determining}
\bibinfo{author}{Benali, A.}, \bibinfo{author}{Russo, A.}, \bibinfo{author}{S{\'a}, A.C.}, \bibinfo{author}{Pinto, R.M.}, \bibinfo{author}{Price, O.}, \bibinfo{author}{Koutsias, N.}, \bibinfo{author}{Pereira, J.M.}, \bibinfo{year}{2016}.
\newblock \bibinfo{title}{Determining fire dates and locating ignition points with satellite data}.
\newblock \bibinfo{journal}{Remote Sensing} \bibinfo{volume}{8}, \bibinfo{pages}{326}.
%Type = Article
\bibitem[{Bergado et~al.(2021)Bergado, Persello, Reinke and Stein}]{bergado_predicting_2021}
\bibinfo{author}{Bergado, J.R.}, \bibinfo{author}{Persello, C.}, \bibinfo{author}{Reinke, K.}, \bibinfo{author}{Stein, A.}, \bibinfo{year}{2021}.
\newblock \bibinfo{title}{Predicting wildfire burns from big geodata using deep learning}.
\newblock \bibinfo{journal}{Safety Science} \bibinfo{volume}{140}, \bibinfo{pages}{105276}.
\newblock \URLprefix \url{https://www.sciencedirect.com/science/article/pii/S0925753521001211}, \DOIprefix\doi{10.1016/j.ssci.2021.105276}.
%Type = Article
\bibitem[{Bergonse et~al.(2021)Bergonse, Oliveira, Gonçalves, Nunes, DaCamara and Zêzere}]{bergonse_predicting_2021}
\bibinfo{author}{Bergonse, R.}, \bibinfo{author}{Oliveira, S.}, \bibinfo{author}{Gonçalves, A.}, \bibinfo{author}{Nunes, S.}, \bibinfo{author}{DaCamara, C.}, \bibinfo{author}{Zêzere, J.L.}, \bibinfo{year}{2021}.
\newblock \bibinfo{title}{Predicting burnt areas during the summer season in {Portugal} by combining wildfire susceptibility and spring meteorological conditions}.
\newblock \bibinfo{journal}{Geomatics, Natural Hazards and Risk} \bibinfo{volume}{12}, \bibinfo{pages}{1039--1057}.
\newblock \URLprefix \url{https://doi.org/10.1080/19475705.2021.1909664}, \DOIprefix\doi{10.1080/19475705.2021.1909664}. \bibinfo{note}{publisher: Taylor \& Francis \_eprint: https://doi.org/10.1080/19475705.2021.1909664}.
%Type = Article
\bibitem[{Bessie and Johnson(1995)}]{bessie1995relative}
\bibinfo{author}{Bessie, W.}, \bibinfo{author}{Johnson, E.}, \bibinfo{year}{1995}.
\newblock \bibinfo{title}{The relative importance of fuels and weather on fire behavior in subalpine forests}.
\newblock \bibinfo{journal}{Ecology} \bibinfo{volume}{76}, \bibinfo{pages}{747--762}.
%Type = Article
\bibitem[{Bhowmik et~al.(2023)Bhowmik, Jung, Aguilera, Prunicki and Nadeau}]{bhowmik_multi-modal_2023}
\bibinfo{author}{Bhowmik, R.T.}, \bibinfo{author}{Jung, Y.S.}, \bibinfo{author}{Aguilera, J.A.}, \bibinfo{author}{Prunicki, M.}, \bibinfo{author}{Nadeau, K.}, \bibinfo{year}{2023}.
\newblock \bibinfo{title}{A multi-modal wildfire prediction and early-warning system based on a novel machine learning framework}.
\newblock \bibinfo{journal}{Journal of Environmental Management} \bibinfo{volume}{341}, \bibinfo{pages}{117908}.
\newblock \URLprefix \url{https://www.sciencedirect.com/science/article/pii/S0301479723006965}, \DOIprefix\doi{10.1016/j.jenvman.2023.117908}.
%Type = Article
\bibitem[{Bi et~al.(2023)Bi, Xie, Zhang, Chen, Gu and Tian}]{bi2023accurate}
\bibinfo{author}{Bi, K.}, \bibinfo{author}{Xie, L.}, \bibinfo{author}{Zhang, H.}, \bibinfo{author}{Chen, X.}, \bibinfo{author}{Gu, X.}, \bibinfo{author}{Tian, Q.}, \bibinfo{year}{2023}.
\newblock \bibinfo{title}{Accurate medium-range global weather forecasting with 3d neural networks}.
\newblock \bibinfo{journal}{Nature} \bibinfo{volume}{619}, \bibinfo{pages}{533--538}.
%Type = Article
\bibitem[{Bianchi and Defossé(2015)}]{bianchi_live_2015}
\bibinfo{author}{Bianchi, L.O.}, \bibinfo{author}{Defossé, G.E.}, \bibinfo{year}{2015}.
\newblock \bibinfo{title}{Live fuel moisture content and leaf ignition of forest species in {Andean} {Patagonia}, {Argentina}}.
\newblock \bibinfo{journal}{International Journal of Wildland Fire} \bibinfo{volume}{24}, \bibinfo{pages}{340--348}.
\newblock \URLprefix \url{https://www.publish.csiro.au/wf/WF13099}, \DOIprefix\doi{10.1071/WF13099}. \bibinfo{note}{publisher: CSIRO PUBLISHING}.
%Type = Article
\bibitem[{Bj{\aa}nes et~al.(2021)Bj{\aa}nes, De~La~Fuente and Mena}]{bjaanes2021deep}
\bibinfo{author}{Bj{\aa}nes, A.}, \bibinfo{author}{De~La~Fuente, R.}, \bibinfo{author}{Mena, P.}, \bibinfo{year}{2021}.
\newblock \bibinfo{title}{A deep learning ensemble model for wildfire susceptibility mapping}.
\newblock \bibinfo{journal}{Ecological Informatics} \bibinfo{volume}{65}, \bibinfo{pages}{101397}.
%Type = Article
\bibitem[{Bommer et~al.(2024)Bommer, Kretschmer, Hedstr{\"o}m, Bareeva and H{\"o}hne}]{bommer2024finding}
\bibinfo{author}{Bommer, P.L.}, \bibinfo{author}{Kretschmer, M.}, \bibinfo{author}{Hedstr{\"o}m, A.}, \bibinfo{author}{Bareeva, D.}, \bibinfo{author}{H{\"o}hne, M.M.C.}, \bibinfo{year}{2024}.
\newblock \bibinfo{title}{Finding the right xai method—a guide for the evaluation and ranking of explainable ai methods in climate science}.
\newblock \bibinfo{journal}{Artificial Intelligence for the Earth Systems} \bibinfo{volume}{3}, \bibinfo{pages}{e230074}.
%Type = Article
\bibitem[{Bond et~al.(2005)Bond, Woodward and Midgley}]{bond2005global}
\bibinfo{author}{Bond, W.J.}, \bibinfo{author}{Woodward, F.I.}, \bibinfo{author}{Midgley, G.F.}, \bibinfo{year}{2005}.
\newblock \bibinfo{title}{The global distribution of ecosystems in a world without fire}.
\newblock \bibinfo{journal}{New phytologist} \bibinfo{volume}{165}, \bibinfo{pages}{525--538}.
%Type = Article
\bibitem[{Boroujeni et~al.(2024)Boroujeni, Razi, Khoshdel, Afghah, Coen, O’Neill, Fule, Watts, Kokolakis and Vamvoudakis}]{boroujeni2024comprehensive}
\bibinfo{author}{Boroujeni, S.P.H.}, \bibinfo{author}{Razi, A.}, \bibinfo{author}{Khoshdel, S.}, \bibinfo{author}{Afghah, F.}, \bibinfo{author}{Coen, J.L.}, \bibinfo{author}{O’Neill, L.}, \bibinfo{author}{Fule, P.}, \bibinfo{author}{Watts, A.}, \bibinfo{author}{Kokolakis, N.M.T.}, \bibinfo{author}{Vamvoudakis, K.G.}, \bibinfo{year}{2024}.
\newblock \bibinfo{title}{A comprehensive survey of research towards ai-enabled unmanned aerial systems in pre-, active-, and post-wildfire management}.
\newblock \bibinfo{journal}{Information Fusion} , \bibinfo{pages}{102369}.
%Type = Article
\bibitem[{Botequim et~al.(2019)Botequim, Fernandes, Borges, Gonz{\'a}lez-Ferreiro and Guerra-Hern{\'a}ndez}]{botequim2019improving}
\bibinfo{author}{Botequim, B.}, \bibinfo{author}{Fernandes, P.M.}, \bibinfo{author}{Borges, J.G.}, \bibinfo{author}{Gonz{\'a}lez-Ferreiro, E.}, \bibinfo{author}{Guerra-Hern{\'a}ndez, J.}, \bibinfo{year}{2019}.
\newblock \bibinfo{title}{Improving silvicultural practices for mediterranean forests through fire behaviour modelling using lidar-derived canopy fuel characteristics}.
\newblock \bibinfo{journal}{International Journal of Wildland Fire} \bibinfo{volume}{28}, \bibinfo{pages}{823--839}.
%Type = Article
\bibitem[{Bouguettaya et~al.(2022)Bouguettaya, Zarzour, Taberkit and Kechida}]{bouguettaya2022review}
\bibinfo{author}{Bouguettaya, A.}, \bibinfo{author}{Zarzour, H.}, \bibinfo{author}{Taberkit, A.M.}, \bibinfo{author}{Kechida, A.}, \bibinfo{year}{2022}.
\newblock \bibinfo{title}{A review on early wildfire detection from unmanned aerial vehicles using deep learning-based computer vision algorithms}.
\newblock \bibinfo{journal}{Signal Processing} \bibinfo{volume}{190}, \bibinfo{pages}{108309}.
%Type = Article
\bibitem[{Bowd et~al.(2021)Bowd, Blair and Lindenmayer}]{bowd2021prior}
\bibinfo{author}{Bowd, E.J.}, \bibinfo{author}{Blair, D.P.}, \bibinfo{author}{Lindenmayer, D.B.}, \bibinfo{year}{2021}.
\newblock \bibinfo{title}{Prior disturbance legacy effects on plant recovery post-high-severity wildfire}.
\newblock \bibinfo{journal}{Ecosphere} \bibinfo{volume}{12}, \bibinfo{pages}{e03480}.
%Type = Article
\bibitem[{Bowyer and Danson(2004)}]{bowyer2004sensitivity}
\bibinfo{author}{Bowyer, P.}, \bibinfo{author}{Danson, F.}, \bibinfo{year}{2004}.
\newblock \bibinfo{title}{Sensitivity of spectral reflectance to variation in live fuel moisture content at leaf and canopy level}.
\newblock \bibinfo{journal}{Remote Sensing of Environment} \bibinfo{volume}{92}, \bibinfo{pages}{297--308}.
%Type = Techreport
\bibitem[{Bradshaw et~al.(1984)Bradshaw, Deeming, Burgan and Cohen}]{Bradshaw_1984}
\bibinfo{author}{Bradshaw, L.S.}, \bibinfo{author}{Deeming, J.E.}, \bibinfo{author}{Burgan, R.E.}, \bibinfo{author}{Cohen, J.D.}, \bibinfo{year}{1984}.
\newblock \bibinfo{title}{The 1978 National Fire-Danger Rating System: technical documentation}.
\newblock \bibinfo{type}{Technical Report}. U.S. Department of Agriculture, Forest Service, Intermountain Forest and Range Experiment Station.
\newblock \URLprefix \url{http://dx.doi.org/10.2737/INT-GTR-169}, \DOIprefix\doi{10.2737/int-gtr-169}.
%Type = Article
\bibitem[{Braun et~al.(2010)Braun, Jones, Lee, Woolford and Wotton}]{braun2010forest}
\bibinfo{author}{Braun, W.J.}, \bibinfo{author}{Jones, B.L.}, \bibinfo{author}{Lee, J.S.}, \bibinfo{author}{Woolford, D.G.}, \bibinfo{author}{Wotton, B.M.}, \bibinfo{year}{2010}.
\newblock \bibinfo{title}{Forest fire risk assessment: an illustrative example from ontario, canada}.
\newblock \bibinfo{journal}{Journal of Probability and Statistics} \bibinfo{volume}{2010}, \bibinfo{pages}{823018}.
%Type = Article
\bibitem[{Bright et~al.(2017)Bright, Hudak, Meddens, Hawbaker, Briggs and Kennedy}]{bright2017prediction}
\bibinfo{author}{Bright, B.C.}, \bibinfo{author}{Hudak, A.T.}, \bibinfo{author}{Meddens, A.J.}, \bibinfo{author}{Hawbaker, T.J.}, \bibinfo{author}{Briggs, J.S.}, \bibinfo{author}{Kennedy, R.E.}, \bibinfo{year}{2017}.
\newblock \bibinfo{title}{Prediction of forest canopy and surface fuels from lidar and satellite time series data in a bark beetle-affected forest}.
\newblock \bibinfo{journal}{Forests} \bibinfo{volume}{8}, \bibinfo{pages}{322}.
%Type = Misc
\bibitem[{Burge et~al.(2021)Burge, Bonanni, Ihme and Hu}]{burge2021convolutionallstmneuralnetworks}
\bibinfo{author}{Burge, J.}, \bibinfo{author}{Bonanni, M.}, \bibinfo{author}{Ihme, M.}, \bibinfo{author}{Hu, L.}, \bibinfo{year}{2021}.
\newblock \bibinfo{title}{Convolutional lstm neural networks for modeling wildland fire dynamics}.
\newblock \URLprefix \url{https://arxiv.org/abs/2012.06679}, \href{http://arxiv.org/abs/2012.06679}{{\tt arXiv:2012.06679}}.
%Type = Article
\bibitem[{Burge et~al.(2023)Burge, Bonanni, Hu and Ihme}]{burge2023recurrent}
\bibinfo{author}{Burge, J.}, \bibinfo{author}{Bonanni, M.R.}, \bibinfo{author}{Hu, R.L.}, \bibinfo{author}{Ihme, M.}, \bibinfo{year}{2023}.
\newblock \bibinfo{title}{Recurrent convolutional deep neural networks for modeling time-resolved wildfire spread behavior}.
\newblock \bibinfo{journal}{Fire Technology} \bibinfo{volume}{59}, \bibinfo{pages}{3327--3354}.
%Type = Article
\bibitem[{Byrne et~al.(2024)Byrne, Liu, Bowman, Pascolini-Campbell, Chatterjee, Pandey, Miyazaki, van~der Werf, Wunch, Wennberg et~al.}]{byrne2024carbon}
\bibinfo{author}{Byrne, B.}, \bibinfo{author}{Liu, J.}, \bibinfo{author}{Bowman, K.W.}, \bibinfo{author}{Pascolini-Campbell, M.}, \bibinfo{author}{Chatterjee, A.}, \bibinfo{author}{Pandey, S.}, \bibinfo{author}{Miyazaki, K.}, \bibinfo{author}{van~der Werf, G.R.}, \bibinfo{author}{Wunch, D.}, \bibinfo{author}{Wennberg, P.O.}, et~al., \bibinfo{year}{2024}.
\newblock \bibinfo{title}{Carbon emissions from the 2023 canadian wildfires}.
\newblock \bibinfo{journal}{Nature} , \bibinfo{pages}{1--5}.
%Type = Article
\bibitem[{Caccamo et~al.(2011)Caccamo, Chisholm, Bradstock, Puotinen and Pippen}]{caccamo2011monitoring}
\bibinfo{author}{Caccamo, G.}, \bibinfo{author}{Chisholm, L.}, \bibinfo{author}{Bradstock, R.}, \bibinfo{author}{Puotinen, M.L.}, \bibinfo{author}{Pippen, B.}, \bibinfo{year}{2011}.
\newblock \bibinfo{title}{Monitoring live fuel moisture content of heathland, shrubland and sclerophyll forest in south-eastern australia using modis data}.
\newblock \bibinfo{journal}{International Journal of Wildland Fire} \bibinfo{volume}{21}, \bibinfo{pages}{257--269}.
%Type = Inproceedings
\bibitem[{Cahuantzi et~al.(2023)Cahuantzi, Chen and G{\"u}ttel}]{cahuantzi2023comparison}
\bibinfo{author}{Cahuantzi, R.}, \bibinfo{author}{Chen, X.}, \bibinfo{author}{G{\"u}ttel, S.}, \bibinfo{year}{2023}.
\newblock \bibinfo{title}{A comparison of lstm and gru networks for learning symbolic sequences}, in: \bibinfo{booktitle}{Science and Information Conference}, \bibinfo{organization}{Springer}. pp. \bibinfo{pages}{771--785}.
%Type = Article
\bibitem[{Calvi{\~n}o-Cancela et~al.(2016)Calvi{\~n}o-Cancela, Chas-Amil, Garc{\'\i}a-Mart{\'\i}nez and Touza}]{calvino2016wildfire}
\bibinfo{author}{Calvi{\~n}o-Cancela, M.}, \bibinfo{author}{Chas-Amil, M.L.}, \bibinfo{author}{Garc{\'\i}a-Mart{\'\i}nez, E.D.}, \bibinfo{author}{Touza, J.}, \bibinfo{year}{2016}.
\newblock \bibinfo{title}{Wildfire risk associated with different vegetation types within and outside wildland-urban interfaces}.
\newblock \bibinfo{journal}{Forest Ecology and Management} \bibinfo{volume}{372}, \bibinfo{pages}{1--9}.
%Type = Article
\bibitem[{Calvi{\~n}o-Cancela et~al.(2017)Calvi{\~n}o-Cancela, Chas-Amil, Garc{\'\i}a-Mart{\'\i}nez and Touza}]{calvino2017interacting}
\bibinfo{author}{Calvi{\~n}o-Cancela, M.}, \bibinfo{author}{Chas-Amil, M.L.}, \bibinfo{author}{Garc{\'\i}a-Mart{\'\i}nez, E.D.}, \bibinfo{author}{Touza, J.}, \bibinfo{year}{2017}.
\newblock \bibinfo{title}{Interacting effects of topography, vegetation, human activities and wildland-urban interfaces on wildfire ignition risk}.
\newblock \bibinfo{journal}{Forest Ecology and Management} \bibinfo{volume}{397}, \bibinfo{pages}{10--17}.
%Type = Article
\bibitem[{Canisius and Chen(2007)}]{canisius_retrieving_2007}
\bibinfo{author}{Canisius, F.}, \bibinfo{author}{Chen, J.M.}, \bibinfo{year}{2007}.
\newblock \bibinfo{title}{Retrieving forest background reflectance in a boreal region from {Multi}-angle {Imaging} {SpectroRadiometer} ({MISR}) data}.
\newblock \bibinfo{journal}{Remote Sensing of Environment} \bibinfo{volume}{107}, \bibinfo{pages}{312--321}.
\newblock \URLprefix \url{https://www.sciencedirect.com/science/article/pii/S0034425706004184}, \DOIprefix\doi{10.1016/j.rse.2006.07.023}.
%Type = Article
\bibitem[{Cao et~al.(2024)Cao, Zhou, Yu, Rao, Wu, Li and Zhu}]{cao2024forest}
\bibinfo{author}{Cao, Y.}, \bibinfo{author}{Zhou, X.}, \bibinfo{author}{Yu, Y.}, \bibinfo{author}{Rao, S.}, \bibinfo{author}{Wu, Y.}, \bibinfo{author}{Li, C.}, \bibinfo{author}{Zhu, Z.}, \bibinfo{year}{2024}.
\newblock \bibinfo{title}{Forest fire prediction based on time series networks and remote sensing images}.
\newblock \bibinfo{journal}{Forests} \bibinfo{volume}{15}, \bibinfo{pages}{1221}.
%Type = Article
\bibitem[{Capps et~al.(2021)Capps, Zhuang, Liu, Rolinski and Qu}]{capps2021modelling}
\bibinfo{author}{Capps, S.B.}, \bibinfo{author}{Zhuang, W.}, \bibinfo{author}{Liu, R.}, \bibinfo{author}{Rolinski, T.}, \bibinfo{author}{Qu, X.}, \bibinfo{year}{2021}.
\newblock \bibinfo{title}{Modelling chamise fuel moisture content across california: A machine learning approach}.
\newblock \bibinfo{journal}{International Journal of Wildland Fire} \bibinfo{volume}{31}, \bibinfo{pages}{136--148}.
%Type = Inproceedings
\bibitem[{Caron et~al.(2024)Caron, Guyeux and Aynes}]{10.1145/3651671.3651708}
\bibinfo{author}{Caron, N.}, \bibinfo{author}{Guyeux, C.}, \bibinfo{author}{Aynes, B.}, \bibinfo{year}{2024}.
\newblock \bibinfo{title}{Predicting wildfire events with calibrated probabilities}, in: \bibinfo{booktitle}{Proceedings of the 2024 16th International Conference on Machine Learning and Computing}, \bibinfo{publisher}{Association for Computing Machinery}, \bibinfo{address}{New York, NY, USA}. p. \bibinfo{pages}{168–175}.
\newblock \URLprefix \url{https://doi.org/10.1145/3651671.3651708}, \DOIprefix\doi{10.1145/3651671.3651708}.
%Type = Article
\bibitem[{Castelvecchi(2016)}]{castelvecchi2016can}
\bibinfo{author}{Castelvecchi, D.}, \bibinfo{year}{2016}.
\newblock \bibinfo{title}{Can we open the black box of ai?}
\newblock \bibinfo{journal}{Nature News} \bibinfo{volume}{538}, \bibinfo{pages}{20}.
%Type = Article
\bibitem[{Castro et~al.(2003)Castro, Tudela and Sebastià}]{castro_modeling_2003}
\bibinfo{author}{Castro, F.X.}, \bibinfo{author}{Tudela, A.}, \bibinfo{author}{Sebastià, M.T.}, \bibinfo{year}{2003}.
\newblock \bibinfo{title}{Modeling moisture content in shrubs to predict fire risk in {Catalonia} ({Spain})}.
\newblock \bibinfo{journal}{Agricultural and Forest Meteorology} \bibinfo{volume}{116}, \bibinfo{pages}{49--59}.
\newblock \URLprefix \url{https://www.sciencedirect.com/science/article/pii/S0168192302002484}, \DOIprefix\doi{10.1016/S0168-1923(02)00248-4}.
%Type = Article
\bibitem[{Catchpole et~al.(2001)Catchpole, Catchpole, Viney, McCaw and Marsden-Smedley}]{catchpole2001estimating}
\bibinfo{author}{Catchpole, E.}, \bibinfo{author}{Catchpole, W.}, \bibinfo{author}{Viney, N.}, \bibinfo{author}{McCaw, W.}, \bibinfo{author}{Marsden-Smedley, J.}, \bibinfo{year}{2001}.
\newblock \bibinfo{title}{Estimating fuel response time and predicting fuel moisture content from field data}.
\newblock \bibinfo{journal}{International Journal of Wildland Fire} \bibinfo{volume}{10}, \bibinfo{pages}{215--222}.
%Type = Article
\bibitem[{Catry et~al.(2009)Catry, Rego, Ba{\c{c}}{\~a}o and Moreira}]{catry2009modeling}
\bibinfo{author}{Catry, F.X.}, \bibinfo{author}{Rego, F.C.}, \bibinfo{author}{Ba{\c{c}}{\~a}o, F.L.}, \bibinfo{author}{Moreira, F.}, \bibinfo{year}{2009}.
\newblock \bibinfo{title}{Modeling and mapping wildfire ignition risk in portugal}.
\newblock \bibinfo{journal}{International Journal of Wildland Fire} \bibinfo{volume}{18}, \bibinfo{pages}{921--931}.
%Type = Article
\bibitem[{Cawson et~al.(2020)Cawson, Nyman, Schunk, Sheridan, Duff, Gibos, Bovill, Conedera, Pezzatti and Menzel}]{cawson2020estimation}
\bibinfo{author}{Cawson, J.G.}, \bibinfo{author}{Nyman, P.}, \bibinfo{author}{Schunk, C.}, \bibinfo{author}{Sheridan, G.J.}, \bibinfo{author}{Duff, T.J.}, \bibinfo{author}{Gibos, K.}, \bibinfo{author}{Bovill, W.D.}, \bibinfo{author}{Conedera, M.}, \bibinfo{author}{Pezzatti, G.B.}, \bibinfo{author}{Menzel, A.}, \bibinfo{year}{2020}.
\newblock \bibinfo{title}{Estimation of surface dead fine fuel moisture using automated fuel moisture sticks across a range of forests worldwide}.
\newblock \bibinfo{journal}{International Journal of Wildland Fire} \bibinfo{volume}{29}, \bibinfo{pages}{548--559}.
%Type = Article
\bibitem[{Chan et~al.(2022)Chan, Leow, Bea, Cheng, Phoong, Hong and Chen}]{math10081283}
\bibinfo{author}{Chan, J.Y.L.}, \bibinfo{author}{Leow, S.M.H.}, \bibinfo{author}{Bea, K.T.}, \bibinfo{author}{Cheng, W.K.}, \bibinfo{author}{Phoong, S.W.}, \bibinfo{author}{Hong, Z.W.}, \bibinfo{author}{Chen, Y.L.}, \bibinfo{year}{2022}.
\newblock \bibinfo{title}{Mitigating the multicollinearity problem and its machine learning approach: A review}.
\newblock \bibinfo{journal}{Mathematics} \bibinfo{volume}{10}.
\newblock \URLprefix \url{https://www.mdpi.com/2227-7390/10/8/1283}, \DOIprefix\doi{10.3390/math10081283}.
%Type = Article
\bibitem[{Chaparro et~al.(2024)Chaparro, Jagdhuber, Piles, Jonard, Fluhrer, Vall-llossera, Camps, L{\'o}pez-Mart{\'\i}nez, Fern{\'a}ndez-Mor{\'a}n, Baur et~al.}]{chaparro2024vegetation}
\bibinfo{author}{Chaparro, D.}, \bibinfo{author}{Jagdhuber, T.}, \bibinfo{author}{Piles, M.}, \bibinfo{author}{Jonard, F.}, \bibinfo{author}{Fluhrer, A.}, \bibinfo{author}{Vall-llossera, M.}, \bibinfo{author}{Camps, A.}, \bibinfo{author}{L{\'o}pez-Mart{\'\i}nez, C.}, \bibinfo{author}{Fern{\'a}ndez-Mor{\'a}n, R.}, \bibinfo{author}{Baur, M.}, et~al., \bibinfo{year}{2024}.
\newblock \bibinfo{title}{Vegetation moisture estimation in the western united states using radiometer-radar-lidar synergy}.
\newblock \bibinfo{journal}{Remote Sensing of Environment} \bibinfo{volume}{303}, \bibinfo{pages}{113993}.
%Type = Misc
\bibitem[{Chavalithumrong et~al.(2021)Chavalithumrong, Yoon and Voulgaris}]{chavalithumrong_learning_2021}
\bibinfo{author}{Chavalithumrong, A.}, \bibinfo{author}{Yoon, H.J.}, \bibinfo{author}{Voulgaris, P.}, \bibinfo{year}{2021}.
\newblock \bibinfo{title}{Learning {Wildfire} {Model} from {Incomplete} {State} {Observations}}.
\newblock \URLprefix \url{http://arxiv.org/abs/2111.14038}. \bibinfo{note}{arXiv:2111.14038 [cs]}.
%Type = Article
\bibitem[{Chen and Jin(2022)}]{Chen_2022}
\bibinfo{author}{Chen, B.}, \bibinfo{author}{Jin, Y.}, \bibinfo{year}{2022}.
\newblock \bibinfo{title}{Spatial patterns and drivers for wildfire ignitions in california}.
\newblock \bibinfo{journal}{Environmental Research Letters} \bibinfo{volume}{17}, \bibinfo{pages}{055004}.
\newblock \URLprefix \url{https://dx.doi.org/10.1088/1748-9326/ac60da}, \DOIprefix\doi{10.1088/1748-9326/ac60da}.
%Type = Article
\bibitem[{Chen et~al.(2024a)Chen, Cheng, Hu, Kasoar and Arcucci}]{chen2024explainable}
\bibinfo{author}{Chen, D.}, \bibinfo{author}{Cheng, S.}, \bibinfo{author}{Hu, J.}, \bibinfo{author}{Kasoar, M.}, \bibinfo{author}{Arcucci, R.}, \bibinfo{year}{2024}a.
\newblock \bibinfo{title}{Explainable global wildfire prediction models using graph neural networks}.
\newblock \bibinfo{journal}{arXiv preprint arXiv:2402.07152} .
%Type = Article
\bibitem[{Chen et~al.(2022)Chen, Yang, Peng, Chen and Ge}]{chen2022knowledge}
\bibinfo{author}{Chen, J.}, \bibinfo{author}{Yang, Y.}, \bibinfo{author}{Peng, L.}, \bibinfo{author}{Chen, L.}, \bibinfo{author}{Ge, X.}, \bibinfo{year}{2022}.
\newblock \bibinfo{title}{Knowledge graph representation learning-based forest fire prediction}.
\newblock \bibinfo{journal}{Remote Sensing} \bibinfo{volume}{14}, \bibinfo{pages}{4391}.
%Type = Article
\bibitem[{Chen et~al.(2023)Chen, Wang, Zheng, You and Zeng}]{Chen31122023}
\bibinfo{author}{Chen, M.}, \bibinfo{author}{Wang, Y.}, \bibinfo{author}{Zheng, Z.}, \bibinfo{author}{You, X.}, \bibinfo{author}{Zeng, Y.}, \bibinfo{year}{2023}.
\newblock \bibinfo{title}{Modeling the forest fire risk by incorporating a new human activity factor from nighttime light data}.
\newblock \bibinfo{journal}{Geocarto International} \bibinfo{volume}{38}, \bibinfo{pages}{2289454}.
%Type = Article
\bibitem[{Chen et~al.(2024b)Chen, He, Li, Fan, Yin, Zhang and Zhang}]{chen_estimation_2024}
\bibinfo{author}{Chen, R.}, \bibinfo{author}{He, B.}, \bibinfo{author}{Li, Y.}, \bibinfo{author}{Fan, C.}, \bibinfo{author}{Yin, J.}, \bibinfo{author}{Zhang, H.}, \bibinfo{author}{Zhang, Y.}, \bibinfo{year}{2024}b.
\newblock \bibinfo{title}{Estimation of potential wildfire behavior characteristics to assess wildfire danger in southwest {China} using deep learning schemes}.
\newblock \bibinfo{journal}{Journal of Environmental Management} \bibinfo{volume}{351}, \bibinfo{pages}{120005}.
\newblock \URLprefix \url{https://linkinghub.elsevier.com/retrieve/pii/S0301479723027937}, \DOIprefix\doi{10.1016/j.jenvman.2023.120005}.
%Type = Article
\bibitem[{Chen et~al.(2024c)Chen, He, Li, Zhang, Liao, Fan, Yin and Zhang}]{chen2024incorporating}
\bibinfo{author}{Chen, R.}, \bibinfo{author}{He, B.}, \bibinfo{author}{Li, Y.}, \bibinfo{author}{Zhang, Y.}, \bibinfo{author}{Liao, Z.}, \bibinfo{author}{Fan, C.}, \bibinfo{author}{Yin, J.}, \bibinfo{author}{Zhang, H.}, \bibinfo{year}{2024}c.
\newblock \bibinfo{title}{Incorporating fire spread simulation and machine learning algorithms to estimate crown fire potential for pine forests in sichuan, china}.
\newblock \bibinfo{journal}{International Journal of Applied Earth Observation and Geoinformation} \bibinfo{volume}{132}, \bibinfo{pages}{104080}.
%Type = Article
\bibitem[{Chen et~al.(2019)Chen, Panahi, Khosravi, Pourghasemi, Rezaie and Parvinnezhad}]{chen_spatial_2019}
\bibinfo{author}{Chen, W.}, \bibinfo{author}{Panahi, M.}, \bibinfo{author}{Khosravi, K.}, \bibinfo{author}{Pourghasemi, H.R.}, \bibinfo{author}{Rezaie, F.}, \bibinfo{author}{Parvinnezhad, D.}, \bibinfo{year}{2019}.
\newblock \bibinfo{title}{Spatial prediction of groundwater potentiality using {ANFIS} ensembled with teaching-learning-based and biogeography-based optimization}.
\newblock \bibinfo{journal}{Journal of Hydrology} \bibinfo{volume}{572}, \bibinfo{pages}{435--448}.
\newblock \URLprefix \url{https://www.sciencedirect.com/science/article/pii/S0022169419302604}, \DOIprefix\doi{10.1016/j.jhydrol.2019.03.013}.
%Type = Article
\bibitem[{Chen et~al.(2016)Chen, Morton, Andela, Giglio and Randerson}]{chen_how_2016}
\bibinfo{author}{Chen, Y.}, \bibinfo{author}{Morton, D.C.}, \bibinfo{author}{Andela, N.}, \bibinfo{author}{Giglio, L.}, \bibinfo{author}{Randerson, J.T.}, \bibinfo{year}{2016}.
\newblock \bibinfo{title}{How much global burned area can be forecast on seasonal time scales using sea surface temperatures?}
\newblock \bibinfo{journal}{Environmental Research Letters} \bibinfo{volume}{11}, \bibinfo{pages}{045001}.
\newblock \URLprefix \url{https://dx.doi.org/10.1088/1748-9326/11/4/045001}, \DOIprefix\doi{10.1088/1748-9326/11/4/045001}. \bibinfo{note}{publisher: IOP Publishing}.
%Type = Article
\bibitem[{Chen et~al.(2017)Chen, Zhu, Yebra, Harris and Tapper}]{chen_development_2017}
\bibinfo{author}{Chen, Y.}, \bibinfo{author}{Zhu, X.}, \bibinfo{author}{Yebra, M.}, \bibinfo{author}{Harris, S.}, \bibinfo{author}{Tapper, N.}, \bibinfo{year}{2017}.
\newblock \bibinfo{title}{Development of a predictive model for estimating forest surface fuel load in {Australian} eucalypt forests with {LiDAR} data}.
\newblock \bibinfo{journal}{Environmental Modelling \& Software} \bibinfo{volume}{97}, \bibinfo{pages}{61--71}.
\newblock \URLprefix \url{https://www.sciencedirect.com/science/article/pii/S1364815216304418}, \DOIprefix\doi{10.1016/j.envsoft.2017.07.007}.
%Type = Article
\bibitem[{Cheng et~al.(2024)Cheng, Chassagnon, Kasoar, Guo and Arcucci}]{cheng2024deep}
\bibinfo{author}{Cheng, S.}, \bibinfo{author}{Chassagnon, H.}, \bibinfo{author}{Kasoar, M.}, \bibinfo{author}{Guo, Y.}, \bibinfo{author}{Arcucci, R.}, \bibinfo{year}{2024}.
\newblock \bibinfo{title}{Deep learning surrogate models of jules-inferno for wildfire prediction on a global scale}.
\newblock \bibinfo{journal}{IEEE Transactions on Emerging Topics in Computational Intelligence} , \bibinfo{pages}{1--11}\DOIprefix\doi{10.1109/TETCI.2024.3445450}.
%Type = Article
\bibitem[{Cheng et~al.(2022)Cheng, Prentice, Huang, Jin, Guo and Arcucci}]{cheng2022data}
\bibinfo{author}{Cheng, S.}, \bibinfo{author}{Prentice, I.C.}, \bibinfo{author}{Huang, Y.}, \bibinfo{author}{Jin, Y.}, \bibinfo{author}{Guo, Y.K.}, \bibinfo{author}{Arcucci, R.}, \bibinfo{year}{2022}.
\newblock \bibinfo{title}{Data-driven surrogate model with latent data assimilation: Application to wildfire forecasting}.
\newblock \bibinfo{journal}{Journal of Computational Physics} \bibinfo{volume}{464}, \bibinfo{pages}{111302}.
%Type = Article
\bibitem[{Cheng and Wang(2008)}]{cheng2008integrated}
\bibinfo{author}{Cheng, T.}, \bibinfo{author}{Wang, J.}, \bibinfo{year}{2008}.
\newblock \bibinfo{title}{Integrated spatio-temporal data mining for forest fire prediction}.
\newblock \bibinfo{journal}{Transactions in GIS} \bibinfo{volume}{12}, \bibinfo{pages}{591--611}.
%Type = Article
\bibitem[{Chicas et~al.(2022)Chicas, {\O}stergaard~Nielsen, Valdez and Chen}]{chicas2022modelling}
\bibinfo{author}{Chicas, S.D.}, \bibinfo{author}{{\O}stergaard~Nielsen, J.}, \bibinfo{author}{Valdez, M.C.}, \bibinfo{author}{Chen, C.F.}, \bibinfo{year}{2022}.
\newblock \bibinfo{title}{Modelling wildfire susceptibility in belize’s ecosystems and protected areas using machine learning and knowledge-based methods}.
\newblock \bibinfo{journal}{Geocarto International} \bibinfo{volume}{37}, \bibinfo{pages}{15823--15846}.
%Type = Inproceedings
\bibitem[{Chino et~al.(2015)Chino, Avalhais, Rodrigues and Traina}]{chino2015bowfire}
\bibinfo{author}{Chino, D.Y.}, \bibinfo{author}{Avalhais, L.P.}, \bibinfo{author}{Rodrigues, J.F.}, \bibinfo{author}{Traina, A.J.}, \bibinfo{year}{2015}.
\newblock \bibinfo{title}{Bowfire: detection of fire in still images by integrating pixel color and texture analysis}, in: \bibinfo{booktitle}{2015 28th SIBGRAPI conference on graphics, patterns and images}, \bibinfo{organization}{IEEE}. pp. \bibinfo{pages}{95--102}.
%Type = Inproceedings
\bibitem[{Chng et~al.(2024)Chng, Zheng, Han, Qiu and Huang}]{Chng_2024_CVPR}
\bibinfo{author}{Chng, Y.X.}, \bibinfo{author}{Zheng, H.}, \bibinfo{author}{Han, Y.}, \bibinfo{author}{Qiu, X.}, \bibinfo{author}{Huang, G.}, \bibinfo{year}{2024}.
\newblock \bibinfo{title}{Mask grounding for referring image segmentation}, in: \bibinfo{booktitle}{Proceedings of the IEEE/CVF Conference on Computer Vision and Pattern Recognition (CVPR)}, pp. \bibinfo{pages}{26573--26583}.
%Type = Article
\bibitem[{Chung et~al.(2014)Chung, Gulcehre, Cho and Bengio}]{chung2014empirical}
\bibinfo{author}{Chung, J.}, \bibinfo{author}{Gulcehre, C.}, \bibinfo{author}{Cho, K.}, \bibinfo{author}{Bengio, Y.}, \bibinfo{year}{2014}.
\newblock \bibinfo{title}{Empirical evaluation of gated recurrent neural networks on sequence modeling}.
\newblock \bibinfo{journal}{arXiv preprint arXiv:1412.3555} .
%Type = Article
\bibitem[{Chuvieco et~al.(2012)Chuvieco, Aguado, Jurdao, Pettinari, Yebra, Salas, Hantson, de~la Riva, Ibarra, Rodrigues et~al.}]{chuvieco2012integrating}
\bibinfo{author}{Chuvieco, E.}, \bibinfo{author}{Aguado, I.}, \bibinfo{author}{Jurdao, S.}, \bibinfo{author}{Pettinari, M.L.}, \bibinfo{author}{Yebra, M.}, \bibinfo{author}{Salas, J.}, \bibinfo{author}{Hantson, S.}, \bibinfo{author}{de~la Riva, J.}, \bibinfo{author}{Ibarra, P.}, \bibinfo{author}{Rodrigues, M.}, et~al., \bibinfo{year}{2012}.
\newblock \bibinfo{title}{Integrating geospatial information into fire risk assessment}.
\newblock \bibinfo{journal}{International journal of wildland fire} \bibinfo{volume}{23}, \bibinfo{pages}{606--619}.
%Type = Article
\bibitem[{Chuvieco et~al.(2020)Chuvieco, Aguado, Salas, Garc{\'\i}a, Yebra and Oliva}]{chuvieco2020satellite}
\bibinfo{author}{Chuvieco, E.}, \bibinfo{author}{Aguado, I.}, \bibinfo{author}{Salas, J.}, \bibinfo{author}{Garc{\'\i}a, M.}, \bibinfo{author}{Yebra, M.}, \bibinfo{author}{Oliva, P.}, \bibinfo{year}{2020}.
\newblock \bibinfo{title}{Satellite remote sensing contributions to wildland fire science and management}.
\newblock \bibinfo{journal}{Current Forestry Reports} \bibinfo{volume}{6}, \bibinfo{pages}{81--96}.
%Type = Article
\bibitem[{Chuvieco et~al.(2004)Chuvieco, Cocero, Riano, Martin, Mart{\i}nez-Vega, De~La~Riva and P{\'e}rez}]{chuvieco2004combining}
\bibinfo{author}{Chuvieco, E.}, \bibinfo{author}{Cocero, D.}, \bibinfo{author}{Riano, D.}, \bibinfo{author}{Martin, P.}, \bibinfo{author}{Mart{\i}nez-Vega, J.}, \bibinfo{author}{De~La~Riva, J.}, \bibinfo{author}{P{\'e}rez, F.}, \bibinfo{year}{2004}.
\newblock \bibinfo{title}{Combining ndvi and surface temperature for the estimation of live fuel moisture content in forest fire danger rating}.
\newblock \bibinfo{journal}{Remote Sensing of Environment} \bibinfo{volume}{92}, \bibinfo{pages}{322--331}.
%Type = Article
\bibitem[{Chuvieco and Congalton(1989)}]{chuvieco1989application}
\bibinfo{author}{Chuvieco, E.}, \bibinfo{author}{Congalton, R.G.}, \bibinfo{year}{1989}.
\newblock \bibinfo{title}{Application of remote sensing and geographic information systems to forest fire hazard mapping}.
\newblock \bibinfo{journal}{Remote sensing of Environment} \bibinfo{volume}{29}, \bibinfo{pages}{147--159}.
%Type = Incollection
\bibitem[{Chuvieco et~al.(1999)Chuvieco, Deshayes, Stach, Cocero and Riaño}]{chuvieco_short-term_1999}
\bibinfo{author}{Chuvieco, E.}, \bibinfo{author}{Deshayes, M.}, \bibinfo{author}{Stach, N.}, \bibinfo{author}{Cocero, D.}, \bibinfo{author}{Riaño, D.}, \bibinfo{year}{1999}.
\newblock \bibinfo{title}{Short-term fire risk: foliage moisture content estimation from satellite data}, in: \bibinfo{editor}{Chuvieco, E.} (Ed.), \bibinfo{booktitle}{Remote {Sensing} of {Large} {Wildfires}: in the {European} {Mediterranean} {Basin}}. \bibinfo{publisher}{Springer}, \bibinfo{address}{Berlin, Heidelberg}, pp. \bibinfo{pages}{17--38}.
\newblock \URLprefix \url{https://doi.org/10.1007/978-3-642-60164-4_3}, \DOIprefix\doi{10.1007/978-3-642-60164-4_3}.
%Type = Article
\bibitem[{Chuvieco et~al.(2002)Chuvieco, Ria{\~n}o, Aguado and Cocero}]{chuvieco2002estimation}
\bibinfo{author}{Chuvieco, E.}, \bibinfo{author}{Ria{\~n}o, D.}, \bibinfo{author}{Aguado, I.}, \bibinfo{author}{Cocero, D.}, \bibinfo{year}{2002}.
\newblock \bibinfo{title}{Estimation of fuel moisture content from multitemporal analysis of landsat thematic mapper reflectance data: applications in fire danger assessment}.
\newblock \bibinfo{journal}{International journal of remote sensing} \bibinfo{volume}{23}, \bibinfo{pages}{2145--2162}.
%Type = Article
\bibitem[{Clark et~al.(2009)Clark, Skowronski, Hom, Duveneck, Pan, Van~Tuyl, Cole, Patterson and Maurer}]{clark2009decision}
\bibinfo{author}{Clark, K.L.}, \bibinfo{author}{Skowronski, N.}, \bibinfo{author}{Hom, J.}, \bibinfo{author}{Duveneck, M.}, \bibinfo{author}{Pan, Y.}, \bibinfo{author}{Van~Tuyl, S.}, \bibinfo{author}{Cole, J.}, \bibinfo{author}{Patterson, M.}, \bibinfo{author}{Maurer, S.}, \bibinfo{year}{2009}.
\newblock \bibinfo{title}{Decision support tools to improve the effectiveness of hazardous fuel reduction treatments in the new jersey pine barrens}.
\newblock \bibinfo{journal}{International journal of wildland fire} \bibinfo{volume}{18}, \bibinfo{pages}{268--277}.
%Type = Article
\bibitem[{Clarke et~al.(2016)Clarke, Pitman, Kala, Carouge, Haverd and Evans}]{clarke2016investigation}
\bibinfo{author}{Clarke, H.}, \bibinfo{author}{Pitman, A.J.}, \bibinfo{author}{Kala, J.}, \bibinfo{author}{Carouge, C.}, \bibinfo{author}{Haverd, V.}, \bibinfo{author}{Evans, J.P.}, \bibinfo{year}{2016}.
\newblock \bibinfo{title}{An investigation of future fuel load and fire weather in australia}.
\newblock \bibinfo{journal}{Climatic Change} \bibinfo{volume}{139}, \bibinfo{pages}{591--605}.
%Type = Article
\bibitem[{Coen et~al.(2013)Coen, Cameron, Michalakes, Patton, Riggan and Yedinak}]{coen_wrf-fire_2013}
\bibinfo{author}{Coen, J.L.}, \bibinfo{author}{Cameron, M.}, \bibinfo{author}{Michalakes, J.}, \bibinfo{author}{Patton, E.G.}, \bibinfo{author}{Riggan, P.J.}, \bibinfo{author}{Yedinak, K.M.}, \bibinfo{year}{2013}.
\newblock \bibinfo{title}{{WRF}-{Fire}: {Coupled} {Weather}–{Wildland} {Fire} {Modeling} with the {Weather} {Research} and {Forecasting} {Model}}.
\newblock \bibinfo{journal}{Journal of Applied Meteorology and Climatology} \bibinfo{volume}{52}, \bibinfo{pages}{16--38}.
\newblock \URLprefix \url{https://journals.ametsoc.org/view/journals/apme/52/1/jamc-d-12-023.1.xml}, \DOIprefix\doi{10.1175/JAMC-D-12-023.1}. \bibinfo{note}{publisher: American Meteorological Society Section: Journal of Applied Meteorology and Climatology}.
%Type = Book
\bibitem[{Cohen(1985)}]{cohen1985national}
\bibinfo{author}{Cohen, J.D.}, \bibinfo{year}{1985}.
\newblock \bibinfo{title}{The national fire-danger rating system: basic equations}. volume~\bibinfo{volume}{82}.
\newblock \bibinfo{publisher}{US Department of Agriculture, Forest Service, Pacific Southwest Forest and~…}.
%Type = Article
\bibitem[{Comesa{\~n}a-Cebral et~al.(2024)Comesa{\~n}a-Cebral, Mart{\'\i}nez-S{\'a}nchez, Su{\'a}rez-Fern{\'a}ndez and Arias}]{comesana2024wildfire}
\bibinfo{author}{Comesa{\~n}a-Cebral, L.}, \bibinfo{author}{Mart{\'\i}nez-S{\'a}nchez, J.}, \bibinfo{author}{Su{\'a}rez-Fern{\'a}ndez, G.}, \bibinfo{author}{Arias, P.}, \bibinfo{year}{2024}.
\newblock \bibinfo{title}{Wildfire response of forest species from multispectral lidar data. a deep learning approach with synthetic data}.
\newblock \bibinfo{journal}{Ecological Informatics} \bibinfo{volume}{81}, \bibinfo{pages}{102612}.
%Type = Article
\bibitem[{Coops(2002)}]{coops2002eucalypt}
\bibinfo{author}{Coops, N.C.}, \bibinfo{year}{2002}.
\newblock \bibinfo{title}{Eucalypt forest structure and synthetic aperture radar backscatter: a theoretical analysis}.
\newblock \bibinfo{journal}{Trees} \bibinfo{volume}{16}, \bibinfo{pages}{28--46}.
%Type = Article
\bibitem[{Coops et~al.(2021)Coops, Tompalski, Goodbody, Queinnec, Luther, Bolton, White, Wulder, van Lier and Hermosilla}]{coops2021modelling}
\bibinfo{author}{Coops, N.C.}, \bibinfo{author}{Tompalski, P.}, \bibinfo{author}{Goodbody, T.R.}, \bibinfo{author}{Queinnec, M.}, \bibinfo{author}{Luther, J.E.}, \bibinfo{author}{Bolton, D.K.}, \bibinfo{author}{White, J.C.}, \bibinfo{author}{Wulder, M.A.}, \bibinfo{author}{van Lier, O.R.}, \bibinfo{author}{Hermosilla, T.}, \bibinfo{year}{2021}.
\newblock \bibinfo{title}{Modelling lidar-derived estimates of forest attributes over space and time: A review of approaches and future trends}.
\newblock \bibinfo{journal}{Remote Sensing of Environment} \bibinfo{volume}{260}, \bibinfo{pages}{112477}.
%Type = Article
\bibitem[{Coskuner(2022)}]{coskuner2022assessing}
\bibinfo{author}{Coskuner, K.A.}, \bibinfo{year}{2022}.
\newblock \bibinfo{title}{Assessing the performance of modis and viirs active fire products in the monitoring of wildfires: a case study in turkey}.
\newblock \bibinfo{journal}{iForest-Biogeosciences and Forestry} \bibinfo{volume}{15}, \bibinfo{pages}{85}.
%Type = Article
\bibitem[{Cruz et~al.(2003)Cruz, Alexander and Wakimoto}]{cruz2003assessing}
\bibinfo{author}{Cruz, M.G.}, \bibinfo{author}{Alexander, M.E.}, \bibinfo{author}{Wakimoto, R.H.}, \bibinfo{year}{2003}.
\newblock \bibinfo{title}{Assessing canopy fuel stratum characteristics in crown fire prone fuel types of western north america}.
\newblock \bibinfo{journal}{International Journal of Wildland Fire} \bibinfo{volume}{12}, \bibinfo{pages}{39--50}.
%Type = Article
\bibitem[{Cruz et~al.(2004)Cruz, Alexander and Wakimoto}]{cruz2004modeling}
\bibinfo{author}{Cruz, M.G.}, \bibinfo{author}{Alexander, M.E.}, \bibinfo{author}{Wakimoto, R.H.}, \bibinfo{year}{2004}.
\newblock \bibinfo{title}{Modeling the likelihood of crown fire occurrence in conifer forest stands}.
\newblock \bibinfo{journal}{Forest Science} \bibinfo{volume}{50}, \bibinfo{pages}{640--658}.
%Type = Article
\bibitem[{Cruz et~al.(2005)Cruz, Alexander and Wakimoto}]{cruz2005development}
\bibinfo{author}{Cruz, M.G.}, \bibinfo{author}{Alexander, M.E.}, \bibinfo{author}{Wakimoto, R.H.}, \bibinfo{year}{2005}.
\newblock \bibinfo{title}{Development and testing of models for predicting crown fire rate of spread in conifer forest stands}.
\newblock \bibinfo{journal}{Canadian Journal of Forest Research} \bibinfo{volume}{35}, \bibinfo{pages}{1626--1639}.
%Type = Article
\bibitem[{Cunill~Camprubi et~al.(2022)Cunill~Camprubi, Gonz{\'a}lez-Moreno and Resco~de Dios}]{cunill2022live}
\bibinfo{author}{Cunill~Camprubi, A.}, \bibinfo{author}{Gonz{\'a}lez-Moreno, P.}, \bibinfo{author}{Resco~de Dios, V.}, \bibinfo{year}{2022}.
\newblock \bibinfo{title}{Live fuel moisture content mapping in the mediterranean basin using random forests and combining modis spectral and thermal data}.
\newblock \bibinfo{journal}{Remote Sensing} \bibinfo{volume}{14}, \bibinfo{pages}{3162}.
%Type = Article
\bibitem[{Cunningham et~al.(2024)Cunningham, Williamson and Bowman}]{cunningham2024increasing}
\bibinfo{author}{Cunningham, C.X.}, \bibinfo{author}{Williamson, G.J.}, \bibinfo{author}{Bowman, D.M.}, \bibinfo{year}{2024}.
\newblock \bibinfo{title}{Increasing frequency and intensity of the most extreme wildfires on earth}.
\newblock \bibinfo{journal}{Nature ecology \& evolution} \bibinfo{volume}{8}, \bibinfo{pages}{1420--1425}.
%Type = Inproceedings
\bibitem[{Dal~Pozzolo et~al.(2015)Dal~Pozzolo, Caelen, Johnson and Bontempi}]{dal2015calibrating}
\bibinfo{author}{Dal~Pozzolo, A.}, \bibinfo{author}{Caelen, O.}, \bibinfo{author}{Johnson, R.A.}, \bibinfo{author}{Bontempi, G.}, \bibinfo{year}{2015}.
\newblock \bibinfo{title}{Calibrating probability with undersampling for unbalanced classification}, in: \bibinfo{booktitle}{2015 IEEE symposium series on computational intelligence}, \bibinfo{organization}{IEEE}. pp. \bibinfo{pages}{159--166}.
%Type = Article
\bibitem[{Danson and Bowyer(2004)}]{danson2004estimating}
\bibinfo{author}{Danson, F.M.}, \bibinfo{author}{Bowyer, P.}, \bibinfo{year}{2004}.
\newblock \bibinfo{title}{Estimating live fuel moisture content from remotely sensed reflectance}.
\newblock \bibinfo{journal}{Remote Sensing of Environment} \bibinfo{volume}{92}, \bibinfo{pages}{309--321}.
%Type = Article
\bibitem[{Dasgupta et~al.(2007)Dasgupta, Qu, Hao and Bhoi}]{dasgupta2007evaluating}
\bibinfo{author}{Dasgupta, S.}, \bibinfo{author}{Qu, J.J.}, \bibinfo{author}{Hao, X.}, \bibinfo{author}{Bhoi, S.}, \bibinfo{year}{2007}.
\newblock \bibinfo{title}{Evaluating remotely sensed live fuel moisture estimations for fire behavior predictions in georgia, usa}.
\newblock \bibinfo{journal}{Remote Sensing of Environment} \bibinfo{volume}{108}, \bibinfo{pages}{138--150}.
%Type = Article
\bibitem[{De~Roo et~al.(2001)De~Roo, Du, Ulaby and Dobson}]{de2001semi}
\bibinfo{author}{De~Roo, R.D.}, \bibinfo{author}{Du, Y.}, \bibinfo{author}{Ulaby, F.T.}, \bibinfo{author}{Dobson, M.C.}, \bibinfo{year}{2001}.
\newblock \bibinfo{title}{A semi-empirical backscattering model at l-band and c-band for a soybean canopy with soil moisture inversion}.
\newblock \bibinfo{journal}{IEEE Transactions on Geoscience and Remote Sensing} \bibinfo{volume}{39}, \bibinfo{pages}{864--872}.
%Type = Book
\bibitem[{Deeming(1972)}]{deeming1972national}
\bibinfo{author}{Deeming, J.E.}, \bibinfo{year}{1972}.
\newblock \bibinfo{title}{National fire-danger rating system}. volume~\bibinfo{volume}{84}.
\newblock \bibinfo{publisher}{Rocky Mountain Forest and Range Experiment Station, Forest Service, US~…}.
%Type = Article
\bibitem[{Deltell et~al.(2024)Deltell, Santana and Jaime~Baeza}]{deltell2024allometric}
\bibinfo{author}{Deltell, L.}, \bibinfo{author}{Santana, V.M.}, \bibinfo{author}{Jaime~Baeza, M.}, \bibinfo{year}{2024}.
\newblock \bibinfo{title}{Allometric equations to calculate living and dead fuel loads in mediterranean species}.
\newblock \bibinfo{journal}{European Journal of Forest Research} \bibinfo{volume}{143}, \bibinfo{pages}{739--749}.
%Type = Article
\bibitem[{Deng et~al.(2023)Deng, Wang, Gu, Chen, Liu, Xie, Weng, Ding and Li}]{deng2023wildfire}
\bibinfo{author}{Deng, J.}, \bibinfo{author}{Wang, W.}, \bibinfo{author}{Gu, G.}, \bibinfo{author}{Chen, Z.}, \bibinfo{author}{Liu, J.}, \bibinfo{author}{Xie, G.}, \bibinfo{author}{Weng, S.}, \bibinfo{author}{Ding, L.}, \bibinfo{author}{Li, C.}, \bibinfo{year}{2023}.
\newblock \bibinfo{title}{Wildfire susceptibility prediction using a multisource and spatiotemporal cooperative approach}.
\newblock \bibinfo{journal}{Earth Science Informatics} \bibinfo{volume}{16}, \bibinfo{pages}{3511--3529}.
%Type = Article
\bibitem[{Deng et~al.(2022)Deng, Volkamer, Wang, Snider, Kille and Romero-Alvarez}]{deng2022wildfire}
\bibinfo{author}{Deng, M.}, \bibinfo{author}{Volkamer, R.M.}, \bibinfo{author}{Wang, Z.}, \bibinfo{author}{Snider, J.R.}, \bibinfo{author}{Kille, N.}, \bibinfo{author}{Romero-Alvarez, L.J.}, \bibinfo{year}{2022}.
\newblock \bibinfo{title}{Wildfire smoke observations in the western united states from the airborne wyoming cloud lidar during the bb-flux project. part ii: Vertical structure and plume injection height}.
\newblock \bibinfo{journal}{Journal of Atmospheric and Oceanic Technology} \bibinfo{volume}{39}, \bibinfo{pages}{559--572}.
%Type = Misc
\bibitem[{{Department of Agriculture Forestry and Fisheries South Africa}(2013)}]{southafrica2013fire}
\bibinfo{author}{{Department of Agriculture Forestry and Fisheries South Africa}}, \bibinfo{year}{2013}.
\newblock \bibinfo{title}{{Notice 1099 of 2013: Publication of the fire danger rating system for general information in terms of section(9)1 on the National Veld and Forest Fire Act 1998 (Act No. 101, 1998)}}.
\newblock \bibinfo{note}{Department of Agriculture Forestry and Fisheries, South Africa}.
%Type = Article
\bibitem[{Depicker et~al.(2020)Depicker, De~Baets and Baetens}]{depicker2020wildfire}
\bibinfo{author}{Depicker, A.}, \bibinfo{author}{De~Baets, B.}, \bibinfo{author}{Baetens, J.M.}, \bibinfo{year}{2020}.
\newblock \bibinfo{title}{Wildfire ignition probability in belgium}.
\newblock \bibinfo{journal}{Natural hazards and earth system sciences} \bibinfo{volume}{20}, \bibinfo{pages}{363--376}.
%Type = Inproceedings
\bibitem[{Devlin et~al.(2019)Devlin, Chang, Lee and Toutanova}]{devlin-etal-2019-bert}
\bibinfo{author}{Devlin, J.}, \bibinfo{author}{Chang, M.W.}, \bibinfo{author}{Lee, K.}, \bibinfo{author}{Toutanova, K.}, \bibinfo{year}{2019}.
\newblock \bibinfo{title}{{BERT}: Pre-training of deep bidirectional transformers for language understanding}, in: \bibinfo{editor}{Burstein, J.}, \bibinfo{editor}{Doran, C.}, \bibinfo{editor}{Solorio, T.} (Eds.), \bibinfo{booktitle}{Proceedings of the 2019 Conference of the North {A}merican Chapter of the Association for Computational Linguistics: Human Language Technologies, Volume 1 (Long and Short Papers)}, \bibinfo{publisher}{Association for Computational Linguistics}, \bibinfo{address}{Minneapolis, Minnesota}. pp. \bibinfo{pages}{4171--4186}.
\newblock \URLprefix \url{https://aclanthology.org/N19-1423}, \DOIprefix\doi{10.18653/v1/N19-1423}.
%Type = Article
\bibitem[{Dong et~al.(2024)Dong, Zhao, Wang, Tian and Li}]{dong2024deep}
\bibinfo{author}{Dong, Z.}, \bibinfo{author}{Zhao, F.}, \bibinfo{author}{Wang, G.}, \bibinfo{author}{Tian, Y.}, \bibinfo{author}{Li, H.}, \bibinfo{year}{2024}.
\newblock \bibinfo{title}{A deep learning framework: Predicting fire radiative power from the combination of polar-orbiting and geostationary satellite data during wildfire spread}.
\newblock \bibinfo{journal}{IEEE Journal of Selected Topics in Applied Earth Observations and Remote Sensing} .
%Type = Article
\bibitem[{Donovan et~al.(2020)Donovan, Wonkka, Wedin and Twidwell}]{donovan2020land}
\bibinfo{author}{Donovan, V.M.}, \bibinfo{author}{Wonkka, C.L.}, \bibinfo{author}{Wedin, D.A.}, \bibinfo{author}{Twidwell, D.}, \bibinfo{year}{2020}.
\newblock \bibinfo{title}{Land-use type as a driver of large wildfire occurrence in the us great plains}.
\newblock \bibinfo{journal}{Remote Sensing} \bibinfo{volume}{12}, \bibinfo{pages}{1869}.
%Type = Article
\bibitem[{Dorigo et~al.(2017)Dorigo, Wagner, Albergel, Albrecht, Balsamo, Brocca, Chung, Ertl, Forkel, Gruber et~al.}]{dorigo2017esa}
\bibinfo{author}{Dorigo, W.}, \bibinfo{author}{Wagner, W.}, \bibinfo{author}{Albergel, C.}, \bibinfo{author}{Albrecht, F.}, \bibinfo{author}{Balsamo, G.}, \bibinfo{author}{Brocca, L.}, \bibinfo{author}{Chung, D.}, \bibinfo{author}{Ertl, M.}, \bibinfo{author}{Forkel, M.}, \bibinfo{author}{Gruber, A.}, et~al., \bibinfo{year}{2017}.
\newblock \bibinfo{title}{Esa cci soil moisture for improved earth system understanding: State-of-the art and future directions}.
\newblock \bibinfo{journal}{Remote Sensing of Environment} \bibinfo{volume}{203}, \bibinfo{pages}{185--215}.
%Type = Article
\bibitem[{Duff et~al.(2012)Duff, Bell and York}]{duff_predicting_2012}
\bibinfo{author}{Duff, T.J.}, \bibinfo{author}{Bell, T.L.}, \bibinfo{author}{York, A.}, \bibinfo{year}{2012}.
\newblock \bibinfo{title}{Predicting continuous variation in forest fuel load using biophysical models: a case study in south-eastern {Australia}}.
\newblock \bibinfo{journal}{International Journal of Wildland Fire} \bibinfo{volume}{22}, \bibinfo{pages}{318--332}.
\newblock \URLprefix \url{https://www.publish.csiro.au/wf/WF11087}, \DOIprefix\doi{10.1071/WF11087}. \bibinfo{note}{publisher: CSIRO PUBLISHING}.
%Type = Article
\bibitem[{Duff et~al.(2017)Duff, Keane, Penman and Tolhurst}]{duff2017revisiting}
\bibinfo{author}{Duff, T.J.}, \bibinfo{author}{Keane, R.E.}, \bibinfo{author}{Penman, T.D.}, \bibinfo{author}{Tolhurst, K.G.}, \bibinfo{year}{2017}.
\newblock \bibinfo{title}{Revisiting wildland fire fuel quantification methods: the challenge of understanding a dynamic, biotic entity}.
\newblock \bibinfo{journal}{Forests} \bibinfo{volume}{8}, \bibinfo{pages}{351}.
%Type = Article
\bibitem[{Dutta et~al.(2013)Dutta, Aryal, Das and Kirkpatrick}]{dutta2013deep}
\bibinfo{author}{Dutta, R.}, \bibinfo{author}{Aryal, J.}, \bibinfo{author}{Das, A.}, \bibinfo{author}{Kirkpatrick, J.B.}, \bibinfo{year}{2013}.
\newblock \bibinfo{title}{Deep cognitive imaging systems enable estimation of continental-scale fire incidence from climate data}.
\newblock \bibinfo{journal}{Scientific reports} \bibinfo{volume}{3}, \bibinfo{pages}{3188}.
%Type = Inproceedings
\bibitem[{Dzulhijjah et~al.(2023)Dzulhijjah, Majid, Alwanda, Kusuma, Zakaria, Kusrini and Kusnawi}]{10455879}
\bibinfo{author}{Dzulhijjah, D.A.}, \bibinfo{author}{Majid, M.N.}, \bibinfo{author}{Alwanda, A.Y.}, \bibinfo{author}{Kusuma, D.C.}, \bibinfo{author}{Zakaria, F.}, \bibinfo{author}{Kusrini, K.}, \bibinfo{author}{Kusnawi, K.}, \bibinfo{year}{2023}.
\newblock \bibinfo{title}{Comparative analysis of hybrid long short-term memory models for fire danger index forecasting with weather data}, in: \bibinfo{booktitle}{2023 6th International Conference on Information and Communications Technology (ICOIACT)}, pp. \bibinfo{pages}{165--170}.
\newblock \DOIprefix\doi{10.1109/ICOIACT59844.2023.10455879}.
%Type = Article
\bibitem[{D’Este et~al.(2021)D’Este, Elia, Giannico, Spano, Lafortezza and Sanesi}]{d2021machine}
\bibinfo{author}{D’Este, M.}, \bibinfo{author}{Elia, M.}, \bibinfo{author}{Giannico, V.}, \bibinfo{author}{Spano, G.}, \bibinfo{author}{Lafortezza, R.}, \bibinfo{author}{Sanesi, G.}, \bibinfo{year}{2021}.
\newblock \bibinfo{title}{Machine learning techniques for fine dead fuel load estimation using multi-source remote sensing data}.
\newblock \bibinfo{journal}{Remote Sensing} \bibinfo{volume}{13}, \bibinfo{pages}{1658}.
%Type = Article
\bibitem[{Eames et~al.(2021)Eames, Russell-Smith, Yates, Edwards, Vernooij, Ribeiro, Steinbruch and van~der Werf}]{eames2021instantaneous}
\bibinfo{author}{Eames, T.}, \bibinfo{author}{Russell-Smith, J.}, \bibinfo{author}{Yates, C.}, \bibinfo{author}{Edwards, A.}, \bibinfo{author}{Vernooij, R.}, \bibinfo{author}{Ribeiro, N.}, \bibinfo{author}{Steinbruch, F.}, \bibinfo{author}{van~der Werf, G.R.}, \bibinfo{year}{2021}.
\newblock \bibinfo{title}{Instantaneous pre-fire biomass and fuel load measurements from multi-spectral uas mapping in southern african savannas}.
\newblock \bibinfo{journal}{Fire} \bibinfo{volume}{4}, \bibinfo{pages}{2}.
%Type = Article
\bibitem[{Eddin et~al.(2023)Eddin, Roscher and Gall}]{eddin2023location}
\bibinfo{author}{Eddin, M.H.S.}, \bibinfo{author}{Roscher, R.}, \bibinfo{author}{Gall, J.}, \bibinfo{year}{2023}.
\newblock \bibinfo{title}{Location-aware adaptive normalization: a deep learning approach for wildfire danger forecasting}.
\newblock \bibinfo{journal}{IEEE Transactions on Geoscience and Remote Sensing} \bibinfo{volume}{61}, \bibinfo{pages}{1--18}.
%Type = Article
\bibitem[{El-Madafri et~al.(2023)El-Madafri, Pe{\~n}a and Olmedo-Torre}]{el2023wildfire}
\bibinfo{author}{El-Madafri, I.}, \bibinfo{author}{Pe{\~n}a, M.}, \bibinfo{author}{Olmedo-Torre, N.}, \bibinfo{year}{2023}.
\newblock \bibinfo{title}{The wildfire dataset: Enhancing deep learning-based forest fire detection with a diverse evolving open-source dataset focused on data representativeness and a novel multi-task learning approach}.
\newblock \bibinfo{journal}{Forests} \bibinfo{volume}{14}, \bibinfo{pages}{1697}.
%Type = Article
\bibitem[{Elia et~al.(2020)Elia, Giannico, Spano, Lafortezza and Sanesi}]{elia2020likelihood}
\bibinfo{author}{Elia, M.}, \bibinfo{author}{Giannico, V.}, \bibinfo{author}{Spano, G.}, \bibinfo{author}{Lafortezza, R.}, \bibinfo{author}{Sanesi, G.}, \bibinfo{year}{2020}.
\newblock \bibinfo{title}{Likelihood and frequency of recurrent fire ignitions in highly urbanised mediterranean landscapes}.
\newblock \bibinfo{journal}{International journal of wildland fire} \bibinfo{volume}{29}, \bibinfo{pages}{120--131}.
%Type = Inproceedings
\bibitem[{Elkan(2001)}]{elkan2001foundations}
\bibinfo{author}{Elkan, C.}, \bibinfo{year}{2001}.
\newblock \bibinfo{title}{The foundations of cost-sensitive learning}, in: \bibinfo{booktitle}{International joint conference on artificial intelligence}, \bibinfo{organization}{Lawrence Erlbaum Associates Ltd}. pp. \bibinfo{pages}{973--978}.
%Type = Phdthesis
\bibitem[{Erhan et~al.(2009)Erhan, Bengio, Courville and Vincent}]{erhan2009visualizing}
\bibinfo{author}{Erhan, D.}, \bibinfo{author}{Bengio, Y.}, \bibinfo{author}{Courville, A.}, \bibinfo{author}{Vincent, P.}, \bibinfo{year}{2009}.
\newblock \bibinfo{title}{Visualizing higher-layer features of a deep network}.
\newblock Ph.D. thesis. University of Montreal.
%Type = Article
\bibitem[{Erni et~al.(2024)Erni, Wang, Swystun, Taylor, Parisien, Robinne, Eddy, Oliver, Armitage and Flannigan}]{erni2024mapping}
\bibinfo{author}{Erni, S.}, \bibinfo{author}{Wang, X.}, \bibinfo{author}{Swystun, T.}, \bibinfo{author}{Taylor, S.W.}, \bibinfo{author}{Parisien, M.A.}, \bibinfo{author}{Robinne, F.N.}, \bibinfo{author}{Eddy, B.}, \bibinfo{author}{Oliver, J.}, \bibinfo{author}{Armitage, B.}, \bibinfo{author}{Flannigan, M.D.}, \bibinfo{year}{2024}.
\newblock \bibinfo{title}{Mapping wildfire hazard, vulnerability, and risk to canadian communities}.
\newblock \bibinfo{journal}{International Journal of Disaster Risk Reduction} \bibinfo{volume}{101}, \bibinfo{pages}{104221}.
%Type = Article
\bibitem[{Espeholt et~al.(2022)Espeholt, Agrawal, S{\o}nderby, Kumar, Heek, Bromberg, Gazen, Carver, Andrychowicz, Hickey et~al.}]{espeholt2022deep}
\bibinfo{author}{Espeholt, L.}, \bibinfo{author}{Agrawal, S.}, \bibinfo{author}{S{\o}nderby, C.}, \bibinfo{author}{Kumar, M.}, \bibinfo{author}{Heek, J.}, \bibinfo{author}{Bromberg, C.}, \bibinfo{author}{Gazen, C.}, \bibinfo{author}{Carver, R.}, \bibinfo{author}{Andrychowicz, M.}, \bibinfo{author}{Hickey, J.}, et~al., \bibinfo{year}{2022}.
\newblock \bibinfo{title}{Deep learning for twelve hour precipitation forecasts}.
\newblock \bibinfo{journal}{Nature communications} \bibinfo{volume}{13}, \bibinfo{pages}{1--10}.
%Type = Article
\bibitem[{Fan and He(2021)}]{fan2021physics}
\bibinfo{author}{Fan, C.}, \bibinfo{author}{He, B.}, \bibinfo{year}{2021}.
\newblock \bibinfo{title}{A physics-guided deep learning model for 10-h dead fuel moisture content estimation}.
\newblock \bibinfo{journal}{Forests} \bibinfo{volume}{12}, \bibinfo{pages}{933}.
%Type = Article
\bibitem[{Fan et~al.(2024)Fan, He, Yin, Chen and Zhang}]{fan2024process}
\bibinfo{author}{Fan, C.}, \bibinfo{author}{He, B.}, \bibinfo{author}{Yin, J.}, \bibinfo{author}{Chen, R.}, \bibinfo{author}{Zhang, H.}, \bibinfo{year}{2024}.
\newblock \bibinfo{title}{Process-based and geostationary meteorological satellite-enhanced dead fuel moisture content estimation}.
\newblock \bibinfo{journal}{GIScience \& Remote Sensing} \bibinfo{volume}{61}, \bibinfo{pages}{2324556}.
%Type = Article
\bibitem[{Fan et~al.(2018)Fan, Wigneron, Xiao, Al-Yaari, Wen, Martin-StPaul, Dupuy, Pimont, Al~Bitar, Fernandez-Moran et~al.}]{fan2018evaluation}
\bibinfo{author}{Fan, L.}, \bibinfo{author}{Wigneron, J.P.}, \bibinfo{author}{Xiao, Q.}, \bibinfo{author}{Al-Yaari, A.}, \bibinfo{author}{Wen, J.}, \bibinfo{author}{Martin-StPaul, N.}, \bibinfo{author}{Dupuy, J.L.}, \bibinfo{author}{Pimont, F.}, \bibinfo{author}{Al~Bitar, A.}, \bibinfo{author}{Fernandez-Moran, R.}, et~al., \bibinfo{year}{2018}.
\newblock \bibinfo{title}{Evaluation of microwave remote sensing for monitoring live fuel moisture content in the mediterranean region}.
\newblock \bibinfo{journal}{Remote Sensing of Environment} \bibinfo{volume}{205}, \bibinfo{pages}{210--223}.
%Type = Techreport
\bibitem[{FAO(2006)}]{FAO_human_2006}
\bibinfo{author}{FAO}, \bibinfo{year}{2006}.
\newblock \bibinfo{title}{Fire management-Global assessment 2006}.
\newblock \bibinfo{type}{Technical Report}. Food and Agriculture Organization of the United Nations.
%Type = Inproceedings
\bibitem[{Feng et~al.(2021)Feng, Hu, Zhang and Lu}]{Feng_2021_CVPR}
\bibinfo{author}{Feng, G.}, \bibinfo{author}{Hu, Z.}, \bibinfo{author}{Zhang, L.}, \bibinfo{author}{Lu, H.}, \bibinfo{year}{2021}.
\newblock \bibinfo{title}{Encoder fusion network with co-attention embedding for referring image segmentation}, in: \bibinfo{booktitle}{Proceedings of the IEEE/CVF Conference on Computer Vision and Pattern Recognition (CVPR)}, pp. \bibinfo{pages}{15506--15515}.
%Type = Article
\bibitem[{Fensholt and Sandholt(2003)}]{fensholt_derivation_2003}
\bibinfo{author}{Fensholt, R.}, \bibinfo{author}{Sandholt, I.}, \bibinfo{year}{2003}.
\newblock \bibinfo{title}{Derivation of a shortwave infrared water stress index from {MODIS} near- and shortwave infrared data in a semiarid environment}.
\newblock \bibinfo{journal}{Remote Sensing of Environment} \bibinfo{volume}{87}, \bibinfo{pages}{111--121}.
\newblock \URLprefix \url{https://www.sciencedirect.com/science/article/pii/S0034425703001895}, \DOIprefix\doi{10.1016/j.rse.2003.07.002}.
%Type = Article
\bibitem[{Ferguson et~al.(2002)Ferguson, Ruthford, McKay, Wright, Wright and Ottmar}]{ferguson2002measuring}
\bibinfo{author}{Ferguson, S.A.}, \bibinfo{author}{Ruthford, J.E.}, \bibinfo{author}{McKay, S.J.}, \bibinfo{author}{Wright, D.}, \bibinfo{author}{Wright, C.}, \bibinfo{author}{Ottmar, R.}, \bibinfo{year}{2002}.
\newblock \bibinfo{title}{Measuring moisture dynamics to predict fire severity in longleaf pine forests}.
\newblock \bibinfo{journal}{International Journal of Wildland Fire} \bibinfo{volume}{11}, \bibinfo{pages}{267--279}.
%Type = Article
\bibitem[{Fernandes and Botelho(2003)}]{fernandes2003review}
\bibinfo{author}{Fernandes, P.M.}, \bibinfo{author}{Botelho, H.S.}, \bibinfo{year}{2003}.
\newblock \bibinfo{title}{A review of prescribed burning effectiveness in fire hazard reduction}.
\newblock \bibinfo{journal}{International Journal of wildland fire} \bibinfo{volume}{12}, \bibinfo{pages}{117--128}.
%Type = Article
\bibitem[{Fern{\'a}ndez-Garc{\'\i}a et~al.(2022)Fern{\'a}ndez-Garc{\'\i}a, Beltr{\'a}n-Marcos, Fern{\'a}ndez-Guisuraga, Marcos and Calvo}]{fernandez2022predicting}
\bibinfo{author}{Fern{\'a}ndez-Garc{\'\i}a, V.}, \bibinfo{author}{Beltr{\'a}n-Marcos, D.}, \bibinfo{author}{Fern{\'a}ndez-Guisuraga, J.M.}, \bibinfo{author}{Marcos, E.}, \bibinfo{author}{Calvo, L.}, \bibinfo{year}{2022}.
\newblock \bibinfo{title}{Predicting potential wildfire severity across southern europe with global data sources}.
\newblock \bibinfo{journal}{Science of the total environment} \bibinfo{volume}{829}, \bibinfo{pages}{154729}.
%Type = Article
\bibitem[{Fillmore et~al.(2024)Fillmore, McCaffrey, Bean, Evans, Iniguez, Thode, Smith and Thompson}]{fillmore2024factors}
\bibinfo{author}{Fillmore, S.D.}, \bibinfo{author}{McCaffrey, S.}, \bibinfo{author}{Bean, R.}, \bibinfo{author}{Evans, A.M.}, \bibinfo{author}{Iniguez, J.}, \bibinfo{author}{Thode, A.}, \bibinfo{author}{Smith, A.M.}, \bibinfo{author}{Thompson, M.P.}, \bibinfo{year}{2024}.
\newblock \bibinfo{title}{Factors influencing wildfire management decisions after the 2009 us federal policy update}.
\newblock \bibinfo{journal}{International journal of wildland fire} \bibinfo{volume}{33}.
%Type = Book
\bibitem[{Finney(1998)}]{finney1998farsite}
\bibinfo{author}{Finney, M.A.}, \bibinfo{year}{1998}.
\newblock \bibinfo{title}{FARSITE, Fire Area Simulator--model development and evaluation}.
\newblock \bibinfo{number}{4}, \bibinfo{publisher}{US Department of Agriculture, Forest Service, Rocky Mountain Research Station}.
%Type = Article
\bibitem[{Finney(2002)}]{finney2002fire}
\bibinfo{author}{Finney, M.A.}, \bibinfo{year}{2002}.
\newblock \bibinfo{title}{Fire growth using minimum travel time methods}.
\newblock \bibinfo{journal}{Canadian Journal of Forest Research} \bibinfo{volume}{32}, \bibinfo{pages}{1420--1424}.
%Type = Article
\bibitem[{Flannigan et~al.(2013)Flannigan, Cantin, De~Groot, Wotton, Newbery and Gowman}]{flannigan2013global}
\bibinfo{author}{Flannigan, M.}, \bibinfo{author}{Cantin, A.S.}, \bibinfo{author}{De~Groot, W.J.}, \bibinfo{author}{Wotton, M.}, \bibinfo{author}{Newbery, A.}, \bibinfo{author}{Gowman, L.M.}, \bibinfo{year}{2013}.
\newblock \bibinfo{title}{Global wildland fire season severity in the 21st century}.
\newblock \bibinfo{journal}{Forest Ecology and Management} \bibinfo{volume}{294}, \bibinfo{pages}{54--61}.
%Type = Book
\bibitem[{{Food and Agriculture Organization of the United Nations (FAO)}(2011)}]{fao2011state}
\bibinfo{author}{{Food and Agriculture Organization of the United Nations (FAO)}}, \bibinfo{year}{2011}.
\newblock \bibinfo{title}{{State of the World's Forests 2011}}.
\newblock \bibinfo{publisher}{{Food and Agriculture Organization of the United Nations}}, \bibinfo{address}{Rome}.
%Type = Article
\bibitem[{Forkel et~al.(2023)Forkel, Schmidt, Zotta, Dorigo and Yebra}]{forkel_estimating_2023}
\bibinfo{author}{Forkel, M.}, \bibinfo{author}{Schmidt, L.}, \bibinfo{author}{Zotta, R.M.}, \bibinfo{author}{Dorigo, W.}, \bibinfo{author}{Yebra, M.}, \bibinfo{year}{2023}.
\newblock \bibinfo{title}{Estimating leaf moisture content at global scale from passive microwave satellite observations of vegetation optical depth}.
\newblock \bibinfo{journal}{Hydrology and Earth System Sciences} \bibinfo{volume}{27}, \bibinfo{pages}{39--68}.
\newblock \URLprefix \url{https://hess.copernicus.org/articles/27/39/2023/}, \DOIprefix\doi{10.5194/hess-27-39-2023}. \bibinfo{note}{publisher: Copernicus GmbH}.
%Type = Book
\bibitem[{Frandsen et~al.(1979)Frandsen, Andrews, Forest and Range Experiment Station~(Ogden}]{bhlitem136922}
\bibinfo{author}{Frandsen, W.H.}, \bibinfo{author}{Andrews, P.L.}, \bibinfo{author}{Forest, I.}, \bibinfo{author}{Range Experiment Station~(Ogden, U.}, \bibinfo{year}{1979}.
\newblock \bibinfo{title}{Fire behavior in nonuniform fuels}. volume \bibinfo{volume}{no.232}.
\newblock \bibinfo{publisher}{Ogden, Utah, Intermountain Forest and Range Experiment Station, Forest Service, U.S. Dept. of Agriculture, 1979}.
\newblock \URLprefix \url{https://www.biodiversitylibrary.org/item/136922}. \bibinfo{note}{https://www.biodiversitylibrary.org/bibliography/68702}.
%Type = Book
\bibitem[{Fuller(1991)}]{fuller1991forest}
\bibinfo{author}{Fuller, M.}, \bibinfo{year}{1991}.
\newblock \bibinfo{title}{Forest fires: an introduction to wildland fire behavior, management, firefighting, and prevention.}
\newblock \bibinfo{publisher}{John Wiley \& Sons, Inc}.
%Type = Article
\bibitem[{Fusco et~al.(2019)Fusco, Finn, Abatzoglou, Balch, Dadashi and Bradley}]{fusco2019detection}
\bibinfo{author}{Fusco, E.J.}, \bibinfo{author}{Finn, J.T.}, \bibinfo{author}{Abatzoglou, J.T.}, \bibinfo{author}{Balch, J.K.}, \bibinfo{author}{Dadashi, S.}, \bibinfo{author}{Bradley, B.A.}, \bibinfo{year}{2019}.
\newblock \bibinfo{title}{Detection rates and biases of fire observations from modis and agency reports in the conterminous united states}.
\newblock \bibinfo{journal}{Remote Sensing of Environment} \bibinfo{volume}{220}, \bibinfo{pages}{30--40}.
%Type = Article
\bibitem[{Gao et~al.(2020)Gao, Huang, Zhang, Han, Wang, Zhang and Lin}]{gao2020short}
\bibinfo{author}{Gao, S.}, \bibinfo{author}{Huang, Y.}, \bibinfo{author}{Zhang, S.}, \bibinfo{author}{Han, J.}, \bibinfo{author}{Wang, G.}, \bibinfo{author}{Zhang, M.}, \bibinfo{author}{Lin, Q.}, \bibinfo{year}{2020}.
\newblock \bibinfo{title}{Short-term runoff prediction with gru and lstm networks without requiring time step optimization during sample generation}.
\newblock \bibinfo{journal}{Journal of Hydrology} \bibinfo{volume}{589}, \bibinfo{pages}{125188}.
%Type = Article
\bibitem[{Gao et~al.(2021)Gao, Wang and Zhou}]{gao2021stock}
\bibinfo{author}{Gao, Y.}, \bibinfo{author}{Wang, R.}, \bibinfo{author}{Zhou, E.}, \bibinfo{year}{2021}.
\newblock \bibinfo{title}{Stock prediction based on optimized lstm and gru models}.
\newblock \bibinfo{journal}{Scientific Programming} \bibinfo{volume}{2021}, \bibinfo{pages}{4055281}.
%Type = Article
\bibitem[{Garc{\'\i}a et~al.(2020)Garc{\'\i}a, Ria{\~n}o, Yebra, Salas, Cardil, Monedero, Ramirez, Mart{\'\i}n, Vilar, Gajardo et~al.}]{garcia2020live}
\bibinfo{author}{Garc{\'\i}a, M.}, \bibinfo{author}{Ria{\~n}o, D.}, \bibinfo{author}{Yebra, M.}, \bibinfo{author}{Salas, J.}, \bibinfo{author}{Cardil, A.}, \bibinfo{author}{Monedero, S.}, \bibinfo{author}{Ramirez, J.}, \bibinfo{author}{Mart{\'\i}n, M.P.}, \bibinfo{author}{Vilar, L.}, \bibinfo{author}{Gajardo, J.}, et~al., \bibinfo{year}{2020}.
\newblock \bibinfo{title}{A live fuel moisture content product from landsat tm satellite time series for implementation in fire behavior models}.
\newblock \bibinfo{journal}{Remote Sensing} \bibinfo{volume}{12}, \bibinfo{pages}{1714}.
%Type = Article
\bibitem[{Ge et~al.(2022)Ge, Yang, Peng, Chen, Li, Zhang and Chen}]{ge2022spatio}
\bibinfo{author}{Ge, X.}, \bibinfo{author}{Yang, Y.}, \bibinfo{author}{Peng, L.}, \bibinfo{author}{Chen, L.}, \bibinfo{author}{Li, W.}, \bibinfo{author}{Zhang, W.}, \bibinfo{author}{Chen, J.}, \bibinfo{year}{2022}.
\newblock \bibinfo{title}{Spatio-temporal knowledge graph based forest fire prediction with multi source heterogeneous data}.
\newblock \bibinfo{journal}{Remote Sensing} \bibinfo{volume}{14}, \bibinfo{pages}{3496}.
%Type = Article
\bibitem[{Gentilucci et~al.(2024)Gentilucci, Barbieri, Younes, Rihab and Pambianchi}]{gentilucci_analysis_2024}
\bibinfo{author}{Gentilucci, M.}, \bibinfo{author}{Barbieri, M.}, \bibinfo{author}{Younes, H.}, \bibinfo{author}{Rihab, H.}, \bibinfo{author}{Pambianchi, G.}, \bibinfo{year}{2024}.
\newblock \bibinfo{title}{Analysis of {Wildfire} {Susceptibility} by {Weight} of {Evidence}, {Using} {Geomorphological} and {Environmental} {Factors} in the {Marche} {Region}, {Central} {Italy}}.
\newblock \bibinfo{journal}{Geosciences} \bibinfo{volume}{14}, \bibinfo{pages}{112}.
\newblock \URLprefix \url{https://www.mdpi.com/2076-3263/14/5/112}, \DOIprefix\doi{10.3390/geosciences14050112}. \bibinfo{note}{number: 5 Publisher: Multidisciplinary Digital Publishing Institute}.
%Type = Article
\bibitem[{Gerard et~al.(2023)Gerard, Zhao and Sullivan}]{gerard2023wildfirespreadts}
\bibinfo{author}{Gerard, S.}, \bibinfo{author}{Zhao, Y.}, \bibinfo{author}{Sullivan, J.}, \bibinfo{year}{2023}.
\newblock \bibinfo{title}{Wildfirespreadts: A dataset of multi-modal time series for wildfire spread prediction}.
\newblock \bibinfo{journal}{Advances in Neural Information Processing Systems} \bibinfo{volume}{36}, \bibinfo{pages}{74515--74529}.
%Type = Article
\bibitem[{Gers et~al.(2000)Gers, Schmidhuber and Cummins}]{gers2000learning}
\bibinfo{author}{Gers, F.A.}, \bibinfo{author}{Schmidhuber, J.}, \bibinfo{author}{Cummins, F.}, \bibinfo{year}{2000}.
\newblock \bibinfo{title}{Learning to forget: Continual prediction with lstm}.
\newblock \bibinfo{journal}{Neural computation} \bibinfo{volume}{12}, \bibinfo{pages}{2451--2471}.
%Type = Inproceedings
\bibitem[{Ghorbanzadeh and Blaschke(2018)}]{ghorbanzadeh_wildfire_2018}
\bibinfo{author}{Ghorbanzadeh, O.}, \bibinfo{author}{Blaschke, T.}, \bibinfo{year}{2018}.
\newblock \bibinfo{title}{Wildfire {Susceptibility} {Evaluation} {By} {Integrating} an {Analytical} {Network} {Process} {Approach} {Into} {GIS}-{Based} {Analyses}}, in: \bibinfo{booktitle}{Wildfire {Susceptibility} {Evaluation} {By} {Integrating} an {Analytical} {Network} {Process} {Approach} {Into} {GIS}-{Based} {Analyses}}.
\newblock \URLprefix \url{https://uni-salzburg.elsevierpure.com/en/publications/wildfire-susceptibility-evaluation-by-integrating-an-analytical-n}.
%Type = Article
\bibitem[{Ghorbanzadeh et~al.(2019)Ghorbanzadeh, Blaschke, Gholamnia and Aryal}]{ghorbanzadeh_forest_2019}
\bibinfo{author}{Ghorbanzadeh, O.}, \bibinfo{author}{Blaschke, T.}, \bibinfo{author}{Gholamnia, K.}, \bibinfo{author}{Aryal, J.}, \bibinfo{year}{2019}.
\newblock \bibinfo{title}{Forest {Fire} {Susceptibility} and {Risk} {Mapping} {Using} {Social}/{Infrastructural} {Vulnerability} and {Environmental} {Variables}}.
\newblock \bibinfo{journal}{Fire} \bibinfo{volume}{2}, \bibinfo{pages}{50}.
\newblock \URLprefix \url{https://www.mdpi.com/2571-6255/2/3/50}, \DOIprefix\doi{10.3390/fire2030050}. \bibinfo{note}{number: 3 Publisher: Multidisciplinary Digital Publishing Institute}.
%Type = Article
\bibitem[{Giglio et~al.(2018)Giglio, Boschetti, Roy, Humber and Justice}]{giglio2018collection}
\bibinfo{author}{Giglio, L.}, \bibinfo{author}{Boschetti, L.}, \bibinfo{author}{Roy, D.P.}, \bibinfo{author}{Humber, M.L.}, \bibinfo{author}{Justice, C.O.}, \bibinfo{year}{2018}.
\newblock \bibinfo{title}{The collection 6 modis burned area mapping algorithm and product}.
\newblock \bibinfo{journal}{Remote sensing of environment} \bibinfo{volume}{217}, \bibinfo{pages}{72--85}.
%Type = Article
\bibitem[{Giglio et~al.(2003)Giglio, Descloitres, Justice and Kaufman}]{GIGLIO2003273}
\bibinfo{author}{Giglio, L.}, \bibinfo{author}{Descloitres, J.}, \bibinfo{author}{Justice, C.O.}, \bibinfo{author}{Kaufman, Y.J.}, \bibinfo{year}{2003}.
\newblock \bibinfo{title}{An enhanced contextual fire detection algorithm for modis}.
\newblock \bibinfo{journal}{Remote Sensing of Environment} \bibinfo{volume}{87}, \bibinfo{pages}{273--282}.
\newblock \URLprefix \url{https://www.sciencedirect.com/science/article/pii/S0034425703001846}, \DOIprefix\doi{https://doi.org/10.1016/S0034-4257(03)00184-6}.
%Type = Article
\bibitem[{Giglio et~al.(2013)Giglio, Randerson and Van Der~Werf}]{giglio2013analysis}
\bibinfo{author}{Giglio, L.}, \bibinfo{author}{Randerson, J.T.}, \bibinfo{author}{Van Der~Werf, G.R.}, \bibinfo{year}{2013}.
\newblock \bibinfo{title}{Analysis of daily, monthly, and annual burned area using the fourth-generation global fire emissions database (gfed4)}.
\newblock \bibinfo{journal}{Journal of Geophysical Research: Biogeosciences} \bibinfo{volume}{118}, \bibinfo{pages}{317--328}.
%Type = Article
\bibitem[{Giglio et~al.(2016)Giglio, Schroeder and Justice}]{GIGLIO201631}
\bibinfo{author}{Giglio, L.}, \bibinfo{author}{Schroeder, W.}, \bibinfo{author}{Justice, C.O.}, \bibinfo{year}{2016}.
\newblock \bibinfo{title}{The collection 6 modis active fire detection algorithm and fire products}.
\newblock \bibinfo{journal}{Remote Sensing of Environment} \bibinfo{volume}{178}, \bibinfo{pages}{31--41}.
\newblock \URLprefix \url{https://www.sciencedirect.com/science/article/pii/S0034425716300827}, \DOIprefix\doi{https://doi.org/10.1016/j.rse.2016.02.054}.
%Type = Article
\bibitem[{Gilreath(2006)}]{gilreath_validation_2006}
\bibinfo{author}{Gilreath, J.}, \bibinfo{year}{2006}.
\newblock \bibinfo{title}{Validation of {Variables} for the {Creation} of a {Descriptive} {Fire} {Potential} {Model} for the {Southeastern} {Fire} {District} of {Mississippi}}.
\newblock \bibinfo{journal}{Theses and Dissertations} \URLprefix \url{https://scholarsjunction.msstate.edu/td/4942}.
%Type = Article
\bibitem[{Gincheva et~al.(2024)Gincheva, Pausas, Edwards, Provenzale, Cerd{\`a}, Hanes, Roy{\'e}, Chuvieco, Mouillot, Vissio et~al.}]{gincheva2024monthly}
\bibinfo{author}{Gincheva, A.}, \bibinfo{author}{Pausas, J.G.}, \bibinfo{author}{Edwards, A.}, \bibinfo{author}{Provenzale, A.}, \bibinfo{author}{Cerd{\`a}, A.}, \bibinfo{author}{Hanes, C.}, \bibinfo{author}{Roy{\'e}, D.}, \bibinfo{author}{Chuvieco, E.}, \bibinfo{author}{Mouillot, F.}, \bibinfo{author}{Vissio, G.}, et~al., \bibinfo{year}{2024}.
\newblock \bibinfo{title}{A monthly gridded burned area database of national wildland fire data}.
\newblock \bibinfo{journal}{Scientific data} \bibinfo{volume}{11}, \bibinfo{pages}{352}.
%Type = Techreport
\bibitem[{{Glossary of Wildland Fire Terminology}(2006)}]{NWCG_2006}
\bibinfo{author}{{Glossary of Wildland Fire Terminology}}, \bibinfo{year}{2006}.
\newblock \bibinfo{title}{Glossary of Wildland Fire Terminology}.
\newblock \bibinfo{type}{Technical Report}. National Wildfire Coordinating Group.
%Type = Article
\bibitem[{Gonzalez-Dugo et~al.(2012)Gonzalez-Dugo, Zarco-Tejada, Berni, Suarez, Goldhamer and Fereres}]{gonzalez2012almond}
\bibinfo{author}{Gonzalez-Dugo, V.}, \bibinfo{author}{Zarco-Tejada, P.}, \bibinfo{author}{Berni, J.A.}, \bibinfo{author}{Suarez, L.}, \bibinfo{author}{Goldhamer, D.}, \bibinfo{author}{Fereres, E.}, \bibinfo{year}{2012}.
\newblock \bibinfo{title}{Almond tree canopy temperature reveals intra-crown variability that is water stress-dependent}.
\newblock \bibinfo{journal}{Agricultural and Forest Meteorology} \bibinfo{volume}{154}, \bibinfo{pages}{156--165}.
%Type = Book
\bibitem[{Good and Hardin(2012)}]{good2012common}
\bibinfo{author}{Good, P.I.}, \bibinfo{author}{Hardin, J.W.}, \bibinfo{year}{2012}.
\newblock \bibinfo{title}{Common errors in statistics (and how to avoid them)}.
\newblock \bibinfo{publisher}{John Wiley \& Sons}.
%Type = Inproceedings
\bibitem[{Gopu et~al.(2023)Gopu, Ramakrishnan, Balasubramanian and Srinidhi}]{gopu2023comparative}
\bibinfo{author}{Gopu, A.}, \bibinfo{author}{Ramakrishnan, A.}, \bibinfo{author}{Balasubramanian, G.}, \bibinfo{author}{Srinidhi, K.}, \bibinfo{year}{2023}.
\newblock \bibinfo{title}{A comparative study on forest fire prediction using arima, sarima, lstm, and gru methods}, in: \bibinfo{booktitle}{2023 IEEE International Conference on Contemporary Computing and Communications (InC4)}, \bibinfo{organization}{IEEE}. pp. \bibinfo{pages}{1--5}.
%Type = Article
\bibitem[{Gould et~al.(2011)Gould, McCaw and Cheney}]{gould2011quantifying}
\bibinfo{author}{Gould, J.S.}, \bibinfo{author}{McCaw, W.L.}, \bibinfo{author}{Cheney, N.P.}, \bibinfo{year}{2011}.
\newblock \bibinfo{title}{Quantifying fine fuel dynamics and structure in dry eucalypt forest (eucalyptus marginata) in western australia for fire management}.
\newblock \bibinfo{journal}{Forest ecology and management} \bibinfo{volume}{262}, \bibinfo{pages}{531--546}.
%Type = Article
\bibitem[{Greenwell et~al.(2017)}]{greenwell2017pdp}
\bibinfo{author}{Greenwell, B.M.}, et~al., \bibinfo{year}{2017}.
\newblock \bibinfo{title}{pdp: An r package for constructing partial dependence plots.}
\newblock \bibinfo{journal}{R J.} \bibinfo{volume}{9}, \bibinfo{pages}{421}.
%Type = Article
\bibitem[{de~Groot et~al.(2013)de~Groot, Cantin, Flannigan, Soja, Gowman and Newbery}]{de2013comparison}
\bibinfo{author}{de~Groot, W.J.}, \bibinfo{author}{Cantin, A.S.}, \bibinfo{author}{Flannigan, M.D.}, \bibinfo{author}{Soja, A.J.}, \bibinfo{author}{Gowman, L.M.}, \bibinfo{author}{Newbery, A.}, \bibinfo{year}{2013}.
\newblock \bibinfo{title}{A comparison of canadian and russian boreal forest fire regimes}.
\newblock \bibinfo{journal}{Forest Ecology and Management} \bibinfo{volume}{294}, \bibinfo{pages}{23--34}.
%Type = Book
\bibitem[{Group et~al.(1992)Group, Science and Directorate}]{canada1992development}
\bibinfo{author}{Group, C.F.C.F.D.}, \bibinfo{author}{Science, C.F.C.}, \bibinfo{author}{Directorate, S.D.}, \bibinfo{year}{1992}.
\newblock \bibinfo{title}{Development and structure of the Canadian forest fire behavior prediction system}. volume~\bibinfo{volume}{3}.
\newblock \bibinfo{publisher}{Forestry Canada, Science and Sustainable Development Directorate}.
%Type = Misc
\bibitem[{Gu and Dao(2023)}]{gu_mamba_2023}
\bibinfo{author}{Gu, A.}, \bibinfo{author}{Dao, T.}, \bibinfo{year}{2023}.
\newblock \bibinfo{title}{Mamba: {Linear}-{Time} {Sequence} {Modeling} with {Selective} {State} {Spaces}}.
\newblock \URLprefix \url{http://arxiv.org/abs/2312.00752}, \DOIprefix\doi{10.48550/arXiv.2312.00752}. \bibinfo{note}{arXiv:2312.00752 [cs]}.
%Type = Article
\bibitem[{Gu et~al.(2021)Gu, Goel and R{\'e}}]{gu2021efficiently}
\bibinfo{author}{Gu, A.}, \bibinfo{author}{Goel, K.}, \bibinfo{author}{R{\'e}, C.}, \bibinfo{year}{2021}.
\newblock \bibinfo{title}{Efficiently modeling long sequences with structured state spaces}.
\newblock \bibinfo{journal}{arXiv preprint arXiv:2111.00396} .
%Type = Article
\bibitem[{Guerra-Hern{\'a}ndez et~al.(2024)Guerra-Hern{\'a}ndez, Pereira, Stovall and Pascual}]{guerra2024impact}
\bibinfo{author}{Guerra-Hern{\'a}ndez, J.}, \bibinfo{author}{Pereira, J.M.}, \bibinfo{author}{Stovall, A.}, \bibinfo{author}{Pascual, A.}, \bibinfo{year}{2024}.
\newblock \bibinfo{title}{Impact of fire severity on forest structure and biomass stocks using nasa gedi data. insights from the 2020 and 2021 wildfire season in spain and portugal}.
\newblock \bibinfo{journal}{Science of Remote Sensing} \bibinfo{volume}{9}, \bibinfo{pages}{100134}.
%Type = Article
\bibitem[{Guk et~al.(2024)Guk, Bar-Massada, Yebra, Scortechini and Levin}]{10638113}
\bibinfo{author}{Guk, E.}, \bibinfo{author}{Bar-Massada, A.}, \bibinfo{author}{Yebra, M.}, \bibinfo{author}{Scortechini, G.}, \bibinfo{author}{Levin, N.}, \bibinfo{year}{2024}.
\newblock \bibinfo{title}{Examining the transferability of remote-sensing-based models of live fuel moisture content for predicting wildfire characteristics}.
\newblock \bibinfo{journal}{IEEE Journal of Selected Topics in Applied Earth Observations and Remote Sensing} \bibinfo{volume}{17}, \bibinfo{pages}{14762--14776}.
\newblock \DOIprefix\doi{10.1109/JSTARS.2024.3445138}.
%Type = Article
\bibitem[{Guo et~al.(2016)Guo, Wang, Su, Liang, Wang, Lin and Liu}]{guo_what_2016}
\bibinfo{author}{Guo, F.}, \bibinfo{author}{Wang, G.}, \bibinfo{author}{Su, Z.}, \bibinfo{author}{Liang, H.}, \bibinfo{author}{Wang, W.}, \bibinfo{author}{Lin, F.}, \bibinfo{author}{Liu, A.}, \bibinfo{year}{2016}.
\newblock \bibinfo{title}{What drives forest fire in {Fujian}, {China}? {Evidence} from logistic regression and {Random} {Forests}}.
\newblock \bibinfo{journal}{International Journal of Wildland Fire} \bibinfo{volume}{25}, \bibinfo{pages}{505--519}.
\newblock \URLprefix \url{https://www.publish.csiro.au/wf/WF15121}, \DOIprefix\doi{10.1071/WF15121}. \bibinfo{note}{publisher: CSIRO PUBLISHING}.
%Type = Book
\bibitem[{Hair~Jr et~al.(1986)Hair~Jr, Anderson and Tatham}]{hair1986multivariate}
\bibinfo{author}{Hair~Jr, J.F.}, \bibinfo{author}{Anderson, R.E.}, \bibinfo{author}{Tatham, R.L.}, \bibinfo{year}{1986}.
\newblock \bibinfo{title}{Multivariate data analysis with readings}.
\newblock \bibinfo{publisher}{Macmillan Publishing Co., Inc.}
%Type = Article
\bibitem[{Hall et~al.(2024)Hall, Argueta, Zubkova, Chen, Randerson and Giglio}]{hall2024glocab}
\bibinfo{author}{Hall, J.V.}, \bibinfo{author}{Argueta, F.}, \bibinfo{author}{Zubkova, M.}, \bibinfo{author}{Chen, Y.}, \bibinfo{author}{Randerson, J.T.}, \bibinfo{author}{Giglio, L.}, \bibinfo{year}{2024}.
\newblock \bibinfo{title}{Glocab: global cropland burned area from mid-2002 to 2020}.
\newblock \bibinfo{journal}{Earth System Science Data} \bibinfo{volume}{16}, \bibinfo{pages}{867--885}.
%Type = Article
\bibitem[{Hall and Burke(2006)}]{hall2006considerations}
\bibinfo{author}{Hall, S.A.}, \bibinfo{author}{Burke, I.C.}, \bibinfo{year}{2006}.
\newblock \bibinfo{title}{Considerations for characterizing fuels as inputs for fire behavior models}.
\newblock \bibinfo{journal}{Forest Ecology and Management} \bibinfo{volume}{227}, \bibinfo{pages}{102--114}.
%Type = Article
\bibitem[{Hanes et~al.(2021)Hanes, Wang and de~Groot}]{hanes2021dead}
\bibinfo{author}{Hanes, C.C.}, \bibinfo{author}{Wang, X.}, \bibinfo{author}{de~Groot, W.J.}, \bibinfo{year}{2021}.
\newblock \bibinfo{title}{Dead and down woody debris fuel loads in canadian forests}.
\newblock \bibinfo{journal}{International journal of wildland fire} \bibinfo{volume}{30}, \bibinfo{pages}{871--885}.
%Type = Article
\bibitem[{Hanes et~al.(2019)Hanes, Wang, Jain, Parisien, Little and Flannigan}]{hanes2019fire}
\bibinfo{author}{Hanes, C.C.}, \bibinfo{author}{Wang, X.}, \bibinfo{author}{Jain, P.}, \bibinfo{author}{Parisien, M.A.}, \bibinfo{author}{Little, J.M.}, \bibinfo{author}{Flannigan, M.D.}, \bibinfo{year}{2019}.
\newblock \bibinfo{title}{Fire-regime changes in canada over the last half century}.
\newblock \bibinfo{journal}{Canadian Journal of Forest Research} \bibinfo{volume}{49}, \bibinfo{pages}{256--269}.
%Type = Article
\bibitem[{Hantson et~al.(2013)Hantson, Padilla, Corti and Chuvieco}]{hantson_strengths_2013}
\bibinfo{author}{Hantson, S.}, \bibinfo{author}{Padilla, M.}, \bibinfo{author}{Corti, D.}, \bibinfo{author}{Chuvieco, E.}, \bibinfo{year}{2013}.
\newblock \bibinfo{title}{Strengths and weaknesses of {MODIS} hotspots to characterize global fire occurrence}.
\newblock \bibinfo{journal}{Remote Sensing of Environment} \bibinfo{volume}{131}, \bibinfo{pages}{152--159}.
\newblock \URLprefix \url{https://www.sciencedirect.com/science/article/pii/S0034425712004610}, \DOIprefix\doi{10.1016/j.rse.2012.12.004}.
%Type = Article
\bibitem[{Hao and Qu(2007)}]{hao2007retrieval}
\bibinfo{author}{Hao, X.}, \bibinfo{author}{Qu, J.J.}, \bibinfo{year}{2007}.
\newblock \bibinfo{title}{Retrieval of real-time live fuel moisture content using modis measurements}.
\newblock \bibinfo{journal}{Remote Sensing of Environment} \bibinfo{volume}{108}, \bibinfo{pages}{130--137}.
%Type = Article
\bibitem[{Haque et~al.(2021)Haque, Azad, Hossain, Ahmed, Uddin and Hossain}]{haque2021wildfire}
\bibinfo{author}{Haque, M.K.}, \bibinfo{author}{Azad, M.A.K.}, \bibinfo{author}{Hossain, M.Y.}, \bibinfo{author}{Ahmed, T.}, \bibinfo{author}{Uddin, M.}, \bibinfo{author}{Hossain, M.M.}, \bibinfo{year}{2021}.
\newblock \bibinfo{title}{Wildfire in australia during 2019-2020, its impact on health, biodiversity and environment with some proposals for risk management: a review}.
\newblock \bibinfo{journal}{Journal of Environmental Protection} \bibinfo{volume}{12}, \bibinfo{pages}{391--414}.
%Type = Article
\bibitem[{Hardy(2005)}]{hardy2005wildland}
\bibinfo{author}{Hardy, C.C.}, \bibinfo{year}{2005}.
\newblock \bibinfo{title}{Wildland fire hazard and risk: Problems, definitions, and context}.
\newblock \bibinfo{journal}{Forest ecology and management} \bibinfo{volume}{211}, \bibinfo{pages}{73--82}.
%Type = Misc
\bibitem[{Hargrove et~al.(2022)Hargrove, Kumar, Norman and Hoffman}]{william_w_hargrove_2022_6522071}
\bibinfo{author}{Hargrove, W.W.}, \bibinfo{author}{Kumar, J.}, \bibinfo{author}{Norman, S.P.}, \bibinfo{author}{Hoffman, F.M.}, \bibinfo{year}{2022}.
\newblock \bibinfo{title}{Statistically determined global fire regimes (gfrs) empirically characterized using historical modis hotspots}.
\newblock \URLprefix \url{https://doi.org/10.5281/zenodo.6522071}, \DOIprefix\doi{10.5281/zenodo.6522071}.
%Type = Article
\bibitem[{Harris et~al.(2023)Harris, Taylor, Kassa, Leta and Powell}]{harris2023humans}
\bibinfo{author}{Harris, L.B.}, \bibinfo{author}{Taylor, A.H.}, \bibinfo{author}{Kassa, H.}, \bibinfo{author}{Leta, S.}, \bibinfo{author}{Powell, B.}, \bibinfo{year}{2023}.
\newblock \bibinfo{title}{Humans and climate modulate fire activity across ethiopia}.
\newblock \bibinfo{journal}{Fire Ecology} \bibinfo{volume}{19}, \bibinfo{pages}{15}.
%Type = Article
\bibitem[{Harrison et~al.(2021)Harrison, Prentice, Bloomfield, Dong, Forkel, Forrest, Ningthoujam, Pellegrini, Shen, Baudena et~al.}]{harrison2021understanding}
\bibinfo{author}{Harrison, S.P.}, \bibinfo{author}{Prentice, I.C.}, \bibinfo{author}{Bloomfield, K.J.}, \bibinfo{author}{Dong, N.}, \bibinfo{author}{Forkel, M.}, \bibinfo{author}{Forrest, M.}, \bibinfo{author}{Ningthoujam, R.K.}, \bibinfo{author}{Pellegrini, A.}, \bibinfo{author}{Shen, Y.}, \bibinfo{author}{Baudena, M.}, et~al., \bibinfo{year}{2021}.
\newblock \bibinfo{title}{Understanding and modelling wildfire regimes: an ecological perspective}.
\newblock \bibinfo{journal}{Environmental Research Letters} \bibinfo{volume}{16}, \bibinfo{pages}{125008}.
%Type = Article
\bibitem[{He et~al.(2024)He, Fan, Li, Li, Gao, Li and Zeng}]{he2024deep}
\bibinfo{author}{He, Z.}, \bibinfo{author}{Fan, G.}, \bibinfo{author}{Li, Z.}, \bibinfo{author}{Li, S.}, \bibinfo{author}{Gao, L.}, \bibinfo{author}{Li, X.}, \bibinfo{author}{Zeng, Z.C.}, \bibinfo{year}{2024}.
\newblock \bibinfo{title}{Deep learning modeling of human activity affected wildfire risk by incorporating structural features: A case study in eastern china}.
\newblock \bibinfo{journal}{Ecological Indicators} \bibinfo{volume}{160}, \bibinfo{pages}{111946}.
%Type = Article
\bibitem[{Helman et~al.(2015)Helman, Lensky, Tessler and Osem}]{helman_phenology-based_2015}
\bibinfo{author}{Helman, D.}, \bibinfo{author}{Lensky, I.M.}, \bibinfo{author}{Tessler, N.}, \bibinfo{author}{Osem, Y.}, \bibinfo{year}{2015}.
\newblock \bibinfo{title}{A {Phenology}-{Based} {Method} for {Monitoring} {Woody} and {Herbaceous} {Vegetation} in {Mediterranean} {Forests} from {NDVI} {Time} {Series}}.
\newblock \bibinfo{journal}{Remote Sensing} \bibinfo{volume}{7}, \bibinfo{pages}{12314--12335}.
\newblock \URLprefix \url{https://www.mdpi.com/2072-4292/7/9/12314}, \DOIprefix\doi{10.3390/rs70912314}. \bibinfo{note}{number: 9 Publisher: Multidisciplinary Digital Publishing Institute}.
%Type = Article
\bibitem[{Hermosilla et~al.(2014)Hermosilla, Ruiz, Kazakova, Coops and Moskal}]{Hermosilla2014}
\bibinfo{author}{Hermosilla, T.}, \bibinfo{author}{Ruiz, L.A.}, \bibinfo{author}{Kazakova, A.N.}, \bibinfo{author}{Coops, N.C.}, \bibinfo{author}{Moskal, L.M.}, \bibinfo{year}{2014}.
\newblock \bibinfo{title}{Estimation of forest structure and canopy fuel parameters from small-footprint full-waveform lidar data}.
\newblock \bibinfo{journal}{International Journal of Wildland Fire} \bibinfo{volume}{23}, \bibinfo{pages}{224--233}.
%Type = Article
\bibitem[{Hertelendy et~al.(2024)Hertelendy, Howard, Sorensen, Ranse, Eboreime, Henderson, Tochkin and Ciottone}]{hertelendy2024seasons}
\bibinfo{author}{Hertelendy, A.J.}, \bibinfo{author}{Howard, C.}, \bibinfo{author}{Sorensen, C.}, \bibinfo{author}{Ranse, J.}, \bibinfo{author}{Eboreime, E.}, \bibinfo{author}{Henderson, S.}, \bibinfo{author}{Tochkin, J.}, \bibinfo{author}{Ciottone, G.}, \bibinfo{year}{2024}.
\newblock \bibinfo{title}{Seasons of smoke and fire: preparing health systems for improved performance before, during, and after wildfires}.
\newblock \bibinfo{journal}{The Lancet Planetary Health} \bibinfo{volume}{8}, \bibinfo{pages}{e588--e602}.
%Type = Article
\bibitem[{Hillman et~al.(2021)Hillman, Wallace, Reinke and Jones}]{hillman2021comparison}
\bibinfo{author}{Hillman, S.}, \bibinfo{author}{Wallace, L.}, \bibinfo{author}{Reinke, K.}, \bibinfo{author}{Jones, S.}, \bibinfo{year}{2021}.
\newblock \bibinfo{title}{A comparison between tls and uas lidar to represent eucalypt crown fuel characteristics}.
\newblock \bibinfo{journal}{ISPRS Journal of Photogrammetry and Remote Sensing} \bibinfo{volume}{181}, \bibinfo{pages}{295--307}.
%Type = Article
\bibitem[{Hochreiter(1997)}]{hochreiter1997long}
\bibinfo{author}{Hochreiter, S.}, \bibinfo{year}{1997}.
\newblock \bibinfo{title}{Long short-term memory}.
\newblock \bibinfo{journal}{Neural Computation MIT-Press} .
%Type = Article
\bibitem[{Hodges and Lattimer(2019)}]{hodges2019wildland}
\bibinfo{author}{Hodges, J.L.}, \bibinfo{author}{Lattimer, B.Y.}, \bibinfo{year}{2019}.
\newblock \bibinfo{title}{Wildland fire spread modeling using convolutional neural networks}.
\newblock \bibinfo{journal}{Fire technology} \bibinfo{volume}{55}, \bibinfo{pages}{2115--2142}.
%Type = Article
\bibitem[{Hoe et~al.(2018)Hoe, Dunn and Temesgen}]{hoe2018multitemporal}
\bibinfo{author}{Hoe, M.S.}, \bibinfo{author}{Dunn, C.J.}, \bibinfo{author}{Temesgen, H.}, \bibinfo{year}{2018}.
\newblock \bibinfo{title}{Multitemporal lidar improves estimates of fire severity in forested landscapes}.
\newblock \bibinfo{journal}{International Journal of Wildland Fire} \bibinfo{volume}{27}, \bibinfo{pages}{581--594}.
%Type = Article
\bibitem[{Holder et~al.(2023)Holder, Ahmed, Vukovich and Rao}]{holder2023hazardous}
\bibinfo{author}{Holder, A.L.}, \bibinfo{author}{Ahmed, A.}, \bibinfo{author}{Vukovich, J.M.}, \bibinfo{author}{Rao, V.}, \bibinfo{year}{2023}.
\newblock \bibinfo{title}{Hazardous air pollutant emissions estimates from wildfires in the wildland urban interface}.
\newblock \bibinfo{journal}{PNAS nexus} \bibinfo{volume}{2}, \bibinfo{pages}{pgad186}.
%Type = Article
\bibitem[{Holdrege et~al.(2024)Holdrege, Schlaepfer, Palmquist, Crist, Doherty, Lauenroth, Remington, Riley, Short, Tull, Wiechman and Bradford}]{holdrege_wildfire_2024}
\bibinfo{author}{Holdrege, M.C.}, \bibinfo{author}{Schlaepfer, D.R.}, \bibinfo{author}{Palmquist, K.A.}, \bibinfo{author}{Crist, M.}, \bibinfo{author}{Doherty, K.E.}, \bibinfo{author}{Lauenroth, W.K.}, \bibinfo{author}{Remington, T.E.}, \bibinfo{author}{Riley, K.}, \bibinfo{author}{Short, K.C.}, \bibinfo{author}{Tull, J.C.}, \bibinfo{author}{Wiechman, L.A.}, \bibinfo{author}{Bradford, J.B.}, \bibinfo{year}{2024}.
\newblock \bibinfo{title}{Wildfire probability estimated from recent climate and fine fuels across the big sagebrush region}.
\newblock \bibinfo{journal}{Fire Ecology} \bibinfo{volume}{20}, \bibinfo{pages}{22}.
\newblock \URLprefix \url{https://doi.org/10.1186/s42408-024-00252-4}, \DOIprefix\doi{10.1186/s42408-024-00252-4}.
%Type = Article
\bibitem[{Hollis et~al.(2024)Hollis, Matthews, Fox-Hughes, Grootemaat, Heemstra, Kenny and Sauvage}]{hollis2024introduction}
\bibinfo{author}{Hollis, J.J.}, \bibinfo{author}{Matthews, S.}, \bibinfo{author}{Fox-Hughes, P.}, \bibinfo{author}{Grootemaat, S.}, \bibinfo{author}{Heemstra, S.}, \bibinfo{author}{Kenny, B.J.}, \bibinfo{author}{Sauvage, S.}, \bibinfo{year}{2024}.
\newblock \bibinfo{title}{Introduction to the australian fire danger rating system}.
\newblock \bibinfo{journal}{International Journal of Wildland Fire} \bibinfo{volume}{33}.
%Type = Article
\bibitem[{Hong et~al.(2019)Hong, Jaafari and Zenner}]{hong_predicting_2019}
\bibinfo{author}{Hong, H.}, \bibinfo{author}{Jaafari, A.}, \bibinfo{author}{Zenner, E.K.}, \bibinfo{year}{2019}.
\newblock \bibinfo{title}{Predicting spatial patterns of wildfire susceptibility in the {Huichang} {County}, {China}: {An} integrated model to analysis of landscape indicators}.
\newblock \bibinfo{journal}{Ecological Indicators} \bibinfo{volume}{101}, \bibinfo{pages}{878--891}.
\newblock \URLprefix \url{https://www.sciencedirect.com/science/article/pii/S1470160X19300792}, \DOIprefix\doi{10.1016/j.ecolind.2019.01.056}.
%Type = Article
\bibitem[{Horton et~al.(2024)Horton, Johnson, Baris, Jagdhuber, Bindlish, Park and Al-Khaldi}]{horton2024wildfire}
\bibinfo{author}{Horton, D.}, \bibinfo{author}{Johnson, J.T.}, \bibinfo{author}{Baris, I.}, \bibinfo{author}{Jagdhuber, T.}, \bibinfo{author}{Bindlish, R.}, \bibinfo{author}{Park, J.}, \bibinfo{author}{Al-Khaldi, M.M.}, \bibinfo{year}{2024}.
\newblock \bibinfo{title}{Wildfire threshold detection and progression monitoring using an improved radar vegetation index in california}.
\newblock \bibinfo{journal}{Remote Sensing} \bibinfo{volume}{16}, \bibinfo{pages}{3050}.
%Type = Article
\bibitem[{Hu et~al.(2023a)Hu, Tanchak and Wang}]{hu_developing_2023}
\bibinfo{author}{Hu, P.}, \bibinfo{author}{Tanchak, R.}, \bibinfo{author}{Wang, Q.}, \bibinfo{year}{2023}a.
\newblock \bibinfo{title}{Developing {Risk} {Assessment} {Framework} for {Wildfire} in the {United} {States} – {A} {Deep} {Learning} {Approach} to {Safety} and {Sustainability}}.
\newblock \bibinfo{journal}{Journal of Safety and Sustainability} \URLprefix \url{https://www.sciencedirect.com/science/article/pii/S2949926723000033}, \DOIprefix\doi{10.1016/j.jsasus.2023.09.002}.
%Type = Article
\bibitem[{Hu et~al.(2020)Hu, Li, Pan, Li, Tao and Du}]{8960629}
\bibinfo{author}{Hu, W.S.}, \bibinfo{author}{Li, H.C.}, \bibinfo{author}{Pan, L.}, \bibinfo{author}{Li, W.}, \bibinfo{author}{Tao, R.}, \bibinfo{author}{Du, Q.}, \bibinfo{year}{2020}.
\newblock \bibinfo{title}{Spatial–spectral feature extraction via deep convlstm neural networks for hyperspectral image classification}.
\newblock \bibinfo{journal}{IEEE Transactions on Geoscience and Remote Sensing} \bibinfo{volume}{58}, \bibinfo{pages}{4237--4250}.
\newblock \DOIprefix\doi{10.1109/TGRS.2019.2961947}.
%Type = Article
\bibitem[{Hu et~al.(2024)Hu, Wen, Zhang, Yuen and Zhong}]{10553270}
\bibinfo{author}{Hu, X.}, \bibinfo{author}{Wen, H.}, \bibinfo{author}{Zhang, P.}, \bibinfo{author}{Yuen, K.V.}, \bibinfo{author}{Zhong, P.}, \bibinfo{year}{2024}.
\newblock \bibinfo{title}{Near real-time burned area progression mapping with multispectral data using ensemble learning}.
\newblock \bibinfo{journal}{IEEE Geoscience and Remote Sensing Letters} \bibinfo{volume}{21}, \bibinfo{pages}{1--5}.
\newblock \DOIprefix\doi{10.1109/LGRS.2024.3412173}.
%Type = Article
\bibitem[{Hu et~al.(2023b)Hu, Zhang, Ban and Rahnemoonfar}]{hu2023gan}
\bibinfo{author}{Hu, X.}, \bibinfo{author}{Zhang, P.}, \bibinfo{author}{Ban, Y.}, \bibinfo{author}{Rahnemoonfar, M.}, \bibinfo{year}{2023}b.
\newblock \bibinfo{title}{Gan-based sar and optical image translation for wildfire impact assessment using multi-source remote sensing data}.
\newblock \bibinfo{journal}{Remote Sensing of Environment} \bibinfo{volume}{289}, \bibinfo{pages}{113522}.
%Type = Article
\bibitem[{Huang et~al.(2024)Huang, Lor{\'\i}a-Salazar, Deng, Lee and Holmes}]{huang2024assessment}
\bibinfo{author}{Huang, J.}, \bibinfo{author}{Lor{\'\i}a-Salazar, S.M.}, \bibinfo{author}{Deng, M.}, \bibinfo{author}{Lee, J.}, \bibinfo{author}{Holmes, H.A.}, \bibinfo{year}{2024}.
\newblock \bibinfo{title}{Assessment of smoke plume height products derived from multisource satellite observations using lidar-derived height metrics for wildfires in the western us}.
\newblock \bibinfo{journal}{Atmospheric Chemistry and Physics} \bibinfo{volume}{24}, \bibinfo{pages}{3673--3698}.
%Type = Article
\bibitem[{Huidobro et~al.(2024)Huidobro, Giessen and Burns}]{huidobro2024and}
\bibinfo{author}{Huidobro, G.}, \bibinfo{author}{Giessen, L.}, \bibinfo{author}{Burns, S.L.}, \bibinfo{year}{2024}.
\newblock \bibinfo{title}{And it burns, burns, burns, the ring-of-fire: Reviewing and harmonizing terminology on wildfire management and policy}.
\newblock \bibinfo{journal}{Environmental Science \& Policy} \bibinfo{volume}{157}, \bibinfo{pages}{103776}.
%Type = Article
\bibitem[{Hunt~Jr et~al.(1987)Hunt~Jr, Rock and Nobel}]{hunt1987measurement}
\bibinfo{author}{Hunt~Jr, E.R.}, \bibinfo{author}{Rock, B.N.}, \bibinfo{author}{Nobel, P.S.}, \bibinfo{year}{1987}.
\newblock \bibinfo{title}{Measurement of leaf relative water content by infrared reflectance}.
\newblock \bibinfo{journal}{Remote sensing of environment} \bibinfo{volume}{22}, \bibinfo{pages}{429--435}.
%Type = Article
\bibitem[{Huot et~al.(2022)Huot, Hu, Goyal, Sankar, Ihme and Chen}]{huot2022next}
\bibinfo{author}{Huot, F.}, \bibinfo{author}{Hu, R.L.}, \bibinfo{author}{Goyal, N.}, \bibinfo{author}{Sankar, T.}, \bibinfo{author}{Ihme, M.}, \bibinfo{author}{Chen, Y.F.}, \bibinfo{year}{2022}.
\newblock \bibinfo{title}{Next day wildfire spread: A machine learning dataset to predict wildfire spreading from remote-sensing data}.
\newblock \bibinfo{journal}{IEEE Transactions on Geoscience and Remote Sensing} \bibinfo{volume}{60}, \bibinfo{pages}{1--13}.
%Type = Misc
\bibitem[{Huot et~al.(2021)Huot, Hu, Ihme, Wang, Burge, Lu, Hickey, Chen and Anderson}]{huot_deep_2021}
\bibinfo{author}{Huot, F.}, \bibinfo{author}{Hu, R.L.}, \bibinfo{author}{Ihme, M.}, \bibinfo{author}{Wang, Q.}, \bibinfo{author}{Burge, J.}, \bibinfo{author}{Lu, T.}, \bibinfo{author}{Hickey, J.}, \bibinfo{author}{Chen, Y.F.}, \bibinfo{author}{Anderson, J.}, \bibinfo{year}{2021}.
\newblock \bibinfo{title}{Deep {Learning} {Models} for {Predicting} {Wildfires} from {Historical} {Remote}-{Sensing} {Data}}.
\newblock \URLprefix \url{http://arxiv.org/abs/2010.07445}. \bibinfo{note}{arXiv:2010.07445 [cs]}.
%Type = Article
\bibitem[{Iban and Aksu(2024)}]{iban2024shap}
\bibinfo{author}{Iban, M.C.}, \bibinfo{author}{Aksu, O.}, \bibinfo{year}{2024}.
\newblock \bibinfo{title}{Shap-driven explainable artificial intelligence framework for wildfire susceptibility mapping using modis active fire pixels: An in-depth interpretation of contributing factors in izmir, t{\"u}rkiye}.
\newblock \bibinfo{journal}{Remote Sensing} \bibinfo{volume}{16}, \bibinfo{pages}{2842}.
%Type = Article
\bibitem[{Iban and Sekertekin(2022)}]{iban2022machine}
\bibinfo{author}{Iban, M.C.}, \bibinfo{author}{Sekertekin, A.}, \bibinfo{year}{2022}.
\newblock \bibinfo{title}{Machine learning based wildfire susceptibility mapping using remotely sensed fire data and gis: A case study of adana and mersin provinces, turkey}.
\newblock \bibinfo{journal}{Ecological Informatics} \bibinfo{volume}{69}, \bibinfo{pages}{101647}.
%Type = Article
\bibitem[{Jaafari et~al.(2017)Jaafari, Gholami and Zenner}]{jaafari_bayesian_2017}
\bibinfo{author}{Jaafari, A.}, \bibinfo{author}{Gholami, D.M.}, \bibinfo{author}{Zenner, E.K.}, \bibinfo{year}{2017}.
\newblock \bibinfo{title}{A {Bayesian} modeling of wildfire probability in the {Zagros} {Mountains}, {Iran}}.
\newblock \bibinfo{journal}{Ecological Informatics} \bibinfo{volume}{39}, \bibinfo{pages}{32--44}.
\newblock \URLprefix \url{https://linkinghub.elsevier.com/retrieve/pii/S1574954117300912}, \DOIprefix\doi{10.1016/j.ecoinf.2017.03.003}.
%Type = Article
\bibitem[{Jaafari et~al.(2019)Jaafari, Mafi-Gholami, Thai~Pham and Tien~Bui}]{jaafari_wildfire_2019}
\bibinfo{author}{Jaafari, A.}, \bibinfo{author}{Mafi-Gholami, D.}, \bibinfo{author}{Thai~Pham, B.}, \bibinfo{author}{Tien~Bui, D.}, \bibinfo{year}{2019}.
\newblock \bibinfo{title}{Wildfire {Probability} {Mapping}: {Bivariate} vs. {Multivariate} {Statistics}}.
\newblock \bibinfo{journal}{Remote Sensing} \bibinfo{volume}{11}, \bibinfo{pages}{618}.
\newblock \URLprefix \url{https://www.mdpi.com/2072-4292/11/6/618}, \DOIprefix\doi{10.3390/rs11060618}. \bibinfo{note}{number: 6 Publisher: Multidisciplinary Digital Publishing Institute}.
%Type = Inproceedings
\bibitem[{Jain and Gandhi(2022)}]{jain-gandhi-2022-comprehensive}
\bibinfo{author}{Jain, K.}, \bibinfo{author}{Gandhi, V.}, \bibinfo{year}{2022}.
\newblock \bibinfo{title}{Comprehensive multi-modal interactions for referring image segmentation}, in: \bibinfo{editor}{Muresan, S.}, \bibinfo{editor}{Nakov, P.}, \bibinfo{editor}{Villavicencio, A.} (Eds.), \bibinfo{booktitle}{Findings of the Association for Computational Linguistics: ACL 2022}, \bibinfo{publisher}{Association for Computational Linguistics}, \bibinfo{address}{Dublin, Ireland}. pp. \bibinfo{pages}{3427--3435}.
\newblock \URLprefix \url{https://aclanthology.org/2022.findings-acl.270}, \DOIprefix\doi{10.18653/v1/2022.findings-acl.270}.
%Type = Article
\bibitem[{Jain et~al.(2024)Jain, Barber, Taylor, Whitman, Castellanos~Acuna, Boulanger, Chavard{\`e}s, Chen, Englefield, Flannigan et~al.}]{jain2024drivers}
\bibinfo{author}{Jain, P.}, \bibinfo{author}{Barber, Q.E.}, \bibinfo{author}{Taylor, S.W.}, \bibinfo{author}{Whitman, E.}, \bibinfo{author}{Castellanos~Acuna, D.}, \bibinfo{author}{Boulanger, Y.}, \bibinfo{author}{Chavard{\`e}s, R.D.}, \bibinfo{author}{Chen, J.}, \bibinfo{author}{Englefield, P.}, \bibinfo{author}{Flannigan, M.}, et~al., \bibinfo{year}{2024}.
\newblock \bibinfo{title}{Drivers and impacts of the record-breaking 2023 wildfire season in canada}.
\newblock \bibinfo{journal}{Nature Communications} \bibinfo{volume}{15}, \bibinfo{pages}{6764}.
%Type = Article
\bibitem[{Jain et~al.(2020)Jain, Coogan, Subramanian, Crowley, Taylor and Flannigan}]{jain_review_2020}
\bibinfo{author}{Jain, P.}, \bibinfo{author}{Coogan, S.C.}, \bibinfo{author}{Subramanian, S.G.}, \bibinfo{author}{Crowley, M.}, \bibinfo{author}{Taylor, S.}, \bibinfo{author}{Flannigan, M.D.}, \bibinfo{year}{2020}.
\newblock \bibinfo{title}{A review of machine learning applications in wildfire science and management}.
\newblock \bibinfo{journal}{Environmental Reviews} \bibinfo{volume}{28}, \bibinfo{pages}{478--505}.
\newblock \URLprefix \url{https://cdnsciencepub.com/doi/10.1139/er-2020-0019}, \DOIprefix\doi{10.1139/er-2020-0019}. \bibinfo{note}{publisher: NRC Research Press}.
%Type = Article
\bibitem[{Jalilian and Jouibary(2023)}]{jalilian2023forest}
\bibinfo{author}{Jalilian, S.}, \bibinfo{author}{Jouibary, S.S.}, \bibinfo{year}{2023}.
\newblock \bibinfo{title}{Forest wildfire risk mapping, performance comparison of machine learning algorithms}.
\newblock \bibinfo{journal}{Research Square} .
%Type = Article
\bibitem[{Jenks and Caspall(1971)}]{jenks1971error}
\bibinfo{author}{Jenks, G.F.}, \bibinfo{author}{Caspall, F.C.}, \bibinfo{year}{1971}.
\newblock \bibinfo{title}{Error on choroplethic maps: definition, measurement, reduction}.
\newblock \bibinfo{journal}{Annals of the Association of American Geographers} \bibinfo{volume}{61}, \bibinfo{pages}{217--244}.
%Type = Article
\bibitem[{Ji et~al.(2024a)Ji, Du, Dang, Gao and Zhang}]{ji2024survey}
\bibinfo{author}{Ji, L.}, \bibinfo{author}{Du, Y.}, \bibinfo{author}{Dang, Y.}, \bibinfo{author}{Gao, W.}, \bibinfo{author}{Zhang, H.}, \bibinfo{year}{2024}a.
\newblock \bibinfo{title}{A survey of methods for addressing the challenges of referring image segmentation}.
\newblock \bibinfo{journal}{Neurocomputing} \bibinfo{volume}{583}, \bibinfo{pages}{127599}.
%Type = Article
\bibitem[{Ji et~al.(2024b)Ji, Wang, Li, Liu and Bai}]{ji2024global}
\bibinfo{author}{Ji, Y.}, \bibinfo{author}{Wang, D.}, \bibinfo{author}{Li, Q.}, \bibinfo{author}{Liu, T.}, \bibinfo{author}{Bai, Y.}, \bibinfo{year}{2024}b.
\newblock \bibinfo{title}{Global wildfire danger predictions based on deep learning taking into account static and dynamic variables}.
\newblock \bibinfo{journal}{Forests} \bibinfo{volume}{15}, \bibinfo{pages}{216}.
%Type = Article
\bibitem[{Jia et~al.(2019)Jia, Kim, Nghiem and Kafatos}]{jia2019estimating}
\bibinfo{author}{Jia, S.}, \bibinfo{author}{Kim, S.H.}, \bibinfo{author}{Nghiem, S.V.}, \bibinfo{author}{Kafatos, M.}, \bibinfo{year}{2019}.
\newblock \bibinfo{title}{Estimating live fuel moisture using smap l-band radiometer soil moisture for southern california, usa}.
\newblock \bibinfo{journal}{Remote Sensing} \bibinfo{volume}{11}, \bibinfo{pages}{1575}.
%Type = Article
\bibitem[{Jiang et~al.(2024a)Jiang, Sweet, Blougouras, Brenning, Li, Reichstein, Denzler, Shangguan, Yu, Huang et~al.}]{jiang2024interpretable}
\bibinfo{author}{Jiang, S.}, \bibinfo{author}{Sweet, L.b.}, \bibinfo{author}{Blougouras, G.}, \bibinfo{author}{Brenning, A.}, \bibinfo{author}{Li, W.}, \bibinfo{author}{Reichstein, M.}, \bibinfo{author}{Denzler, J.}, \bibinfo{author}{Shangguan, W.}, \bibinfo{author}{Yu, G.}, \bibinfo{author}{Huang, F.}, et~al., \bibinfo{year}{2024}a.
\newblock \bibinfo{title}{How interpretable machine learning can benefit process understanding in the geosciences}.
\newblock \bibinfo{journal}{Earth's Future} \bibinfo{volume}{12}, \bibinfo{pages}{e2024EF004540}.
%Type = Article
\bibitem[{Jiang et~al.(2023)Jiang, Qiao, Su, Li, Meng, Wu, Quan, Wang and Wang}]{jiang2023wfnet}
\bibinfo{author}{Jiang, W.}, \bibinfo{author}{Qiao, Y.}, \bibinfo{author}{Su, G.}, \bibinfo{author}{Li, X.}, \bibinfo{author}{Meng, Q.}, \bibinfo{author}{Wu, H.}, \bibinfo{author}{Quan, W.}, \bibinfo{author}{Wang, J.}, \bibinfo{author}{Wang, F.}, \bibinfo{year}{2023}.
\newblock \bibinfo{title}{Wfnet: A hierarchical convolutional neural network for wildfire spread prediction}.
\newblock \bibinfo{journal}{Environmental Modelling \& Software} \bibinfo{volume}{170}, \bibinfo{pages}{105841}.
%Type = Article
\bibitem[{Jiang et~al.(2024b)Jiang, Qiao, Zheng, Zhou, Jiang, Meng, Su, Zhong and Wang}]{jiang_wildfire_2024}
\bibinfo{author}{Jiang, W.}, \bibinfo{author}{Qiao, Y.}, \bibinfo{author}{Zheng, X.}, \bibinfo{author}{Zhou, J.}, \bibinfo{author}{Jiang, J.}, \bibinfo{author}{Meng, Q.}, \bibinfo{author}{Su, G.}, \bibinfo{author}{Zhong, S.}, \bibinfo{author}{Wang, F.}, \bibinfo{year}{2024}b.
\newblock \bibinfo{title}{Wildfire risk assessment using deep learning in {Guangdong} {Province}, {China}}.
\newblock \bibinfo{journal}{International Journal of Applied Earth Observation and Geoinformation} \bibinfo{volume}{128}, \bibinfo{pages}{103750}.
\newblock \URLprefix \url{https://www.sciencedirect.com/science/article/pii/S1569843224001043}, \DOIprefix\doi{10.1016/j.jag.2024.103750}.
%Type = Article
\bibitem[{Jiang et~al.(2022)Jiang, Wang, Su, Li, Wang, Zheng, Wang and Meng}]{jiang2022modeling}
\bibinfo{author}{Jiang, W.}, \bibinfo{author}{Wang, F.}, \bibinfo{author}{Su, G.}, \bibinfo{author}{Li, X.}, \bibinfo{author}{Wang, G.}, \bibinfo{author}{Zheng, X.}, \bibinfo{author}{Wang, T.}, \bibinfo{author}{Meng, Q.}, \bibinfo{year}{2022}.
\newblock \bibinfo{title}{Modeling wildfire spread with an irregular graph network}.
\newblock \bibinfo{journal}{Fire} \bibinfo{volume}{5}, \bibinfo{pages}{185}.
%Type = Article
\bibitem[{Jin et~al.(2020a)Jin, Wang, Zhu, Feng, Huang and Hu}]{jin2020urban}
\bibinfo{author}{Jin, G.}, \bibinfo{author}{Wang, Q.}, \bibinfo{author}{Zhu, C.}, \bibinfo{author}{Feng, Y.}, \bibinfo{author}{Huang, J.}, \bibinfo{author}{Hu, X.}, \bibinfo{year}{2020}a.
\newblock \bibinfo{title}{Urban fire situation forecasting: Deep sequence learning with spatio-temporal dynamics}.
\newblock \bibinfo{journal}{Applied Soft Computing} \bibinfo{volume}{97}, \bibinfo{pages}{106730}.
%Type = Inproceedings
\bibitem[{Jin et~al.(2020b)Jin, Zhu, Chen, Sha, Hu and Huang}]{jin2020ufsp}
\bibinfo{author}{Jin, G.}, \bibinfo{author}{Zhu, C.}, \bibinfo{author}{Chen, X.}, \bibinfo{author}{Sha, H.}, \bibinfo{author}{Hu, X.}, \bibinfo{author}{Huang, J.}, \bibinfo{year}{2020}b.
\newblock \bibinfo{title}{Ufsp-net: a neural network with spatio-temporal information fusion for urban fire situation prediction}, in: \bibinfo{booktitle}{IOP Conference Series: Materials Science and Engineering}, \bibinfo{organization}{IOP Publishing}. p. \bibinfo{pages}{012050}.
%Type = Article
\bibitem[{Jolly et~al.(2015)Jolly, Cochrane, Freeborn, Holden, Brown, Williamson and Bowman}]{jolly2015climate}
\bibinfo{author}{Jolly, W.M.}, \bibinfo{author}{Cochrane, M.A.}, \bibinfo{author}{Freeborn, P.H.}, \bibinfo{author}{Holden, Z.A.}, \bibinfo{author}{Brown, T.J.}, \bibinfo{author}{Williamson, G.J.}, \bibinfo{author}{Bowman, D.M.}, \bibinfo{year}{2015}.
\newblock \bibinfo{title}{Climate-induced variations in global wildfire danger from 1979 to 2013}.
\newblock \bibinfo{journal}{Nature communications} \bibinfo{volume}{6}, \bibinfo{pages}{7537}.
%Type = Article
\bibitem[{Jolly and Johnson(2018)}]{jolly2018pyro}
\bibinfo{author}{Jolly, W.M.}, \bibinfo{author}{Johnson, D.M.}, \bibinfo{year}{2018}.
\newblock \bibinfo{title}{Pyro-ecophysiology: shifting the paradigm of live wildland fuel research}.
\newblock \bibinfo{journal}{Fire} \bibinfo{volume}{1}, \bibinfo{pages}{8}.
%Type = Misc
\bibitem[{Jones et~al.(2024a)Jones, Brambleby, Andela, van~der Werf, Parrington and Giglio}]{jones202data}
\bibinfo{author}{Jones, M.W.}, \bibinfo{author}{Brambleby, E.}, \bibinfo{author}{Andela, N.}, \bibinfo{author}{van~der Werf, G.R.}, \bibinfo{author}{Parrington, M.}, \bibinfo{author}{Giglio, L.}, \bibinfo{year}{2024}a.
\newblock \bibinfo{title}{State of wildfires 2023-24: Anomalies in burned area, fire emissions, and individual fire characteristics by continent, biome, country, and administrative region}.
\newblock \DOIprefix\doi{10.5281/zenodo.11400539}. \bibinfo{note}{[data set]}.
%Type = Article
\bibitem[{Jones et~al.(2024b)Jones, Kelley, Burton, Di~Giuseppe, Barbosa, Brambleby, Hartley, Lombardi, Mataveli, McNorton et~al.}]{jones2024state}
\bibinfo{author}{Jones, M.W.}, \bibinfo{author}{Kelley, D.I.}, \bibinfo{author}{Burton, C.A.}, \bibinfo{author}{Di~Giuseppe, F.}, \bibinfo{author}{Barbosa, M.L.F.}, \bibinfo{author}{Brambleby, E.}, \bibinfo{author}{Hartley, A.J.}, \bibinfo{author}{Lombardi, A.}, \bibinfo{author}{Mataveli, G.}, \bibinfo{author}{McNorton, J.R.}, et~al., \bibinfo{year}{2024}b.
\newblock \bibinfo{title}{State of wildfires 2023--2024}.
\newblock \bibinfo{journal}{Earth System Science Data} \bibinfo{volume}{16}, \bibinfo{pages}{3601--3685}.
%Type = Book
\bibitem[{Josephson and Josephson(1996)}]{josephson1996abductive}
\bibinfo{author}{Josephson, J.R.}, \bibinfo{author}{Josephson, S.G.}, \bibinfo{year}{1996}.
\newblock \bibinfo{title}{Abductive inference: Computation, philosophy, technology}.
\newblock \bibinfo{publisher}{Cambridge University Press}.
%Type = Article
\bibitem[{Kadir et~al.(2023)Kadir, Kung, AlMansour, Irie, Rosa and Fauzi}]{kadir2023wildfire}
\bibinfo{author}{Kadir, E.A.}, \bibinfo{author}{Kung, H.T.}, \bibinfo{author}{AlMansour, A.A.}, \bibinfo{author}{Irie, H.}, \bibinfo{author}{Rosa, S.L.}, \bibinfo{author}{Fauzi, S.S.M.}, \bibinfo{year}{2023}.
\newblock \bibinfo{title}{Wildfire hotspots forecasting and mapping for environmental monitoring based on the long short-term memory networks deep learning algorithm}.
\newblock \bibinfo{journal}{Environments} \bibinfo{volume}{10}, \bibinfo{pages}{124}.
%Type = Article
\bibitem[{Kalamkar et~al.(2023)}]{kalamkar2023multimodal}
\bibinfo{author}{Kalamkar, S.}, et~al., \bibinfo{year}{2023}.
\newblock \bibinfo{title}{Multimodal image fusion: A systematic review}.
\newblock \bibinfo{journal}{Decision Analytics Journal} , \bibinfo{pages}{100327}.
%Type = Article
\bibitem[{Kalantar et~al.(2020)Kalantar, Ueda, Idrees, Janizadeh, Ahmadi and Shabani}]{kalantar_forest_2020}
\bibinfo{author}{Kalantar, B.}, \bibinfo{author}{Ueda, N.}, \bibinfo{author}{Idrees, M.O.}, \bibinfo{author}{Janizadeh, S.}, \bibinfo{author}{Ahmadi, K.}, \bibinfo{author}{Shabani, F.}, \bibinfo{year}{2020}.
\newblock \bibinfo{title}{Forest {Fire} {Susceptibility} {Prediction} {Based} on {Machine} {Learning} {Models} with {Resampling} {Algorithms} on {Remote} {Sensing} {Data}}.
\newblock \bibinfo{journal}{Remote Sensing} \bibinfo{volume}{12}, \bibinfo{pages}{3682}.
\newblock \URLprefix \url{https://www.mdpi.com/2072-4292/12/22/3682}, \DOIprefix\doi{10.3390/rs12223682}. \bibinfo{note}{number: 22 Publisher: Multidisciplinary Digital Publishing Institute}.
%Type = Article
\bibitem[{Kanwal et~al.(2023)Kanwal, Rafaqat, Iqbal and Weiguo}]{kanwal_data-driven_2023}
\bibinfo{author}{Kanwal, R.}, \bibinfo{author}{Rafaqat, W.}, \bibinfo{author}{Iqbal, M.}, \bibinfo{author}{Weiguo, S.}, \bibinfo{year}{2023}.
\newblock \bibinfo{title}{Data-{Driven} {Approaches} for {Wildfire} {Mapping} and {Prediction} {Assessment} {Using} a {Convolutional} {Neural} {Network} ({CNN}).}
\newblock \bibinfo{journal}{Remote Sensing} \bibinfo{volume}{15}, \bibinfo{pages}{NA--NA}.
\newblock \URLprefix \url{https://go.gale.com/ps/i.do?p=AONE&sw=w&issn=20724292&v=2.1&it=r&id=GALE%7CA772535779&sid=googleScholar&linkaccess=abs}, \DOIprefix\doi{10.3390/rs15215099}. \bibinfo{note}{publisher: MDPI AG}.
%Type = Article
\bibitem[{Karniadakis et~al.(2021)Karniadakis, Kevrekidis, Lu, Perdikaris, Wang and Yang}]{karniadakis2021physics}
\bibinfo{author}{Karniadakis, G.E.}, \bibinfo{author}{Kevrekidis, I.G.}, \bibinfo{author}{Lu, L.}, \bibinfo{author}{Perdikaris, P.}, \bibinfo{author}{Wang, S.}, \bibinfo{author}{Yang, L.}, \bibinfo{year}{2021}.
\newblock \bibinfo{title}{Physics-informed machine learning}.
\newblock \bibinfo{journal}{Nature Reviews Physics} \bibinfo{volume}{3}, \bibinfo{pages}{422--440}.
%Type = Article
\bibitem[{Katrutsa and Strijov(2017)}]{katrutsa2017comprehensive}
\bibinfo{author}{Katrutsa, A.}, \bibinfo{author}{Strijov, V.}, \bibinfo{year}{2017}.
\newblock \bibinfo{title}{Comprehensive study of feature selection methods to solve multicollinearity problem according to evaluation criteria}.
\newblock \bibinfo{journal}{Expert Systems with Applications} \bibinfo{volume}{76}, \bibinfo{pages}{1--11}.
%Type = Article
\bibitem[{Keane et~al.(2001)Keane, Burgan and Wagtendonk}]{keane_mapping_2001}
\bibinfo{author}{Keane, R.E.}, \bibinfo{author}{Burgan, R.}, \bibinfo{author}{Wagtendonk, J.v.}, \bibinfo{year}{2001}.
\newblock \bibinfo{title}{Mapping wildland fuels for fire management across multiple scales: {Integrating} remote sensing, {GIS}, and biophysical modeling}.
\newblock \bibinfo{journal}{International Journal of Wildland Fire} \bibinfo{volume}{10}, \bibinfo{pages}{301--319}.
\newblock \URLprefix \url{https://www.publish.csiro.au/wf/wf01028}, \DOIprefix\doi{10.1071/wf01028}. \bibinfo{note}{publisher: CSIRO PUBLISHING}.
%Type = Article
\bibitem[{Keenan et~al.(2015)Keenan, Reams, Achard, de~Freitas, Grainger and Lindquist}]{keenan2015dynamics}
\bibinfo{author}{Keenan, R.J.}, \bibinfo{author}{Reams, G.A.}, \bibinfo{author}{Achard, F.}, \bibinfo{author}{de~Freitas, J.V.}, \bibinfo{author}{Grainger, A.}, \bibinfo{author}{Lindquist, E.}, \bibinfo{year}{2015}.
\newblock \bibinfo{title}{Dynamics of global forest area: Results from the fao global forest resources assessment 2015}.
\newblock \bibinfo{journal}{Forest ecology and management} \bibinfo{volume}{352}, \bibinfo{pages}{9--20}.
%Type = Article
\bibitem[{Khalaf et~al.(2024)Khalaf, Jouibary and Jahdi}]{khalaf2024performance}
\bibinfo{author}{Khalaf, M.W.A.}, \bibinfo{author}{Jouibary, S.S.}, \bibinfo{author}{Jahdi, R.}, \bibinfo{year}{2024}.
\newblock \bibinfo{title}{Performance analysis of convlstm, flammap, and ca algorithms to predict wildfire spread in golestan national park, ne iran}.
\newblock \bibinfo{journal}{Environmental Modeling \& Assessment} , \bibinfo{pages}{1--14}.
%Type = Article
\bibitem[{Khan et~al.(2023)Khan, Khan, Azhar, Khan, Lee and Javed}]{khan2023multimodal}
\bibinfo{author}{Khan, S.U.}, \bibinfo{author}{Khan, M.A.}, \bibinfo{author}{Azhar, M.}, \bibinfo{author}{Khan, F.}, \bibinfo{author}{Lee, Y.}, \bibinfo{author}{Javed, M.}, \bibinfo{year}{2023}.
\newblock \bibinfo{title}{Multimodal medical image fusion towards future research: A review}.
\newblock \bibinfo{journal}{Journal of King Saud University-Computer and Information Sciences} \bibinfo{volume}{35}, \bibinfo{pages}{101733}.
%Type = Article
\bibitem[{Kiefer and Zhong(2015)}]{kiefer_role_2015}
\bibinfo{author}{Kiefer, M.T.}, \bibinfo{author}{Zhong, S.}, \bibinfo{year}{2015}.
\newblock \bibinfo{title}{The role of forest cover and valley geometry in cold-air pool evolution}.
\newblock \bibinfo{journal}{Journal of Geophysical Research: Atmospheres} \bibinfo{volume}{120}, \bibinfo{pages}{8693--8711}.
\newblock \URLprefix \url{https://onlinelibrary.wiley.com/doi/abs/10.1002/2014JD022998}, \DOIprefix\doi{10.1002/2014JD022998}. \bibinfo{note}{\_eprint: https://agupubs.onlinelibrary.wiley.com/doi/pdf/10.1002/2014JD022998}.
%Type = Article
\bibitem[{Kilgore and Sando(1975)}]{kilgore1975crown}
\bibinfo{author}{Kilgore, B.M.}, \bibinfo{author}{Sando, R.W.}, \bibinfo{year}{1975}.
\newblock \bibinfo{title}{Crown-fire potential in a sequoia forest after prescribed burning}.
\newblock \bibinfo{journal}{Forest Science} \bibinfo{volume}{21}, \bibinfo{pages}{83--87}.
%Type = Article
\bibitem[{Kim et~al.(2016)Kim, Chung and Lee}]{kim2016exploring}
\bibinfo{author}{Kim, D.W.}, \bibinfo{author}{Chung, W.}, \bibinfo{author}{Lee, B.}, \bibinfo{year}{2016}.
\newblock \bibinfo{title}{Exploring tree crown spacing and slope interaction effects on fire behavior with a physics-based fire model}.
\newblock \bibinfo{journal}{Forest Science and Technology} \bibinfo{volume}{12}, \bibinfo{pages}{167--175}.
%Type = Article
\bibitem[{Kim et~al.(2012)Kim, Jackson, Bindlish, Lee and Hong}]{6112792}
\bibinfo{author}{Kim, Y.}, \bibinfo{author}{Jackson, T.}, \bibinfo{author}{Bindlish, R.}, \bibinfo{author}{Lee, H.}, \bibinfo{author}{Hong, S.}, \bibinfo{year}{2012}.
\newblock \bibinfo{title}{Radar vegetation index for estimating the vegetation water content of rice and soybean}.
\newblock \bibinfo{journal}{IEEE Geoscience and Remote Sensing Letters} \bibinfo{volume}{9}, \bibinfo{pages}{564--568}.
\newblock \DOIprefix\doi{10.1109/LGRS.2011.2174772}.
%Type = Article
\bibitem[{Knipling(1970)}]{knipling1970physical}
\bibinfo{author}{Knipling, E.B.}, \bibinfo{year}{1970}.
\newblock \bibinfo{title}{Physical and physiological basis for the reflectance of visible and near-infrared radiation from vegetation}.
\newblock \bibinfo{journal}{Remote sensing of environment} \bibinfo{volume}{1}, \bibinfo{pages}{155--159}.
%Type = Article
\bibitem[{Kochkov et~al.(2024)Kochkov, Yuval, Langmore, Norgaard, Smith, Mooers, Kl{\"o}wer, Lottes, Rasp, D{\"u}ben et~al.}]{kochkov2024neural}
\bibinfo{author}{Kochkov, D.}, \bibinfo{author}{Yuval, J.}, \bibinfo{author}{Langmore, I.}, \bibinfo{author}{Norgaard, P.}, \bibinfo{author}{Smith, J.}, \bibinfo{author}{Mooers, G.}, \bibinfo{author}{Kl{\"o}wer, M.}, \bibinfo{author}{Lottes, J.}, \bibinfo{author}{Rasp, S.}, \bibinfo{author}{D{\"u}ben, P.}, et~al., \bibinfo{year}{2024}.
\newblock \bibinfo{title}{Neural general circulation models for weather and climate}.
\newblock \bibinfo{journal}{Nature} \bibinfo{volume}{632}, \bibinfo{pages}{1060--1066}.
%Type = Inproceedings
\bibitem[{Kondylatos et~al.(2023)Kondylatos, Prapas, Camps-Valls and Papoutsis}]{kondylatos2023mesogeos}
\bibinfo{author}{Kondylatos, S.}, \bibinfo{author}{Prapas, I.}, \bibinfo{author}{Camps-Valls, G.}, \bibinfo{author}{Papoutsis, I.}, \bibinfo{year}{2023}.
\newblock \bibinfo{title}{Mesogeos: A multi-purpose dataset for data-driven wildfire modeling in the mediterranean}, in: \bibinfo{booktitle}{Thirty-seventh Conference on Neural Information Processing Systems Datasets and Benchmarks Track}.
\newblock \URLprefix \url{https://openreview.net/forum?id=VH1vxapUTs}.
%Type = Article
\bibitem[{Kondylatos et~al.(2022)Kondylatos, Prapas, Ronco, Papoutsis, Camps-Valls, Piles, Fern\'{a}ndez-Torres and Carvalhais}]{kondylatos_wildfire_2022}
\bibinfo{author}{Kondylatos, S.}, \bibinfo{author}{Prapas, I.}, \bibinfo{author}{Ronco, M.}, \bibinfo{author}{Papoutsis, I.}, \bibinfo{author}{Camps-Valls, G.}, \bibinfo{author}{Piles, M.}, \bibinfo{author}{Fern\'{a}ndez-Torres, M.A.}, \bibinfo{author}{Carvalhais, N.}, \bibinfo{year}{2022}.
\newblock \bibinfo{title}{Wildfire {Danger} {Prediction} and {Understanding} {With} {Deep} {Learning}}.
\newblock \bibinfo{journal}{Geophysical Research Letters} \bibinfo{volume}{49}, \bibinfo{pages}{e2022GL099368}.
\newblock \DOIprefix\doi{10.1029/2022GL099368}. \bibinfo{note}{\_eprint: https://onlinelibrary.wiley.com/doi/pdf/10.1029/2022GL099368}.
%Type = Book
\bibitem[{Kozlowski et~al.(1990)Kozlowski, Kramer and Pallardy}]{kozlowskiphy}
\bibinfo{author}{Kozlowski, T.T.}, \bibinfo{author}{Kramer, P.J.}, \bibinfo{author}{Pallardy, S.G.}, \bibinfo{year}{1990}.
\newblock \bibinfo{title}{{The Physiological Ecology of Woody Plants}}.
\newblock \bibinfo{publisher}{Academic Press}.
\newblock \DOIprefix\doi{10.1093/treephys/8.2.213}.
%Type = Article
\bibitem[{Kramer et~al.(2014)Kramer, Collins, Kelly and Stephens}]{kramer2014quantifying}
\bibinfo{author}{Kramer, H.A.}, \bibinfo{author}{Collins, B.M.}, \bibinfo{author}{Kelly, M.}, \bibinfo{author}{Stephens, S.L.}, \bibinfo{year}{2014}.
\newblock \bibinfo{title}{Quantifying ladder fuels: A new approach using lidar}.
\newblock \bibinfo{journal}{Forests} \bibinfo{volume}{5}, \bibinfo{pages}{1432--1453}.
%Type = Incollection
\bibitem[{Kriedemann and Barrs(1983)}]{kriedemann_photosynthetic_1983}
\bibinfo{author}{Kriedemann, P.E.}, \bibinfo{author}{Barrs, H.D.}, \bibinfo{year}{1983}.
\newblock \bibinfo{title}{Photosynthetic {Adaptation} to {Water} {Stress} and {Implications} for {Drought} {Resistance}}, in: \bibinfo{booktitle}{Crop {Reactions} {To} {Water} {And} {Temperature} {Stresses} {In} {Humid}, {Temperate} {Climates}}. \bibinfo{publisher}{CRC Press}.
\newblock \bibinfo{note}{Num Pages: 30}.
%Type = Article
\bibitem[{Krizhevsky et~al.(2012)Krizhevsky, Sutskever and Hinton}]{krizhevsky2012imagenet}
\bibinfo{author}{Krizhevsky, A.}, \bibinfo{author}{Sutskever, I.}, \bibinfo{author}{Hinton, G.E.}, \bibinfo{year}{2012}.
\newblock \bibinfo{title}{Imagenet classification with deep convolutional neural networks}.
\newblock \bibinfo{journal}{Advances in neural information processing systems} \bibinfo{volume}{25}.
%Type = Article
\bibitem[{Krueger et~al.(2022)Krueger, Levi, Achieng, Bolten, Carlson, Coops, Holden, Magi, Rigden and Ochsner}]{krueger_using_2022}
\bibinfo{author}{Krueger, E.S.}, \bibinfo{author}{Levi, M.R.}, \bibinfo{author}{Achieng, K.O.}, \bibinfo{author}{Bolten, J.D.}, \bibinfo{author}{Carlson, J.D.}, \bibinfo{author}{Coops, N.C.}, \bibinfo{author}{Holden, Z.A.}, \bibinfo{author}{Magi, B.I.}, \bibinfo{author}{Rigden, A.J.}, \bibinfo{author}{Ochsner, T.E.}, \bibinfo{year}{2022}.
\newblock \bibinfo{title}{Using soil moisture information to better understand and predict wildfire danger: a review of recent developments and outstanding questions}.
\newblock \bibinfo{journal}{International Journal of Wildland Fire} \bibinfo{volume}{32}, \bibinfo{pages}{111--132}.
\newblock \URLprefix \url{https://www.publish.csiro.au/wf/WF22056}, \DOIprefix\doi{10.1071/WF22056}. \bibinfo{note}{publisher: CSIRO PUBLISHING}.
%Type = Inproceedings
\bibitem[{Kurth et~al.(2023)Kurth, Subramanian, Harrington, Pathak, Mardani, Hall, Miele, Kashinath and Anandkumar}]{fourcast2023}
\bibinfo{author}{Kurth, T.}, \bibinfo{author}{Subramanian, S.}, \bibinfo{author}{Harrington, P.}, \bibinfo{author}{Pathak, J.}, \bibinfo{author}{Mardani, M.}, \bibinfo{author}{Hall, D.}, \bibinfo{author}{Miele, A.}, \bibinfo{author}{Kashinath, K.}, \bibinfo{author}{Anandkumar, A.}, \bibinfo{year}{2023}.
\newblock \bibinfo{title}{Fourcastnet: Accelerating global high-resolution weather forecasting using adaptive fourier neural operators}, in: \bibinfo{booktitle}{Proceedings of the Platform for Advanced Scientific Computing Conference}, \bibinfo{publisher}{Association for Computing Machinery}, \bibinfo{address}{New York, NY, USA}.
\newblock \URLprefix \url{https://doi.org/10.1145/3592979.3593412}, \DOIprefix\doi{10.1145/3592979.3593412}.
%Type = Article
\bibitem[{Labenski et~al.(2023)Labenski, Ewald, Schmidtlein, Heinsch and Fassnacht}]{labenski_quantifying_2023}
\bibinfo{author}{Labenski, P.}, \bibinfo{author}{Ewald, M.}, \bibinfo{author}{Schmidtlein, S.}, \bibinfo{author}{Heinsch, F.A.}, \bibinfo{author}{Fassnacht, F.E.}, \bibinfo{year}{2023}.
\newblock \bibinfo{title}{Quantifying surface fuels for fire modelling in temperate forests using airborne lidar and {Sentinel}-2: potential and limitations}.
\newblock \bibinfo{journal}{Remote Sensing of Environment} \bibinfo{volume}{295}, \bibinfo{pages}{113711}.
\newblock \URLprefix \url{https://www.sciencedirect.com/science/article/pii/S0034425723002626}, \DOIprefix\doi{10.1016/j.rse.2023.113711}.
%Type = Inproceedings
\bibitem[{Lall and Mathibela(2016)}]{lall2016application}
\bibinfo{author}{Lall, S.}, \bibinfo{author}{Mathibela, B.}, \bibinfo{year}{2016}.
\newblock \bibinfo{title}{The application of artificial neural networks for wildfire risk prediction}, in: \bibinfo{booktitle}{2016 International Conference on Robotics and Automation for Humanitarian Applications (RAHA)}, \bibinfo{organization}{IEEE}. pp. \bibinfo{pages}{1--6}.
%Type = Article
\bibitem[{Lam et~al.(2023)Lam, Sanchez-Gonzalez, Willson, Wirnsberger, Fortunato, Alet, Ravuri, Ewalds, Eaton-Rosen, Hu et~al.}]{lam2023learning}
\bibinfo{author}{Lam, R.}, \bibinfo{author}{Sanchez-Gonzalez, A.}, \bibinfo{author}{Willson, M.}, \bibinfo{author}{Wirnsberger, P.}, \bibinfo{author}{Fortunato, M.}, \bibinfo{author}{Alet, F.}, \bibinfo{author}{Ravuri, S.}, \bibinfo{author}{Ewalds, T.}, \bibinfo{author}{Eaton-Rosen, Z.}, \bibinfo{author}{Hu, W.}, et~al., \bibinfo{year}{2023}.
\newblock \bibinfo{title}{Learning skillful medium-range global weather forecasting}.
\newblock \bibinfo{journal}{Science} \bibinfo{volume}{382}, \bibinfo{pages}{1416--1421}.
%Type = Misc
\bibitem[{Lang et~al.(2024)Lang, Alexe, Chantry, Dramsch, Pinault, Raoult, Clare, Lessig, Maier-Gerber, Magnusson, Bouallègue, Nemesio, Dueben, Brown, Pappenberger and Rabier}]{lang2024aifsecmwfsdatadriven}
\bibinfo{author}{Lang, S.}, \bibinfo{author}{Alexe, M.}, \bibinfo{author}{Chantry, M.}, \bibinfo{author}{Dramsch, J.}, \bibinfo{author}{Pinault, F.}, \bibinfo{author}{Raoult, B.}, \bibinfo{author}{Clare, M.C.A.}, \bibinfo{author}{Lessig, C.}, \bibinfo{author}{Maier-Gerber, M.}, \bibinfo{author}{Magnusson, L.}, \bibinfo{author}{Bouallègue, Z.B.}, \bibinfo{author}{Nemesio, A.P.}, \bibinfo{author}{Dueben, P.D.}, \bibinfo{author}{Brown, A.}, \bibinfo{author}{Pappenberger, F.}, \bibinfo{author}{Rabier, F.}, \bibinfo{year}{2024}.
\newblock \bibinfo{title}{Aifs -- ecmwf's data-driven forecasting system}.
\newblock \URLprefix \url{https://arxiv.org/abs/2406.01465}, \href{http://arxiv.org/abs/2406.01465}{{\tt arXiv:2406.01465}}.
%Type = Article
\bibitem[{Lecina-Diaz et~al.(2014)Lecina-Diaz, Alvarez and Retana}]{lecina-diaz_extreme_2014}
\bibinfo{author}{Lecina-Diaz, J.}, \bibinfo{author}{Alvarez, A.}, \bibinfo{author}{Retana, J.}, \bibinfo{year}{2014}.
\newblock \bibinfo{title}{Extreme {Fire} {Severity} {Patterns} in {Topographic}, {Convective} and {Wind}-{Driven} {Historical} {Wildfires} of {Mediterranean} {Pine} {Forests}}.
\newblock \bibinfo{journal}{PLOS ONE} \bibinfo{volume}{9}, \bibinfo{pages}{e85127}.
\newblock \URLprefix \url{https://journals.plos.org/plosone/article?id=10.1371/journal.pone.0085127}, \DOIprefix\doi{10.1371/journal.pone.0085127}. \bibinfo{note}{publisher: Public Library of Science}.
%Type = Article
\bibitem[{LeCun et~al.(2015)LeCun, Bengio and Hinton}]{lecun2015deep}
\bibinfo{author}{LeCun, Y.}, \bibinfo{author}{Bengio, Y.}, \bibinfo{author}{Hinton, G.}, \bibinfo{year}{2015}.
\newblock \bibinfo{title}{Deep learning}.
\newblock \bibinfo{journal}{nature} \bibinfo{volume}{521}, \bibinfo{pages}{436--444}.
%Type = Article
\bibitem[{Lee et~al.(2022)Lee, Lee, Ryu, Kwon, Seo and Kim}]{lee2022prediction}
\bibinfo{author}{Lee, S.J.}, \bibinfo{author}{Lee, Y.J.}, \bibinfo{author}{Ryu, J.Y.}, \bibinfo{author}{Kwon, C.G.}, \bibinfo{author}{Seo, K.W.}, \bibinfo{author}{Kim, S.Y.}, \bibinfo{year}{2022}.
\newblock \bibinfo{title}{Prediction of wildfire fuel load for pinus densiflora stands in south korea based on the forest-growth model}.
\newblock \bibinfo{journal}{Forests} \bibinfo{volume}{13}, \bibinfo{pages}{1372}.
%Type = Article
\bibitem[{Leite et~al.(2024)Leite, Amaral, Neigh, Cosenza, Klauberg, Hudak, Arag{\~a}o, Morton, Coffield, McCabe et~al.}]{leite2024leveraging}
\bibinfo{author}{Leite, R.V.}, \bibinfo{author}{Amaral, C.}, \bibinfo{author}{Neigh, C.S.}, \bibinfo{author}{Cosenza, D.N.}, \bibinfo{author}{Klauberg, C.}, \bibinfo{author}{Hudak, A.T.}, \bibinfo{author}{Arag{\~a}o, L.}, \bibinfo{author}{Morton, D.C.}, \bibinfo{author}{Coffield, S.}, \bibinfo{author}{McCabe, T.}, et~al., \bibinfo{year}{2024}.
\newblock \bibinfo{title}{Leveraging the next generation of spaceborne earth observations for fuel monitoring and wildland fire management}.
\newblock \bibinfo{journal}{Remote Sensing in Ecology and Conservation} .
%Type = Article
\bibitem[{Leite et~al.(2022)Leite, Silva, Broadbent, Do~Amaral, Liesenberg, De~Almeida, Mohan, Godinho, Cardil, Hamamura et~al.}]{leite2022large}
\bibinfo{author}{Leite, R.V.}, \bibinfo{author}{Silva, C.A.}, \bibinfo{author}{Broadbent, E.N.}, \bibinfo{author}{Do~Amaral, C.H.}, \bibinfo{author}{Liesenberg, V.}, \bibinfo{author}{De~Almeida, D.R.A.}, \bibinfo{author}{Mohan, M.}, \bibinfo{author}{Godinho, S.}, \bibinfo{author}{Cardil, A.}, \bibinfo{author}{Hamamura, C.}, et~al., \bibinfo{year}{2022}.
\newblock \bibinfo{title}{Large scale multi-layer fuel load characterization in tropical savanna using gedi spaceborne lidar data}.
\newblock \bibinfo{journal}{Remote Sensing of Environment} \bibinfo{volume}{268}, \bibinfo{pages}{112764}.
%Type = Article
\bibitem[{Lelis et~al.(2024)Lelis, Roncal, Silveira, De~Aquino, Marcondes, Marques, Loubach, Verri, Curtis and De~Souza}]{10681400}
\bibinfo{author}{Lelis, C.A.S.}, \bibinfo{author}{Roncal, J.J.}, \bibinfo{author}{Silveira, L.}, \bibinfo{author}{De~Aquino, R.D.G.}, \bibinfo{author}{Marcondes, C.A.C.}, \bibinfo{author}{Marques, J.}, \bibinfo{author}{Loubach, D.S.}, \bibinfo{author}{Verri, F.A.N.}, \bibinfo{author}{Curtis, V.V.}, \bibinfo{author}{De~Souza, D.G.}, \bibinfo{year}{2024}.
\newblock \bibinfo{title}{Drone-based ai system for wildfire monitoring and risk prediction}.
\newblock \bibinfo{journal}{IEEE Access} \bibinfo{volume}{12}, \bibinfo{pages}{139865--139882}.
\newblock \DOIprefix\doi{10.1109/ACCESS.2024.3462436}.
%Type = Article
\bibitem[{Leuenberger et~al.(2018)Leuenberger, Parente, Tonini, Pereira and Kanevski}]{leuenberger2018wildfire}
\bibinfo{author}{Leuenberger, M.}, \bibinfo{author}{Parente, J.}, \bibinfo{author}{Tonini, M.}, \bibinfo{author}{Pereira, M.G.}, \bibinfo{author}{Kanevski, M.}, \bibinfo{year}{2018}.
\newblock \bibinfo{title}{Wildfire susceptibility mapping: Deterministic vs. stochastic approaches}.
\newblock \bibinfo{journal}{Environmental Modelling \& Software} \bibinfo{volume}{101}, \bibinfo{pages}{194--203}.
%Type = Inproceedings
\bibitem[{Li and Rad(2024)}]{li2024wildfire}
\bibinfo{author}{Li, B.S.}, \bibinfo{author}{Rad, R.}, \bibinfo{year}{2024}.
\newblock \bibinfo{title}{Wildfire spread prediction in north america using satellite imagery and vision transformer}, in: \bibinfo{booktitle}{2024 IEEE Conference on Artificial Intelligence (CAI)}, \bibinfo{organization}{IEEE}. pp. \bibinfo{pages}{1536--1541}.
%Type = Inproceedings
\bibitem[{Li et~al.(2023a)Li, Zhang, Xu, Liu, Zhang, Ni and Shum}]{Li_2023_CVPR}
\bibinfo{author}{Li, F.}, \bibinfo{author}{Zhang, H.}, \bibinfo{author}{Xu, H.}, \bibinfo{author}{Liu, S.}, \bibinfo{author}{Zhang, L.}, \bibinfo{author}{Ni, L.M.}, \bibinfo{author}{Shum, H.Y.}, \bibinfo{year}{2023}a.
\newblock \bibinfo{title}{Mask dino: Towards a unified transformer-based framework for object detection and segmentation}, in: \bibinfo{booktitle}{Proceedings of the IEEE/CVF Conference on Computer Vision and Pattern Recognition (CVPR)}, pp. \bibinfo{pages}{3041--3050}.
%Type = Article
\bibitem[{Li et~al.(2023b)Li, Zhu, Riley, Zhao, Xu, Yuan, Chen, Wu, Gui, Gong et~al.}]{li2023attentionfire_v1}
\bibinfo{author}{Li, F.}, \bibinfo{author}{Zhu, Q.}, \bibinfo{author}{Riley, W.J.}, \bibinfo{author}{Zhao, L.}, \bibinfo{author}{Xu, L.}, \bibinfo{author}{Yuan, K.}, \bibinfo{author}{Chen, M.}, \bibinfo{author}{Wu, H.}, \bibinfo{author}{Gui, Z.}, \bibinfo{author}{Gong, J.}, et~al., \bibinfo{year}{2023}b.
\newblock \bibinfo{title}{Attentionfire\_v1. 0: interpretable machine learning fire model for burned-area predictions over tropics}.
\newblock \bibinfo{journal}{Geoscientific Model Development} \bibinfo{volume}{16}, \bibinfo{pages}{869--884}.
%Type = Article
\bibitem[{Li et~al.(2022a)Li, Xu, Yi and Liu}]{li_predictive_2022}
\bibinfo{author}{Li, W.}, \bibinfo{author}{Xu, Q.}, \bibinfo{author}{Yi, J.}, \bibinfo{author}{Liu, J.}, \bibinfo{year}{2022}a.
\newblock \bibinfo{title}{Predictive model of spatial scale of forest fire driving factors: a case study of {Yunnan} {Province}, {China}}.
\newblock \bibinfo{journal}{Scientific Reports} \bibinfo{volume}{12}, \bibinfo{pages}{19029}.
\newblock \DOIprefix\doi{10.1038/s41598-022-23697-6}. \bibinfo{note}{publisher: Nature Publishing Group}.
%Type = Article
\bibitem[{Li et~al.(2023c)Li, Wang, Sun, Wang, Li and Li}]{li2023predicting}
\bibinfo{author}{Li, X.}, \bibinfo{author}{Wang, X.}, \bibinfo{author}{Sun, S.}, \bibinfo{author}{Wang, Y.}, \bibinfo{author}{Li, S.}, \bibinfo{author}{Li, D.}, \bibinfo{year}{2023}c.
\newblock \bibinfo{title}{Predicting the wildland fire spread using a mixed-input cnn model with both channel and spatial attention mechanisms}.
\newblock \bibinfo{journal}{Fire Technology} \bibinfo{volume}{59}, \bibinfo{pages}{2683--2717}.
%Type = Article
\bibitem[{Li et~al.(2022b)Li, Chen, He and Veraverbeke}]{li_forest_2022}
\bibinfo{author}{Li, Y.}, \bibinfo{author}{Chen, R.}, \bibinfo{author}{He, B.}, \bibinfo{author}{Veraverbeke, S.}, \bibinfo{year}{2022}b.
\newblock \bibinfo{title}{Forest foliage fuel load estimation from multi-sensor spatiotemporal features}.
\newblock \bibinfo{journal}{International Journal of Applied Earth Observation and Geoinformation} \bibinfo{volume}{115}, \bibinfo{pages}{103101}.
\newblock \URLprefix \url{https://www.sciencedirect.com/science/article/pii/S1569843222002898}, \DOIprefix\doi{10.1016/j.jag.2022.103101}.
%Type = Inproceedings
\bibitem[{Li et~al.(2021a)Li, He, Kong, Xu, Zhang and Quan}]{9553105}
\bibinfo{author}{Li, Y.}, \bibinfo{author}{He, B.}, \bibinfo{author}{Kong, P.}, \bibinfo{author}{Xu, H.}, \bibinfo{author}{Zhang, Q.}, \bibinfo{author}{Quan, X.}, \bibinfo{year}{2021}a.
\newblock \bibinfo{title}{Estimation of forest surface dead fuel loads based on multi-source remote sensing data}, in: \bibinfo{booktitle}{2021 IEEE International Geoscience and Remote Sensing Symposium IGARSS}, pp. \bibinfo{pages}{6893--6896}.
\newblock \DOIprefix\doi{10.1109/IGARSS47720.2021.9553105}.
%Type = Article
\bibitem[{Li et~al.(2021b)Li, Quan, Liao and He}]{li2021forest}
\bibinfo{author}{Li, Y.}, \bibinfo{author}{Quan, X.}, \bibinfo{author}{Liao, Z.}, \bibinfo{author}{He, B.}, \bibinfo{year}{2021}b.
\newblock \bibinfo{title}{Forest fuel loads estimation from landsat etm+ and alos palsar data}.
\newblock \bibinfo{journal}{Remote Sensing} \bibinfo{volume}{13}, \bibinfo{pages}{1189}.
%Type = Article
\bibitem[{Li et~al.(2021c)Li, Huang, Li and Xu}]{li_wildland_2021}
\bibinfo{author}{Li, Z.}, \bibinfo{author}{Huang, Y.}, \bibinfo{author}{Li, X.}, \bibinfo{author}{Xu, L.}, \bibinfo{year}{2021}c.
\newblock \bibinfo{title}{Wildland {Fire} {Burned} {Areas} {Prediction} {Using} {Long} {Short}-{Term} {Memory} {Neural} {Network} with {Attention} {Mechanism}}.
\newblock \bibinfo{journal}{Fire Technology} \bibinfo{volume}{57}, \bibinfo{pages}{1--23}.
\newblock \URLprefix \url{https://doi.org/10.1007/s10694-020-01028-3}, \DOIprefix\doi{10.1007/s10694-020-01028-3}.
%Type = Article
\bibitem[{Li et~al.(2020)Li, Shi, Vogelmann, Hawbaker and Peterson}]{li2020assessment}
\bibinfo{author}{Li, Z.}, \bibinfo{author}{Shi, H.}, \bibinfo{author}{Vogelmann, J.E.}, \bibinfo{author}{Hawbaker, T.J.}, \bibinfo{author}{Peterson, B.}, \bibinfo{year}{2020}.
\newblock \bibinfo{title}{Assessment of fire fuel load dynamics in shrubland ecosystems in the western united states using modis products}.
\newblock \bibinfo{journal}{Remote Sensing} \bibinfo{volume}{12}, \bibinfo{pages}{1911}.
%Type = Article
\bibitem[{Liang et~al.(2019)Liang, Zhang and Wang}]{liang_neural_2019}
\bibinfo{author}{Liang, H.}, \bibinfo{author}{Zhang, M.}, \bibinfo{author}{Wang, H.}, \bibinfo{year}{2019}.
\newblock \bibinfo{title}{A {Neural} {Network} {Model} for {Wildfire} {Scale} {Prediction} {Using} {Meteorological} {Factors}}.
\newblock \bibinfo{journal}{IEEE Access} \bibinfo{volume}{7}, \bibinfo{pages}{176746--176755}.
\newblock \URLprefix \url{https://ieeexplore.ieee.org/document/8924693}, \DOIprefix\doi{10.1109/ACCESS.2019.2957837}. \bibinfo{note}{conference Name: IEEE Access}.
%Type = Article
\bibitem[{Liang et~al.(2023)Liang, Duncanson, Silva and Sedano}]{liang2023quantifying}
\bibinfo{author}{Liang, M.}, \bibinfo{author}{Duncanson, L.}, \bibinfo{author}{Silva, J.A.}, \bibinfo{author}{Sedano, F.}, \bibinfo{year}{2023}.
\newblock \bibinfo{title}{Quantifying aboveground biomass dynamics from charcoal degradation in mozambique using gedi lidar and landsat}.
\newblock \bibinfo{journal}{Remote sensing of environment} \bibinfo{volume}{284}, \bibinfo{pages}{113367}.
%Type = Article
\bibitem[{Liao and Valliant(2012)}]{liao_variance_2012}
\bibinfo{author}{Liao, D.}, \bibinfo{author}{Valliant, R.}, \bibinfo{year}{2012}.
\newblock \bibinfo{title}{Variance inflation factors in the analysis of complex survey data}.
\newblock \bibinfo{journal}{Survey Methodology} \bibinfo{volume}{38}, \bibinfo{pages}{53--62}.
%Type = Article
\bibitem[{Lin et~al.(2024)Lin, Giannico, Lafortezza, Sanesi and Elia}]{lin2024use}
\bibinfo{author}{Lin, D.}, \bibinfo{author}{Giannico, V.}, \bibinfo{author}{Lafortezza, R.}, \bibinfo{author}{Sanesi, G.}, \bibinfo{author}{Elia, M.}, \bibinfo{year}{2024}.
\newblock \bibinfo{title}{Use of airborne lidar to predict fine dead fuel load in mediterranean forest stands of southern europe}.
\newblock \bibinfo{journal}{Fire Ecology} \bibinfo{volume}{20}, \bibinfo{pages}{58}.
%Type = Article
\bibitem[{Liu et~al.(2023)Liu, Zeng, Ren, Li, Zhang, Yang, Li, Yang, Su, Zhu et~al.}]{liu2023grounding}
\bibinfo{author}{Liu, S.}, \bibinfo{author}{Zeng, Z.}, \bibinfo{author}{Ren, T.}, \bibinfo{author}{Li, F.}, \bibinfo{author}{Zhang, H.}, \bibinfo{author}{Yang, J.}, \bibinfo{author}{Li, C.}, \bibinfo{author}{Yang, J.}, \bibinfo{author}{Su, H.}, \bibinfo{author}{Zhu, J.}, et~al., \bibinfo{year}{2023}.
\newblock \bibinfo{title}{Grounding dino: Marrying dino with grounded pre-training for open-set object detection}.
\newblock \bibinfo{journal}{arXiv preprint arXiv:2303.05499} .
%Type = Article
\bibitem[{Liu et~al.(2021a)Liu, Lin and Feng}]{liu2021short}
\bibinfo{author}{Liu, X.}, \bibinfo{author}{Lin, Z.}, \bibinfo{author}{Feng, Z.}, \bibinfo{year}{2021}a.
\newblock \bibinfo{title}{Short-term offshore wind speed forecast by seasonal arima-a comparison against gru and lstm}.
\newblock \bibinfo{journal}{Energy} \bibinfo{volume}{227}, \bibinfo{pages}{120492}.
%Type = Article
\bibitem[{Liu et~al.(2010)Liu, Stanturf and Goodrick}]{liu_wildfire_2010}
\bibinfo{author}{Liu, Y.}, \bibinfo{author}{Stanturf, J.}, \bibinfo{author}{Goodrick, S.}, \bibinfo{year}{2010}.
\newblock \bibinfo{title}{Wildfire potential evaluation during a drought event with a regional climate model and {NDVI}}.
\newblock \bibinfo{journal}{Ecological Informatics} \bibinfo{volume}{5}, \bibinfo{pages}{418--428}.
\newblock \URLprefix \url{https://www.sciencedirect.com/science/article/pii/S1574954110000531}, \DOIprefix\doi{10.1016/j.ecoinf.2010.04.001}.
%Type = Inproceedings
\bibitem[{Liu et~al.(2021b)Liu, Lin, Cao, Hu, Wei, Zhang, Lin and Guo}]{Liu_2021_ICCV}
\bibinfo{author}{Liu, Z.}, \bibinfo{author}{Lin, Y.}, \bibinfo{author}{Cao, Y.}, \bibinfo{author}{Hu, H.}, \bibinfo{author}{Wei, Y.}, \bibinfo{author}{Zhang, Z.}, \bibinfo{author}{Lin, S.}, \bibinfo{author}{Guo, B.}, \bibinfo{year}{2021}b.
\newblock \bibinfo{title}{Swin transformer: Hierarchical vision transformer using shifted windows}, in: \bibinfo{booktitle}{Proceedings of the IEEE/CVF International Conference on Computer Vision (ICCV)}, pp. \bibinfo{pages}{10012--10022}.
%Type = Article
\bibitem[{Lizundia-Loiola et~al.(2020)Lizundia-Loiola, Ot{\'o}n, Ramo and Chuvieco}]{lizundia2020spatio}
\bibinfo{author}{Lizundia-Loiola, J.}, \bibinfo{author}{Ot{\'o}n, G.}, \bibinfo{author}{Ramo, R.}, \bibinfo{author}{Chuvieco, E.}, \bibinfo{year}{2020}.
\newblock \bibinfo{title}{A spatio-temporal active-fire clustering approach for global burned area mapping at 250 m from modis data}.
\newblock \bibinfo{journal}{Remote Sensing of Environment} \bibinfo{volume}{236}, \bibinfo{pages}{111493}.
%Type = Article
\bibitem[{Lopes~Queiroz et~al.(2020)Lopes~Queiroz, McDermid, Linke, Hopkinson and Kariyeva}]{lopes2020estimating}
\bibinfo{author}{Lopes~Queiroz, G.}, \bibinfo{author}{McDermid, G.J.}, \bibinfo{author}{Linke, J.}, \bibinfo{author}{Hopkinson, C.}, \bibinfo{author}{Kariyeva, J.}, \bibinfo{year}{2020}.
\newblock \bibinfo{title}{Estimating coarse woody debris volume using image analysis and multispectral lidar}.
\newblock \bibinfo{journal}{Forests} \bibinfo{volume}{11}, \bibinfo{pages}{141}.
%Type = Article
\bibitem[{Lopez et~al.(2024)Lopez, Avila, VanderRoest, Roth, Fendorf and Borch}]{lopez2024molecular}
\bibinfo{author}{Lopez, A.M.}, \bibinfo{author}{Avila, C.C.E.}, \bibinfo{author}{VanderRoest, J.P.}, \bibinfo{author}{Roth, H.K.}, \bibinfo{author}{Fendorf, S.}, \bibinfo{author}{Borch, T.}, \bibinfo{year}{2024}.
\newblock \bibinfo{title}{Molecular insights and impacts of wildfire-induced soil chemical changes}.
\newblock \bibinfo{journal}{Nature Reviews Earth \& Environment} , \bibinfo{pages}{1--16}.
%Type = Article
\bibitem[{Lu and Wei(2021a)}]{lu2021evaluation}
\bibinfo{author}{Lu, Y.}, \bibinfo{author}{Wei, C.}, \bibinfo{year}{2021}a.
\newblock \bibinfo{title}{Evaluation of microwave soil moisture data for monitoring live fuel moisture content (lfmc) over the coterminous united states}.
\newblock \bibinfo{journal}{Science of The Total Environment} \bibinfo{volume}{771}, \bibinfo{pages}{145410}.
%Type = Article
\bibitem[{Lu and Wei(2021b)}]{lu_evaluation_2021}
\bibinfo{author}{Lu, Y.}, \bibinfo{author}{Wei, C.}, \bibinfo{year}{2021}b.
\newblock \bibinfo{title}{Evaluation of microwave soil moisture data for monitoring live fuel moisture content ({LFMC}) over the coterminous {United} {States}}.
\newblock \bibinfo{journal}{Science of The Total Environment} \bibinfo{volume}{771}, \bibinfo{pages}{145410}.
\newblock \URLprefix \url{https://www.sciencedirect.com/science/article/pii/S0048969721004782}, \DOIprefix\doi{10.1016/j.scitotenv.2021.145410}.
%Type = Inproceedings
\bibitem[{Lundberg and Lee(2017)}]{lundberg_unified_2017}
\bibinfo{author}{Lundberg, S.M.}, \bibinfo{author}{Lee, S.I.}, \bibinfo{year}{2017}.
\newblock \bibinfo{title}{A {Unified} {Approach} to {Interpreting} {Model} {Predictions}}, in: \bibinfo{booktitle}{Advances in {Neural} {Information} {Processing} {Systems}}, \bibinfo{publisher}{Curran Associates, Inc.}
\newblock \URLprefix \url{https://proceedings.neurips.cc/paper/2017/hash/8a20a8621978632d76c43dfd28b67767-Abstract.html}.
%Type = Article
\bibitem[{Luo et~al.(2018)Luo, Zhai, Su, Ma, Kelly and Guo}]{luo2018simple}
\bibinfo{author}{Luo, L.}, \bibinfo{author}{Zhai, Q.}, \bibinfo{author}{Su, Y.}, \bibinfo{author}{Ma, Q.}, \bibinfo{author}{Kelly, M.}, \bibinfo{author}{Guo, Q.}, \bibinfo{year}{2018}.
\newblock \bibinfo{title}{Simple method for direct crown base height estimation of individual conifer trees using airborne lidar data}.
\newblock \bibinfo{journal}{Optics express} \bibinfo{volume}{26}, \bibinfo{pages}{A562--A578}.
%Type = Article
\bibitem[{Malik et~al.(2021)Malik, Rao, Puppala, Koouri, Thota, Liu, Chiao and Gao}]{malik_data-driven_2021}
\bibinfo{author}{Malik, A.}, \bibinfo{author}{Rao, M.R.}, \bibinfo{author}{Puppala, N.}, \bibinfo{author}{Koouri, P.}, \bibinfo{author}{Thota, V.A.K.}, \bibinfo{author}{Liu, Q.}, \bibinfo{author}{Chiao, S.}, \bibinfo{author}{Gao, J.}, \bibinfo{year}{2021}.
\newblock \bibinfo{title}{Data-{Driven} {Wildfire} {Risk} {Prediction} in {Northern} {California}}.
\newblock \bibinfo{journal}{Atmosphere} \bibinfo{volume}{12}, \bibinfo{pages}{109}.
\newblock \URLprefix \url{https://www.mdpi.com/2073-4433/12/1/109}, \DOIprefix\doi{10.3390/atmos12010109}. \bibinfo{note}{number: 1 Publisher: Multidisciplinary Digital Publishing Institute}.
%Type = Article
\bibitem[{Mansuy et~al.(2019)Mansuy, Miller, Parisien, Parks, Batllori and Moritz}]{mansuy_contrasting_2019}
\bibinfo{author}{Mansuy, N.}, \bibinfo{author}{Miller, C.}, \bibinfo{author}{Parisien, M.A.}, \bibinfo{author}{Parks, S.A.}, \bibinfo{author}{Batllori, E.}, \bibinfo{author}{Moritz, M.A.}, \bibinfo{year}{2019}.
\newblock \bibinfo{title}{Contrasting human influences and macro-environmental factors on fire activity inside and outside protected areas of {North} {America}}.
\newblock \bibinfo{journal}{Environmental Research Letters} \bibinfo{volume}{14}, \bibinfo{pages}{064007}.
\newblock \URLprefix \url{https://dx.doi.org/10.1088/1748-9326/ab1bc5}, \DOIprefix\doi{10.1088/1748-9326/ab1bc5}. \bibinfo{note}{publisher: IOP Publishing}.
%Type = Article
\bibitem[{Marcozzi et~al.(2023)Marcozzi, Johnson, Parsons, Flanary, Seielstad and Downs}]{marcozzi2023application}
\bibinfo{author}{Marcozzi, A.A.}, \bibinfo{author}{Johnson, J.V.}, \bibinfo{author}{Parsons, R.A.}, \bibinfo{author}{Flanary, S.J.}, \bibinfo{author}{Seielstad, C.A.}, \bibinfo{author}{Downs, J.Z.}, \bibinfo{year}{2023}.
\newblock \bibinfo{title}{Application of lidar derived fuel cells to wildfire modeling at laboratory scale}.
\newblock \bibinfo{journal}{Fire} \bibinfo{volume}{6}, \bibinfo{pages}{394}.
%Type = Article
\bibitem[{Marino et~al.(2020)Marino, Yebra, Guill{\'e}n-Climent, Algeet, Tom{\'e}, Madrigal, Guijarro and Hernando}]{marino2020investigating}
\bibinfo{author}{Marino, E.}, \bibinfo{author}{Yebra, M.}, \bibinfo{author}{Guill{\'e}n-Climent, M.}, \bibinfo{author}{Algeet, N.}, \bibinfo{author}{Tom{\'e}, J.L.}, \bibinfo{author}{Madrigal, J.}, \bibinfo{author}{Guijarro, M.}, \bibinfo{author}{Hernando, C.}, \bibinfo{year}{2020}.
\newblock \bibinfo{title}{Investigating live fuel moisture content estimation in fire-prone shrubland from remote sensing using empirical modelling and rtm simulations}.
\newblock \bibinfo{journal}{Remote Sensing} \bibinfo{volume}{12}, \bibinfo{pages}{2251}.
%Type = Article
\bibitem[{Marjani et~al.(2023)Marjani, Ahmadi and Mahdianpari}]{marjani2023firepred}
\bibinfo{author}{Marjani, M.}, \bibinfo{author}{Ahmadi, S.A.}, \bibinfo{author}{Mahdianpari, M.}, \bibinfo{year}{2023}.
\newblock \bibinfo{title}{Firepred: A hybrid multi-temporal convolutional neural network model for wildfire spread prediction}.
\newblock \bibinfo{journal}{Ecological Informatics} \bibinfo{volume}{78}, \bibinfo{pages}{102282}.
%Type = Article
\bibitem[{Marjani et~al.(2024a)Marjani, Mahdianpari, Ahmadi, Hemmati, Mohammadimanesh and Mesgari}]{marjani2024application}
\bibinfo{author}{Marjani, M.}, \bibinfo{author}{Mahdianpari, M.}, \bibinfo{author}{Ahmadi, S.A.}, \bibinfo{author}{Hemmati, E.}, \bibinfo{author}{Mohammadimanesh, F.}, \bibinfo{author}{Mesgari, M.S.}, \bibinfo{year}{2024}a.
\newblock \bibinfo{title}{Application of explainable artificial intelligence in predicting wildfire spread: an aspp-enabled cnn approach}.
\newblock \bibinfo{journal}{IEEE Geoscience and Remote Sensing Letters} .
%Type = Article
\bibitem[{Marjani et~al.(2024b)Marjani, Mahdianpari and Mohammadimanesh}]{marjani2024cnn}
\bibinfo{author}{Marjani, M.}, \bibinfo{author}{Mahdianpari, M.}, \bibinfo{author}{Mohammadimanesh, F.}, \bibinfo{year}{2024}b.
\newblock \bibinfo{title}{Cnn-bilstm: A novel deep learning model for near-real-time daily wildfire spread prediction}.
\newblock \bibinfo{journal}{Remote Sensing} \bibinfo{volume}{16}, \bibinfo{pages}{1467}.
%Type = Article
\bibitem[{Marris(2023)}]{marris2023hawaii}
\bibinfo{author}{Marris, E.}, \bibinfo{year}{2023}.
\newblock \bibinfo{title}{Hawaii wildfires: did scientists expect maui to burn?}
\newblock \bibinfo{journal}{Nature} .
%Type = Article
\bibitem[{Martell et~al.(1989)Martell, Bevilacqua and Stocks}]{martell_modelling_1989}
\bibinfo{author}{Martell, D.L.}, \bibinfo{author}{Bevilacqua, E.}, \bibinfo{author}{Stocks, B.J.}, \bibinfo{year}{1989}.
\newblock \bibinfo{title}{Modelling seasonal variation in daily people-caused forest fire occurrence}.
\newblock \bibinfo{journal}{Canadian Journal of Forest Research} \bibinfo{volume}{19}, \bibinfo{pages}{1555--1563}.
\newblock \URLprefix \url{https://cdnsciencepub.com/doi/10.1139/x89-237}, \DOIprefix\doi{10.1139/x89-237}. \bibinfo{note}{publisher: NRC Research Press}.
%Type = Article
\bibitem[{Martínez et~al.(2009)Martínez, Vega-Garcia and Chuvieco}]{martinez_human-caused_2009}
\bibinfo{author}{Martínez, J.}, \bibinfo{author}{Vega-Garcia, C.}, \bibinfo{author}{Chuvieco, E.}, \bibinfo{year}{2009}.
\newblock \bibinfo{title}{Human-caused wildfire risk rating for prevention planning in {Spain}}.
\newblock \bibinfo{journal}{Journal of Environmental Management} \bibinfo{volume}{90}, \bibinfo{pages}{1241--1252}.
\newblock \URLprefix \url{https://www.sciencedirect.com/science/article/pii/S0301479708001758}, \DOIprefix\doi{10.1016/j.jenvman.2008.07.005}.
%Type = Incollection
\bibitem[{Masrur and Yu(2023)}]{MASRUR2023119}
\bibinfo{author}{Masrur, A.}, \bibinfo{author}{Yu, M.}, \bibinfo{year}{2023}.
\newblock \bibinfo{title}{Chapter 6 - spatiotemporal attention convlstm networks for predicting and physically interpreting wildfire spread}, in: \bibinfo{editor}{Sun, Z.}, \bibinfo{editor}{Cristea, N.}, \bibinfo{editor}{Rivas, P.} (Eds.), \bibinfo{booktitle}{Artificial Intelligence in Earth Science}. \bibinfo{publisher}{Elsevier}, pp. \bibinfo{pages}{119--156}.
\newblock \URLprefix \url{https://www.sciencedirect.com/science/article/pii/B9780323917377000098}, \DOIprefix\doi{https://doi.org/10.1016/B978-0-323-91737-7.00009-8}.
%Type = Article
\bibitem[{Masrur et~al.(2024)Masrur, Yu and Taylor}]{masrur2024capturing}
\bibinfo{author}{Masrur, A.}, \bibinfo{author}{Yu, M.}, \bibinfo{author}{Taylor, A.}, \bibinfo{year}{2024}.
\newblock \bibinfo{title}{Capturing and interpreting wildfire spread dynamics: attention-based spatiotemporal models using convlstm networks}.
\newblock \bibinfo{journal}{Ecological Informatics} \bibinfo{volume}{82}, \bibinfo{pages}{102760}.
%Type = Article
\bibitem[{Matthews(1997)}]{matthews1997global}
\bibinfo{author}{Matthews, E.}, \bibinfo{year}{1997}.
\newblock \bibinfo{title}{Global litter production, pools, and turnover times: Estimates from measurement data and regression models}.
\newblock \bibinfo{journal}{Journal of Geophysical Research: Atmospheres} \bibinfo{volume}{102}, \bibinfo{pages}{18771--18800}.
%Type = Article
\bibitem[{Matthews(2013)}]{matthews2013dead}
\bibinfo{author}{Matthews, S.}, \bibinfo{year}{2013}.
\newblock \bibinfo{title}{Dead fuel moisture research: 1991--2012}.
\newblock \bibinfo{journal}{International journal of wildland fire} \bibinfo{volume}{23}, \bibinfo{pages}{78--92}.
%Type = Article
\bibitem[{Mbow et~al.(2004)Mbow, Go\i{}\"ta and B\'{e}ni\'{e}}]{mbow_spectral_2004}
\bibinfo{author}{Mbow, C.}, \bibinfo{author}{Go\i{}\"ta, K.}, \bibinfo{author}{B\'{e}ni\'{e}, G.B.}, \bibinfo{year}{2004}.
\newblock \bibinfo{title}{Spectral indices and fire behavior simulation for fire risk assessment in savanna ecosystems}.
\newblock \bibinfo{journal}{Remote Sensing of Environment} \bibinfo{volume}{91}, \bibinfo{pages}{1--13}.
\newblock \DOIprefix\doi{10.1016/j.rse.2003.10.019}.
%Type = Techreport
\bibitem[{McArthur(1958)}]{mcarthur1958the}
\bibinfo{author}{McArthur, A.G.}, \bibinfo{year}{1958}.
\newblock \bibinfo{title}{The preparation and use of fire danger tables.}
\newblock \bibinfo{type}{Technical Report}. Forestry and Timber Bureau: Canberra, Australia.
%Type = Techreport
\bibitem[{McArthur(1966)}]{mcarthur1966weather}
\bibinfo{author}{McArthur, A.G.}, \bibinfo{year}{1966}.
\newblock \bibinfo{title}{Weather and grassland fire behaviour.}
\newblock \bibinfo{type}{Technical Report}. Department of National Development, Forestry and Timber Bureau: Canberra, Australia.
%Type = Article
\bibitem[{McCandless et~al.(2020)McCandless, Kosovic and Petzke}]{mccandless2020enhancing}
\bibinfo{author}{McCandless, T.C.}, \bibinfo{author}{Kosovic, B.}, \bibinfo{author}{Petzke, W.}, \bibinfo{year}{2020}.
\newblock \bibinfo{title}{Enhancing wildfire spread modelling by building a gridded fuel moisture content product with machine learning}.
\newblock \bibinfo{journal}{Machine Learning: Science and Technology} \bibinfo{volume}{1}, \bibinfo{pages}{035010}.
%Type = Article
\bibitem[{McLauchlan et~al.(2020)McLauchlan, Higuera, Miesel, Rogers, Schweitzer, Shuman, Tepley, Varner, Veblen, Adalsteinsson et~al.}]{mclauchlan2020fire}
\bibinfo{author}{McLauchlan, K.K.}, \bibinfo{author}{Higuera, P.E.}, \bibinfo{author}{Miesel, J.}, \bibinfo{author}{Rogers, B.M.}, \bibinfo{author}{Schweitzer, J.}, \bibinfo{author}{Shuman, J.K.}, \bibinfo{author}{Tepley, A.J.}, \bibinfo{author}{Varner, J.M.}, \bibinfo{author}{Veblen, T.T.}, \bibinfo{author}{Adalsteinsson, S.A.}, et~al., \bibinfo{year}{2020}.
\newblock \bibinfo{title}{Fire as a fundamental ecological process: Research advances and frontiers}.
\newblock \bibinfo{journal}{Journal of Ecology} \bibinfo{volume}{108}, \bibinfo{pages}{2047--2069}.
%Type = Article
\bibitem[{McNorton and Di~Giuseppe(2024)}]{mcnorton_global_2024}
\bibinfo{author}{McNorton, J.R.}, \bibinfo{author}{Di~Giuseppe, F.}, \bibinfo{year}{2024}.
\newblock \bibinfo{title}{A global fuel characteristic model and dataset for wildfire prediction}.
\newblock \bibinfo{journal}{Biogeosciences} \bibinfo{volume}{21}, \bibinfo{pages}{279--300}.
\newblock \DOIprefix\doi{10.5194/bg-21-279-2024}. \bibinfo{note}{publisher: Copernicus GmbH}.
%Type = Article
\bibitem[{Meikle and Heine(1987)}]{meikle1987fire}
\bibinfo{author}{Meikle, S.}, \bibinfo{author}{Heine, J.}, \bibinfo{year}{1987}.
\newblock \bibinfo{title}{A fire danger index system for the transvaal lowveld and adjoining escarpment areas}.
\newblock \bibinfo{journal}{South African Forestry Journal} \bibinfo{volume}{143}, \bibinfo{pages}{55--56}.
%Type = Article
\bibitem[{Mell et~al.(2007)Mell, Jenkins, Gould and Cheney}]{mell_physics-based_2007}
\bibinfo{author}{Mell, W.}, \bibinfo{author}{Jenkins, M.A.}, \bibinfo{author}{Gould, J.}, \bibinfo{author}{Cheney, P.}, \bibinfo{year}{2007}.
\newblock \bibinfo{title}{A physics-based approach to modelling grassland fires}.
\newblock \bibinfo{journal}{International Journal of Wildland Fire} \bibinfo{volume}{16}, \bibinfo{pages}{1--22}.
\newblock \URLprefix \url{https://www.publish.csiro.au/wf/WF06002}, \DOIprefix\doi{10.1071/WF06002}. \bibinfo{note}{publisher: CSIRO PUBLISHING}.
%Type = Article
\bibitem[{Menning and Stephens(2007)}]{menning2007fire}
\bibinfo{author}{Menning, K.M.}, \bibinfo{author}{Stephens, S.L.}, \bibinfo{year}{2007}.
\newblock \bibinfo{title}{Fire climbing in the forest: a semiqualitative, semiquantitative approach to assessing ladder fuel hazards}.
\newblock \bibinfo{journal}{Western Journal of Applied Forestry} \bibinfo{volume}{22}, \bibinfo{pages}{88--93}.
%Type = Inproceedings
\bibitem[{Meyer(2019)}]{Meyer2019CHAPTER2S}
\bibinfo{author}{Meyer, F.J.}, \bibinfo{year}{2019}.
\newblock \bibinfo{title}{Chapter 2 spaceborne synthetic aperture radar : Principles , data access , and basic processing techniques}.
\newblock \URLprefix \url{https://api.semanticscholar.org/CorpusID:201649124}.
%Type = Article
\bibitem[{Miao et~al.(2023)Miao, Li, Mu, He, Ma, Chen, Wei and Gao}]{miao_time_2023}
\bibinfo{author}{Miao, X.}, \bibinfo{author}{Li, J.}, \bibinfo{author}{Mu, Y.}, \bibinfo{author}{He, C.}, \bibinfo{author}{Ma, Y.}, \bibinfo{author}{Chen, J.}, \bibinfo{author}{Wei, W.}, \bibinfo{author}{Gao, D.}, \bibinfo{year}{2023}.
\newblock \bibinfo{title}{Time {Series} {Forest} {Fire} {Prediction} {Based} on {Improved} {Transformer}}.
\newblock \bibinfo{journal}{Forests} \bibinfo{volume}{14}, \bibinfo{pages}{1596}.
\newblock \URLprefix \url{https://www.mdpi.com/1999-4907/14/8/1596}, \DOIprefix\doi{10.3390/f14081596}. \bibinfo{note}{number: 8 Publisher: Multidisciplinary Digital Publishing Institute}.
%Type = Misc
\bibitem[{Michail et~al.(2024)Michail, Panagiotou, Davalas, Prapas, Kondylatos, Bountos and Papoutsis}]{michail_seasonal_2024}
\bibinfo{author}{Michail, D.}, \bibinfo{author}{Panagiotou, L.I.}, \bibinfo{author}{Davalas, C.}, \bibinfo{author}{Prapas, I.}, \bibinfo{author}{Kondylatos, S.}, \bibinfo{author}{Bountos, N.I.}, \bibinfo{author}{Papoutsis, I.}, \bibinfo{year}{2024}.
\newblock \bibinfo{title}{Seasonal {Fire} {Prediction} using {Spatio}-{Temporal} {Deep} {Neural} {Networks}}.
\newblock \URLprefix \url{http://arxiv.org/abs/2404.06437}. \bibinfo{note}{arXiv:2404.06437 [cs]}.
%Type = Article
\bibitem[{Millard et~al.(2018)Millard, Thompson, Parisien and Richardson}]{millard2018soil}
\bibinfo{author}{Millard, K.}, \bibinfo{author}{Thompson, D.K.}, \bibinfo{author}{Parisien, M.A.}, \bibinfo{author}{Richardson, M.}, \bibinfo{year}{2018}.
\newblock \bibinfo{title}{Soil moisture monitoring in a temperate peatland using multi-sensor remote sensing and linear mixed effects}.
\newblock \bibinfo{journal}{Remote Sensing} \bibinfo{volume}{10}, \bibinfo{pages}{903}.
%Type = Article
\bibitem[{Miller and Ager(2012)}]{miller2012review}
\bibinfo{author}{Miller, C.}, \bibinfo{author}{Ager, A.A.}, \bibinfo{year}{2012}.
\newblock \bibinfo{title}{A review of recent advances in risk analysis for wildfire management}.
\newblock \bibinfo{journal}{International journal of wildland fire} \bibinfo{volume}{22}, \bibinfo{pages}{1--14}.
%Type = Article
\bibitem[{Miller et~al.(2008)Miller, Parisien, Ager and Finney}]{miller2008evaluating}
\bibinfo{author}{Miller, C.}, \bibinfo{author}{Parisien, M.A.}, \bibinfo{author}{Ager, A.}, \bibinfo{author}{Finney, M.}, \bibinfo{year}{2008}.
\newblock \bibinfo{title}{Evaluating spatially-explicit burn probabilities for strategic fire management planning}.
\newblock \bibinfo{journal}{WIT Transactions on Ecology and the Environment} \bibinfo{volume}{119}, \bibinfo{pages}{245--252}.
%Type = Article
\bibitem[{Miller et~al.(2023)Miller, Zhu, Yebra, Rüdiger and Webb}]{miller_projecting_2023}
\bibinfo{author}{Miller, L.}, \bibinfo{author}{Zhu, L.}, \bibinfo{author}{Yebra, M.}, \bibinfo{author}{Rüdiger, C.}, \bibinfo{author}{Webb, G.I.}, \bibinfo{year}{2023}.
\newblock \bibinfo{title}{Projecting live fuel moisture content via deep learning}.
\newblock \bibinfo{journal}{International Journal of Wildland Fire} \bibinfo{volume}{32}, \bibinfo{pages}{709--727}.
\newblock \URLprefix \url{https://www.publish.csiro.au/wf/WF22188}, \DOIprefix\doi{10.1071/WF22188}. \bibinfo{note}{publisher: CSIRO PUBLISHING}.
%Type = Article
\bibitem[{Moayedi and Khasmakhi(2023)}]{moayedi2023wildfire}
\bibinfo{author}{Moayedi, H.}, \bibinfo{author}{Khasmakhi, M.A.S.A.}, \bibinfo{year}{2023}.
\newblock \bibinfo{title}{Wildfire susceptibility mapping using two empowered machine learning algorithms}.
\newblock \bibinfo{journal}{Stochastic Environmental Research and Risk Assessment} \bibinfo{volume}{37}, \bibinfo{pages}{49--72}.
%Type = Inproceedings
\bibitem[{Molnar et~al.(2020)Molnar, K{\"o}nig, Herbinger, Freiesleben, Dandl, Scholbeck, Casalicchio, Grosse-Wentrup and Bischl}]{molnar2020general}
\bibinfo{author}{Molnar, C.}, \bibinfo{author}{K{\"o}nig, G.}, \bibinfo{author}{Herbinger, J.}, \bibinfo{author}{Freiesleben, T.}, \bibinfo{author}{Dandl, S.}, \bibinfo{author}{Scholbeck, C.A.}, \bibinfo{author}{Casalicchio, G.}, \bibinfo{author}{Grosse-Wentrup, M.}, \bibinfo{author}{Bischl, B.}, \bibinfo{year}{2020}.
\newblock \bibinfo{title}{General pitfalls of model-agnostic interpretation methods for machine learning models}, in: \bibinfo{booktitle}{International Workshop on Extending Explainable AI Beyond Deep Models and Classifiers}, \bibinfo{organization}{Springer}. pp. \bibinfo{pages}{39--68}.
%Type = Article
\bibitem[{Moritz et~al.(2014)Moritz, Batllori, Bradstock, Gill, Handmer, Hessburg, Leonard, McCaffrey, Odion, Schoennagel et~al.}]{moritz2014learning}
\bibinfo{author}{Moritz, M.A.}, \bibinfo{author}{Batllori, E.}, \bibinfo{author}{Bradstock, R.A.}, \bibinfo{author}{Gill, A.M.}, \bibinfo{author}{Handmer, J.}, \bibinfo{author}{Hessburg, P.F.}, \bibinfo{author}{Leonard, J.}, \bibinfo{author}{McCaffrey, S.}, \bibinfo{author}{Odion, D.C.}, \bibinfo{author}{Schoennagel, T.}, et~al., \bibinfo{year}{2014}.
\newblock \bibinfo{title}{Learning to coexist with wildfire}.
\newblock \bibinfo{journal}{Nature} \bibinfo{volume}{515}, \bibinfo{pages}{58--66}.
%Type = Article
\bibitem[{Mou et~al.(2020)Mou, Hua, Jin and Zhu}]{mou2020era}
\bibinfo{author}{Mou, L.}, \bibinfo{author}{Hua, Y.}, \bibinfo{author}{Jin, P.}, \bibinfo{author}{Zhu, X.X.}, \bibinfo{year}{2020}.
\newblock \bibinfo{title}{Era: A data set and deep learning benchmark for event recognition in aerial videos [software and data sets]}.
\newblock \bibinfo{journal}{IEEE Geoscience and Remote Sensing Magazine} \bibinfo{volume}{8}, \bibinfo{pages}{125--133}.
%Type = Article
\bibitem[{Moumgiakmas et~al.(2021)Moumgiakmas, Samatas and Papakostas}]{moumgiakmas2021computer}
\bibinfo{author}{Moumgiakmas, S.S.}, \bibinfo{author}{Samatas, G.G.}, \bibinfo{author}{Papakostas, G.A.}, \bibinfo{year}{2021}.
\newblock \bibinfo{title}{Computer vision for fire detection on uavs—from software to hardware}.
\newblock \bibinfo{journal}{Future Internet} \bibinfo{volume}{13}, \bibinfo{pages}{200}.
%Type = Article
\bibitem[{Mulverhill et~al.(2024)Mulverhill, Coops, Boulanger, Hoffman, Cardinal~Christianson, Daniels, Flamand-Hubert, Wotherspoon and Achim}]{mulverhill2024wildfires}
\bibinfo{author}{Mulverhill, C.}, \bibinfo{author}{Coops, N.C.}, \bibinfo{author}{Boulanger, Y.}, \bibinfo{author}{Hoffman, K.M.}, \bibinfo{author}{Cardinal~Christianson, A.}, \bibinfo{author}{Daniels, L.D.}, \bibinfo{author}{Flamand-Hubert, M.}, \bibinfo{author}{Wotherspoon, A.R.}, \bibinfo{author}{Achim, A.}, \bibinfo{year}{2024}.
\newblock \bibinfo{title}{Wildfires are spreading fast in canada—we must strengthen forests for the future}.
\newblock \bibinfo{journal}{Nature} \bibinfo{volume}{633}, \bibinfo{pages}{282--285}.
%Type = Article
\bibitem[{Myoung et~al.(2018)Myoung, Kim, Nghiem, Jia, Whitney and Kafatos}]{myoung2018estimating}
\bibinfo{author}{Myoung, B.}, \bibinfo{author}{Kim, S.H.}, \bibinfo{author}{Nghiem, S.V.}, \bibinfo{author}{Jia, S.}, \bibinfo{author}{Whitney, K.}, \bibinfo{author}{Kafatos, M.C.}, \bibinfo{year}{2018}.
\newblock \bibinfo{title}{Estimating live fuel moisture from modis satellite data for wildfire danger assessment in southern california usa}.
\newblock \bibinfo{journal}{Remote Sensing} \bibinfo{volume}{10}, \bibinfo{pages}{87}.
%Type = Article
\bibitem[{Myroniuk et~al.(2023)Myroniuk, Zibtsev, Bogomolov, Goldammer, Soshenskyi, Levchenko and Matsala}]{myroniuk2023combining}
\bibinfo{author}{Myroniuk, V.}, \bibinfo{author}{Zibtsev, S.}, \bibinfo{author}{Bogomolov, V.}, \bibinfo{author}{Goldammer, J.G.}, \bibinfo{author}{Soshenskyi, O.}, \bibinfo{author}{Levchenko, V.}, \bibinfo{author}{Matsala, M.}, \bibinfo{year}{2023}.
\newblock \bibinfo{title}{Combining landsat time series and gedi data for improved characterization of fuel types and canopy metrics in wildfire simulation}.
\newblock \bibinfo{journal}{Journal of Environmental Management} \bibinfo{volume}{345}, \bibinfo{pages}{118736}.
%Type = Inproceedings
\bibitem[{Naderpour et~al.(2020)Naderpour, Rizeei and Ramezani}]{naderpour_wildfire_2020}
\bibinfo{author}{Naderpour, M.}, \bibinfo{author}{Rizeei, H.M.}, \bibinfo{author}{Ramezani, F.}, \bibinfo{year}{2020}.
\newblock \bibinfo{title}{Wildfire {Prediction}: {Handling} {Uncertainties} {Using} {Integrated} {Bayesian} {Networks} and {Fuzzy} {Logic}}, in: \bibinfo{booktitle}{2020 {IEEE} {International} {Conference} on {Fuzzy} {Systems} ({FUZZ}-{IEEE})}, pp. \bibinfo{pages}{1--7}.
\newblock \URLprefix \url{https://ieeexplore.ieee.org/document/9177700}, \DOIprefix\doi{10.1109/FUZZ48607.2020.9177700}. \bibinfo{note}{iSSN: 1558-4739}.
%Type = Article
\bibitem[{Nami et~al.(2018)Nami, Jaafari, Fallah and Nabiuni}]{nami_spatial_2018}
\bibinfo{author}{Nami, M.H.}, \bibinfo{author}{Jaafari, A.}, \bibinfo{author}{Fallah, M.}, \bibinfo{author}{Nabiuni, S.}, \bibinfo{year}{2018}.
\newblock \bibinfo{title}{Spatial prediction of wildfire probability in the {Hyrcanian} ecoregion using evidential belief function model and {GIS}}.
\newblock \bibinfo{journal}{International Journal of Environmental Science and Technology} \bibinfo{volume}{15}, \bibinfo{pages}{373--384}.
\newblock \URLprefix \url{https://doi.org/10.1007/s13762-017-1371-6}, \DOIprefix\doi{10.1007/s13762-017-1371-6}.
%Type = Article
\bibitem[{Nasir et~al.(2023)Nasir, Kansal, Barneih, Al-Shaltone, Bonny, Al-Shabi and Al~Shammaa}]{nasir2023multi}
\bibinfo{author}{Nasir, N.}, \bibinfo{author}{Kansal, A.}, \bibinfo{author}{Barneih, F.}, \bibinfo{author}{Al-Shaltone, O.}, \bibinfo{author}{Bonny, T.}, \bibinfo{author}{Al-Shabi, M.}, \bibinfo{author}{Al~Shammaa, A.}, \bibinfo{year}{2023}.
\newblock \bibinfo{title}{Multi-modal image classification of covid-19 cases using computed tomography and x-rays scans}.
\newblock \bibinfo{journal}{Intelligent Systems with Applications} \bibinfo{volume}{17}, \bibinfo{pages}{200160}.
%Type = Inproceedings
\bibitem[{Natekar et~al.(2021)Natekar, Patil, Nair and Roychowdhury}]{9456113}
\bibinfo{author}{Natekar, S.}, \bibinfo{author}{Patil, S.}, \bibinfo{author}{Nair, A.}, \bibinfo{author}{Roychowdhury, S.}, \bibinfo{year}{2021}.
\newblock \bibinfo{title}{Forest fire prediction using lstm}, in: \bibinfo{booktitle}{2021 2nd International Conference for Emerging Technology (INCET)}, pp. \bibinfo{pages}{1--5}.
\newblock \DOIprefix\doi{10.1109/INCET51464.2021.9456113}.
%Type = Article
\bibitem[{Nelson~Jr(2000)}]{nelson_jr_prediction_2000}
\bibinfo{author}{Nelson~Jr, R.M.}, \bibinfo{year}{2000}.
\newblock \bibinfo{title}{Prediction of diurnal change in 10-h fuel stick moisture content}.
\newblock \bibinfo{journal}{Canadian Journal of Forest Research} \bibinfo{volume}{30}, \bibinfo{pages}{1071--1087}.
\newblock \URLprefix \url{https://cdnsciencepub.com/doi/abs/10.1139/x00-032}, \DOIprefix\doi{10.1139/x00-032}. \bibinfo{note}{publisher: NRC Research Press}.
%Type = Article
\bibitem[{Nieto et~al.(2010)Nieto, Aguado, Chuvieco and Sandholt}]{nieto2010dead}
\bibinfo{author}{Nieto, H.}, \bibinfo{author}{Aguado, I.}, \bibinfo{author}{Chuvieco, E.}, \bibinfo{author}{Sandholt, I.}, \bibinfo{year}{2010}.
\newblock \bibinfo{title}{Dead fuel moisture estimation with msg--seviri data. retrieval of meteorological data for the calculation of the equilibrium moisture content}.
\newblock \bibinfo{journal}{Agricultural and Forest Meteorology} \bibinfo{volume}{150}, \bibinfo{pages}{861--870}.
%Type = Article
\bibitem[{Nolan et~al.(2016)Nolan, de~Dios, Boer, Caccamo, Goulden and Bradstock}]{nolan2016predicting}
\bibinfo{author}{Nolan, R.H.}, \bibinfo{author}{de~Dios, V.R.}, \bibinfo{author}{Boer, M.M.}, \bibinfo{author}{Caccamo, G.}, \bibinfo{author}{Goulden, M.L.}, \bibinfo{author}{Bradstock, R.A.}, \bibinfo{year}{2016}.
\newblock \bibinfo{title}{Predicting dead fine fuel moisture at regional scales using vapour pressure deficit from modis and gridded weather data}.
\newblock \bibinfo{journal}{Remote Sensing of Environment} \bibinfo{volume}{174}, \bibinfo{pages}{100--108}.
%Type = Article
\bibitem[{Nolan et~al.(2022)Nolan, Price, Samson, Jenkins, Rahmani and Boer}]{nolan2022framework}
\bibinfo{author}{Nolan, R.H.}, \bibinfo{author}{Price, O.F.}, \bibinfo{author}{Samson, S.A.}, \bibinfo{author}{Jenkins, M.E.}, \bibinfo{author}{Rahmani, S.}, \bibinfo{author}{Boer, M.M.}, \bibinfo{year}{2022}.
\newblock \bibinfo{title}{Framework for assessing live fine fuel loads and biomass consumption during fire}.
\newblock \bibinfo{journal}{Forest Ecology and Management} \bibinfo{volume}{504}, \bibinfo{pages}{119830}.
%Type = Article
\bibitem[{Nur et~al.(2022)Nur, Kim and Lee}]{nur2022creation}
\bibinfo{author}{Nur, A.S.}, \bibinfo{author}{Kim, Y.J.}, \bibinfo{author}{Lee, C.W.}, \bibinfo{year}{2022}.
\newblock \bibinfo{title}{Creation of wildfire susceptibility maps in plumas national forest using insar coherence, deep learning, and metaheuristic optimization approaches}.
\newblock \bibinfo{journal}{Remote Sensing} \bibinfo{volume}{14}, \bibinfo{pages}{4416}.
%Type = Article
\bibitem[{Nur et~al.(2023)Nur, Kim, Lee and Lee}]{nur_spatial_2023}
\bibinfo{author}{Nur, A.S.}, \bibinfo{author}{Kim, Y.J.}, \bibinfo{author}{Lee, J.H.}, \bibinfo{author}{Lee, C.W.}, \bibinfo{year}{2023}.
\newblock \bibinfo{title}{Spatial {Prediction} of {Wildfire} {Susceptibility} {Using} {Hybrid} {Machine} {Learning} {Models} {Based} on {Support} {Vector} {Regression} in {Sydney}, {Australia}}.
\newblock \bibinfo{journal}{Remote Sensing} \bibinfo{volume}{15}, \bibinfo{pages}{760}.
\newblock \URLprefix \url{https://www.mdpi.com/2072-4292/15/3/760}, \DOIprefix\doi{10.3390/rs15030760}. \bibinfo{note}{number: 3 Publisher: Multidisciplinary Digital Publishing Institute}.
%Type = Inproceedings
\bibitem[{Oak et~al.(2024)Oak, Nazre, Naigaonkar, Sawant and Joshi}]{oak2024novel}
\bibinfo{author}{Oak, O.}, \bibinfo{author}{Nazre, R.}, \bibinfo{author}{Naigaonkar, S.}, \bibinfo{author}{Sawant, S.}, \bibinfo{author}{Joshi, A.}, \bibinfo{year}{2024}.
\newblock \bibinfo{title}{A novel transfer learning based cnn model for wildfire susceptibility prediction}, in: \bibinfo{booktitle}{2024 5th International Conference for Emerging Technology (INCET)}, \bibinfo{organization}{IEEE}. pp. \bibinfo{pages}{1--6}.
%Type = Article
\bibitem[{Oliveira et~al.(2021a)Oliveira, Gon{\c{c}}alves and Z{\^e}zere}]{oliveira2021reassessing}
\bibinfo{author}{Oliveira, S.}, \bibinfo{author}{Gon{\c{c}}alves, A.}, \bibinfo{author}{Z{\^e}zere, J.L.}, \bibinfo{year}{2021}a.
\newblock \bibinfo{title}{Reassessing wildfire susceptibility and hazard for mainland portugal}.
\newblock \bibinfo{journal}{Science of the total environment} \bibinfo{volume}{762}, \bibinfo{pages}{143121}.
%Type = Article
\bibitem[{Oliveira et~al.(2021b)Oliveira, Rocha and Sá}]{oliveira_wildfire_2021}
\bibinfo{author}{Oliveira, S.}, \bibinfo{author}{Rocha, J.}, \bibinfo{author}{Sá, A.}, \bibinfo{year}{2021}b.
\newblock \bibinfo{title}{Wildfire risk modeling}.
\newblock \bibinfo{journal}{Current Opinion in Environmental Science \& Health} \bibinfo{volume}{23}, \bibinfo{pages}{100274}.
\newblock \URLprefix \url{https://www.sciencedirect.com/science/article/pii/S2468584421000465}, \DOIprefix\doi{10.1016/j.coesh.2021.100274}.
%Type = Article
\bibitem[{de~Oliveira-J{\'u}nior et~al.(2024)de~Oliveira-J{\'u}nior, Mendes, Szabo, Singh, Jamjareegulgarn, Cardoso, Bertalan, da~Silva, da~Rosa Ferraz~Jardim, da~Silva et~al.}]{de2024impact}
\bibinfo{author}{de~Oliveira-J{\'u}nior, J.F.}, \bibinfo{author}{Mendes, D.}, \bibinfo{author}{Szabo, S.}, \bibinfo{author}{Singh, S.K.}, \bibinfo{author}{Jamjareegulgarn, P.}, \bibinfo{author}{Cardoso, K.R.A.}, \bibinfo{author}{Bertalan, L.}, \bibinfo{author}{da~Silva, M.V.}, \bibinfo{author}{da~Rosa Ferraz~Jardim, A.M.}, \bibinfo{author}{da~Silva, J.L.B.}, et~al., \bibinfo{year}{2024}.
\newblock \bibinfo{title}{Impact of the el ni{\~n}o on fire dynamics on the african continent}.
\newblock \bibinfo{journal}{Earth Systems and Environment} \bibinfo{volume}{8}, \bibinfo{pages}{45--61}.
%Type = Misc
\bibitem[{Oquab et~al.(2024)Oquab, Darcet, Moutakanni, Vo, Szafraniec, Khalidov, Fernandez, Haziza, Massa, El-Nouby, Assran, Ballas, Galuba, Howes, Huang, Li, Misra, Rabbat, Sharma, Synnaeve, Xu, Jegou, Mairal, Labatut, Joulin and Bojanowski}]{oquab2024dinov2learningrobustvisual}
\bibinfo{author}{Oquab, M.}, \bibinfo{author}{Darcet, T.}, \bibinfo{author}{Moutakanni, T.}, \bibinfo{author}{Vo, H.}, \bibinfo{author}{Szafraniec, M.}, \bibinfo{author}{Khalidov, V.}, \bibinfo{author}{Fernandez, P.}, \bibinfo{author}{Haziza, D.}, \bibinfo{author}{Massa, F.}, \bibinfo{author}{El-Nouby, A.}, \bibinfo{author}{Assran, M.}, \bibinfo{author}{Ballas, N.}, \bibinfo{author}{Galuba, W.}, \bibinfo{author}{Howes, R.}, \bibinfo{author}{Huang, P.Y.}, \bibinfo{author}{Li, S.W.}, \bibinfo{author}{Misra, I.}, \bibinfo{author}{Rabbat, M.}, \bibinfo{author}{Sharma, V.}, \bibinfo{author}{Synnaeve, G.}, \bibinfo{author}{Xu, H.}, \bibinfo{author}{Jegou, H.}, \bibinfo{author}{Mairal, J.}, \bibinfo{author}{Labatut, P.}, \bibinfo{author}{Joulin, A.}, \bibinfo{author}{Bojanowski, P.}, \bibinfo{year}{2024}.
\newblock \bibinfo{title}{Dinov2: Learning robust visual features without supervision}.
\newblock \URLprefix \url{https://arxiv.org/abs/2304.07193}, \href{http://arxiv.org/abs/2304.07193}{{\tt arXiv:2304.07193}}.
%Type = Article
\bibitem[{Ottmar et~al.(2007)Ottmar, Sandberg, Riccardi and Prichard}]{ottmar2007overview}
\bibinfo{author}{Ottmar, R.D.}, \bibinfo{author}{Sandberg, D.V.}, \bibinfo{author}{Riccardi, C.L.}, \bibinfo{author}{Prichard, S.J.}, \bibinfo{year}{2007}.
\newblock \bibinfo{title}{An overview of the fuel characteristic classification system—quantifying, classifying, and creating fuelbeds for resource planning}.
\newblock \bibinfo{journal}{Canadian Journal of Forest Research} \bibinfo{volume}{37}, \bibinfo{pages}{2383--2393}.
%Type = Article
\bibitem[{Parks and Abatzoglou(2020)}]{parks_warmer_2020}
\bibinfo{author}{Parks, S.A.}, \bibinfo{author}{Abatzoglou, J.T.}, \bibinfo{year}{2020}.
\newblock \bibinfo{title}{Warmer and {Drier} {Fire} {Seasons} {Contribute} to {Increases} in {Area} {Burned} at {High} {Severity} in {Western} {US} {Forests} {From} 1985 to 2017}.
\newblock \bibinfo{journal}{Geophysical Research Letters} \bibinfo{volume}{47}, \bibinfo{pages}{e2020GL089858}.
\newblock \URLprefix \url{https://onlinelibrary.wiley.com/doi/abs/10.1029/2020GL089858}, \DOIprefix\doi{10.1029/2020GL089858}. \bibinfo{note}{\_eprint: https://onlinelibrary.wiley.com/doi/pdf/10.1029/2020GL089858}.
%Type = Article
\bibitem[{Pausas and Keeley(2019)}]{pausas2019wildfires}
\bibinfo{author}{Pausas, J.G.}, \bibinfo{author}{Keeley, J.E.}, \bibinfo{year}{2019}.
\newblock \bibinfo{title}{Wildfires as an ecosystem service}.
\newblock \bibinfo{journal}{Frontiers in Ecology and the Environment} \bibinfo{volume}{17}, \bibinfo{pages}{289--295}.
%Type = Article
\bibitem[{Pelletier et~al.(2023)Pelletier, Millard and Darling}]{pelletier_wildfire_2023}
\bibinfo{author}{Pelletier, N.}, \bibinfo{author}{Millard, K.}, \bibinfo{author}{Darling, S.}, \bibinfo{year}{2023}.
\newblock \bibinfo{title}{Wildfire likelihood in {Canadian} treed peatlands based on remote-sensing time-series of surface conditions}.
\newblock \bibinfo{journal}{Remote Sensing of Environment} \bibinfo{volume}{296}, \bibinfo{pages}{113747}.
\newblock \URLprefix \url{https://www.sciencedirect.com/science/article/pii/S0034425723002985}, \DOIprefix\doi{10.1016/j.rse.2023.113747}.
%Type = Article
\bibitem[{Pellizzaro et~al.(2007)Pellizzaro, Duce, Ventura and Zara}]{pellizzaro2007seasonal}
\bibinfo{author}{Pellizzaro, G.}, \bibinfo{author}{Duce, P.}, \bibinfo{author}{Ventura, A.}, \bibinfo{author}{Zara, P.}, \bibinfo{year}{2007}.
\newblock \bibinfo{title}{Seasonal variations of live moisture content and ignitability in shrubs of the mediterranean basin}.
\newblock \bibinfo{journal}{International Journal of Wildland Fire} \bibinfo{volume}{16}, \bibinfo{pages}{633--641}.
%Type = Article
\bibitem[{Peterson et~al.(2008)Peterson, Roberts and Dennison}]{peterson2008mapping}
\bibinfo{author}{Peterson, S.H.}, \bibinfo{author}{Roberts, D.A.}, \bibinfo{author}{Dennison, P.E.}, \bibinfo{year}{2008}.
\newblock \bibinfo{title}{Mapping live fuel moisture with modis data: A multiple regression approach}.
\newblock \bibinfo{journal}{Remote Sensing of Environment} \bibinfo{volume}{112}, \bibinfo{pages}{4272--4284}.
%Type = Article
\bibitem[{Phelps and Woolford(2021a)}]{phelps_comparing_2021}
\bibinfo{author}{Phelps, N.}, \bibinfo{author}{Woolford, D.G.}, \bibinfo{year}{2021}a.
\newblock \bibinfo{title}{Comparing calibrated statistical and machine learning methods for wildland fire occurrence prediction: a case study of human-caused fires in {Lac} {La} {Biche}, {Alberta}, {Canada}}.
\newblock \bibinfo{journal}{International Journal of Wildland Fire} \bibinfo{volume}{30}, \bibinfo{pages}{850--870}.
\newblock \URLprefix \url{https://www.publish.csiro.au/wf/WF20139}, \DOIprefix\doi{10.1071/WF20139}. \bibinfo{note}{publisher: CSIRO PUBLISHING}.
%Type = Article
\bibitem[{Phelps and Woolford(2021b)}]{phelps2021guidelines}
\bibinfo{author}{Phelps, N.}, \bibinfo{author}{Woolford, D.G.}, \bibinfo{year}{2021}b.
\newblock \bibinfo{title}{Guidelines for effective evaluation and comparison of wildland fire occurrence prediction models}.
\newblock \bibinfo{journal}{International journal of wildland fire} \bibinfo{volume}{30}, \bibinfo{pages}{225--240}.
%Type = Article
\bibitem[{Pisek et~al.(2015)Pisek, Rautiainen, Nikopensius and Raabe}]{pisek_estimation_2015}
\bibinfo{author}{Pisek, J.}, \bibinfo{author}{Rautiainen, M.}, \bibinfo{author}{Nikopensius, M.}, \bibinfo{author}{Raabe, K.}, \bibinfo{year}{2015}.
\newblock \bibinfo{title}{Estimation of seasonal dynamics of understory {NDVI} in northern forests using {MODIS} {BRDF} data: {Semi}-empirical versus physically-based approach}.
\newblock \bibinfo{journal}{Remote Sensing of Environment} \bibinfo{volume}{163}, \bibinfo{pages}{42--47}.
\newblock \URLprefix \url{https://www.sciencedirect.com/science/article/pii/S0034425715000991}, \DOIprefix\doi{10.1016/j.rse.2015.03.003}.
%Type = Article
\bibitem[{Ploton et~al.(2020)Ploton, Mortier, R{\'e}jou-M{\'e}chain, Barbier, Picard, Rossi, Dormann, Cornu, Viennois, Bayol et~al.}]{ploton2020spatial}
\bibinfo{author}{Ploton, P.}, \bibinfo{author}{Mortier, F.}, \bibinfo{author}{R{\'e}jou-M{\'e}chain, M.}, \bibinfo{author}{Barbier, N.}, \bibinfo{author}{Picard, N.}, \bibinfo{author}{Rossi, V.}, \bibinfo{author}{Dormann, C.}, \bibinfo{author}{Cornu, G.}, \bibinfo{author}{Viennois, G.}, \bibinfo{author}{Bayol, N.}, et~al., \bibinfo{year}{2020}.
\newblock \bibinfo{title}{Spatial validation reveals poor predictive performance of large-scale ecological mapping models}.
\newblock \bibinfo{journal}{Nature communications} \bibinfo{volume}{11}, \bibinfo{pages}{4540}.
%Type = Article
\bibitem[{Polivka et~al.(2016)Polivka, Wang, Ellison, Hyer and Ichoku}]{7498622}
\bibinfo{author}{Polivka, T.N.}, \bibinfo{author}{Wang, J.}, \bibinfo{author}{Ellison, L.T.}, \bibinfo{author}{Hyer, E.J.}, \bibinfo{author}{Ichoku, C.M.}, \bibinfo{year}{2016}.
\newblock \bibinfo{title}{Improving nocturnal fire detection with the viirs day–night band}.
\newblock \bibinfo{journal}{IEEE Transactions on Geoscience and Remote Sensing} \bibinfo{volume}{54}, \bibinfo{pages}{5503--5519}.
\newblock \DOIprefix\doi{10.1109/TGRS.2016.2566665}.
%Type = Article
\bibitem[{Pollina et~al.(2013)Pollina, Colle and Charney}]{pollina_climatology_2013}
\bibinfo{author}{Pollina, J.B.}, \bibinfo{author}{Colle, B.A.}, \bibinfo{author}{Charney, J.J.}, \bibinfo{year}{2013}.
\newblock \bibinfo{title}{Climatology and {Meteorological} {Evolution} of {Major} {Wildfire} {Events} over the {Northeast} {United} {States}}.
\newblock \bibinfo{journal}{Weather and Forecasting} \bibinfo{volume}{28}, \bibinfo{pages}{175--193}.
\newblock \URLprefix \url{https://journals.ametsoc.org/view/journals/wefo/28/1/waf-d-12-00009_1.xml}, \DOIprefix\doi{10.1175/WAF-D-12-00009.1}. \bibinfo{note}{publisher: American Meteorological Society Section: Weather and Forecasting}.
%Type = Article
\bibitem[{Ponomarev et~al.(2023)Ponomarev, Zabrodin, Shvetsov and Ponomareva}]{ponomarev2023wildfire}
\bibinfo{author}{Ponomarev, E.I.}, \bibinfo{author}{Zabrodin, A.N.}, \bibinfo{author}{Shvetsov, E.G.}, \bibinfo{author}{Ponomareva, T.V.}, \bibinfo{year}{2023}.
\newblock \bibinfo{title}{Wildfire intensity and fire emissions in siberia}.
\newblock \bibinfo{journal}{Fire} \bibinfo{volume}{6}, \bibinfo{pages}{246}.
%Type = Inproceedings
\bibitem[{Prapas et~al.(2023)Prapas, Bountos, Kondylatos, Michail, Camps-Valls and Papoutsis}]{prapas_televit_2023}
\bibinfo{author}{Prapas, I.}, \bibinfo{author}{Bountos, N.I.}, \bibinfo{author}{Kondylatos, S.}, \bibinfo{author}{Michail, D.}, \bibinfo{author}{Camps-Valls, G.}, \bibinfo{author}{Papoutsis, I.}, \bibinfo{year}{2023}.
\newblock \bibinfo{title}{{TeleViT}: {Teleconnection}-{Driven} {Transformers} {Improve} {Subseasonal} to {Seasonal} {Wildfire} {Forecasting}}, pp. \bibinfo{pages}{3754--3759}.
\newblock \URLprefix \url{https://openaccess.thecvf.com/content/ICCV2023W/AIHADR/html/Prapas_TeleViT_Teleconnection-Driven_Transformers_Improve_Subseasonal_to_Seasonal_Wildfire_Forecasting_ICCVW_2023_paper.html}.
%Type = Misc
\bibitem[{Prapas et~al.(2022)Prapas, Kondylatos and Papoutsis}]{prapas_2022_6475592}
\bibinfo{author}{Prapas, I.}, \bibinfo{author}{Kondylatos, S.}, \bibinfo{author}{Papoutsis, I.}, \bibinfo{year}{2022}.
\newblock \bibinfo{title}{{FireCube: A Daily Datacube for the Modeling and Analysis of Wildfires in Greece}}.
\newblock \URLprefix \url{https://doi.org/10.5281/zenodo.6475592}, \DOIprefix\doi{10.5281/zenodo.6475592}.
%Type = Misc
\bibitem[{Prapas et~al.(2021)Prapas, Kondylatos, Papoutsis, Camps-Valls, Ronco, Ángel Fernández-Torres, Guillem and Carvalhais}]{prapas2021deep}
\bibinfo{author}{Prapas, I.}, \bibinfo{author}{Kondylatos, S.}, \bibinfo{author}{Papoutsis, I.}, \bibinfo{author}{Camps-Valls, G.}, \bibinfo{author}{Ronco, M.}, \bibinfo{author}{Ángel Fernández-Torres, M.}, \bibinfo{author}{Guillem, M.P.}, \bibinfo{author}{Carvalhais, N.}, \bibinfo{year}{2021}.
\newblock \bibinfo{title}{Deep learning methods for daily wildfire danger forecasting}.
\newblock \URLprefix \url{https://arxiv.org/abs/2111.02736}, \href{http://arxiv.org/abs/2111.02736}{{\tt arXiv:2111.02736}}.
%Type = Article
\bibitem[{Preisler et~al.(2004)Preisler, Brillinger, Burgan and Benoit}]{preisler2004probability}
\bibinfo{author}{Preisler, H.K.}, \bibinfo{author}{Brillinger, D.R.}, \bibinfo{author}{Burgan, R.E.}, \bibinfo{author}{Benoit, J.}, \bibinfo{year}{2004}.
\newblock \bibinfo{title}{Probability based models for estimation of wildfire risk}.
\newblock \bibinfo{journal}{International Journal of wildland fire} \bibinfo{volume}{13}, \bibinfo{pages}{133--142}.
%Type = Article
\bibitem[{Price et~al.(2025)Price, Sanchez-Gonzalez, Alet, Andersson, El-Kadi, Masters, Ewalds, Stott, Mohamed, Battaglia et~al.}]{price2025probabilistic}
\bibinfo{author}{Price, I.}, \bibinfo{author}{Sanchez-Gonzalez, A.}, \bibinfo{author}{Alet, F.}, \bibinfo{author}{Andersson, T.R.}, \bibinfo{author}{El-Kadi, A.}, \bibinfo{author}{Masters, D.}, \bibinfo{author}{Ewalds, T.}, \bibinfo{author}{Stott, J.}, \bibinfo{author}{Mohamed, S.}, \bibinfo{author}{Battaglia, P.}, et~al., \bibinfo{year}{2025}.
\newblock \bibinfo{title}{Probabilistic weather forecasting with machine learning}.
\newblock \bibinfo{journal}{Nature} \bibinfo{volume}{637}, \bibinfo{pages}{84--90}.
%Type = Article
\bibitem[{Prior and Eriksen(2013)}]{prior_wildfire_2013}
\bibinfo{author}{Prior, T.}, \bibinfo{author}{Eriksen, C.}, \bibinfo{year}{2013}.
\newblock \bibinfo{title}{Wildfire preparedness, community cohesion and social–ecological systems}.
\newblock \bibinfo{journal}{Global Environmental Change} \bibinfo{volume}{23}, \bibinfo{pages}{1575--1586}.
\newblock \URLprefix \url{https://www.sciencedirect.com/science/article/pii/S0959378013001684}, \DOIprefix\doi{10.1016/j.gloenvcha.2013.09.016}.
%Type = Inproceedings
\bibitem[{Pye et~al.(2003)Pye, Prestemon, Butry and Abt}]{pye2003prescribed}
\bibinfo{author}{Pye, J.M.}, \bibinfo{author}{Prestemon, J.P.}, \bibinfo{author}{Butry, D.T.}, \bibinfo{author}{Abt, K.L.}, \bibinfo{year}{2003}.
\newblock \bibinfo{title}{Prescribed burning and wildfire risk in the 1998 fire season in florida}, in: \bibinfo{booktitle}{In: Omi, Philip N.; Joyce, Linda A., technical editors. Fire, fuel treatments, and ecological restoration: Conference proceedings; 2002 16-18 April; Fort Collins, CO. Proceedings RMRS-P-29. Fort Collins, CO: US Department of Agriculture, Forest Service, Rocky Mountain Research Station. p. 15-26}, pp. \bibinfo{pages}{15--26}.
%Type = Article
\bibitem[{Qadir et~al.(2021)Qadir, Talukdar, Uddin, Ahmad and Goparaju}]{qadir_predicting_2021}
\bibinfo{author}{Qadir, A.}, \bibinfo{author}{Talukdar, N.R.}, \bibinfo{author}{Uddin, M.M.}, \bibinfo{author}{Ahmad, F.}, \bibinfo{author}{Goparaju, L.}, \bibinfo{year}{2021}.
\newblock \bibinfo{title}{Predicting forest fire using multispectral satellite measurements in {Nepal}}.
\newblock \bibinfo{journal}{Remote Sensing Applications: Society and Environment} \bibinfo{volume}{23}, \bibinfo{pages}{100539}.
\newblock \URLprefix \url{https://www.sciencedirect.com/science/article/pii/S2352938521000756}, \DOIprefix\doi{10.1016/j.rsase.2021.100539}.
%Type = Article
\bibitem[{Quan et~al.(2023)Quan, Wang, Xie, He, Resco De~Dios, Yebra, Jiao and Chen}]{quan_improving_2023}
\bibinfo{author}{Quan, X.}, \bibinfo{author}{Wang, W.}, \bibinfo{author}{Xie, Q.}, \bibinfo{author}{He, B.}, \bibinfo{author}{Resco De~Dios, V.}, \bibinfo{author}{Yebra, M.}, \bibinfo{author}{Jiao, M.}, \bibinfo{author}{Chen, R.}, \bibinfo{year}{2023}.
\newblock \bibinfo{title}{Improving wildfire occurrence modelling by integrating time-series features of weather and fuel moisture content}.
\newblock \bibinfo{journal}{Environmental Modelling \& Software} \bibinfo{volume}{170}, \bibinfo{pages}{105840}.
\newblock \DOIprefix\doi{10.1016/j.envsoft.2023.105840}.
%Type = Article
\bibitem[{Quan et~al.(2021a)Quan, Yebra, Ria{\~n}o, He, Lai and Liu}]{quan2021global}
\bibinfo{author}{Quan, X.}, \bibinfo{author}{Yebra, M.}, \bibinfo{author}{Ria{\~n}o, D.}, \bibinfo{author}{He, B.}, \bibinfo{author}{Lai, G.}, \bibinfo{author}{Liu, X.}, \bibinfo{year}{2021}a.
\newblock \bibinfo{title}{Global fuel moisture content mapping from modis}.
\newblock \bibinfo{journal}{International Journal of Applied Earth Observation and Geoinformation} \bibinfo{volume}{101}, \bibinfo{pages}{102354}.
%Type = Article
\bibitem[{Quan et~al.(2021b)Quan, Yebra, Riaño, He, Lai and Liu}]{quan_global_2021}
\bibinfo{author}{Quan, X.}, \bibinfo{author}{Yebra, M.}, \bibinfo{author}{Riaño, D.}, \bibinfo{author}{He, B.}, \bibinfo{author}{Lai, G.}, \bibinfo{author}{Liu, X.}, \bibinfo{year}{2021}b.
\newblock \bibinfo{title}{Global fuel moisture content mapping from {MODIS}}.
\newblock \bibinfo{journal}{International Journal of Applied Earth Observation and Geoinformation} \bibinfo{volume}{101}, \bibinfo{pages}{102354}.
\newblock \URLprefix \url{https://www.sciencedirect.com/science/article/pii/S0303243421000611}, \DOIprefix\doi{10.1016/j.jag.2021.102354}.
%Type = Article
\bibitem[{Rabier(2005)}]{rabier2005overview}
\bibinfo{author}{Rabier, F.}, \bibinfo{year}{2005}.
\newblock \bibinfo{title}{Overview of global data assimilation developments in numerical weather-prediction centres}.
\newblock \bibinfo{journal}{Quarterly Journal of the Royal Meteorological Society: A journal of the atmospheric sciences, applied meteorology and physical oceanography} \bibinfo{volume}{131}, \bibinfo{pages}{3215--3233}.
%Type = Article
\bibitem[{Rahimzadeh-Bajgiran et~al.(2012)Rahimzadeh-Bajgiran, Omasa and Shimizu}]{rahimzadeh2012comparative}
\bibinfo{author}{Rahimzadeh-Bajgiran, P.}, \bibinfo{author}{Omasa, K.}, \bibinfo{author}{Shimizu, Y.}, \bibinfo{year}{2012}.
\newblock \bibinfo{title}{Comparative evaluation of the vegetation dryness index (vdi), the temperature vegetation dryness index (tvdi) and the improved tvdi (itvdi) for water stress detection in semi-arid regions of iran}.
\newblock \bibinfo{journal}{ISPRS Journal of Photogrammetry and Remote Sensing} \bibinfo{volume}{68}, \bibinfo{pages}{1--12}.
%Type = Article
\bibitem[{Rakhmatulina et~al.(2021)Rakhmatulina, Stephens and Thompson}]{rakhmatulina2021soil}
\bibinfo{author}{Rakhmatulina, E.}, \bibinfo{author}{Stephens, S.}, \bibinfo{author}{Thompson, S.}, \bibinfo{year}{2021}.
\newblock \bibinfo{title}{Soil moisture influences on sierra nevada dead fuel moisture content and fire risks}.
\newblock \bibinfo{journal}{Forest Ecology and Management} \bibinfo{volume}{496}, \bibinfo{pages}{119379}.
%Type = Article
\bibitem[{Rao et~al.(2020a)Rao, Williams, Flefil and Konings}]{rao2020sar}
\bibinfo{author}{Rao, K.}, \bibinfo{author}{Williams, A.P.}, \bibinfo{author}{Flefil, J.F.}, \bibinfo{author}{Konings, A.G.}, \bibinfo{year}{2020}a.
\newblock \bibinfo{title}{Sar-enhanced mapping of live fuel moisture content}.
\newblock \bibinfo{journal}{Remote Sensing of Environment} \bibinfo{volume}{245}, \bibinfo{pages}{111797}.
%Type = Article
\bibitem[{Rao et~al.(2020b)Rao, Williams, Flefil and Konings}]{rao_sar-enhanced_2020}
\bibinfo{author}{Rao, K.}, \bibinfo{author}{Williams, A.P.}, \bibinfo{author}{Flefil, J.F.}, \bibinfo{author}{Konings, A.G.}, \bibinfo{year}{2020}b.
\newblock \bibinfo{title}{{SAR}-enhanced mapping of live fuel moisture content}.
\newblock \bibinfo{journal}{Remote Sensing of Environment} \bibinfo{volume}{245}, \bibinfo{pages}{111797}.
\newblock \URLprefix \url{https://www.sciencedirect.com/science/article/pii/S003442572030167X}, \DOIprefix\doi{10.1016/j.rse.2020.111797}.
%Type = Article
\bibitem[{Rashkovetsky et~al.(2021)Rashkovetsky, Mauracher, Langer and Schmitt}]{9468971}
\bibinfo{author}{Rashkovetsky, D.}, \bibinfo{author}{Mauracher, F.}, \bibinfo{author}{Langer, M.}, \bibinfo{author}{Schmitt, M.}, \bibinfo{year}{2021}.
\newblock \bibinfo{title}{Wildfire detection from multisensor satellite imagery using deep semantic segmentation}.
\newblock \bibinfo{journal}{IEEE Journal of Selected Topics in Applied Earth Observations and Remote Sensing} \bibinfo{volume}{14}, \bibinfo{pages}{7001--7016}.
\newblock \DOIprefix\doi{10.1109/JSTARS.2021.3093625}.
%Type = Article
\bibitem[{Ray et~al.(2010)Ray, Nepstad and Brando}]{ray2010predicting}
\bibinfo{author}{Ray, D.}, \bibinfo{author}{Nepstad, D.}, \bibinfo{author}{Brando, P.}, \bibinfo{year}{2010}.
\newblock \bibinfo{title}{Predicting moisture dynamics of fine understory fuels in a moist tropical rainforest system: results of a pilot study undertaken to identify proxy variables useful for rating fire danger}.
\newblock \bibinfo{journal}{New Phytologist} \bibinfo{volume}{187}, \bibinfo{pages}{720--732}.
%Type = Article
\bibitem[{Razavi~Termeh et~al.(2018)Razavi~Termeh, Kornejady, Pourghasemi and Keesstra}]{razavi_termeh_flood_2018}
\bibinfo{author}{Razavi~Termeh, S.V.}, \bibinfo{author}{Kornejady, A.}, \bibinfo{author}{Pourghasemi, H.R.}, \bibinfo{author}{Keesstra, S.}, \bibinfo{year}{2018}.
\newblock \bibinfo{title}{Flood susceptibility mapping using novel ensembles of adaptive neuro fuzzy inference system and metaheuristic algorithms}.
\newblock \bibinfo{journal}{Science of The Total Environment} \bibinfo{volume}{615}, \bibinfo{pages}{438--451}.
\newblock \URLprefix \url{https://www.sciencedirect.com/science/article/pii/S0048969717326141}, \DOIprefix\doi{10.1016/j.scitotenv.2017.09.262}.
%Type = Article
\bibitem[{Rezaie et~al.(2023)Rezaie, Panahi, Bateni, Lee, Jun, Trauernicht and Neale}]{rezaie_development_2023}
\bibinfo{author}{Rezaie, F.}, \bibinfo{author}{Panahi, M.}, \bibinfo{author}{Bateni, S.M.}, \bibinfo{author}{Lee, S.}, \bibinfo{author}{Jun, C.}, \bibinfo{author}{Trauernicht, C.}, \bibinfo{author}{Neale, C.M.U.}, \bibinfo{year}{2023}.
\newblock \bibinfo{title}{Development of novel optimized deep learning algorithms for wildfire modeling: {A} case study of {Maui}, {Hawai}‘i}.
\newblock \bibinfo{journal}{Engineering Applications of Artificial Intelligence} \bibinfo{volume}{125}, \bibinfo{pages}{106699}.
\newblock \URLprefix \url{https://www.sciencedirect.com/science/article/pii/S0952197623008837}, \DOIprefix\doi{10.1016/j.engappai.2023.106699}.
%Type = Article
\bibitem[{Ritter et~al.(2023)Ritter, Hoffman, Battaglia, Linn and Mell}]{ritter2023vertical}
\bibinfo{author}{Ritter, S.M.}, \bibinfo{author}{Hoffman, C.M.}, \bibinfo{author}{Battaglia, M.A.}, \bibinfo{author}{Linn, R.}, \bibinfo{author}{Mell, W.E.}, \bibinfo{year}{2023}.
\newblock \bibinfo{title}{Vertical and horizontal crown fuel continuity influences group-scale ignition and fuel consumption}.
\newblock \bibinfo{journal}{Fire} \bibinfo{volume}{6}, \bibinfo{pages}{321}.
%Type = Article
\bibitem[{Rodriguez-Jimenez et~al.(2023)Rodriguez-Jimenez, Lorenzo, Novo, Acu{\~n}a-Alonso and Alvarez}]{rodriguez2023modelling}
\bibinfo{author}{Rodriguez-Jimenez, F.}, \bibinfo{author}{Lorenzo, H.}, \bibinfo{author}{Novo, A.}, \bibinfo{author}{Acu{\~n}a-Alonso, C.}, \bibinfo{author}{Alvarez, X.}, \bibinfo{year}{2023}.
\newblock \bibinfo{title}{Modelling of live fuel moisture content in different vegetation scenarios during dry periods using meteorological data and spectral indices}.
\newblock \bibinfo{journal}{Forest Ecology and Management} \bibinfo{volume}{546}, \bibinfo{pages}{121378}.
%Type = Article
\bibitem[{Rogger et~al.(2017)Rogger, Agnoletti, Alaoui, Bathurst, Bodner, Borga, Chaplot, Gallart, Glatzel, Hall et~al.}]{rogger2017land}
\bibinfo{author}{Rogger, M.}, \bibinfo{author}{Agnoletti, M.}, \bibinfo{author}{Alaoui, A.}, \bibinfo{author}{Bathurst, J.}, \bibinfo{author}{Bodner, G.}, \bibinfo{author}{Borga, M.}, \bibinfo{author}{Chaplot, V.}, \bibinfo{author}{Gallart, F.}, \bibinfo{author}{Glatzel, G.}, \bibinfo{author}{Hall, J.}, et~al., \bibinfo{year}{2017}.
\newblock \bibinfo{title}{Land use change impacts on floods at the catchment scale: Challenges and opportunities for future research}.
\newblock \bibinfo{journal}{Water resources research} \bibinfo{volume}{53}, \bibinfo{pages}{5209--5219}.
%Type = Article
\bibitem[{Romero-Calcerrada et~al.(2008)Romero-Calcerrada, Novillo, Millington and Gomez-Jimenez}]{romero2008gis}
\bibinfo{author}{Romero-Calcerrada, R.}, \bibinfo{author}{Novillo, C.}, \bibinfo{author}{Millington, J.D.}, \bibinfo{author}{Gomez-Jimenez, I.}, \bibinfo{year}{2008}.
\newblock \bibinfo{title}{Gis analysis of spatial patterns of human-caused wildfire ignition risk in the sw of madrid (central spain)}.
\newblock \bibinfo{journal}{Landscape ecology} \bibinfo{volume}{23}, \bibinfo{pages}{341--354}.
%Type = Article
\bibitem[{R{\"o}sch et~al.(2024)R{\"o}sch, Nolde, Ullmann and Riedlinger}]{rosch2024data}
\bibinfo{author}{R{\"o}sch, M.}, \bibinfo{author}{Nolde, M.}, \bibinfo{author}{Ullmann, T.}, \bibinfo{author}{Riedlinger, T.}, \bibinfo{year}{2024}.
\newblock \bibinfo{title}{Data-driven wildfire spread modeling of european wildfires using a spatiotemporal graph neural network}.
\newblock \bibinfo{journal}{Fire} \bibinfo{volume}{7}, \bibinfo{pages}{207}.
%Type = Techreport
\bibitem[{Rothermel(1972)}]{rothermel1972fire}
\bibinfo{author}{Rothermel, R.C.}, \bibinfo{year}{1972}.
\newblock \bibinfo{title}{A Mathematical Model for Predicting Fire Spread in Wildland Fuels}.
\newblock \bibinfo{type}{Research Paper}. Intermountain Forest \& Range Experiment Station, Forest Service, US Department of Agriculture. \bibinfo{address}{Ogden, UT, USA}.
%Type = Book
\bibitem[{Rothermel et~al.(1986)Rothermel, Wilson, Morris and Sackett}]{rothermel1986modeling}
\bibinfo{author}{Rothermel, R.C.}, \bibinfo{author}{Wilson, R.}, \bibinfo{author}{Morris, G.A.}, \bibinfo{author}{Sackett, S.S.}, \bibinfo{year}{1986}.
\newblock \bibinfo{title}{Modeling Moisture Content of Fine Dead Wildland Fuels:.}
\newblock \bibinfo{publisher}{United States Department of Agriculture, Forest Service, Intermountain~…}.
%Type = Article
\bibitem[{Roy et~al.(2023)Roy, Deria, Hong, Rasti, Plaza and Chanussot}]{10153685}
\bibinfo{author}{Roy, S.K.}, \bibinfo{author}{Deria, A.}, \bibinfo{author}{Hong, D.}, \bibinfo{author}{Rasti, B.}, \bibinfo{author}{Plaza, A.}, \bibinfo{author}{Chanussot, J.}, \bibinfo{year}{2023}.
\newblock \bibinfo{title}{Multimodal fusion transformer for remote sensing image classification}.
\newblock \bibinfo{journal}{IEEE Transactions on Geoscience and Remote Sensing} \bibinfo{volume}{61}, \bibinfo{pages}{1--20}.
\newblock \DOIprefix\doi{10.1109/TGRS.2023.3286826}.
%Type = Article
\bibitem[{Rubí and Gondim(2023)}]{rubi_performance_2023}
\bibinfo{author}{Rubí, J.N.S.}, \bibinfo{author}{Gondim, P.R.L.}, \bibinfo{year}{2023}.
\newblock \bibinfo{title}{A performance comparison of machine learning models for wildfire occurrence risk prediction in the {Brazilian} {Federal} {District} region}.
\newblock \bibinfo{journal}{Environment Systems and Decisions} \URLprefix \url{https://doi.org/10.1007/s10669-023-09921-2}, \DOIprefix\doi{10.1007/s10669-023-09921-2}.
%Type = Article
\bibitem[{Ruffault et~al.(2018)Ruffault, Martin-StPaul, Pimont and Dupuy}]{ruffault2018well}
\bibinfo{author}{Ruffault, J.}, \bibinfo{author}{Martin-StPaul, N.}, \bibinfo{author}{Pimont, F.}, \bibinfo{author}{Dupuy, J.L.}, \bibinfo{year}{2018}.
\newblock \bibinfo{title}{How well do meteorological drought indices predict live fuel moisture content (lfmc)? an assessment for wildfire research and operations in mediterranean ecosystems}.
\newblock \bibinfo{journal}{Agricultural and Forest Meteorology} \bibinfo{volume}{262}, \bibinfo{pages}{391--401}.
%Type = Article
\bibitem[{Saatchi et~al.(2007)Saatchi, Halligan, Despain and Crabtree}]{saatchi_estimation_2007}
\bibinfo{author}{Saatchi, S.}, \bibinfo{author}{Halligan, K.}, \bibinfo{author}{Despain, D.G.}, \bibinfo{author}{Crabtree, R.L.}, \bibinfo{year}{2007}.
\newblock \bibinfo{title}{Estimation of {Forest} {Fuel} {Load} {From} {Radar} {Remote} {Sensing}}.
\newblock \bibinfo{journal}{IEEE Transactions on Geoscience and Remote Sensing} \bibinfo{volume}{45}, \bibinfo{pages}{1726--1740}.
\newblock \URLprefix \url{https://ieeexplore.ieee.org/abstract/document/4215087}, \DOIprefix\doi{10.1109/TGRS.2006.887002}. \bibinfo{note}{conference Name: IEEE Transactions on Geoscience and Remote Sensing}.
%Type = Article
\bibitem[{Sadasivuni et~al.(2013)Sadasivuni, Cooke and Bhushan}]{sadasivuni_wildfire_2013}
\bibinfo{author}{Sadasivuni, R.}, \bibinfo{author}{Cooke, W.H.}, \bibinfo{author}{Bhushan, S.}, \bibinfo{year}{2013}.
\newblock \bibinfo{title}{Wildfire risk prediction in {Southeastern} {Mississippi} using population interaction}.
\newblock \bibinfo{journal}{Ecological Modelling} \bibinfo{volume}{251}, \bibinfo{pages}{297--306}.
\newblock \URLprefix \url{https://www.sciencedirect.com/science/article/pii/S0304380013000070}, \DOIprefix\doi{10.1016/j.ecolmodel.2012.12.024}.
%Type = Article
\bibitem[{Salih et~al.(2024)Salih, Raisi-Estabragh, Galazzo, Radeva, Petersen, Lekadir and Menegaz}]{salih2024perspective}
\bibinfo{author}{Salih, A.M.}, \bibinfo{author}{Raisi-Estabragh, Z.}, \bibinfo{author}{Galazzo, I.B.}, \bibinfo{author}{Radeva, P.}, \bibinfo{author}{Petersen, S.E.}, \bibinfo{author}{Lekadir, K.}, \bibinfo{author}{Menegaz, G.}, \bibinfo{year}{2024}.
\newblock \bibinfo{title}{A perspective on explainable artificial intelligence methods: Shap and lime}.
\newblock \bibinfo{journal}{Advanced Intelligent Systems} , \bibinfo{pages}{2400304}.
%Type = Article
\bibitem[{Salis et~al.(2021)Salis, Arca, Del~Giudice, Palaiologou, Alcasena-Urdiroz, Ager, Fiori, Pellizzaro, Scarpa, Schirru, Ventura, Casula and Duce}]{salis_application_2021}
\bibinfo{author}{Salis, M.}, \bibinfo{author}{Arca, B.}, \bibinfo{author}{Del~Giudice, L.}, \bibinfo{author}{Palaiologou, P.}, \bibinfo{author}{Alcasena-Urdiroz, F.}, \bibinfo{author}{Ager, A.}, \bibinfo{author}{Fiori, M.}, \bibinfo{author}{Pellizzaro, G.}, \bibinfo{author}{Scarpa, C.}, \bibinfo{author}{Schirru, M.}, \bibinfo{author}{Ventura, A.}, \bibinfo{author}{Casula, M.}, \bibinfo{author}{Duce, P.}, \bibinfo{year}{2021}.
\newblock \bibinfo{title}{Application of simulation modeling for wildfire exposure and transmission assessment in {Sardinia}, {Italy}}.
\newblock \bibinfo{journal}{International Journal of Disaster Risk Reduction} \bibinfo{volume}{58}, \bibinfo{pages}{102189}.
\newblock \URLprefix \url{https://linkinghub.elsevier.com/retrieve/pii/S2212420921001552}, \DOIprefix\doi{10.1016/j.ijdrr.2021.102189}.
%Type = Inproceedings
\bibitem[{Santopaolo et~al.(2021)Santopaolo, Saif, Pietrabissa and Giuseppi}]{santopaolo2021forest}
\bibinfo{author}{Santopaolo, A.}, \bibinfo{author}{Saif, S.S.}, \bibinfo{author}{Pietrabissa, A.}, \bibinfo{author}{Giuseppi, A.}, \bibinfo{year}{2021}.
\newblock \bibinfo{title}{Forest fire risk prediction from satellite data with convolutional neural networks}, in: \bibinfo{booktitle}{2021 29th Mediterranean Conference on Control and Automation (MED)}, \bibinfo{organization}{IEEE}. pp. \bibinfo{pages}{360--367}.
%Type = Article
\bibitem[{Sayad et~al.(2019)Sayad, Mousannif and Al~Moatassime}]{sayad_predictive_2019}
\bibinfo{author}{Sayad, Y.O.}, \bibinfo{author}{Mousannif, H.}, \bibinfo{author}{Al~Moatassime, H.}, \bibinfo{year}{2019}.
\newblock \bibinfo{title}{Predictive modeling of wildfires: A new dataset and machine learning approach}.
\newblock \bibinfo{journal}{Fire safety journal} \bibinfo{volume}{104}, \bibinfo{pages}{130--146}.
%Type = Article
\bibitem[{Scheller et~al.(2007)Scheller, Domingo, Sturtevant, Williams, Rudy, Gustafson and Mladenoff}]{scheller2007design}
\bibinfo{author}{Scheller, R.M.}, \bibinfo{author}{Domingo, J.B.}, \bibinfo{author}{Sturtevant, B.R.}, \bibinfo{author}{Williams, J.S.}, \bibinfo{author}{Rudy, A.}, \bibinfo{author}{Gustafson, E.J.}, \bibinfo{author}{Mladenoff, D.J.}, \bibinfo{year}{2007}.
\newblock \bibinfo{title}{Design, development, and application of landis-ii, a spatial landscape simulation model with flexible temporal and spatial resolution}.
\newblock \bibinfo{journal}{ecological modelling} \bibinfo{volume}{201}, \bibinfo{pages}{409--419}.
%Type = Book
\bibitem[{Schroeder and Buck(1977)}]{schroeder1977fire}
\bibinfo{author}{Schroeder, M.J.}, \bibinfo{author}{Buck, C.C.}, \bibinfo{year}{1977}.
\newblock \bibinfo{title}{Fire weather: A guide for application of meteorological information to forest fire control operations}.
\newblock \bibinfo{publisher}{US Forest Service}.
%Type = Article
\bibitem[{Schroeder et~al.(2014)Schroeder, Oliva, Giglio and Csiszar}]{schroeder2014new}
\bibinfo{author}{Schroeder, W.}, \bibinfo{author}{Oliva, P.}, \bibinfo{author}{Giglio, L.}, \bibinfo{author}{Csiszar, I.A.}, \bibinfo{year}{2014}.
\newblock \bibinfo{title}{The new viirs 375 m active fire detection data product: Algorithm description and initial assessment}.
\newblock \bibinfo{journal}{Remote Sensing of Environment} \bibinfo{volume}{143}, \bibinfo{pages}{85--96}.
%Type = Book
\bibitem[{Scott(2001)}]{scott2001assessing}
\bibinfo{author}{Scott, J.H.}, \bibinfo{year}{2001}.
\newblock \bibinfo{title}{Assessing crown fire potential by linking models of surface and crown fire behavior}.
\newblock \bibinfo{number}{29}, \bibinfo{publisher}{US Department of Agriculture, Forest Service, Rocky Mountain Research Station}.
%Type = Article
\bibitem[{Senawi et~al.(2017)Senawi, Wei and Billings}]{senawi2017new}
\bibinfo{author}{Senawi, A.}, \bibinfo{author}{Wei, H.L.}, \bibinfo{author}{Billings, S.A.}, \bibinfo{year}{2017}.
\newblock \bibinfo{title}{A new maximum relevance-minimum multicollinearity (mrmmc) method for feature selection and ranking}.
\newblock \bibinfo{journal}{Pattern Recognition} \bibinfo{volume}{67}, \bibinfo{pages}{47--61}.
%Type = Article
\bibitem[{Shadrin et~al.(2024)Shadrin, Illarionova, Gubanov, Evteeva, Mironenko, Levchunets, Belousov and Burnaev}]{shadrin_wildfire_2024}
\bibinfo{author}{Shadrin, D.}, \bibinfo{author}{Illarionova, S.}, \bibinfo{author}{Gubanov, F.}, \bibinfo{author}{Evteeva, K.}, \bibinfo{author}{Mironenko, M.}, \bibinfo{author}{Levchunets, I.}, \bibinfo{author}{Belousov, R.}, \bibinfo{author}{Burnaev, E.}, \bibinfo{year}{2024}.
\newblock \bibinfo{title}{Wildfire spreading prediction using multimodal data and deep neural network approach}.
\newblock \bibinfo{journal}{Scientific Reports} \bibinfo{volume}{14}, \bibinfo{pages}{2606}.
\newblock \URLprefix \url{https://www.ncbi.nlm.nih.gov/pmc/articles/PMC10831103/}, \DOIprefix\doi{10.1038/s41598-024-52821-x}.
%Type = Inproceedings
\bibitem[{Shah et~al.(2024)Shah, VS and Patel}]{Shah_2024_CVPR}
\bibinfo{author}{Shah, N.A.}, \bibinfo{author}{VS, V.}, \bibinfo{author}{Patel, V.M.}, \bibinfo{year}{2024}.
\newblock \bibinfo{title}{Lqmformer: Language-aware query mask transformer for referring image segmentation}, in: \bibinfo{booktitle}{Proceedings of the IEEE/CVF Conference on Computer Vision and Pattern Recognition (CVPR)}, pp. \bibinfo{pages}{12903--12913}.
%Type = Misc
\bibitem[{Shamsoshoara et~al.(2020)Shamsoshoara, Afghah, Razi, Zheng, Fulé and Blasch}]{qad6-r683-20}
\bibinfo{author}{Shamsoshoara, A.}, \bibinfo{author}{Afghah, F.}, \bibinfo{author}{Razi, A.}, \bibinfo{author}{Zheng, L.}, \bibinfo{author}{Fulé, P.}, \bibinfo{author}{Blasch, E.}, \bibinfo{year}{2020}.
\newblock \bibinfo{title}{The flame dataset: Aerial imagery pile burn detection using drones (uavs)}.
\newblock \URLprefix \url{https://dx.doi.org/10.21227/qad6-r683}, \DOIprefix\doi{10.21227/qad6-r683}.
%Type = Article
\bibitem[{Shapley(1953)}]{shapley1953stochastic}
\bibinfo{author}{Shapley, L.S.}, \bibinfo{year}{1953}.
\newblock \bibinfo{title}{Stochastic games}.
\newblock \bibinfo{journal}{Proceedings of the national academy of sciences} \bibinfo{volume}{39}, \bibinfo{pages}{1095--1100}.
%Type = Misc
\bibitem[{Shi et~al.(2025)Shi, Shirali, Jin, Zhou, Hu, Rangaraj, Wang, Han, Wang, Lall, Wu, Bobadilla and Narasimhan}]{shi2025deeplearningfoundationmodels}
\bibinfo{author}{Shi, J.}, \bibinfo{author}{Shirali, A.}, \bibinfo{author}{Jin, B.}, \bibinfo{author}{Zhou, S.}, \bibinfo{author}{Hu, W.}, \bibinfo{author}{Rangaraj, R.}, \bibinfo{author}{Wang, S.}, \bibinfo{author}{Han, J.}, \bibinfo{author}{Wang, Z.}, \bibinfo{author}{Lall, U.}, \bibinfo{author}{Wu, Y.}, \bibinfo{author}{Bobadilla, L.}, \bibinfo{author}{Narasimhan, G.}, \bibinfo{year}{2025}.
\newblock \bibinfo{title}{Deep learning and foundation models for weather prediction: A survey}.
\newblock \URLprefix \url{https://arxiv.org/abs/2501.06907}, \href{http://arxiv.org/abs/2501.06907}{{\tt arXiv:2501.06907}}.
%Type = Article
\bibitem[{Short(2014)}]{short2014spatial}
\bibinfo{author}{Short, K.C.}, \bibinfo{year}{2014}.
\newblock \bibinfo{title}{A spatial database of wildfires in the united states, 1992-2011}.
\newblock \bibinfo{journal}{Earth System Science Data} \bibinfo{volume}{6}, \bibinfo{pages}{1--27}.
%Type = Inproceedings
\bibitem[{Shvidenko et~al.(2011)Shvidenko, Shchepashchenko, Vaganov, Sukhinin, Maksyutov, McCallum and Lakyda}]{shvidenko2011impact}
\bibinfo{author}{Shvidenko, A.}, \bibinfo{author}{Shchepashchenko, D.}, \bibinfo{author}{Vaganov, E.}, \bibinfo{author}{Sukhinin, A.}, \bibinfo{author}{Maksyutov, S.S.}, \bibinfo{author}{McCallum, I.}, \bibinfo{author}{Lakyda, I.}, \bibinfo{year}{2011}.
\newblock \bibinfo{title}{Impact of wildfire in russia between 1998--2010 on ecosystems and the global carbon budget}, in: \bibinfo{booktitle}{Doklady Earth Sciences}, \bibinfo{organization}{Springer}. pp. \bibinfo{pages}{1678--1682}.
%Type = Inproceedings
\bibitem[{Singla et~al.(2020)Singla, Diao, Mukhopadhyay, Eldawy, Shachter and Kochenderfer}]{singla2020wildfiredb}
\bibinfo{author}{Singla, S.}, \bibinfo{author}{Diao, T.}, \bibinfo{author}{Mukhopadhyay, A.}, \bibinfo{author}{Eldawy, A.}, \bibinfo{author}{Shachter, R.}, \bibinfo{author}{Kochenderfer, M.}, \bibinfo{year}{2020}.
\newblock \bibinfo{title}{Wildfiredb: A spatio-temporal dataset combining wildfire occurrence with relevant covariates}, in: \bibinfo{booktitle}{34th Conference on Neural Information Processing Systems (NeurIPS 2020)}.
%Type = Article
\bibitem[{Skowronski et~al.(2011)Skowronski, Clark, Duveneck and Hom}]{skowronski_three-Dimensional_2011}
\bibinfo{author}{Skowronski, N.S.}, \bibinfo{author}{Clark, K.L.}, \bibinfo{author}{Duveneck, M.}, \bibinfo{author}{Hom, J.}, \bibinfo{year}{2011}.
\newblock \bibinfo{title}{Three-dimensional canopy fuel loading predicted using upward and downward sensing {LiDAR} systems}.
\newblock \bibinfo{journal}{Remote Sensing of Environment} \bibinfo{volume}{115}, \bibinfo{pages}{703--714}.
\newblock \URLprefix \url{https://www.sciencedirect.com/science/article/pii/S0034425710003172}, \DOIprefix\doi{10.1016/j.rse.2010.10.012}.
%Type = Article
\bibitem[{Song and Wang(2020)}]{song_global_2020}
\bibinfo{author}{Song, Y.}, \bibinfo{author}{Wang, Y.}, \bibinfo{year}{2020}.
\newblock \bibinfo{title}{Global {Wildfire} {Outlook} {Forecast} with {Neural} {Networks}}.
\newblock \bibinfo{journal}{Remote Sensing} \bibinfo{volume}{12}, \bibinfo{pages}{2246}.
\newblock \URLprefix \url{https://www.mdpi.com/2072-4292/12/14/2246}, \DOIprefix\doi{10.3390/rs12142246}. \bibinfo{note}{number: 14 Publisher: Multidisciplinary Digital Publishing Institute}.
%Type = Article
\bibitem[{Sow et~al.(2013)Sow, Mbow, H{\'e}ly, Fensholt and Sambou}]{sow2013estimation}
\bibinfo{author}{Sow, M.}, \bibinfo{author}{Mbow, C.}, \bibinfo{author}{H{\'e}ly, C.}, \bibinfo{author}{Fensholt, R.}, \bibinfo{author}{Sambou, B.}, \bibinfo{year}{2013}.
\newblock \bibinfo{title}{Estimation of herbaceous fuel moisture content using vegetation indices and land surface temperature from modis data}.
\newblock \bibinfo{journal}{Remote Sensing} \bibinfo{volume}{5}, \bibinfo{pages}{2617--2638}.
%Type = Article
\bibitem[{Srivastava et~al.(2014)Srivastava, Saran, de~By and Dadhwal}]{srivastava_geo-information_2014}
\bibinfo{author}{Srivastava, S.K.}, \bibinfo{author}{Saran, S.}, \bibinfo{author}{de~By, R.A.}, \bibinfo{author}{Dadhwal, V.K.}, \bibinfo{year}{2014}.
\newblock \bibinfo{title}{A geo-information system approach for forest fire likelihood based on causative and anti-causative factors}.
\newblock \bibinfo{journal}{International Journal of Geographical Information Science} \bibinfo{volume}{28}, \bibinfo{pages}{427--454}.
\newblock \URLprefix \url{https://doi.org/10.1080/13658816.2013.797984}, \DOIprefix\doi{10.1080/13658816.2013.797984}. \bibinfo{note}{publisher: Taylor \& Francis \_eprint: https://doi.org/10.1080/13658816.2013.797984}.
%Type = Article
\bibitem[{Stojanovic et~al.(2016)Stojanovic, nee Voogdt, Webb, Cook and Heinsohn}]{stojanovic2016loss}
\bibinfo{author}{Stojanovic, D.}, \bibinfo{author}{nee Voogdt, J.W.}, \bibinfo{author}{Webb, M.}, \bibinfo{author}{Cook, H.}, \bibinfo{author}{Heinsohn, R.}, \bibinfo{year}{2016}.
\newblock \bibinfo{title}{Loss of habitat for a secondary cavity nesting bird after wildfire}.
\newblock \bibinfo{journal}{Forest Ecology and Management} \bibinfo{volume}{360}, \bibinfo{pages}{235--241}.
%Type = Article
\bibitem[{Stow et~al.(2006)Stow, Niphadkar and Kaiser}]{stow2006time}
\bibinfo{author}{Stow, D.}, \bibinfo{author}{Niphadkar, M.}, \bibinfo{author}{Kaiser, J.}, \bibinfo{year}{2006}.
\newblock \bibinfo{title}{Time series of chaparral live fuel moisture maps derived from modis satellite data}.
\newblock \bibinfo{journal}{International Journal of Wildland Fire} \bibinfo{volume}{15}, \bibinfo{pages}{347--360}.
%Type = Article
\bibitem[{Su et~al.(2021)Su, Zheng, Luo, Tigabu and Guo}]{su_modeling_2021}
\bibinfo{author}{Su, Z.}, \bibinfo{author}{Zheng, L.}, \bibinfo{author}{Luo, S.}, \bibinfo{author}{Tigabu, M.}, \bibinfo{author}{Guo, F.}, \bibinfo{year}{2021}.
\newblock \bibinfo{title}{Modeling wildfire drivers in {Chinese} tropical forest ecosystems using global logistic regression and geographically weighted logistic regression}.
\newblock \bibinfo{journal}{Natural Hazards} \bibinfo{volume}{108}, \bibinfo{pages}{1317--1345}.
\newblock \URLprefix \url{https://doi.org/10.1007/s11069-021-04733-6}, \DOIprefix\doi{10.1007/s11069-021-04733-6}.
%Type = Inproceedings
\bibitem[{Sundararajan et~al.(2017)Sundararajan, Taly and Yan}]{sundararajan2017axiomatic}
\bibinfo{author}{Sundararajan, M.}, \bibinfo{author}{Taly, A.}, \bibinfo{author}{Yan, Q.}, \bibinfo{year}{2017}.
\newblock \bibinfo{title}{Axiomatic attribution for deep networks}, in: \bibinfo{booktitle}{International conference on machine learning}, \bibinfo{organization}{PMLR}. pp. \bibinfo{pages}{3319--3328}.
%Type = Article
\bibitem[{Swetnam and Betancourt(1990)}]{swetnam1990fire}
\bibinfo{author}{Swetnam, T.W.}, \bibinfo{author}{Betancourt, J.L.}, \bibinfo{year}{1990}.
\newblock \bibinfo{title}{Fire-southern oscillation relations in the southwestern united states}.
\newblock \bibinfo{journal}{Science} \bibinfo{volume}{249}, \bibinfo{pages}{1017--1020}.
%Type = Misc
\bibitem[{Sønderby et~al.(2020)Sønderby, Espeholt, Heek, Dehghani, Oliver, Salimans, Agrawal, Hickey and Kalchbrenner}]{sønderby2020metnetneuralweathermodel}
\bibinfo{author}{Sønderby, C.K.}, \bibinfo{author}{Espeholt, L.}, \bibinfo{author}{Heek, J.}, \bibinfo{author}{Dehghani, M.}, \bibinfo{author}{Oliver, A.}, \bibinfo{author}{Salimans, T.}, \bibinfo{author}{Agrawal, S.}, \bibinfo{author}{Hickey, J.}, \bibinfo{author}{Kalchbrenner, N.}, \bibinfo{year}{2020}.
\newblock \bibinfo{title}{Metnet: A neural weather model for precipitation forecasting}.
\newblock \URLprefix \url{https://arxiv.org/abs/2003.12140}, \href{http://arxiv.org/abs/2003.12140}{{\tt arXiv:2003.12140}}.
%Type = Article
\bibitem[{Tanase et~al.(2015)Tanase, Panciera, Lowell and Aponte}]{tanase2015monitoring}
\bibinfo{author}{Tanase, M.}, \bibinfo{author}{Panciera, R.}, \bibinfo{author}{Lowell, K.}, \bibinfo{author}{Aponte, C.}, \bibinfo{year}{2015}.
\newblock \bibinfo{title}{Monitoring live fuel moisture in semiarid environments using l-band radar data}.
\newblock \bibinfo{journal}{International Journal of Wildland Fire} \bibinfo{volume}{24}, \bibinfo{pages}{560--572}.
%Type = Article
\bibitem[{Tanase et~al.(2022)Tanase, Nova, Marino, Aponte, Tom{\'e}, Y{\'a}{\~n}ez, Madrigal, Guijarro and Hernando}]{tanase2022characterizing}
\bibinfo{author}{Tanase, M.A.}, \bibinfo{author}{Nova, J.P.G.}, \bibinfo{author}{Marino, E.}, \bibinfo{author}{Aponte, C.}, \bibinfo{author}{Tom{\'e}, J.L.}, \bibinfo{author}{Y{\'a}{\~n}ez, L.}, \bibinfo{author}{Madrigal, J.}, \bibinfo{author}{Guijarro, M.}, \bibinfo{author}{Hernando, C.}, \bibinfo{year}{2022}.
\newblock \bibinfo{title}{Characterizing live fuel moisture content from active and passive sensors in a mediterranean environment}.
\newblock \bibinfo{journal}{Forests} \bibinfo{volume}{13}, \bibinfo{pages}{1846}.
%Type = Article
\bibitem[{Taneja et~al.(2021)Taneja, Hilton, Wallace, Reinke and Jones}]{taneja_effect_2021}
\bibinfo{author}{Taneja, R.}, \bibinfo{author}{Hilton, J.}, \bibinfo{author}{Wallace, L.}, \bibinfo{author}{Reinke, K.}, \bibinfo{author}{Jones, S.}, \bibinfo{year}{2021}.
\newblock \bibinfo{title}{Effect of fuel spatial resolution on predictive wildfire models}.
\newblock \bibinfo{journal}{International Journal of Wildland Fire} \bibinfo{volume}{30}, \bibinfo{pages}{776--789}.
\newblock \URLprefix \url{https://www.publish.csiro.au/wf/WF20192}, \DOIprefix\doi{10.1071/WF20192}. \bibinfo{note}{publisher: CSIRO PUBLISHING}.
%Type = Article
\bibitem[{Tariq et~al.(2021)Tariq, Shu, Li, Altan, Khan, Baqa and Lu}]{tariq2021quantitative}
\bibinfo{author}{Tariq, A.}, \bibinfo{author}{Shu, H.}, \bibinfo{author}{Li, Q.}, \bibinfo{author}{Altan, O.}, \bibinfo{author}{Khan, M.R.}, \bibinfo{author}{Baqa, M.F.}, \bibinfo{author}{Lu, L.}, \bibinfo{year}{2021}.
\newblock \bibinfo{title}{Quantitative analysis of forest fires in southeastern australia using sar data}.
\newblock \bibinfo{journal}{Remote Sensing} \bibinfo{volume}{13}, \bibinfo{pages}{2386}.
%Type = Article
\bibitem[{Tavakkoli~Piralilou et~al.(2022)Tavakkoli~Piralilou, Einali, Ghorbanzadeh, Nachappa, Gholamnia, Blaschke and Ghamisi}]{tavakkoli_piralilou_google_2022}
\bibinfo{author}{Tavakkoli~Piralilou, S.}, \bibinfo{author}{Einali, G.}, \bibinfo{author}{Ghorbanzadeh, O.}, \bibinfo{author}{Nachappa, T.G.}, \bibinfo{author}{Gholamnia, K.}, \bibinfo{author}{Blaschke, T.}, \bibinfo{author}{Ghamisi, P.}, \bibinfo{year}{2022}.
\newblock \bibinfo{title}{A {Google} {Earth} {Engine} {Approach} for {Wildfire} {Susceptibility} {Prediction} {Fusion} with {Remote} {Sensing} {Data} of {Different} {Spatial} {Resolutions}}.
\newblock \bibinfo{journal}{Remote Sensing} \bibinfo{volume}{14}, \bibinfo{pages}{672}.
\newblock \URLprefix \url{https://www.mdpi.com/2072-4292/14/3/672}, \DOIprefix\doi{10.3390/rs14030672}. \bibinfo{note}{number: 3 Publisher: Multidisciplinary Digital Publishing Institute}.
%Type = Article
\bibitem[{Tesch et~al.(2023)Tesch, Kollet and Garcke}]{tesch2023causal}
\bibinfo{author}{Tesch, T.}, \bibinfo{author}{Kollet, S.}, \bibinfo{author}{Garcke, J.}, \bibinfo{year}{2023}.
\newblock \bibinfo{title}{Causal deep learning models for studying the earth system}.
\newblock \bibinfo{journal}{Geoscientific Model Development} \bibinfo{volume}{16}, \bibinfo{pages}{2149--2166}.
%Type = Article
\bibitem[{Thompson et~al.(2013)Thompson, Vaillant, Haas, Gebert and Stockmann}]{thompson2013quantifying}
\bibinfo{author}{Thompson, M.P.}, \bibinfo{author}{Vaillant, N.M.}, \bibinfo{author}{Haas, J.R.}, \bibinfo{author}{Gebert, K.M.}, \bibinfo{author}{Stockmann, K.D.}, \bibinfo{year}{2013}.
\newblock \bibinfo{title}{Quantifying the potential impacts of fuel treatments on wildfire suppression costs}.
\newblock \bibinfo{journal}{Journal of Forestry} \bibinfo{volume}{111}, \bibinfo{pages}{49--58}.
%Type = Article
\bibitem[{Tolhurst et~al.(2008)Tolhurst, Shields and Chong}]{tolhurst2008phoenix}
\bibinfo{author}{Tolhurst, K.}, \bibinfo{author}{Shields, B.}, \bibinfo{author}{Chong, D.}, \bibinfo{year}{2008}.
\newblock \bibinfo{title}{Phoenix: development and application of a bushfire risk management tool}.
\newblock \bibinfo{journal}{Australian Journal of Emergency Management, The} \bibinfo{volume}{23}, \bibinfo{pages}{47--54}.
%Type = Article
\bibitem[{Toulouse et~al.(2017)Toulouse, Rossi, Campana, Celik and Akhloufi}]{toulouse2017computer}
\bibinfo{author}{Toulouse, T.}, \bibinfo{author}{Rossi, L.}, \bibinfo{author}{Campana, A.}, \bibinfo{author}{Celik, T.}, \bibinfo{author}{Akhloufi, M.A.}, \bibinfo{year}{2017}.
\newblock \bibinfo{title}{Computer vision for wildfire research: An evolving image dataset for processing and analysis}.
\newblock \bibinfo{journal}{Fire Safety Journal} \bibinfo{volume}{92}, \bibinfo{pages}{188--194}.
%Type = Article
\bibitem[{Tran et~al.(2023)Tran, Bateni, Rezaie, Panahi, Jun, Trauernicht and Neale}]{tran_enhancing_2023}
\bibinfo{author}{Tran, T.T.K.}, \bibinfo{author}{Bateni, S.M.}, \bibinfo{author}{Rezaie, F.}, \bibinfo{author}{Panahi, M.}, \bibinfo{author}{Jun, C.}, \bibinfo{author}{Trauernicht, C.}, \bibinfo{author}{Neale, C.M.U.}, \bibinfo{year}{2023}.
\newblock \bibinfo{title}{Enhancing predictive ability of optimized group method of data handling ({GMDH}) method for wildfire susceptibility mapping}.
\newblock \bibinfo{journal}{Agricultural and Forest Meteorology} \bibinfo{volume}{339}, \bibinfo{pages}{109587}.
\newblock \URLprefix \url{https://www.sciencedirect.com/science/article/pii/S0168192323002782}, \DOIprefix\doi{10.1016/j.agrformet.2023.109587}.
%Type = Article
\bibitem[{Tran et~al.(2024)Tran, Janizadeh, Bateni, Jun, Kim, Trauernicht, Rezaie, Giambelluca and Panahi}]{tran2024improving}
\bibinfo{author}{Tran, T.T.K.}, \bibinfo{author}{Janizadeh, S.}, \bibinfo{author}{Bateni, S.M.}, \bibinfo{author}{Jun, C.}, \bibinfo{author}{Kim, D.}, \bibinfo{author}{Trauernicht, C.}, \bibinfo{author}{Rezaie, F.}, \bibinfo{author}{Giambelluca, T.W.}, \bibinfo{author}{Panahi, M.}, \bibinfo{year}{2024}.
\newblock \bibinfo{title}{Improving the prediction of wildfire susceptibility on hawai'i island, hawai'i, using explainable hybrid machine learning models}.
\newblock \bibinfo{journal}{Journal of environmental management} \bibinfo{volume}{351}, \bibinfo{pages}{119724}.
%Type = Article
\bibitem[{Trucchia et~al.(2022)Trucchia, Meschi, Fiorucci, Gollini and Negro}]{trucchia_defining_2022}
\bibinfo{author}{Trucchia, A.}, \bibinfo{author}{Meschi, G.}, \bibinfo{author}{Fiorucci, P.}, \bibinfo{author}{Gollini, A.}, \bibinfo{author}{Negro, D.}, \bibinfo{year}{2022}.
\newblock \bibinfo{title}{Defining {Wildfire} {Susceptibility} {Maps} in {Italy} for {Understanding} {Seasonal} {Wildfire} {Regimes} at the {National} {Level}}.
\newblock \bibinfo{journal}{Fire} \bibinfo{volume}{5}, \bibinfo{pages}{30}.
\newblock \URLprefix \url{https://www.mdpi.com/2571-6255/5/1/30}, \DOIprefix\doi{10.3390/fire5010030}. \bibinfo{note}{number: 1 Publisher: Multidisciplinary Digital Publishing Institute}.
%Type = Article
\bibitem[{Tucker(1979)}]{tucker_red_1979}
\bibinfo{author}{Tucker, C.J.}, \bibinfo{year}{1979}.
\newblock \bibinfo{title}{Red and photographic infrared linear combinations for monitoring vegetation}.
\newblock \bibinfo{journal}{Remote Sensing of Environment} \bibinfo{volume}{8}, \bibinfo{pages}{127--150}.
\newblock \URLprefix \url{https://www.sciencedirect.com/science/article/pii/0034425779900130}, \DOIprefix\doi{10.1016/0034-4257(79)90013-0}.
%Type = Article
\bibitem[{Tymstra et~al.(2010)Tymstra, Bryce, Wotton, Taylor, Armitage et~al.}]{tymstra2010development}
\bibinfo{author}{Tymstra, C.}, \bibinfo{author}{Bryce, R.}, \bibinfo{author}{Wotton, B.}, \bibinfo{author}{Taylor, S.}, \bibinfo{author}{Armitage, O.}, et~al., \bibinfo{year}{2010}.
\newblock \bibinfo{title}{Development and structure of prometheus: the canadian wildland fire growth simulation model}.
\newblock \bibinfo{journal}{Natural Resources Canada, Canadian Forest Service, Northern Forestry Centre, Information Report NOR-X-417.(Edmonton, AB)} .
%Type = Article
\bibitem[{Ujjwal et~al.(2022)Ujjwal, Hilton, Garg and Aryal}]{ujjwal2022probability}
\bibinfo{author}{Ujjwal, K.}, \bibinfo{author}{Hilton, J.}, \bibinfo{author}{Garg, S.}, \bibinfo{author}{Aryal, J.}, \bibinfo{year}{2022}.
\newblock \bibinfo{title}{A probability-based risk metric for operational wildfire risk management}.
\newblock \bibinfo{journal}{Environmental Modelling \& Software} \bibinfo{volume}{148}, \bibinfo{pages}{105286}.
%Type = Article
\bibitem[{Umunnakwe et~al.(2022)Umunnakwe, Parvania, Nguyen, Horel and Davis}]{umunnakwe2022data}
\bibinfo{author}{Umunnakwe, A.}, \bibinfo{author}{Parvania, M.}, \bibinfo{author}{Nguyen, H.}, \bibinfo{author}{Horel, J.D.}, \bibinfo{author}{Davis, K.R.}, \bibinfo{year}{2022}.
\newblock \bibinfo{title}{Data-driven spatio-temporal analysis of wildfire risk to power systems operation}.
\newblock \bibinfo{journal}{IET Generation, Transmission \& Distribution} \bibinfo{volume}{16}, \bibinfo{pages}{2531--2546}.
%Type = Article
\bibitem[{Ustin and Middleton(2021)}]{ustin2021current}
\bibinfo{author}{Ustin, S.L.}, \bibinfo{author}{Middleton, E.M.}, \bibinfo{year}{2021}.
\newblock \bibinfo{title}{Current and near-term advances in earth observation for ecological applications}.
\newblock \bibinfo{journal}{Ecological Processes} \bibinfo{volume}{10}, \bibinfo{pages}{1}.
%Type = Inproceedings
\bibitem[{Vaswani et~al.(2017)Vaswani, Shazeer, Parmar, Uszkoreit, Jones, Gomez, Kaiser and Polosukhin}]{vaswani2017attention}
\bibinfo{author}{Vaswani, A.}, \bibinfo{author}{Shazeer, N.}, \bibinfo{author}{Parmar, N.}, \bibinfo{author}{Uszkoreit, J.}, \bibinfo{author}{Jones, L.}, \bibinfo{author}{Gomez, A.N.}, \bibinfo{author}{Kaiser, L.}, \bibinfo{author}{Polosukhin, I.}, \bibinfo{year}{2017}.
\newblock \bibinfo{title}{Attention is all you need}, in: \bibinfo{booktitle}{Advances in Neural Information Processing Systems}.
%Type = Article
\bibitem[{Vega et~al.(2022)Vega, Arellano-P{\'e}rez, {\'A}lvarez-Gonz{\'a}lez, Fern{\'a}ndez, Jim{\'e}nez, Fern{\'a}ndez-Alonso, Vega-Nieva, Briones-Herrera, Alonso-Rego, Font{\'u}rbel et~al.}]{vega2022modelling}
\bibinfo{author}{Vega, J.A.}, \bibinfo{author}{Arellano-P{\'e}rez, S.}, \bibinfo{author}{{\'A}lvarez-Gonz{\'a}lez, J.G.}, \bibinfo{author}{Fern{\'a}ndez, C.}, \bibinfo{author}{Jim{\'e}nez, E.}, \bibinfo{author}{Fern{\'a}ndez-Alonso, J.M.}, \bibinfo{author}{Vega-Nieva, D.J.}, \bibinfo{author}{Briones-Herrera, C.}, \bibinfo{author}{Alonso-Rego, C.}, \bibinfo{author}{Font{\'u}rbel, T.}, et~al., \bibinfo{year}{2022}.
\newblock \bibinfo{title}{Modelling aboveground biomass and fuel load components at stand level in shrub communities in nw spain}.
\newblock \bibinfo{journal}{Forest Ecology and Management} \bibinfo{volume}{505}, \bibinfo{pages}{119926}.
%Type = Inproceedings
\bibitem[{Verbesselt et~al.(2002)Verbesselt, Fleck, Coppin and Viegas}]{verbesselt_estimation_2002}
\bibinfo{author}{Verbesselt, J.}, \bibinfo{author}{Fleck, S.}, \bibinfo{author}{Coppin, P.}, \bibinfo{author}{Viegas, D.}, \bibinfo{year}{2002}.
\newblock \bibinfo{title}{Estimation of fuel moisture content towards fire risk assessment: a review}.
\newblock \URLprefix \url{https://www.semanticscholar.org/paper/Estimation-of-fuel-moisture-content-towards-fire-a-Verbesselt-Fleck/5d7c76a2a1ee4ca7f5984be7a66866def092df59}.
%Type = Article
\bibitem[{Viegas et~al.(2001)Viegas, Pi{\~n}ol, Viegas and Ogaya}]{viegas2001estimating}
\bibinfo{author}{Viegas, D.}, \bibinfo{author}{Pi{\~n}ol, J.}, \bibinfo{author}{Viegas, M.}, \bibinfo{author}{Ogaya, R.}, \bibinfo{year}{2001}.
\newblock \bibinfo{title}{Estimating live fine fuels moisture content using meteorologically-based indices}.
\newblock \bibinfo{journal}{International Journal of Wildland Fire} \bibinfo{volume}{10}, \bibinfo{pages}{223--240}.
%Type = Article
\bibitem[{Vilar et~al.(2016)Vilar, Camia, San-Miguel-Ayanz and Mart{\'\i}n}]{vilar2016modeling}
\bibinfo{author}{Vilar, L.}, \bibinfo{author}{Camia, A.}, \bibinfo{author}{San-Miguel-Ayanz, J.}, \bibinfo{author}{Mart{\'\i}n, M.P.}, \bibinfo{year}{2016}.
\newblock \bibinfo{title}{Modeling temporal changes in human-caused wildfires in mediterranean europe based on land use-land cover interfaces}.
\newblock \bibinfo{journal}{Forest Ecology and Management} \bibinfo{volume}{378}, \bibinfo{pages}{68--78}.
%Type = Article
\bibitem[{Viney(1991)}]{viney1991review}
\bibinfo{author}{Viney, N.R.}, \bibinfo{year}{1991}.
\newblock \bibinfo{title}{A review of fine fuel moisture modelling}.
\newblock \bibinfo{journal}{International Journal of Wildland Fire} \bibinfo{volume}{1}, \bibinfo{pages}{215--234}.
%Type = Article
\bibitem[{Vinodkumar et~al.(2021a)Vinodkumar, Dharssi, Yebra and Fox-Hughes}]{VINODKUMAR2021108503}
\bibinfo{author}{Vinodkumar, V.}, \bibinfo{author}{Dharssi, I.}, \bibinfo{author}{Yebra, M.}, \bibinfo{author}{Fox-Hughes, P.}, \bibinfo{year}{2021}a.
\newblock \bibinfo{title}{Continental-scale prediction of live fuel moisture content using soil moisture information}.
\newblock \bibinfo{journal}{Agricultural and Forest Meteorology} \bibinfo{volume}{307}, \bibinfo{pages}{108503}.
\newblock \URLprefix \url{https://www.sciencedirect.com/science/article/pii/S0168192321001866}, \DOIprefix\doi{https://doi.org/10.1016/j.agrformet.2021.108503}.
%Type = Article
\bibitem[{Vinodkumar et~al.(2021b)Vinodkumar, Dharssi, Yebra and Fox-Hughes}]{vinodkumar_continental-scale_2021}
\bibinfo{author}{Vinodkumar, V.}, \bibinfo{author}{Dharssi, I.}, \bibinfo{author}{Yebra, M.}, \bibinfo{author}{Fox-Hughes, P.}, \bibinfo{year}{2021}b.
\newblock \bibinfo{title}{Continental-scale prediction of live fuel moisture content using soil moisture information}.
\newblock \bibinfo{journal}{Agricultural and Forest Meteorology} \bibinfo{volume}{307}, \bibinfo{pages}{108503}.
\newblock \URLprefix \url{https://www.sciencedirect.com/science/article/pii/S0168192321001866}, \DOIprefix\doi{10.1016/j.agrformet.2021.108503}.
%Type = Article
\bibitem[{Vissio et~al.(2023)Vissio, Turco and Provenzale}]{vissio_testing_2023}
\bibinfo{author}{Vissio, G.}, \bibinfo{author}{Turco, M.}, \bibinfo{author}{Provenzale, A.}, \bibinfo{year}{2023}.
\newblock \bibinfo{title}{Testing drought indicators for summer burned area prediction in {Italy}}.
\newblock \bibinfo{journal}{Natural Hazards} \bibinfo{volume}{116}, \bibinfo{pages}{1125--1137}.
\newblock \URLprefix \url{https://doi.org/10.1007/s11069-022-05714-z}, \DOIprefix\doi{10.1007/s11069-022-05714-z}.
%Type = Inproceedings
\bibitem[{Viswanathan et~al.(2019)Viswanathan, Anand~Kumar and Soman}]{viswanathan2019sequence}
\bibinfo{author}{Viswanathan, S.}, \bibinfo{author}{Anand~Kumar, M.}, \bibinfo{author}{Soman, K.}, \bibinfo{year}{2019}.
\newblock \bibinfo{title}{A sequence-based machine comprehension modeling using lstm and gru}, in: \bibinfo{booktitle}{Emerging Research in Electronics, Computer Science and Technology: Proceedings of International Conference, ICERECT 2018}, \bibinfo{organization}{Springer}. pp. \bibinfo{pages}{47--55}.
%Type = Article
\bibitem[{Vitolo et~al.(2020)Vitolo, Di~Giuseppe, Barnard, Coughlan, San-Miguel-Ayanz, Libertá and Krzeminski}]{vitolo_era5-based_2020}
\bibinfo{author}{Vitolo, C.}, \bibinfo{author}{Di~Giuseppe, F.}, \bibinfo{author}{Barnard, C.}, \bibinfo{author}{Coughlan, R.}, \bibinfo{author}{San-Miguel-Ayanz, J.}, \bibinfo{author}{Libertá, G.}, \bibinfo{author}{Krzeminski, B.}, \bibinfo{year}{2020}.
\newblock \bibinfo{title}{{ERA5}-based global meteorological wildfire danger maps}.
\newblock \bibinfo{journal}{Scientific Data} \bibinfo{volume}{7}, \bibinfo{pages}{216}.
\newblock \URLprefix \url{https://www.nature.com/articles/s41597-020-0554-z}, \DOIprefix\doi{10.1038/s41597-020-0554-z}. \bibinfo{note}{publisher: Nature Publishing Group}.
%Type = Article
\bibitem[{Wagner(1993)}]{wagner1993prediction}
\bibinfo{author}{Wagner, C.V.}, \bibinfo{year}{1993}.
\newblock \bibinfo{title}{Prediction of crown fire behavior in two stands of jack pine}.
\newblock \bibinfo{journal}{Canadian Journal of Forest Research} \bibinfo{volume}{23}, \bibinfo{pages}{442--449}.
%Type = Article
\bibitem[{Walker et~al.(2020)Walker, Rogers, Veraverbeke, Johnstone, Baltzer, Barrett, Bourgeau-Chavez, Day, de~Groot, Dieleman et~al.}]{walker2020fuel}
\bibinfo{author}{Walker, X.J.}, \bibinfo{author}{Rogers, B.M.}, \bibinfo{author}{Veraverbeke, S.}, \bibinfo{author}{Johnstone, J.F.}, \bibinfo{author}{Baltzer, J.L.}, \bibinfo{author}{Barrett, K.}, \bibinfo{author}{Bourgeau-Chavez, L.}, \bibinfo{author}{Day, N.}, \bibinfo{author}{de~Groot, W.}, \bibinfo{author}{Dieleman, C.}, et~al., \bibinfo{year}{2020}.
\newblock \bibinfo{title}{Fuel availability not fire weather controls boreal wildfire severity and carbon emissions}.
\newblock \bibinfo{journal}{Nature Climate Change} \bibinfo{volume}{10}, \bibinfo{pages}{1130--1136}.
%Type = Inproceedings
\bibitem[{Wang and Ke(2024)}]{Wang_2024_CVPR}
\bibinfo{author}{Wang, J.}, \bibinfo{author}{Ke, L.}, \bibinfo{year}{2024}.
\newblock \bibinfo{title}{Llm-seg: Bridging image segmentation and large language model reasoning}, in: \bibinfo{booktitle}{Proceedings of the IEEE/CVF Conference on Computer Vision and Pattern Recognition (CVPR) Workshops}, pp. \bibinfo{pages}{1765--1774}.
%Type = Article
\bibitem[{Wang et~al.(2020)Wang, Roudini, Hyer, Xu, Zhou, Garcia, Reid, Peterson and {da Silva}}]{WANG2020111466}
\bibinfo{author}{Wang, J.}, \bibinfo{author}{Roudini, S.}, \bibinfo{author}{Hyer, E.J.}, \bibinfo{author}{Xu, X.}, \bibinfo{author}{Zhou, M.}, \bibinfo{author}{Garcia, L.C.}, \bibinfo{author}{Reid, J.S.}, \bibinfo{author}{Peterson, D.A.}, \bibinfo{author}{{da Silva}, A.M.}, \bibinfo{year}{2020}.
\newblock \bibinfo{title}{Detecting nighttime fire combustion phase by hybrid application of visible and infrared radiation from suomi npp viirs}.
\newblock \bibinfo{journal}{Remote Sensing of Environment} \bibinfo{volume}{237}, \bibinfo{pages}{111466}.
\newblock \URLprefix \url{https://www.sciencedirect.com/science/article/pii/S0034425719304857}, \DOIprefix\doi{https://doi.org/10.1016/j.rse.2019.111466}.
%Type = Article
\bibitem[{Wang et~al.(2019)Wang, Quan, He, Yebra, Xing and Liu}]{wang2019assessment}
\bibinfo{author}{Wang, L.}, \bibinfo{author}{Quan, X.}, \bibinfo{author}{He, B.}, \bibinfo{author}{Yebra, M.}, \bibinfo{author}{Xing, M.}, \bibinfo{author}{Liu, X.}, \bibinfo{year}{2019}.
\newblock \bibinfo{title}{Assessment of the dual polarimetric sentinel-1a data for forest fuel moisture content estimation}.
\newblock \bibinfo{journal}{Remote Sensing} \bibinfo{volume}{11}, \bibinfo{pages}{1568}.
%Type = Article
\bibitem[{Wang et~al.(2024)Wang, Ihme, Gazen, Chen and Anderson}]{wang2024firebench}
\bibinfo{author}{Wang, Q.}, \bibinfo{author}{Ihme, M.}, \bibinfo{author}{Gazen, C.}, \bibinfo{author}{Chen, Y.F.}, \bibinfo{author}{Anderson, J.}, \bibinfo{year}{2024}.
\newblock \bibinfo{title}{Firebench: A high-fidelity ensemble simulation framework for exploring wildfire behavior and data-driven modeling}.
\newblock \bibinfo{journal}{arXiv preprint arXiv:2406.08589} .
%Type = Article
\bibitem[{Wang et~al.(2021)Wang, Qian, Leung and Zhang}]{wang_identifying_2021}
\bibinfo{author}{Wang, S.S.C.}, \bibinfo{author}{Qian, Y.}, \bibinfo{author}{Leung, L.R.}, \bibinfo{author}{Zhang, Y.}, \bibinfo{year}{2021}.
\newblock \bibinfo{title}{Identifying {Key} {Drivers} of {Wildfires} in the {Contiguous} {US} {Using} {Machine} {Learning} and {Game} {Theory} {Interpretation}}.
\newblock \bibinfo{journal}{Earth's Future} \bibinfo{volume}{9}, \bibinfo{pages}{e2020EF001910}.
\newblock \URLprefix \url{https://onlinelibrary.wiley.com/doi/abs/10.1029/2020EF001910}, \DOIprefix\doi{10.1029/2020EF001910}. \bibinfo{note}{\_eprint: https://onlinelibrary.wiley.com/doi/pdf/10.1029/2020EF001910}.
%Type = Article
\bibitem[{Wang and Quan(2023)}]{wang2023estimation}
\bibinfo{author}{Wang, W.}, \bibinfo{author}{Quan, X.}, \bibinfo{year}{2023}.
\newblock \bibinfo{title}{Estimation of live fuel moisture content from multiple sources of remotely sensed data}.
\newblock \bibinfo{journal}{IEEE Geoscience and Remote Sensing Letters} .
%Type = Article
\bibitem[{Wang et~al.(2022a)Wang, Xi, Wang, Yang, Wang, Nie and Du}]{9765478}
\bibinfo{author}{Wang, Y.}, \bibinfo{author}{Xi, X.}, \bibinfo{author}{Wang, C.}, \bibinfo{author}{Yang, X.}, \bibinfo{author}{Wang, P.}, \bibinfo{author}{Nie, S.}, \bibinfo{author}{Du, M.}, \bibinfo{year}{2022}a.
\newblock \bibinfo{title}{A novel method based on kernel density for estimating crown base height using uav-borne lidar data}.
\newblock \bibinfo{journal}{IEEE Geoscience and Remote Sensing Letters} \bibinfo{volume}{19}, \bibinfo{pages}{1--5}.
\newblock \DOIprefix\doi{10.1109/LGRS.2022.3171316}.
%Type = Inproceedings
\bibitem[{Wang et~al.(2022b)Wang, Lu, Li, Tao, Guo, Gong and Liu}]{9878961}
\bibinfo{author}{Wang, Z.}, \bibinfo{author}{Lu, Y.}, \bibinfo{author}{Li, Q.}, \bibinfo{author}{Tao, X.}, \bibinfo{author}{Guo, Y.}, \bibinfo{author}{Gong, M.}, \bibinfo{author}{Liu, T.}, \bibinfo{year}{2022}b.
\newblock \bibinfo{title}{Cris: Clip-driven referring image segmentation}, in: \bibinfo{booktitle}{2022 IEEE/CVF Conference on Computer Vision and Pattern Recognition (CVPR)}, pp. \bibinfo{pages}{11676--11685}.
\newblock \DOIprefix\doi{10.1109/CVPR52688.2022.01139}.
%Type = Article
\bibitem[{Waqas et~al.(2024)Waqas, Humphries, Chueasa and Wangwongchai}]{waqas2024artificial}
\bibinfo{author}{Waqas, M.}, \bibinfo{author}{Humphries, U.W.}, \bibinfo{author}{Chueasa, B.}, \bibinfo{author}{Wangwongchai, A.}, \bibinfo{year}{2024}.
\newblock \bibinfo{title}{Artificial intelligence and numerical weather prediction models: A technical survey}.
\newblock \bibinfo{journal}{Natural Hazards Research} .
%Type = Article
\bibitem[{Wing et~al.(2012)Wing, Ritchie, Boston, Cohen, Gitelman and Olsen}]{wing2012prediction}
\bibinfo{author}{Wing, B.M.}, \bibinfo{author}{Ritchie, M.W.}, \bibinfo{author}{Boston, K.}, \bibinfo{author}{Cohen, W.B.}, \bibinfo{author}{Gitelman, A.}, \bibinfo{author}{Olsen, M.J.}, \bibinfo{year}{2012}.
\newblock \bibinfo{title}{Prediction of understory vegetation cover with airborne lidar in an interior ponderosa pine forest}.
\newblock \bibinfo{journal}{Remote Sensing of Environment} \bibinfo{volume}{124}, \bibinfo{pages}{730--741}.
%Type = Article
\bibitem[{Woodall et~al.(2013)Woodall, Walters, Oswalt, Domke, Toney and Gray}]{woodall2013biomass}
\bibinfo{author}{Woodall, C.}, \bibinfo{author}{Walters, B.}, \bibinfo{author}{Oswalt, S.}, \bibinfo{author}{Domke, G.}, \bibinfo{author}{Toney, C.}, \bibinfo{author}{Gray, A.}, \bibinfo{year}{2013}.
\newblock \bibinfo{title}{Biomass and carbon attributes of downed woody materials in forests of the united states}.
\newblock \bibinfo{journal}{Forest Ecology and Management} \bibinfo{volume}{305}, \bibinfo{pages}{48--59}.
%Type = Article
\bibitem[{Wu et~al.(2024)Wu, Li, Li, Ding, Tong and Tao}]{10460426}
\bibinfo{author}{Wu, J.}, \bibinfo{author}{Li, X.}, \bibinfo{author}{Li, X.}, \bibinfo{author}{Ding, H.}, \bibinfo{author}{Tong, Y.}, \bibinfo{author}{Tao, D.}, \bibinfo{year}{2024}.
\newblock \bibinfo{title}{Toward robust referring image segmentation}.
\newblock \bibinfo{journal}{IEEE Transactions on Image Processing} \bibinfo{volume}{33}, \bibinfo{pages}{1782--1794}.
\newblock \DOIprefix\doi{10.1109/TIP.2024.3371348}.
%Type = Article
\bibitem[{Xie et~al.(2022a)Xie, Qi, Hu, Huang, Chen and Zhang}]{xie2022retrieval}
\bibinfo{author}{Xie, J.}, \bibinfo{author}{Qi, T.}, \bibinfo{author}{Hu, W.}, \bibinfo{author}{Huang, H.}, \bibinfo{author}{Chen, B.}, \bibinfo{author}{Zhang, J.}, \bibinfo{year}{2022}a.
\newblock \bibinfo{title}{Retrieval of live fuel moisture content based on multi-source remote sensing data and ensemble deep learning model}.
\newblock \bibinfo{journal}{Remote Sensing} \bibinfo{volume}{14}, \bibinfo{pages}{4378}.
%Type = Article
\bibitem[{Xie et~al.(2022b)Xie, Zhang, Zhan, Li, Shama, Zhan, Wang, Lv, Bao and Wu}]{xie2022wildfire}
\bibinfo{author}{Xie, L.}, \bibinfo{author}{Zhang, R.}, \bibinfo{author}{Zhan, J.}, \bibinfo{author}{Li, S.}, \bibinfo{author}{Shama, A.}, \bibinfo{author}{Zhan, R.}, \bibinfo{author}{Wang, T.}, \bibinfo{author}{Lv, J.}, \bibinfo{author}{Bao, X.}, \bibinfo{author}{Wu, R.}, \bibinfo{year}{2022}b.
\newblock \bibinfo{title}{Wildfire risk assessment in liangshan prefecture, china based on an integration machine learning algorithm}.
\newblock \bibinfo{journal}{Remote Sensing} \bibinfo{volume}{14}, \bibinfo{pages}{4592}.
%Type = Article
\bibitem[{Xu et~al.(2023a)Xu, Zhu and Clifton}]{10123038}
\bibinfo{author}{Xu, P.}, \bibinfo{author}{Zhu, X.}, \bibinfo{author}{Clifton, D.A.}, \bibinfo{year}{2023}a.
\newblock \bibinfo{title}{Multimodal learning with transformers: A survey}.
\newblock \bibinfo{journal}{IEEE Transactions on Pattern Analysis and Machine Intelligence} \bibinfo{volume}{45}, \bibinfo{pages}{12113--12132}.
\newblock \DOIprefix\doi{10.1109/TPAMI.2023.3275156}.
%Type = Misc
\bibitem[{Xu et~al.(2024)Xu, Yang, Wang, Cai, Du and Chen}]{xu2024visualmambasurveynew}
\bibinfo{author}{Xu, R.}, \bibinfo{author}{Yang, S.}, \bibinfo{author}{Wang, Y.}, \bibinfo{author}{Cai, Y.}, \bibinfo{author}{Du, B.}, \bibinfo{author}{Chen, H.}, \bibinfo{year}{2024}.
\newblock \bibinfo{title}{Visual mamba: A survey and new outlooks}.
\newblock \URLprefix \url{https://arxiv.org/abs/2404.18861}, \href{http://arxiv.org/abs/2404.18861}{{\tt arXiv:2404.18861}}.
%Type = Article
\bibitem[{Xu and Wooster(2023)}]{xu2023sentinel}
\bibinfo{author}{Xu, W.}, \bibinfo{author}{Wooster, M.J.}, \bibinfo{year}{2023}.
\newblock \bibinfo{title}{Sentinel-3 slstr active fire (af) detection and frp daytime product-algorithm description and global intercomparison to modis, viirs and landsat af data}.
\newblock \bibinfo{journal}{Science of Remote Sensing} \bibinfo{volume}{7}, \bibinfo{pages}{100087}.
%Type = Inproceedings
\bibitem[{Xu et~al.(2023b)Xu, Chen, Zhang, Song, Wan and Li}]{Xu_2023_ICCV}
\bibinfo{author}{Xu, Z.}, \bibinfo{author}{Chen, Z.}, \bibinfo{author}{Zhang, Y.}, \bibinfo{author}{Song, Y.}, \bibinfo{author}{Wan, X.}, \bibinfo{author}{Li, G.}, \bibinfo{year}{2023}b.
\newblock \bibinfo{title}{Bridging vision and language encoders: Parameter-efficient tuning for referring image segmentation}, in: \bibinfo{booktitle}{Proceedings of the IEEE/CVF International Conference on Computer Vision (ICCV)}, pp. \bibinfo{pages}{17503--17512}.
%Type = Inproceedings
\bibitem[{Yahia et~al.(2023)Yahia, Ghabi and Karoui}]{yahia_prediction_2023}
\bibinfo{author}{Yahia, O.}, \bibinfo{author}{Ghabi, M.}, \bibinfo{author}{Karoui, M.S.}, \bibinfo{year}{2023}.
\newblock \bibinfo{title}{The {Prediction} of {Regional} {Wildfire} {Risk} {Using} {High}-{Resolution} {Remotely} {Sensed} {Soil} {Moisture} {Content} {Estimation}, {Case} {Study}: {Sidi} {Douma} {Forest}, {Saida}, {Algeria}}, in: \bibinfo{booktitle}{{IGARSS} 2023 - 2023 {IEEE} {International} {Geoscience} and {Remote} {Sensing} {Symposium}}, pp. \bibinfo{pages}{3387--3390}.
\newblock \URLprefix \url{https://ieeexplore.ieee.org/abstract/document/10281986}, \DOIprefix\doi{10.1109/IGARSS52108.2023.10281986}. \bibinfo{note}{iSSN: 2153-7003}.
%Type = Misc
\bibitem[{Yan et~al.(2023)Yan, He, Wan and Liu}]{yan2023mmnetmultimasknetworkreferring}
\bibinfo{author}{Yan, Y.}, \bibinfo{author}{He, X.}, \bibinfo{author}{Wan, W.}, \bibinfo{author}{Liu, J.}, \bibinfo{year}{2023}.
\newblock \bibinfo{title}{Mmnet: Multi-mask network for referring image segmentation}.
\newblock \URLprefix \url{https://arxiv.org/abs/2305.14969}, \href{http://arxiv.org/abs/2305.14969}{{\tt arXiv:2305.14969}}.
%Type = Article
\bibitem[{Yang et~al.(2022a)Yang, Fu, Liu, Dong and Liu}]{yang_projections_2022}
\bibinfo{author}{Yang, C.E.}, \bibinfo{author}{Fu, J.S.}, \bibinfo{author}{Liu, Y.}, \bibinfo{author}{Dong, X.}, \bibinfo{author}{Liu, Y.}, \bibinfo{year}{2022}a.
\newblock \bibinfo{title}{Projections of future wildfires impacts on air pollutants and air toxics in a changing climate over the western {United} {States}}.
\newblock \bibinfo{journal}{Environmental Pollution} \bibinfo{volume}{304}, \bibinfo{pages}{119213}.
\newblock \URLprefix \url{https://www.sciencedirect.com/science/article/pii/S0269749122004274}, \DOIprefix\doi{10.1016/j.envpol.2022.119213}.
%Type = Article
\bibitem[{Yang et~al.(2004)Yang, He and Gustafson}]{yang2004hierarchical}
\bibinfo{author}{Yang, J.}, \bibinfo{author}{He, H.S.}, \bibinfo{author}{Gustafson, E.J.}, \bibinfo{year}{2004}.
\newblock \bibinfo{title}{A hierarchical fire frequency model to simulate temporal patterns of fire regimes in landis}.
\newblock \bibinfo{journal}{Ecological Modelling} \bibinfo{volume}{180}, \bibinfo{pages}{119--133}.
%Type = Inproceedings
\bibitem[{Yang et~al.(2022b)Yang, Wang, Tang, Chen, Zhao and Torr}]{Yang_2022_CVPR}
\bibinfo{author}{Yang, Z.}, \bibinfo{author}{Wang, J.}, \bibinfo{author}{Tang, Y.}, \bibinfo{author}{Chen, K.}, \bibinfo{author}{Zhao, H.}, \bibinfo{author}{Torr, P.H.}, \bibinfo{year}{2022}b.
\newblock \bibinfo{title}{Lavt: Language-aware vision transformer for referring image segmentation}, in: \bibinfo{booktitle}{Proceedings of the IEEE/CVF Conference on Computer Vision and Pattern Recognition (CVPR)}, pp. \bibinfo{pages}{18155--18165}.
%Type = Article
\bibitem[{Yao et~al.(2023)Yao, Zhang, Li, Hong and Chanussot}]{10147258}
\bibinfo{author}{Yao, J.}, \bibinfo{author}{Zhang, B.}, \bibinfo{author}{Li, C.}, \bibinfo{author}{Hong, D.}, \bibinfo{author}{Chanussot, J.}, \bibinfo{year}{2023}.
\newblock \bibinfo{title}{Extended vision transformer (exvit) for land use and land cover classification: A multimodal deep learning framework}.
\newblock \bibinfo{journal}{IEEE Transactions on Geoscience and Remote Sensing} \bibinfo{volume}{61}, \bibinfo{pages}{1--15}.
\newblock \DOIprefix\doi{10.1109/TGRS.2023.3284671}.
%Type = Article
\bibitem[{Ye et~al.(2022)Ye, Saide, Hair, Fenn, Shingler, Soja, Gargulinski and Wiggins}]{ye2022assessing}
\bibinfo{author}{Ye, X.}, \bibinfo{author}{Saide, P.E.}, \bibinfo{author}{Hair, J.}, \bibinfo{author}{Fenn, M.}, \bibinfo{author}{Shingler, T.}, \bibinfo{author}{Soja, A.}, \bibinfo{author}{Gargulinski, E.}, \bibinfo{author}{Wiggins, E.}, \bibinfo{year}{2022}.
\newblock \bibinfo{title}{Assessing vertical allocation of wildfire smoke emissions using observational constraints from airborne lidar in the western us}.
\newblock \bibinfo{journal}{Journal of Geophysical Research: Atmospheres} \bibinfo{volume}{127}, \bibinfo{pages}{e2022JD036808}.
%Type = Article
\bibitem[{Yebra et~al.(2008)Yebra, Chuvieco and Riano}]{yebra2008estimation}
\bibinfo{author}{Yebra, M.}, \bibinfo{author}{Chuvieco, E.}, \bibinfo{author}{Riano, D.}, \bibinfo{year}{2008}.
\newblock \bibinfo{title}{Estimation of live fuel moisture content from modis images for fire risk assessment}.
\newblock \bibinfo{journal}{Agricultural and forest meteorology} \bibinfo{volume}{148}, \bibinfo{pages}{523--536}.
%Type = Article
\bibitem[{Yebra et~al.(2013a)Yebra, Dennison, Chuvieco, Ria{\~n}o, Zylstra, Hunt~Jr, Danson, Qi and Jurdao}]{yebra2013global}
\bibinfo{author}{Yebra, M.}, \bibinfo{author}{Dennison, P.E.}, \bibinfo{author}{Chuvieco, E.}, \bibinfo{author}{Ria{\~n}o, D.}, \bibinfo{author}{Zylstra, P.}, \bibinfo{author}{Hunt~Jr, E.R.}, \bibinfo{author}{Danson, F.M.}, \bibinfo{author}{Qi, Y.}, \bibinfo{author}{Jurdao, S.}, \bibinfo{year}{2013}a.
\newblock \bibinfo{title}{A global review of remote sensing of live fuel moisture content for fire danger assessment: Moving towards operational products}.
\newblock \bibinfo{journal}{Remote Sensing of Environment} \bibinfo{volume}{136}, \bibinfo{pages}{455--468}.
%Type = Article
\bibitem[{Yebra et~al.(2013b)Yebra, Dennison, Chuvieco, Riaño, Zylstra, Hunt, Danson, Qi and Jurdao}]{yebra_global_2013}
\bibinfo{author}{Yebra, M.}, \bibinfo{author}{Dennison, P.E.}, \bibinfo{author}{Chuvieco, E.}, \bibinfo{author}{Riaño, D.}, \bibinfo{author}{Zylstra, P.}, \bibinfo{author}{Hunt, E.R.}, \bibinfo{author}{Danson, F.M.}, \bibinfo{author}{Qi, Y.}, \bibinfo{author}{Jurdao, S.}, \bibinfo{year}{2013}b.
\newblock \bibinfo{title}{A global review of remote sensing of live fuel moisture content for fire danger assessment: {Moving} towards operational products}.
\newblock \bibinfo{journal}{Remote Sensing of Environment} \bibinfo{volume}{136}, \bibinfo{pages}{455--468}.
\newblock \URLprefix \url{https://www.sciencedirect.com/science/article/pii/S0034425713001831}, \DOIprefix\doi{10.1016/j.rse.2013.05.029}.
%Type = Article
\bibitem[{Ying et~al.(2021)Ying, Cheng, Shen, Guan, Luo and Peng}]{ying2021relative}
\bibinfo{author}{Ying, L.}, \bibinfo{author}{Cheng, H.}, \bibinfo{author}{Shen, Z.}, \bibinfo{author}{Guan, P.}, \bibinfo{author}{Luo, C.}, \bibinfo{author}{Peng, X.}, \bibinfo{year}{2021}.
\newblock \bibinfo{title}{Relative humidity and agricultural activities dominate wildfire ignitions in yunnan, southwest china: Patterns, thresholds, and implications}.
\newblock \bibinfo{journal}{Agricultural and Forest Meteorology} \bibinfo{volume}{307}, \bibinfo{pages}{108540}.
%Type = Article
\bibitem[{Ying et~al.(2019)Ying, Shen, Yang and Piao}]{ying2019wildfire}
\bibinfo{author}{Ying, L.}, \bibinfo{author}{Shen, Z.}, \bibinfo{author}{Yang, M.}, \bibinfo{author}{Piao, S.}, \bibinfo{year}{2019}.
\newblock \bibinfo{title}{Wildfire detection probability of modis fire products under the constraint of environmental factors: A study based on confirmed ground wildfire records}.
\newblock \bibinfo{journal}{Remote Sensing} \bibinfo{volume}{11}, \bibinfo{pages}{3031}.
%Type = Inproceedings
\bibitem[{Yoon and Voulgaris(2022)}]{10029946_2022}
\bibinfo{author}{Yoon, H.J.}, \bibinfo{author}{Voulgaris, P.}, \bibinfo{year}{2022}.
\newblock \bibinfo{title}{Multi-time predictions of wildfire grid map using remote sensing local data}, in: \bibinfo{booktitle}{2022 IEEE International Conference on Knowledge Graph (ICKG)}, pp. \bibinfo{pages}{365--372}.
\newblock \DOIprefix\doi{10.1109/ICKG55886.2022.00053}.
%Type = Article
\bibitem[{Yu et~al.(2024)Yu, Liu, Chang, Li and Shi}]{10551624}
\bibinfo{author}{Yu, Y.}, \bibinfo{author}{Liu, L.}, \bibinfo{author}{Chang, Z.}, \bibinfo{author}{Li, Y.}, \bibinfo{author}{Shi, K.}, \bibinfo{year}{2024}.
\newblock \bibinfo{title}{Detecting forest fires in southwest china from remote sensing nighttime lights using the random forest classification model}.
\newblock \bibinfo{journal}{IEEE Journal of Selected Topics in Applied Earth Observations and Remote Sensing} \bibinfo{volume}{17}, \bibinfo{pages}{10759--10769}.
\newblock \DOIprefix\doi{10.1109/JSTARS.2024.3410172}.
%Type = Article
\bibitem[{Yuan et~al.(2015)Yuan, Zhang and Liu}]{yuan2015survey}
\bibinfo{author}{Yuan, C.}, \bibinfo{author}{Zhang, Y.}, \bibinfo{author}{Liu, Z.}, \bibinfo{year}{2015}.
\newblock \bibinfo{title}{A survey on technologies for automatic forest fire monitoring, detection, and fighting using unmanned aerial vehicles and remote sensing techniques}.
\newblock \bibinfo{journal}{Canadian journal of forest research} \bibinfo{volume}{45}, \bibinfo{pages}{783--792}.
%Type = Article
\bibitem[{Yuan et~al.(2024)Yuan, Mou, Hua and Zhu}]{10458079}
\bibinfo{author}{Yuan, Z.}, \bibinfo{author}{Mou, L.}, \bibinfo{author}{Hua, Y.}, \bibinfo{author}{Zhu, X.X.}, \bibinfo{year}{2024}.
\newblock \bibinfo{title}{Rrsis: Referring remote sensing image segmentation}.
\newblock \bibinfo{journal}{IEEE Transactions on Geoscience and Remote Sensing} \bibinfo{volume}{62}, \bibinfo{pages}{1--12}.
\newblock \DOIprefix\doi{10.1109/TGRS.2024.3369720}.
%Type = Article
\bibitem[{Yue et~al.(2023)Yue, Ren, Liang, Liang, Lin, Yin and Wei}]{yue2023assessment}
\bibinfo{author}{Yue, W.}, \bibinfo{author}{Ren, C.}, \bibinfo{author}{Liang, Y.}, \bibinfo{author}{Liang, J.}, \bibinfo{author}{Lin, X.}, \bibinfo{author}{Yin, A.}, \bibinfo{author}{Wei, Z.}, \bibinfo{year}{2023}.
\newblock \bibinfo{title}{Assessment of wildfire susceptibility and wildfire threats to ecological environment and urban development based on gis and multi-source data: A case study of guilin, china}.
\newblock \bibinfo{journal}{Remote Sensing} \bibinfo{volume}{15}, \bibinfo{pages}{2659}.
%Type = Article
\bibitem[{Zacharakis and Tsihrintzis(2023)}]{zacharakis_integrated_2023}
\bibinfo{author}{Zacharakis, I.}, \bibinfo{author}{Tsihrintzis, V.A.}, \bibinfo{year}{2023}.
\newblock \bibinfo{title}{Integrated wildfire danger models and factors: {A} review}.
\newblock \bibinfo{journal}{Science of The Total Environment} \bibinfo{volume}{899}, \bibinfo{pages}{165704}.
\newblock \URLprefix \url{https://linkinghub.elsevier.com/retrieve/pii/S0048969723043279}, \DOIprefix\doi{10.1016/j.scitotenv.2023.165704}.
%Type = Article
\bibitem[{Zarzycki and {\L}awry{\'n}czuk(2022)}]{zarzycki2022advanced}
\bibinfo{author}{Zarzycki, K.}, \bibinfo{author}{{\L}awry{\'n}czuk, M.}, \bibinfo{year}{2022}.
\newblock \bibinfo{title}{Advanced predictive control for gru and lstm networks}.
\newblock \bibinfo{journal}{Information Sciences} \bibinfo{volume}{616}, \bibinfo{pages}{229--254}.
%Type = Inproceedings
\bibitem[{Zerveas et~al.(2021)Zerveas, Jayaraman, Patel, Bhamidipaty and Eickhoff}]{zerveas_transformer-based_2021}
\bibinfo{author}{Zerveas, G.}, \bibinfo{author}{Jayaraman, S.}, \bibinfo{author}{Patel, D.}, \bibinfo{author}{Bhamidipaty, A.}, \bibinfo{author}{Eickhoff, C.}, \bibinfo{year}{2021}.
\newblock \bibinfo{title}{A {Transformer}-based {Framework} for {Multivariate} {Time} {Series} {Representation} {Learning}}, in: \bibinfo{booktitle}{Proceedings of the 27th {ACM} {SIGKDD} {Conference} on {Knowledge} {Discovery} \& {Data} {Mining}}, \bibinfo{publisher}{Association for Computing Machinery}, \bibinfo{address}{New York, NY, USA}. pp. \bibinfo{pages}{2114--2124}.
\newblock \URLprefix \url{https://dl.acm.org/doi/10.1145/3447548.3467401}, \DOIprefix\doi{10.1145/3447548.3467401}.
%Type = Article
\bibitem[{Zhang et~al.(2018)Zhang, Pan, Li, Gardiner, Sargent, Hare and Atkinson}]{zhang_hybrid_2018}
\bibinfo{author}{Zhang, C.}, \bibinfo{author}{Pan, X.}, \bibinfo{author}{Li, H.}, \bibinfo{author}{Gardiner, A.}, \bibinfo{author}{Sargent, I.}, \bibinfo{author}{Hare, J.}, \bibinfo{author}{Atkinson, P.M.}, \bibinfo{year}{2018}.
\newblock \bibinfo{title}{A hybrid {MLP}-{CNN} classifier for very fine resolution remotely sensed image classification}.
\newblock \bibinfo{journal}{ISPRS Journal of Photogrammetry and Remote Sensing} \bibinfo{volume}{140}, \bibinfo{pages}{133--144}.
\newblock \URLprefix \url{https://www.sciencedirect.com/science/article/pii/S0924271617300254}, \DOIprefix\doi{10.1016/j.isprsjprs.2017.07.014}.
%Type = Article
\bibitem[{Zhang et~al.(2023a)Zhang, Huang, Gu, Hou, Zhang, Han, Dou and Feng}]{rs15061541}
\bibinfo{author}{Zhang, D.}, \bibinfo{author}{Huang, C.}, \bibinfo{author}{Gu, J.}, \bibinfo{author}{Hou, J.}, \bibinfo{author}{Zhang, Y.}, \bibinfo{author}{Han, W.}, \bibinfo{author}{Dou, P.}, \bibinfo{author}{Feng, Y.}, \bibinfo{year}{2023}a.
\newblock \bibinfo{title}{Real-time wildfire detection algorithm based on viirs fire product and himawari-8 data}.
\newblock \bibinfo{journal}{Remote Sensing} \bibinfo{volume}{15}.
\newblock \URLprefix \url{https://www.mdpi.com/2072-4292/15/6/1541}, \DOIprefix\doi{10.3390/rs15061541}.
%Type = Article
\bibitem[{Zhang et~al.(2019)Zhang, Wang and Liu}]{zhang_forest_2019}
\bibinfo{author}{Zhang, G.}, \bibinfo{author}{Wang, M.}, \bibinfo{author}{Liu, K.}, \bibinfo{year}{2019}.
\newblock \bibinfo{title}{Forest {Fire} {Susceptibility} {Modeling} {Using} a {Convolutional} {Neural} {Network} for {Yunnan} {Province} of {China}}.
\newblock \bibinfo{journal}{International Journal of Disaster Risk Science} \bibinfo{volume}{10}, \bibinfo{pages}{386--403}.
\newblock \URLprefix \url{https://doi.org/10.1007/s13753-019-00233-1}, \DOIprefix\doi{10.1007/s13753-019-00233-1}.
%Type = Article
\bibitem[{Zhang et~al.(2021a)Zhang, Wang and Liu}]{zhang_deep_2021}
\bibinfo{author}{Zhang, G.}, \bibinfo{author}{Wang, M.}, \bibinfo{author}{Liu, K.}, \bibinfo{year}{2021}a.
\newblock \bibinfo{title}{Deep neural networks for global wildfire susceptibility modelling}.
\newblock \bibinfo{journal}{Ecological Indicators} \bibinfo{volume}{127}, \bibinfo{pages}{107735}.
\newblock \URLprefix \url{https://www.sciencedirect.com/science/article/pii/S1470160X21004003}, \DOIprefix\doi{10.1016/j.ecolind.2021.107735}.
%Type = Article
\bibitem[{Zhang et~al.(2022)Zhang, Wang and Liu}]{zhang2022dynamic}
\bibinfo{author}{Zhang, G.}, \bibinfo{author}{Wang, M.}, \bibinfo{author}{Liu, K.}, \bibinfo{year}{2022}.
\newblock \bibinfo{title}{Dynamic prediction of global monthly burned area with hybrid deep neural networks}.
\newblock \bibinfo{journal}{Ecological Applications} \bibinfo{volume}{32}, \bibinfo{pages}{e2610}.
%Type = Article
\bibitem[{Zhang et~al.(2024a)Zhang, Wang, Yang and Liu}]{zhang2024current}
\bibinfo{author}{Zhang, G.}, \bibinfo{author}{Wang, M.}, \bibinfo{author}{Yang, B.}, \bibinfo{author}{Liu, K.}, \bibinfo{year}{2024}a.
\newblock \bibinfo{title}{Current and future patterns of global wildfire based on deep neural networks}.
\newblock \bibinfo{journal}{Earth's Future} \bibinfo{volume}{12}, \bibinfo{pages}{e2023EF004088}.
%Type = Article
\bibitem[{Zhang et~al.(2025)Zhang, Ángel Fernández-Torres, Cohrs and Camps-Valls}]{ZHANG2025104563}
\bibinfo{author}{Zhang, M.}, \bibinfo{author}{Ángel Fernández-Torres, M.}, \bibinfo{author}{Cohrs, K.H.}, \bibinfo{author}{Camps-Valls, G.}, \bibinfo{year}{2025}.
\newblock \bibinfo{title}{Calibration and uncertainty quantification for deep learning-based drought detection}.
\newblock \bibinfo{journal}{International Journal of Applied Earth Observation and Geoinformation} \bibinfo{volume}{140}, \bibinfo{pages}{104563}.
\newblock \URLprefix \url{https://www.sciencedirect.com/science/article/pii/S1569843225002109}, \DOIprefix\doi{https://doi.org/10.1016/j.jag.2025.104563}.
%Type = Article
\bibitem[{Zhang et~al.(2021b)Zhang, Ban and Nascetti}]{zhang2021learning}
\bibinfo{author}{Zhang, P.}, \bibinfo{author}{Ban, Y.}, \bibinfo{author}{Nascetti, A.}, \bibinfo{year}{2021}b.
\newblock \bibinfo{title}{Learning u-net without forgetting for near real-time wildfire monitoring by the fusion of sar and optical time series}.
\newblock \bibinfo{journal}{Remote Sensing of Environment} \bibinfo{volume}{261}, \bibinfo{pages}{112467}.
%Type = Article
\bibitem[{Zhang et~al.(2024b)Zhang, Hu, Ban, Nascetti and Gong}]{zhang2024assessing}
\bibinfo{author}{Zhang, P.}, \bibinfo{author}{Hu, X.}, \bibinfo{author}{Ban, Y.}, \bibinfo{author}{Nascetti, A.}, \bibinfo{author}{Gong, M.}, \bibinfo{year}{2024}b.
\newblock \bibinfo{title}{Assessing sentinel-2, sentinel-1, and alos-2 palsar-2 data for large-scale wildfire-burned area mapping: insights from the 2017--2019 canada wildfires}.
\newblock \bibinfo{journal}{Remote Sensing} \bibinfo{volume}{16}, \bibinfo{pages}{556}.
%Type = Article
\bibitem[{Zhang et~al.(2023b)Zhang, Lan, Ming, Zhu and Lo}]{rs15030598}
\bibinfo{author}{Zhang, X.}, \bibinfo{author}{Lan, M.}, \bibinfo{author}{Ming, J.}, \bibinfo{author}{Zhu, J.}, \bibinfo{author}{Lo, S.}, \bibinfo{year}{2023}b.
\newblock \bibinfo{title}{Spatiotemporal heterogeneity of forest fire occurrence based on remote sensing data: An analysis in anhui, china}.
\newblock \bibinfo{journal}{Remote Sensing} \bibinfo{volume}{15}.
\newblock \URLprefix \url{https://www.mdpi.com/2072-4292/15/3/598}, \DOIprefix\doi{10.3390/rs15030598}.
%Type = Article
\bibitem[{Zhao and Liu(2021)}]{zhao_important_2021}
\bibinfo{author}{Zhao, F.}, \bibinfo{author}{Liu, Y.}, \bibinfo{year}{2021}.
\newblock \bibinfo{title}{Important meteorological predictors for long-range wildfires in {China}}.
\newblock \bibinfo{journal}{Forest Ecology and Management} \bibinfo{volume}{499}, \bibinfo{pages}{119638}.
\newblock \URLprefix \url{https://www.sciencedirect.com/science/article/pii/S0378112721007283}, \DOIprefix\doi{10.1016/j.foreco.2021.119638}.
%Type = Misc
\bibitem[{Zhao et~al.(2024)Zhao, Prapas, Karasante, Xiong, Papoutsis, Camps-Valls and Zhu}]{zhao2024causalgraphneuralnetworks}
\bibinfo{author}{Zhao, S.}, \bibinfo{author}{Prapas, I.}, \bibinfo{author}{Karasante, I.}, \bibinfo{author}{Xiong, Z.}, \bibinfo{author}{Papoutsis, I.}, \bibinfo{author}{Camps-Valls, G.}, \bibinfo{author}{Zhu, X.X.}, \bibinfo{year}{2024}.
\newblock \bibinfo{title}{Causal graph neural networks for wildfire danger prediction}.
\newblock \URLprefix \url{https://arxiv.org/abs/2403.08414}, \href{http://arxiv.org/abs/2403.08414}{{\tt arXiv:2403.08414}}.
%Type = Article
\bibitem[{Zhou et~al.(2023)Zhou, Wang, Garcia, Chen, da~Silva, Wang, Román, Hyer and Miller}]{10081011}
\bibinfo{author}{Zhou, M.}, \bibinfo{author}{Wang, J.}, \bibinfo{author}{Garcia, L.C.}, \bibinfo{author}{Chen, X.}, \bibinfo{author}{da~Silva, A.M.}, \bibinfo{author}{Wang, Z.}, \bibinfo{author}{Román, M.O.}, \bibinfo{author}{Hyer, E.J.}, \bibinfo{author}{Miller, S.D.}, \bibinfo{year}{2023}.
\newblock \bibinfo{title}{Enhancement of nighttime fire detection and combustion efficiency characterization using suomi-npp and noaa-20 viirs instruments}.
\newblock \bibinfo{journal}{IEEE Transactions on Geoscience and Remote Sensing} \bibinfo{volume}{61}, \bibinfo{pages}{1--20}.
\newblock \DOIprefix\doi{10.1109/TGRS.2023.3261664}.
%Type = Article
\bibitem[{Zhu et~al.(2021)Zhu, Webb, Yebra, Scortechini, Miller and Petitjean}]{zhu_live_2021}
\bibinfo{author}{Zhu, L.}, \bibinfo{author}{Webb, G.I.}, \bibinfo{author}{Yebra, M.}, \bibinfo{author}{Scortechini, G.}, \bibinfo{author}{Miller, L.}, \bibinfo{author}{Petitjean, F.}, \bibinfo{year}{2021}.
\newblock \bibinfo{title}{Live fuel moisture content estimation from {MODIS}: {A} deep learning approach}.
\newblock \bibinfo{journal}{ISPRS Journal of Photogrammetry and Remote Sensing} \bibinfo{volume}{179}, \bibinfo{pages}{81--91}.
\newblock \URLprefix \url{https://www.sciencedirect.com/science/article/pii/S0924271621001957}, \DOIprefix\doi{10.1016/j.isprsjprs.2021.07.010}.
%Type = Article
\bibitem[{ZORMPAS et~al.(2017)ZORMPAS, VASILAKOS, ATHANASIS, SOULAKELLIS and KALABOKIDIS}]{zormpas2017dead}
\bibinfo{author}{ZORMPAS, K.}, \bibinfo{author}{VASILAKOS, C.}, \bibinfo{author}{ATHANASIS, N.}, \bibinfo{author}{SOULAKELLIS, N.}, \bibinfo{author}{KALABOKIDIS, K.}, \bibinfo{year}{2017}.
\newblock \bibinfo{title}{Dead fuel moisture content estimation using remote sensing}.
\newblock \bibinfo{journal}{European Journal of Geography} \bibinfo{volume}{8}.
%Type = Article
\bibitem[{Zubkova et~al.(2024)Zubkova, L{\"o}tter, Bronkhorst and Giglio}]{zubkova2024assessment}
\bibinfo{author}{Zubkova, M.}, \bibinfo{author}{L{\"o}tter, M.}, \bibinfo{author}{Bronkhorst, F.}, \bibinfo{author}{Giglio, L.}, \bibinfo{year}{2024}.
\newblock \bibinfo{title}{Assessment of the effectiveness of coarse resolution fire products in monitoring long-term changes in fire regime within protected areas in south africa}.
\newblock \bibinfo{journal}{International Journal of Applied Earth Observation and Geoinformation} \bibinfo{volume}{132}, \bibinfo{pages}{104064}.

\end{thebibliography}
%% else use the following coding to input the bibitems directly in the
%% TeX file.

%%\begin{thebibliography}{00}

%% \bibitem[Author(year)]{label}
%% For example:

%% \bibitem[Aladro et al.(2015)]{Aladro15} Aladro, R., Martín, S., Riquelme, D., et al. 2015, \aas, 579, A101

%%\end{thebibliography}

\end{document}